\documentclass[preprint,12pt]{elsarticle}

\usepackage{amssymb}
\usepackage{relsize}
\usepackage{array}
\usepackage{caption}
\usepackage{subcaption}
\usepackage{amsmath}
  \DeclareMathOperator*{\argmax}{argmax}
  \DeclareMathOperator*{\argmin}{argmin}

\newcolumntype{L}[1]{>{\raggedright\let\newline\\\arraybackslash\hspace{0pt}}m{#1}}
\newcolumntype{C}[1]{>{\centering\let\newline\\\arraybackslash\hspace{0pt}}m{#1}}
\newcolumntype{R}[1]{>{\raggedleft\let\newline\\\arraybackslash\hspace{0pt}}m{#1}}
\newcolumntype{M}[1]{>{\centering\arraybackslash}m{#1}}

\usepackage[margin=2.5cm]{geometry}
\usepackage{booktabs}

\usepackage{tikz}
\usepackage{adjustbox}
\usepackage{amsfonts} 
\usepackage{siunitx}
\usepackage{float}

\usepackage{hyperref}

\newsavebox{\mysavebox}

\journal{Neural Networks}

\begin{document}

\begin{frontmatter}



    \title{A matter of attitude: Focusing on positive and active gradients to boost saliency maps}


\author[inst1]{Oscar Llorente\corref{cor1}}
\ead{oscar.llorente.gonzalez@ericsson.com}

\affiliation[inst1]{organization={Cognitive Software, Ericsson},
            addressline={Retama Ed 1 Torre Suecia}, 
            city={Madrid},
            postcode={28045}, 
            state={Madrid},
            country={Spain}}

\author[inst2]{Jaime Boal}
\ead{jaime.boal@iit.comillas.edu}
\author[inst2]{Eugenio F. Sánchez-Úbeda}
\ead{eugenio.sanchez@iit.comillas.edu}

\affiliation[inst2]{organization={Institute for Research in Technology~(IIT), ICAI School of Engineering, Comillas Pontifical University},
            addressline={Santa Cruz de Marcenado, 26}, 
            city={Madrid},
            postcode={28015}, 
            state={Madrid},
            country={Spain}}

\begin{abstract}
Saliency maps have become one of the most widely used interpretability techniques for convolutional neural networks~(CNN) due to their simplicity and the quality of the insights they provide. However, there are still some doubts about whether these insights are a trustworthy representation of what CNNs use to come up with their predictions. This paper explores how rescuing the sign of the gradients from the saliency map can lead to a deeper understanding of multi-class classification problems. Using both pretrained and trained from scratch CNNs we unveil that considering the sign and the effect not only of the correct class, but also the influence of the other classes, allows to better identify the pixels of the image that the network is really focusing on. Furthermore, how occluding or altering those pixels is expected to affect the outcome also becomes clearer\footnote{All code to replicate our findings will be available here: \url{https://github.com/OscarLlorente/positive_active_saliency_maps}}. 
\end{abstract}




\end{frontmatter}


\section{Introduction}
\label{sec:introduction}
The overwhelming mediatic interest that generative artificial intelligence is drawing lately is fostering the adoption of deep learning models in almost every area of our lives. Unfortunately, the outstanding advances brought by these massive models have yet to be accompanied by an equivalent effort to make them more interpretable. Blindly using complex models without worrying about how their outputs are generated entails risks that must be mitigated if we strive to adopt them in sensitive sectors, as many authorized voices and legislators are already pointing out. 

There are basically two approaches to address this issue: building models that are easier to understand by design and at the same time match the performance of their black box counterparts, or developing techniques to disclose what is going on inside the black boxes. This paper concentrates on the field of computer vision, where there are indeed some attempts to construct interpretable models for object classification~\cite{Zhang2021} and medical applications~\cite{Singh2022}. However, due to their great feature extraction capabilities, far better than reknown traditional engineered features such as SIFT~\cite{Lowe2004}, many modern computer vision solutions still rely on regular convolutional neural networks~(CNNs)~\cite{LeCun1998} as part of their pipeline, whose inner workings are hard to interpret and understand. 

Over the past decade, the research community has produced several techniques that seek to shed some light on how CNNs come up with their predictions. Leaving the visual inspection of the convolutional filters aside, most of the proposals consist in studying the effect of exciting the model with known stimuli and projecting the result back into the input image. 

One family of techniques attempts to approximate the trained network with simpler models. Zhou~et~al.~\cite{Zhou2015} remove information from the input images to obtain a minimal representation that preserves as little visual information as possible without significantly impacting the classification score. This is done by segmenting the image and iteratively discarding those regions that contribute the least. Ribeiro~et~al.~\cite{Ribeiro2016} propose~LIME, an algorithm that approximates any classifier or regressor locally with interpretable models. To deal with images they extract $K$ superpixels and treat them as input binary features for a linear model. Similarly, Frosst and Hinton~\cite{Frosst2017} suggest building a binary soft decision tree from the learned filters.

Another popular approach relies on backpropagation. Deconvolutional Networks or DeconvNets~\cite{Zeiler2014} invert the order of the CNN layers to discover the input pixels responsible for the information encoded in every feature map. DeconvNets allow gathering evidence about the type of features every CNN layer is able to extract, from basic geometries like corners and edges at the beginning to class-specific features as one proceeds deeper into the network. 

Due to their simplicity, and perhaps for being one of the seminal methods in this category, saliency maps~\cite{Simonyan2014} have become one of the most popular local interpretability methods for CNNs. They compute the absolute gradient of the target output with respect to every input feature (i.e.,~pixels on images) and are commonly employed in multi-class classification problems:
\begin{equation}
    \label{eq: cnn output}
    \hat{c} = \argmax_{c \in C}S_c(\boldsymbol{I})
\end{equation}
where $\boldsymbol{I}$ is the input image, $C$ the set of all possible classes, and $S_c$ corresponds to the classification score for a given class $c$. Since images are usually encoded in the RGB color space, it is important to bear in mind that $\boldsymbol{I} \in \mathbb{R}^{\text{channels} \times \text{height} \times \text{width}}$ is a tensor. The original saliency map method is mathematically expressed as
\begin{equation}
    \boldsymbol{M}_{i\, j} = \max_{k} \bigg | \frac{\partial S_{\hat{c}}}{\partial \boldsymbol{I}_{k\, i\, j}} \bigg |
    \label{eq: 3d saliency map}
\end{equation}
$\boldsymbol{M}$ is a 2D map, $k$ indicates a specific channel and, $i$ and $j$, are the row and column of every pixel, respectively. According to~\cite{Simonyan2014}, the brightest points in a saliency map (i.e.,~the derivatives with a larger absolute value) are the pixels that, with a minimal change, should affect the class score the most. As shown in (\ref{eq: 3d saliency map}), the maximum over the three channels is computed to obtain a single value per pixel. Even though this decision may seem arbitrary, it is the convention followed in almost every subsequent paper on saliency maps.

There have been several attempts to reduce the inherent noise of saliency maps like that of Shrikumar~et~al.~\cite{Shrikumar2017}, who suggest multiplying element-wise the input and the gradient to produce sharper representations. However, the fact that on visual inspection the representation resembles the input image is no guarantee of its correctness as put forward in the sanity checks proposed by Adebayo~et~al.~\citep{Adebayo2018}. Apparently, the same happens to varying degrees in other similar methods such as $\epsilon$-LRP, DeepLIFT~\cite{Ancona2018}, or integrated gradients~\cite{Sundarajan2017}. The technique does not highlight what the neural network has learned to pay attention to, but rather tends to augment features from the input image such as edges that may or may not be relevant to the prediction.

This paper proposes several improvements over the regular saliency maps to increase the insights that can be extracted. The contributions can be summarized in three main points:
\begin{itemize}
  \item Instead of taking the absolute value of the gradients, and thus neglecting their sign, we prove that preserving this information enhances interpretability in multi-class classification problems by better identifying which pixels assist or deceive the network.  
  \item The network would want pixels with a positive gradient to have higher intensities and those with negative gradients to be dimmed towards zero. This fact makes occlusion experiments more self-explanatory, since it is easier to understand the meaning of replacing a given pixel with a brighter or darker one, ultimately with white or black.
  \item Typically, only the class of interest is considered when analyzing saliency maps. Based on the gradient sign, a set of metrics have been defined to quantitatively compare the impact on the prediction of a particular class caused by the rest of the classes.
\end{itemize}

The remainder of the document is structured as follows. It starts with a brief discussion of the implications of ignoring the sign of the gradients (\autoref{sec:the sign in saliency maps}). Using the information provided by the sign, \autoref{sec:multi_class_saliency_map} explores the effect that modifying a pixel value has on a multi-class classification problem. Finally, \autoref{sec:experiments} presents the experiments conducted to support the conclusions derived in \autoref{sec:conclusions and future research}. 

\section{The importance of the gradient sign}
\label{sec:the sign in saliency maps}
To the top of our knowledge, there is little research about the meaning or impact of the sign in saliency maps. The only article that briefly discusses this topic is~\cite{Smilkov2017}, which explains that the raw value of gradients (without taking the absolute value) is commonly used in the MNIST dataset~\cite{MNISTHandwrittenDigit}, but not in other datasets like ImageNet~\cite{ImageNet}. Apparently, experimental results suggest that on MNIST raw gradients produce clearer saliency maps and, at the same time, worse representations on ImageNet. Since the latter is the \textit{de facto} standard dataset for CNNs, in general saliency maps are implemented with the absolute value.

However, taking the absolute value of every pixel in the saliency map comes at a cost and some enlightening information is lost. In terms of explainability, the opportunity of knowing which regions of the image should be brighter or darker to improve the classification accuracy is disregarded. Moreover, if both pixels with positive and negative gradients are combined in the same image without any distinctions, the representation can become confusing. Sometimes it may seem as if the model is not able to tell apart regions that should be brighter or darker like, for instance, an object (positive gradient) on an uninformative background (negative). Therefore, two sets of pixels can be distinguished in the image:

\begin{itemize}
    \item Pixels that improve the classification score of the predicted class if their value is \emph{increased}, since they have a positive value (gradient) in the saliency map.
    \item Pixels that improve the classification score of the predicted class if their value is \emph{decreased}, because they have a negative gradient.
\end{itemize}

The advantage of this separation with respect to focusing on the raw gradients and then normalizing their values to represent them on a single image is that in~\cite{Smilkov2017} zero gradients shine at medium intensity after scaling, conveying a misleading idea of importance to those pixels. Instead, we propose creating two different visualizations before taking the absolute value (or the ReLU function, which naturally provides the same result if positive and negative gradients are handled separately):

\begin{itemize}
    \item Positive saliency maps:
    \begin{equation}
        \boldsymbol{M}_{i\, j} = \max_{k} \bigg ( \text{ReLU} \bigg (\frac{\partial S_{\hat{c}}}{\partial \boldsymbol{I}_{k\, i\, j}} \bigg ) \bigg )
        \label{eq: positive saliency map}
    \end{equation}

    \item Negative saliency maps:
    \begin{equation}
      \boldsymbol{M}_{i\, j} = \max_{k} \bigg ( \text{ReLU} \bigg ( - \frac{\partial S_{\hat{c}}}{\partial \boldsymbol{I}_{k\, i\, j}} \bigg ) \bigg )
      \label{eq: negative saliency map}
  \end{equation}
\end{itemize}

\section{Multi-class saliency maps}
\label{sec:multi_class_saliency_map}
All saliency map techniques use the actual class to compute the derivatives. In multi-class classification problems this approach disregards the effect of the rest of the classes. Despite there can be pixel gradients computed with respect to incorrect classes with a higher value, the current techniques do not draw attention to this fact. Whenever this happens, the interpretation of the saliency map changes since if the value of this pixel is increased, the classification score for the incorrect class will improve more than that of the true class, worsening the prediction. 

Taking the absolute value makes things even more undecipherable. Once you lose the sign information, you can no longer determine whether increasing the intensity of a pixel is bound to increase or decrease the score of a given class. In order to extend the scope of positive and negative saliency maps to consider the effect of all the classes, the definitions put forward in the previous section can be restated:
\begin{itemize}
  \item \emph{Active} pixels are those that improve the most the classification score of the predicted class if their value is \emph{increased}, more than that of any other class considered.
    \item Analogously, \emph{inactive} pixels are those that improve the most the score of the predicted class if their value is \emph{decreased}. Their gradient with respect to the actual class is therefore the lowest (the most negative) among all the classes. These pixels are the ones that cause more confusion to the classifier.
\end{itemize}

Based on these definitions, two additional saliency map visualization can be derived:

\begin{itemize}
    \item \emph{Active saliency maps} highlight the pixels that should be increased to improve the classification score of the true class:
    \begin{equation}
      \boldsymbol{M}_{i\, j} = \max_{k}
        \begin{cases}
            \frac{\partial S_{\hat{c}}}{\partial \boldsymbol{I}_{k\, i\, j}} & \text{if } \frac{\partial S_{\hat{c}}}{\partial \boldsymbol{I}_{k\, i\, j}} = \argmax_{c \in C}\frac{\partial S_{c}}{\partial \boldsymbol{I}_{k\, i\, j}} \\
            0 & \text{otherwise}
        \end{cases}
        \label{eq: active saliency map}
    \end{equation}

    \item \emph{Inactive saliency maps} depict the pixels that should be dimmed to enhance the classification score of the correct class:
    
    \begin{equation}
      \boldsymbol{M}_{i\, j} = \max_{k}
        \begin{cases}
            \frac{\partial S_{\hat{c}}}{\partial \boldsymbol{I}_{k\, i\, j}} & \text{if } \frac{\partial S_{\hat{c}}}{\partial \boldsymbol{I}_{k\, i\, j}} = \argmin_{c \in C}\frac{\partial S_{c}}{\partial \boldsymbol{I}_{k\, i\, j}} \\
            0 & \text{otherwise}
        \end{cases}
        \label{eq: inactive saliency map}
    \end{equation}

\end{itemize}

In conclusion, where positive and negative saliency maps provide information about whether increasing or decreasing the value of particular pixels improves the score of the correct class, active and inactive saliency maps go a step further and identify those pixels that should be altered to increase the confidence of the model in the prediction of the true class. 

\section{Experiments}
\label{sec:experiments}
This section evaluates the proposed new saliency map representations both qualitatively and quantitatively on two different datasets: {CIFAR-10}~\cite{CIFAR10CIFAR100Datasets} and Imagenette~\cite{Imagenette2022}. The former is commonly used in image classification and interpretability papers. The latter is a subset of ten ImageNet classes that allows drawing grounded conclusions without requiring an immense computational effort. 

The new saliency maps have been tested against both trained from scratch and pretrained models. Two models have been trained from scratch using CIFAR-10. The first is a basic CNN with several convolutional blocks with either max- or average-pooling and a final linear layer. The second uses the standard ResNet-18 architecture~\cite{heDeepResidualLearning2016}. For Imagenette, in addition to the previous two models, pre-trained versions of ResNet-18 and ConvNeXt~\cite{Liu2022} have also been evaluated. In all cases, the networks were trained during 50~epochs with a learning rate of \num{0.001} using the AdamW optimizer. \autoref{tab:training_results} shows the accuracies obtained. 

\begin{table}[h]
  \centering
  \small
  \caption{Test set accuracy.}
  \label{tab:training_results}
  \begin{tabular}{L{6cm}C{2cm}C{2cm}}
    \toprule
    {} & CIFAR-10 & Imagenette \\
    \midrule
    Basic CNN & 0.6307 & 0.6205 \\
    ResNet-18  & 0.7569 & 0.8357 \\
    ResNet-18 pre-trained  & - & 0.9733 \\
    ConvNext pre-trained & - & 0.9932 \\
    \bottomrule
  \end{tabular}
\end{table}

\subsection{Qualitative evaluation}
\begin{figure}[ht]
  \centering
  \footnotesize
  \newcommand{\scale}{0.28}
  \setlength{\tabcolsep}{2pt}
  \begin{tabular}{cccccc}
  Image & Original & Positive & Negative & Active & Inactive \\
  
  \includegraphics[scale=\scale]{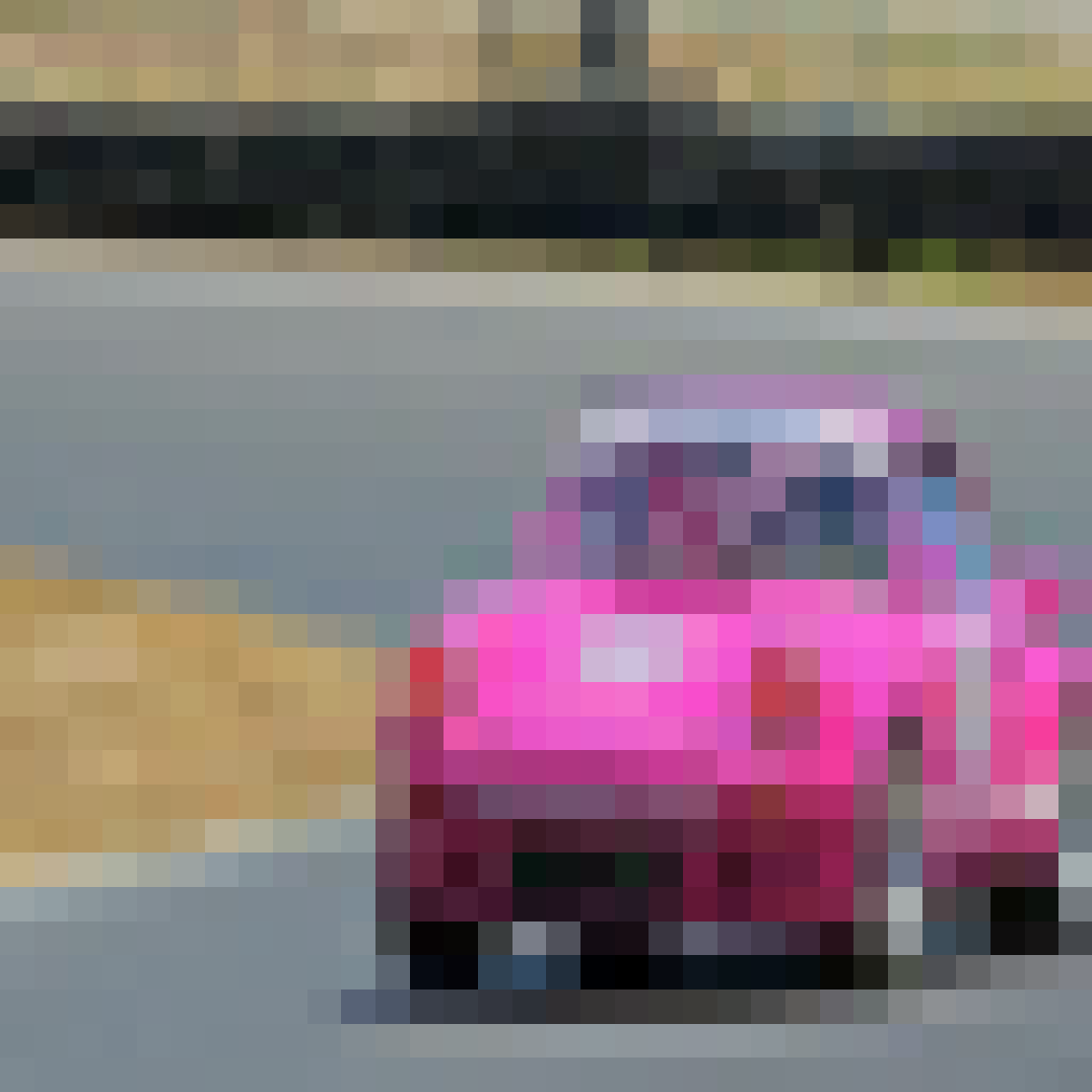} &
  \includegraphics[scale=\scale]{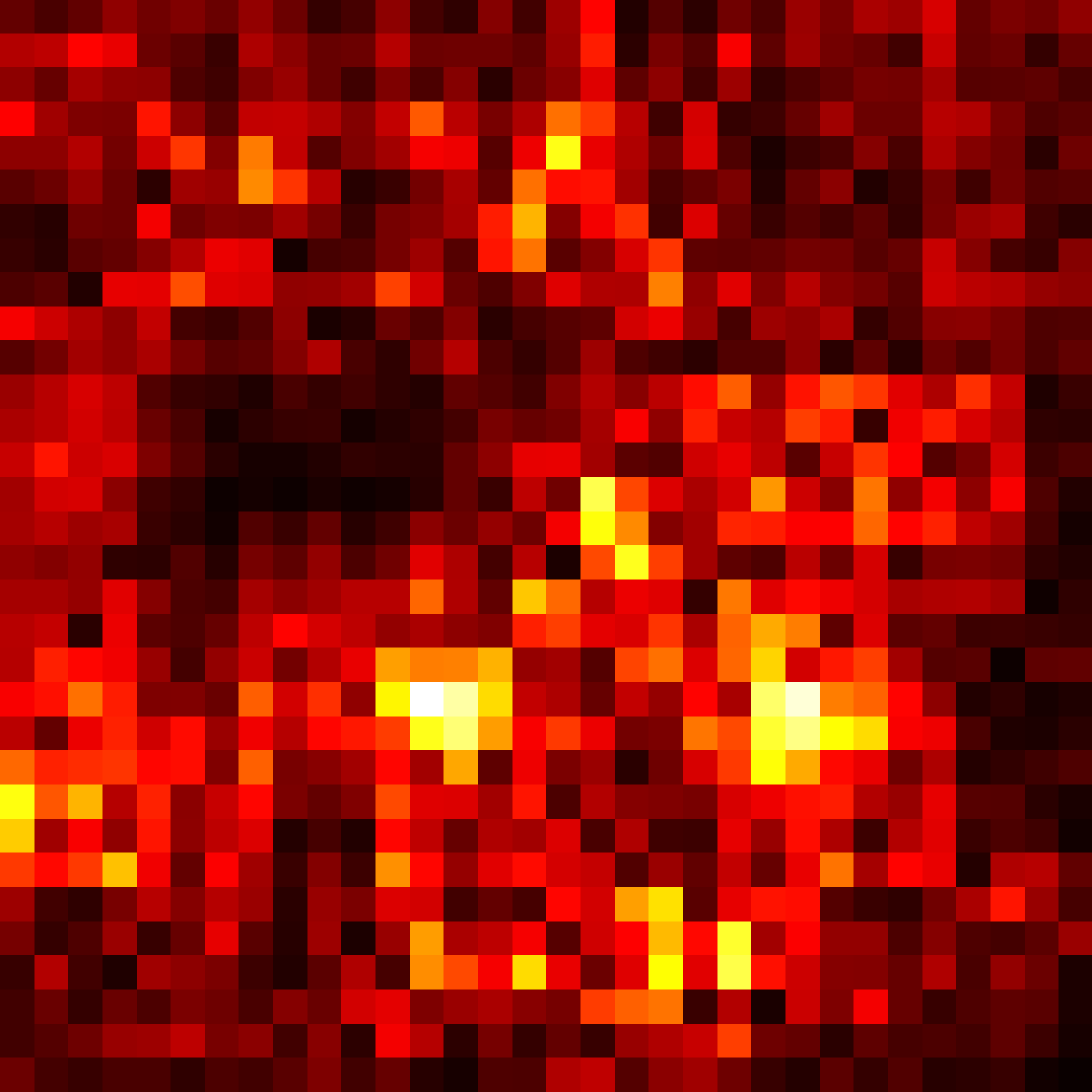} & 
  \includegraphics[scale=\scale]{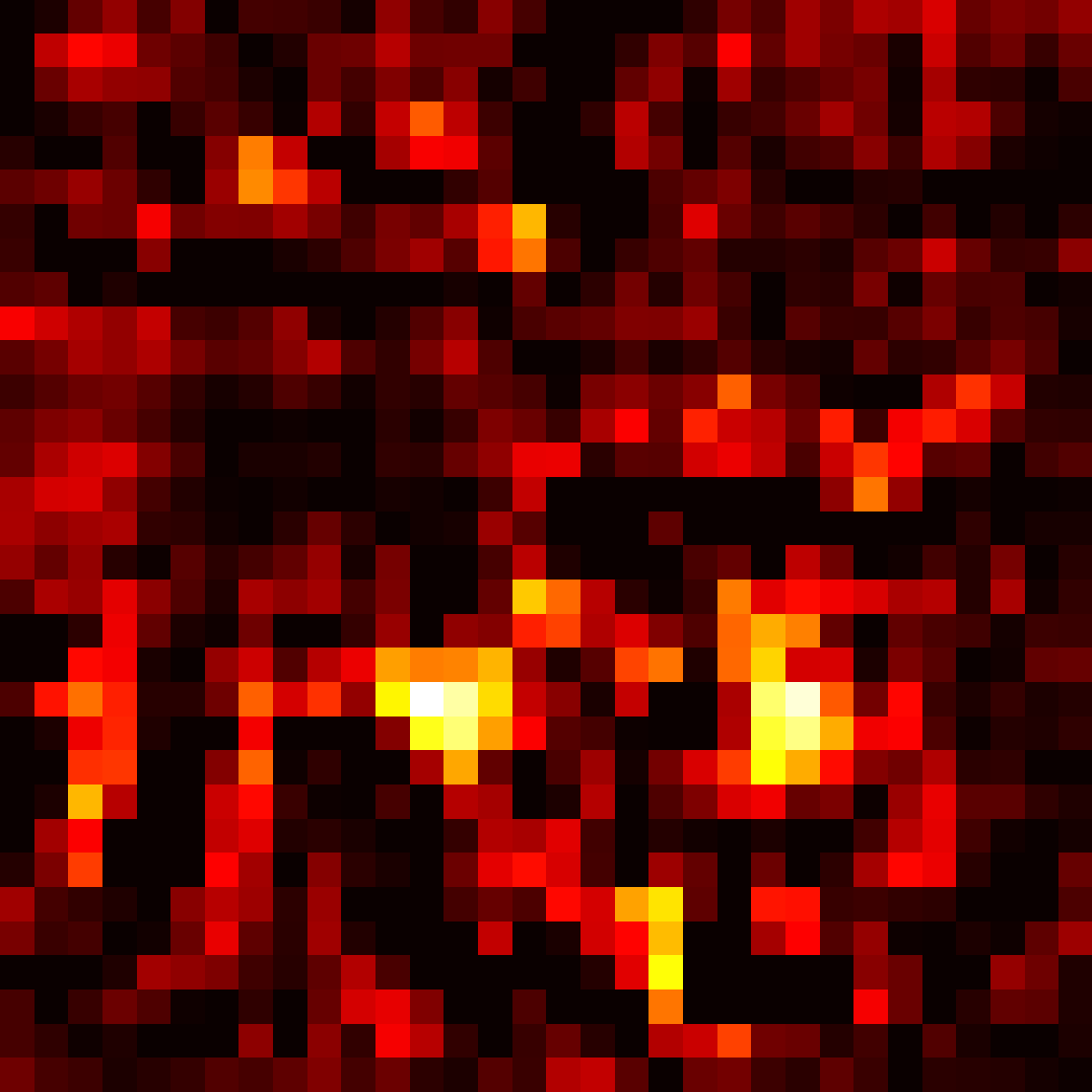} & 
  \includegraphics[scale=\scale]{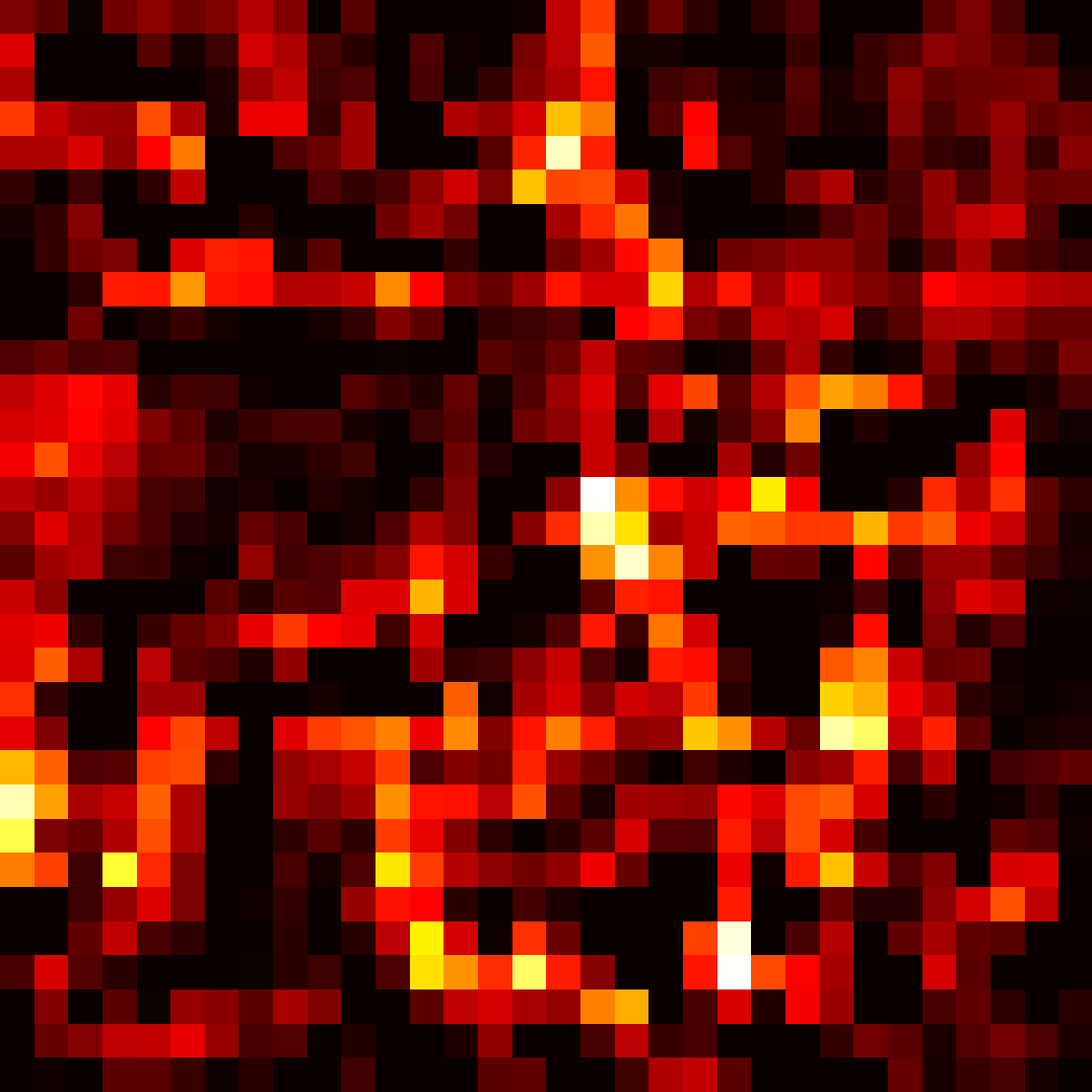} & 
  \includegraphics[scale=\scale]{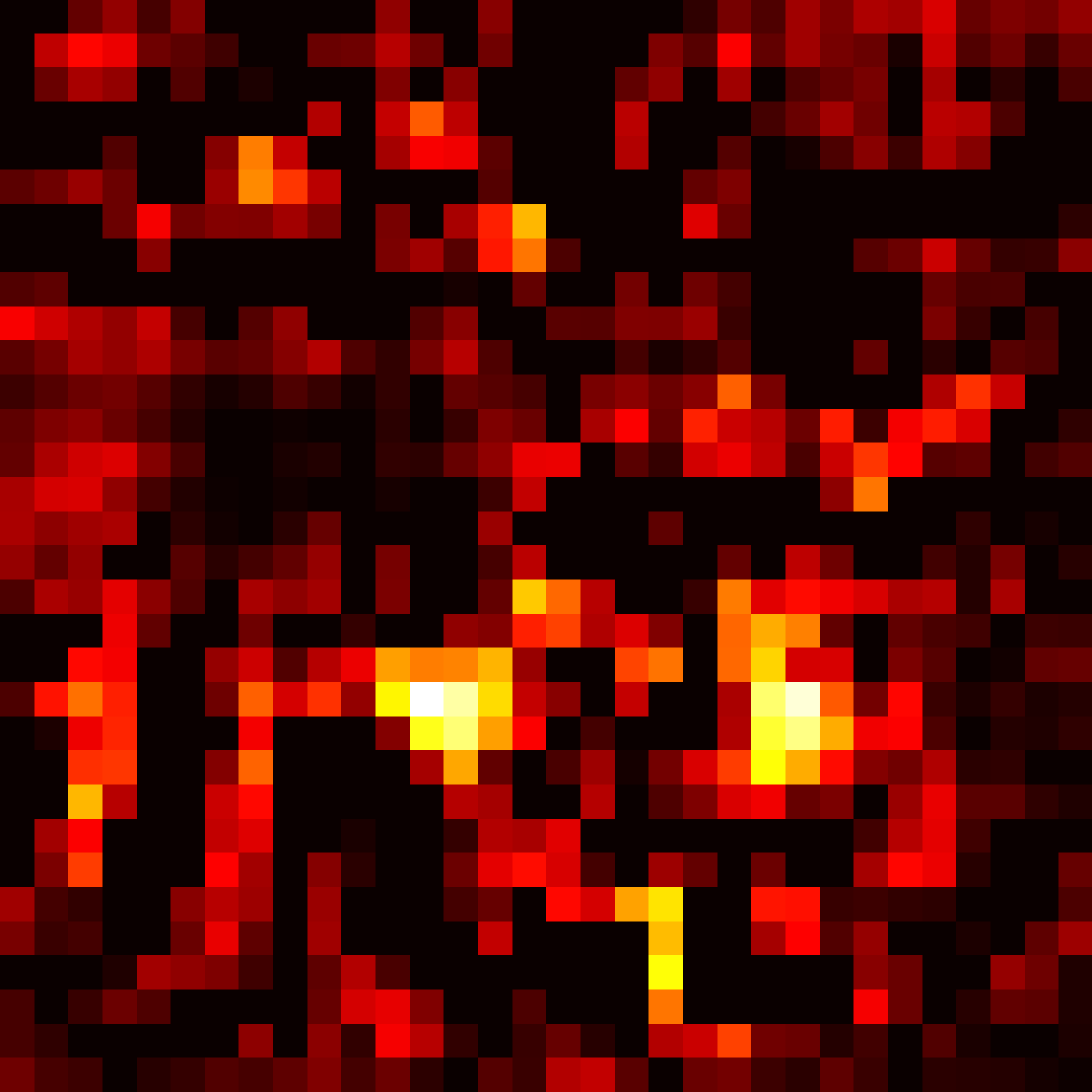} & 
  \includegraphics[scale=\scale]{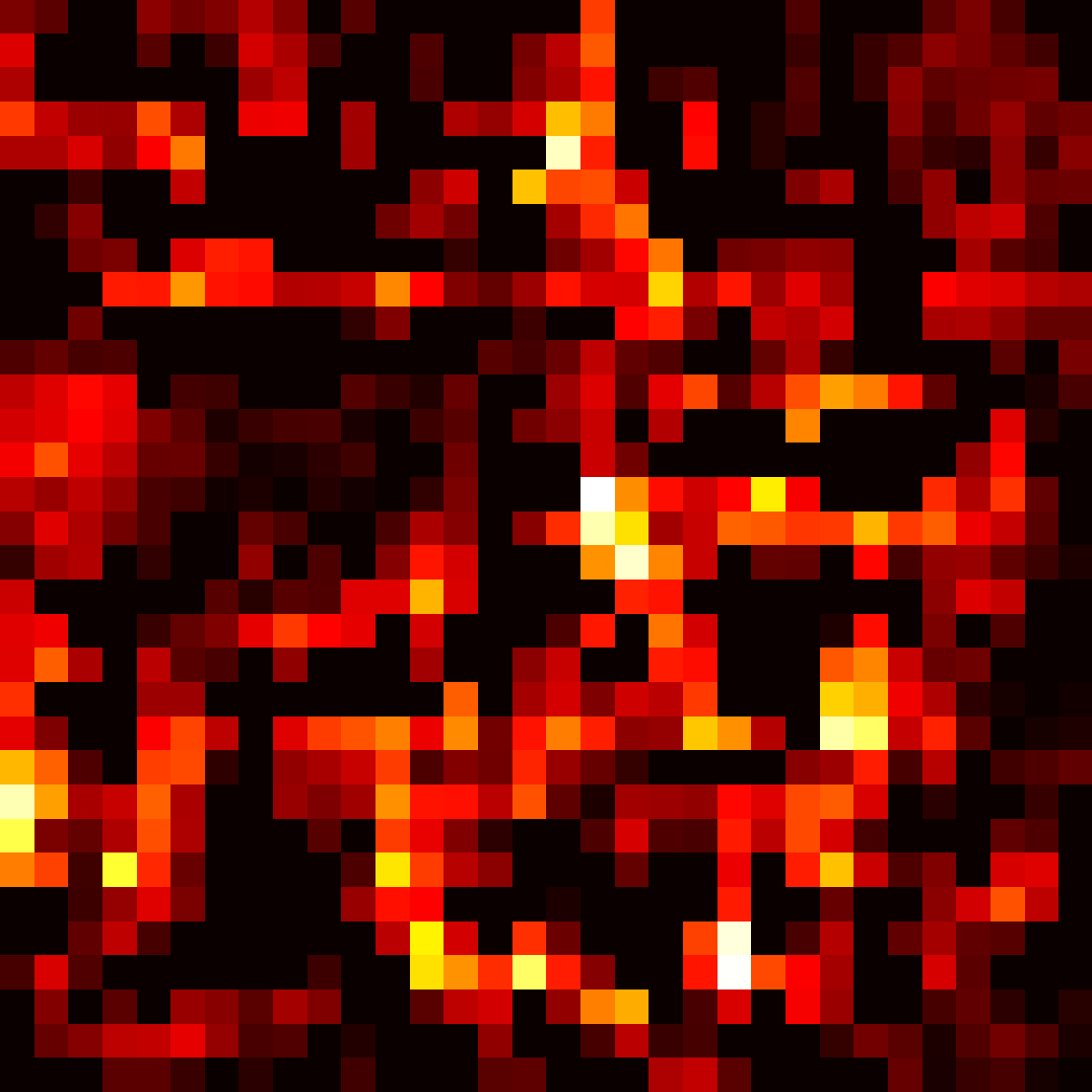} \\
  
  \includegraphics[scale=\scale]{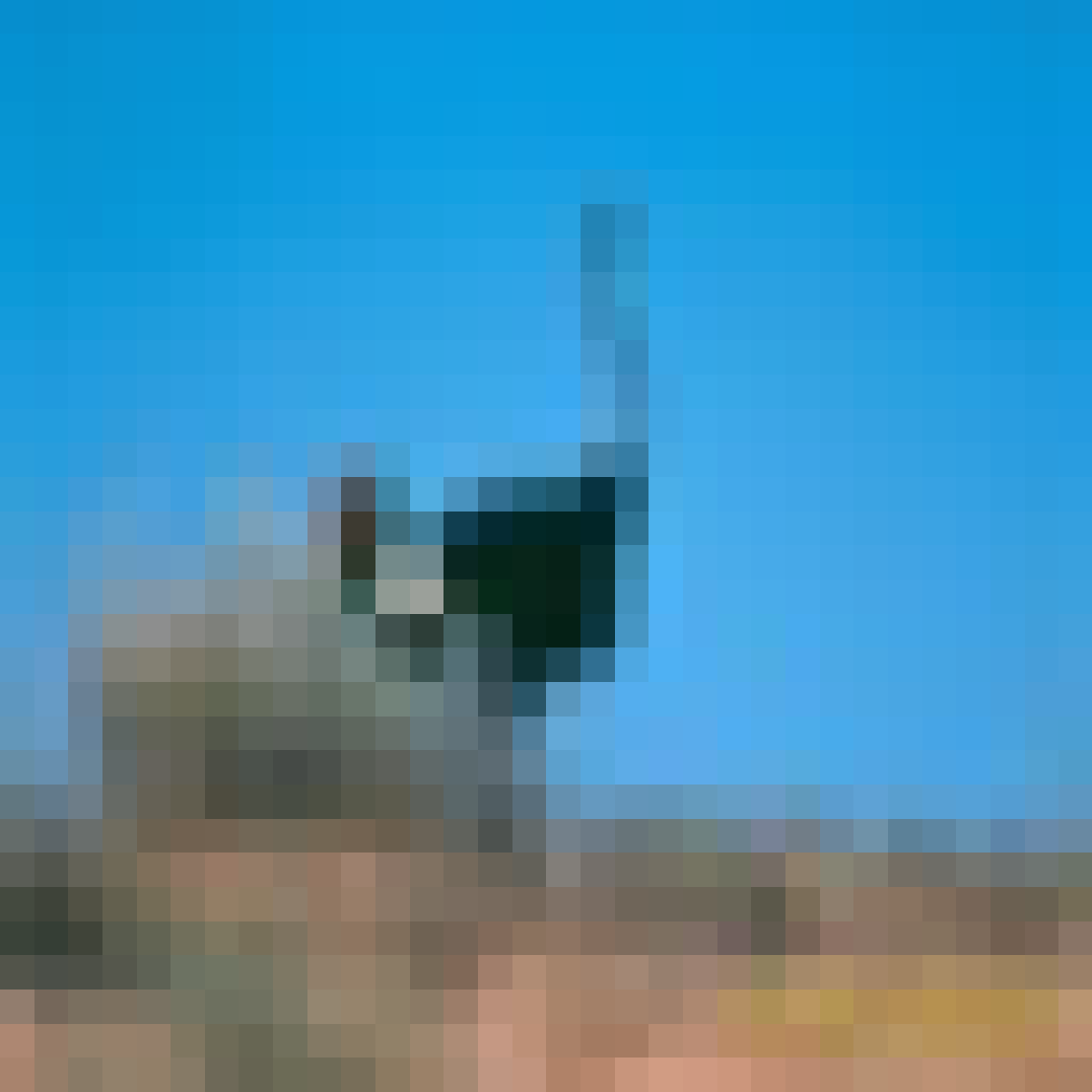} &
  \includegraphics[scale=\scale]{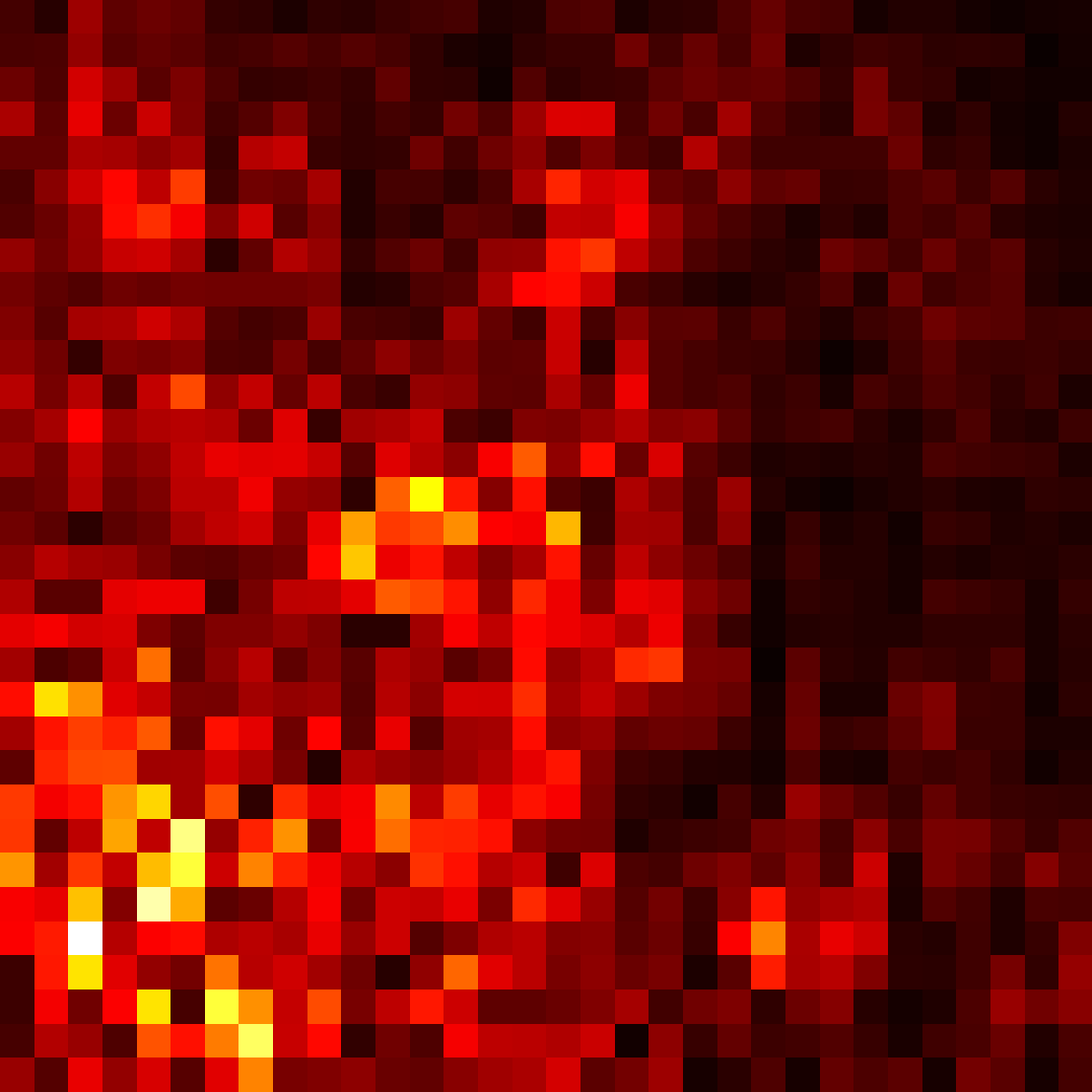} & 
  \includegraphics[scale=\scale]{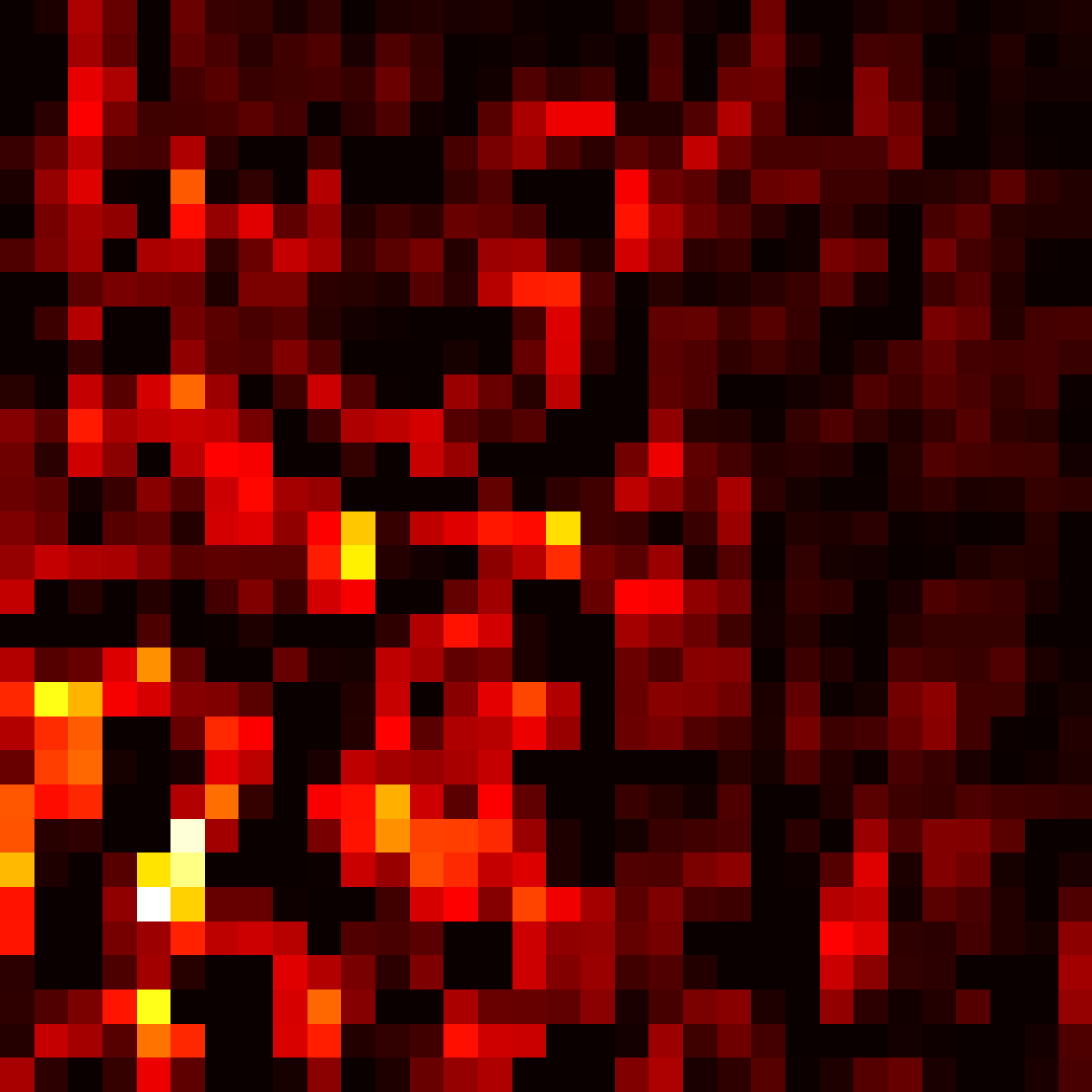} & 
  \includegraphics[scale=\scale]{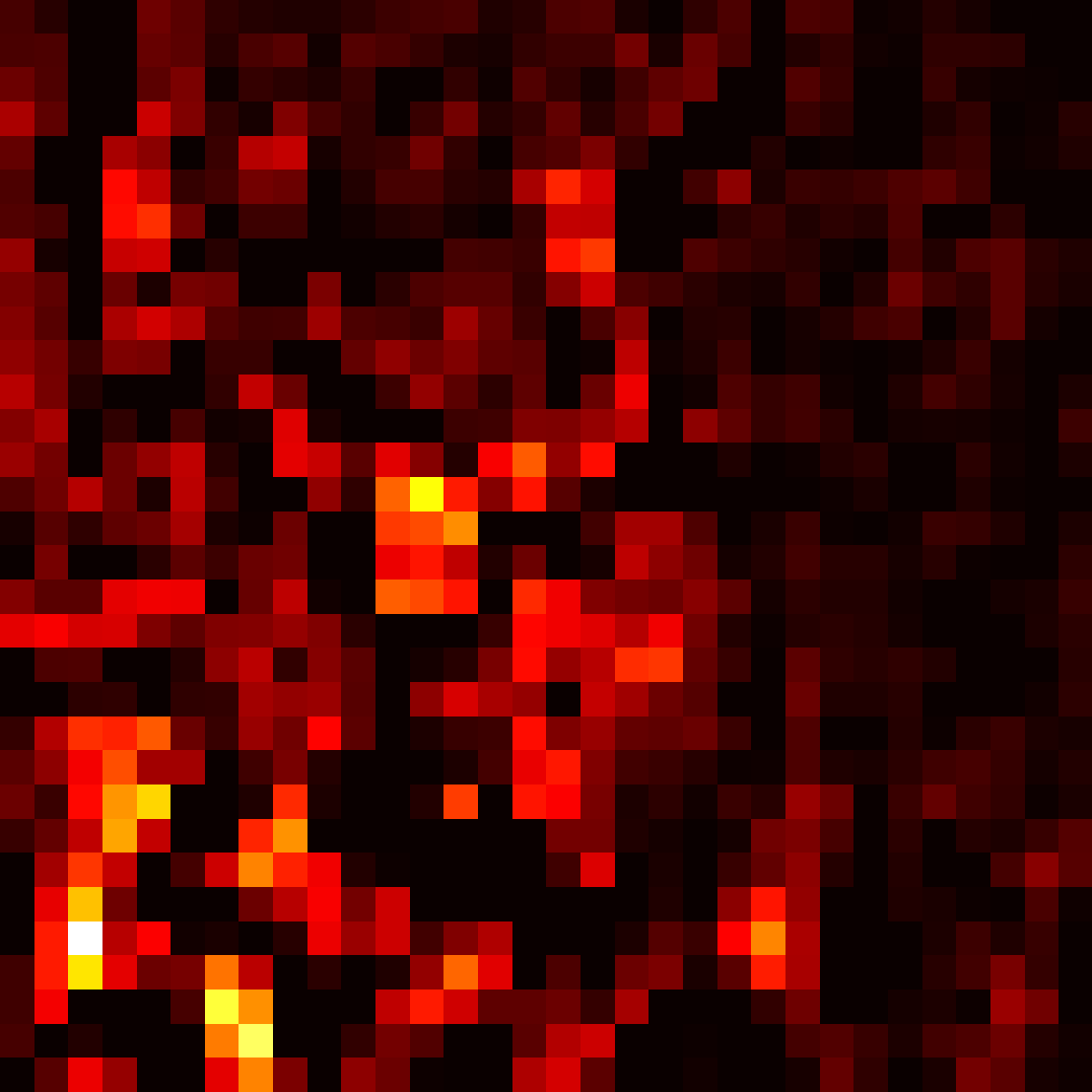} & 
  \includegraphics[scale=\scale]{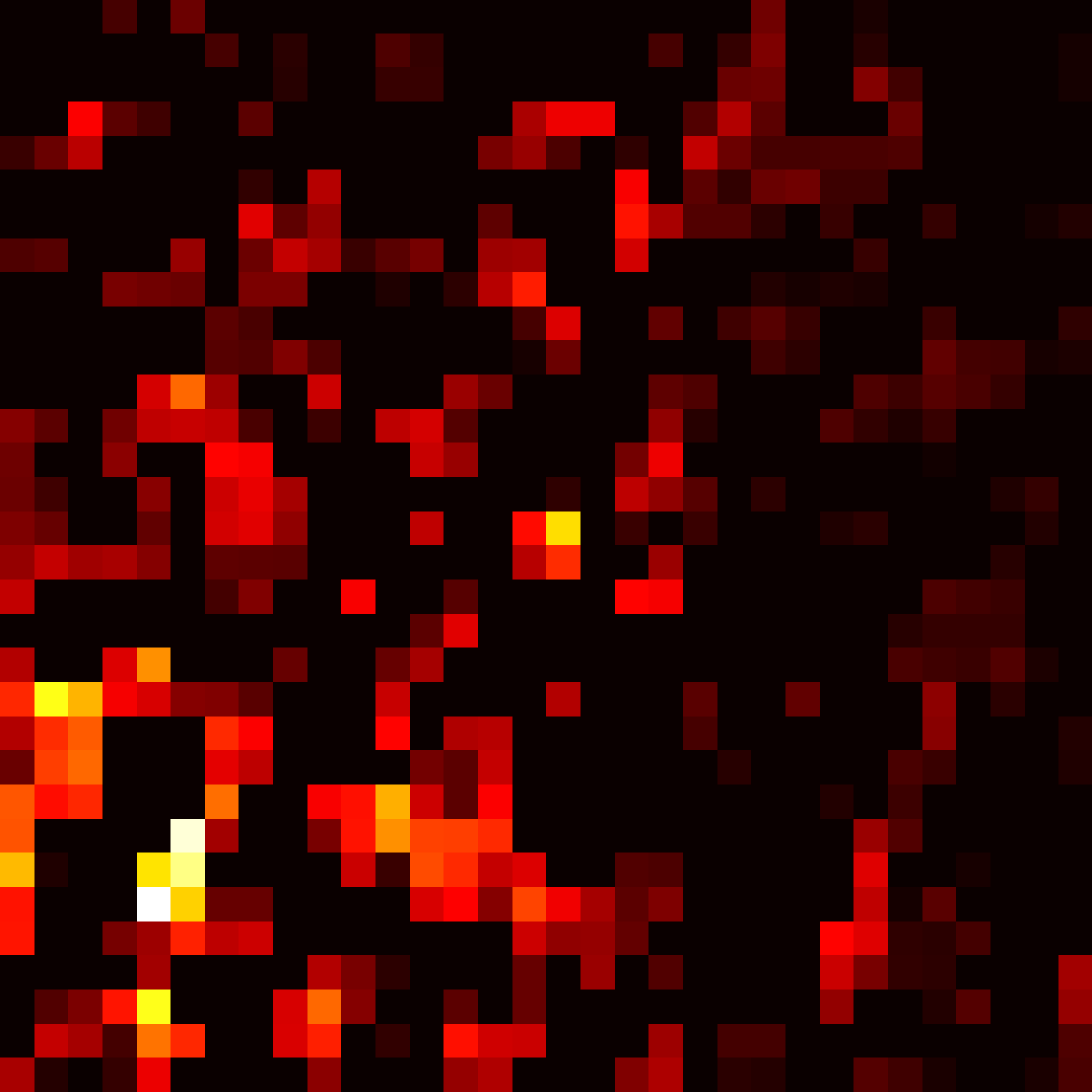} & 
  \includegraphics[scale=\scale]{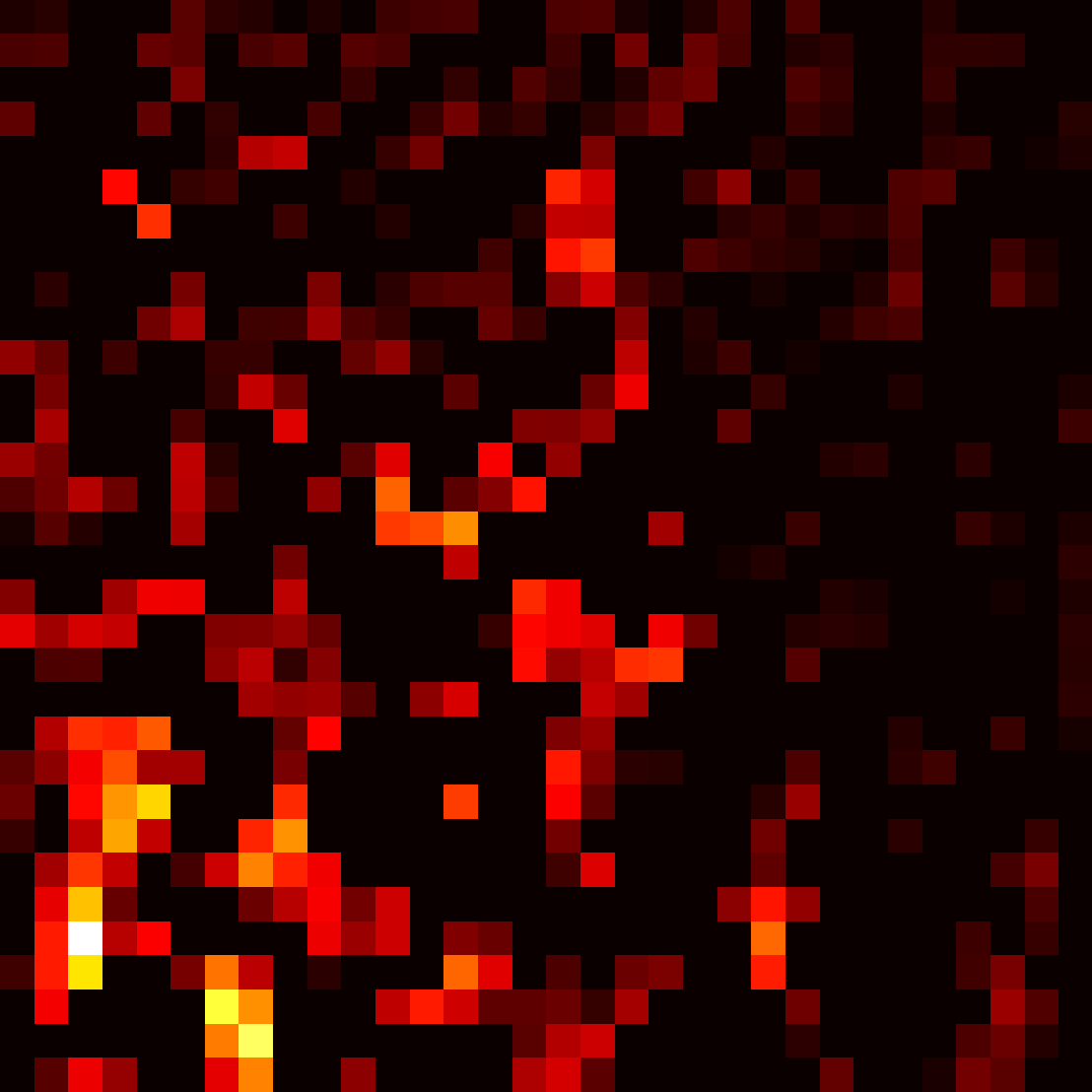} \\
  \end{tabular}
  \caption{Comparison saliency maps CNN CIFAR-10.}
  \label{fig: comparison saliency maps cnn cifar10}
\end{figure}

\begin{figure}[ht]
  \centering
  \footnotesize
  \newcommand{\scale}{0.28}
  \setlength{\tabcolsep}{2pt}
  \begin{tabular}{cccccc}
  Image & Original & Positive & Negative & Active & Inactive \\

  \includegraphics[scale=\scale]{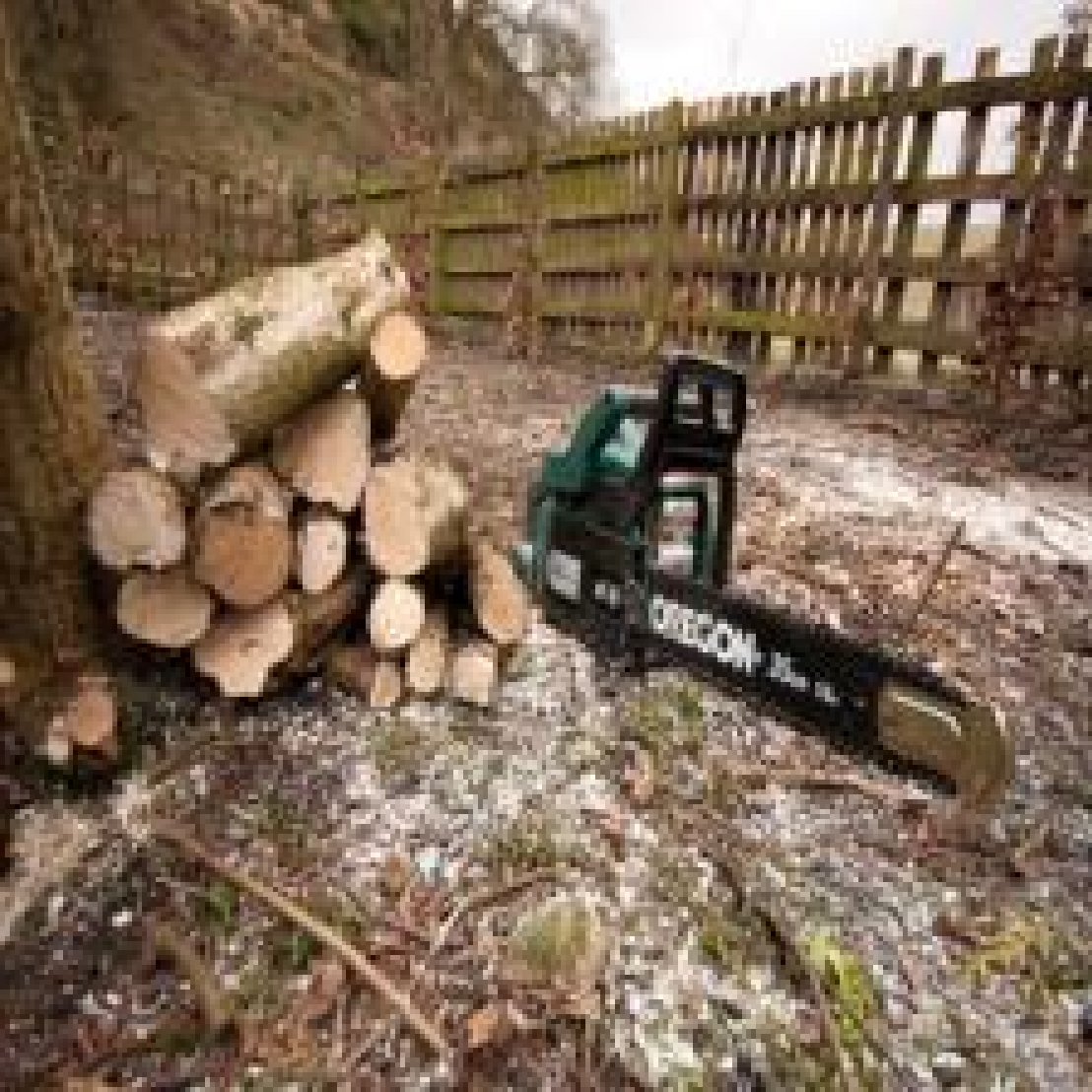} & 
  \includegraphics[scale=\scale]{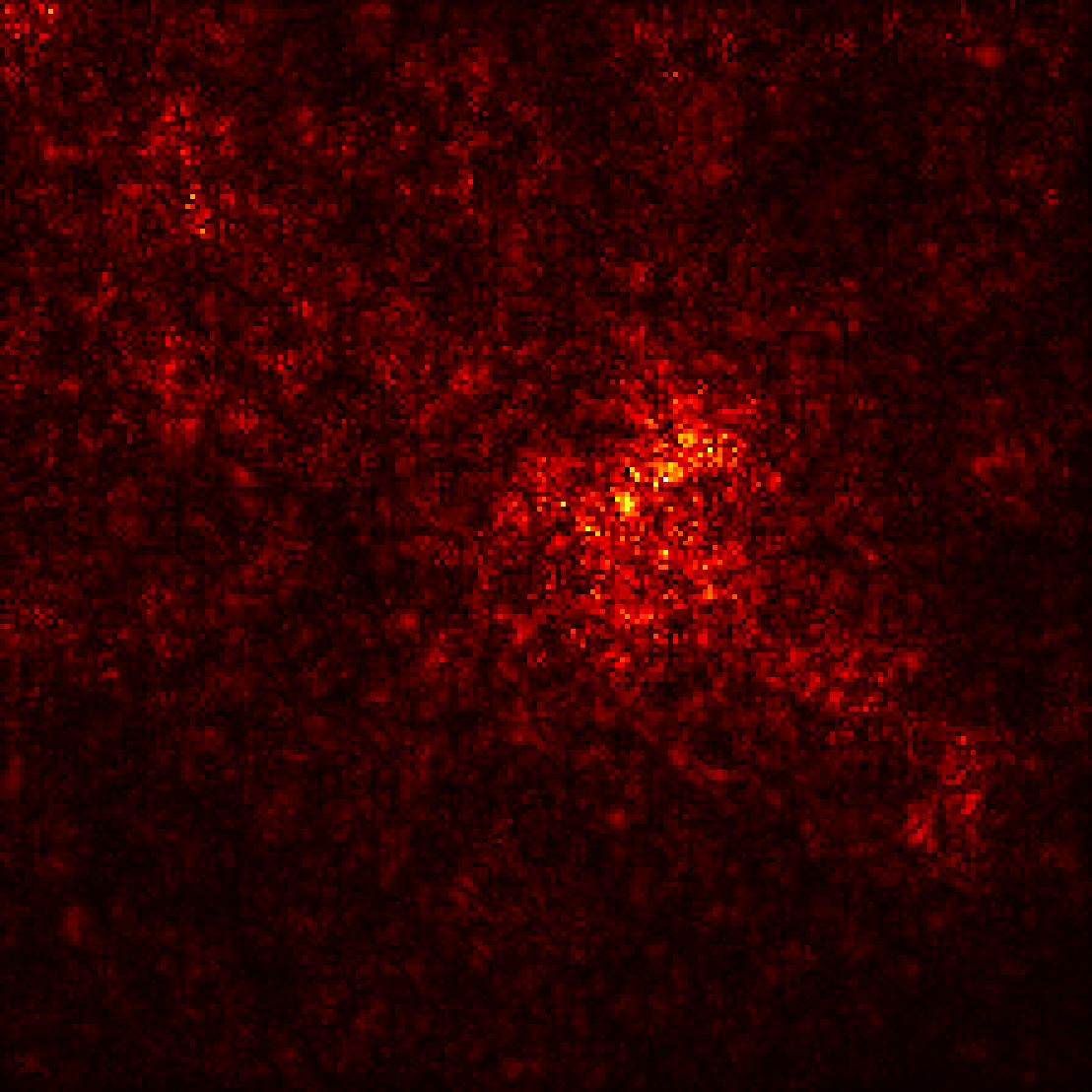} & 
  \includegraphics[scale=\scale]{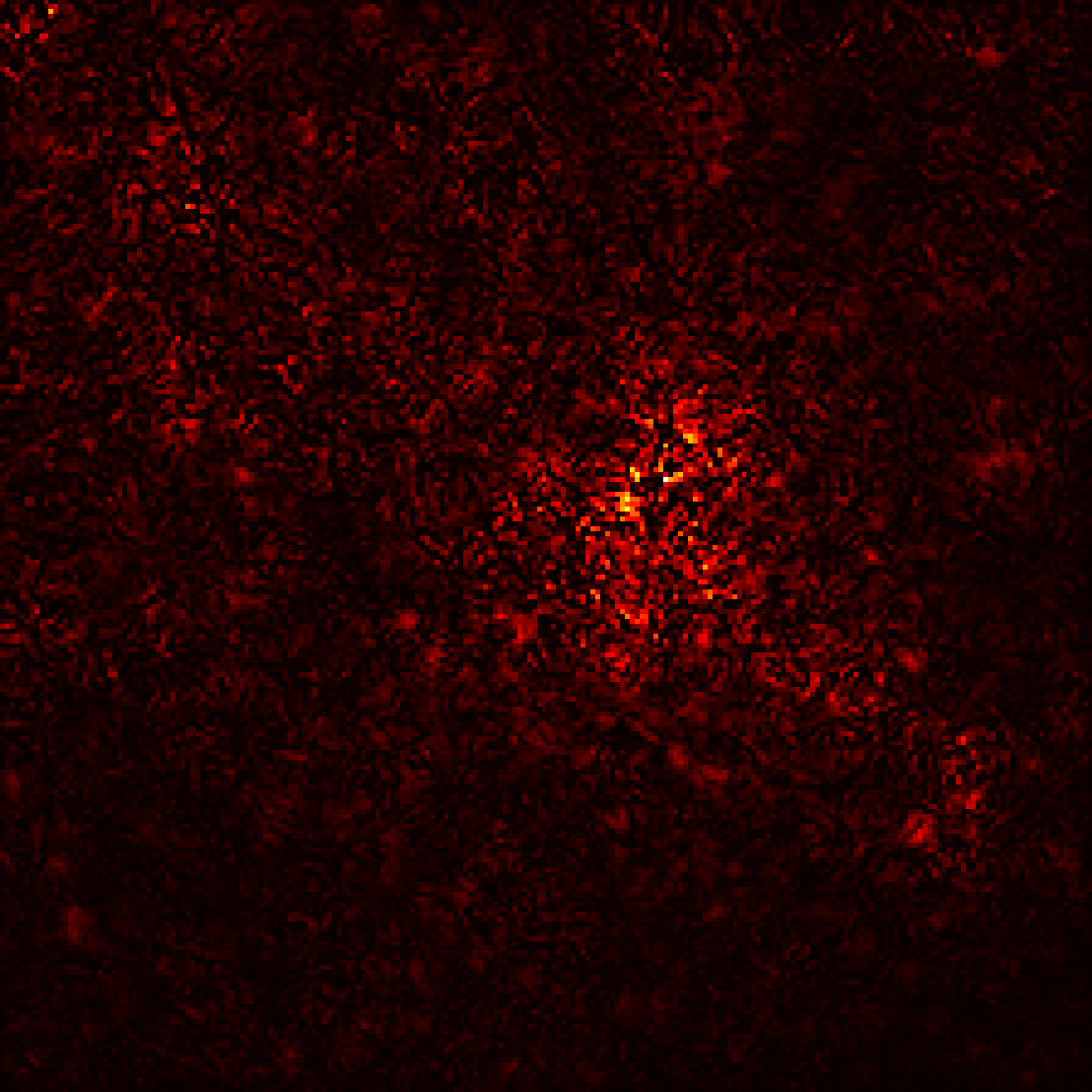} & 
  \includegraphics[scale=\scale]{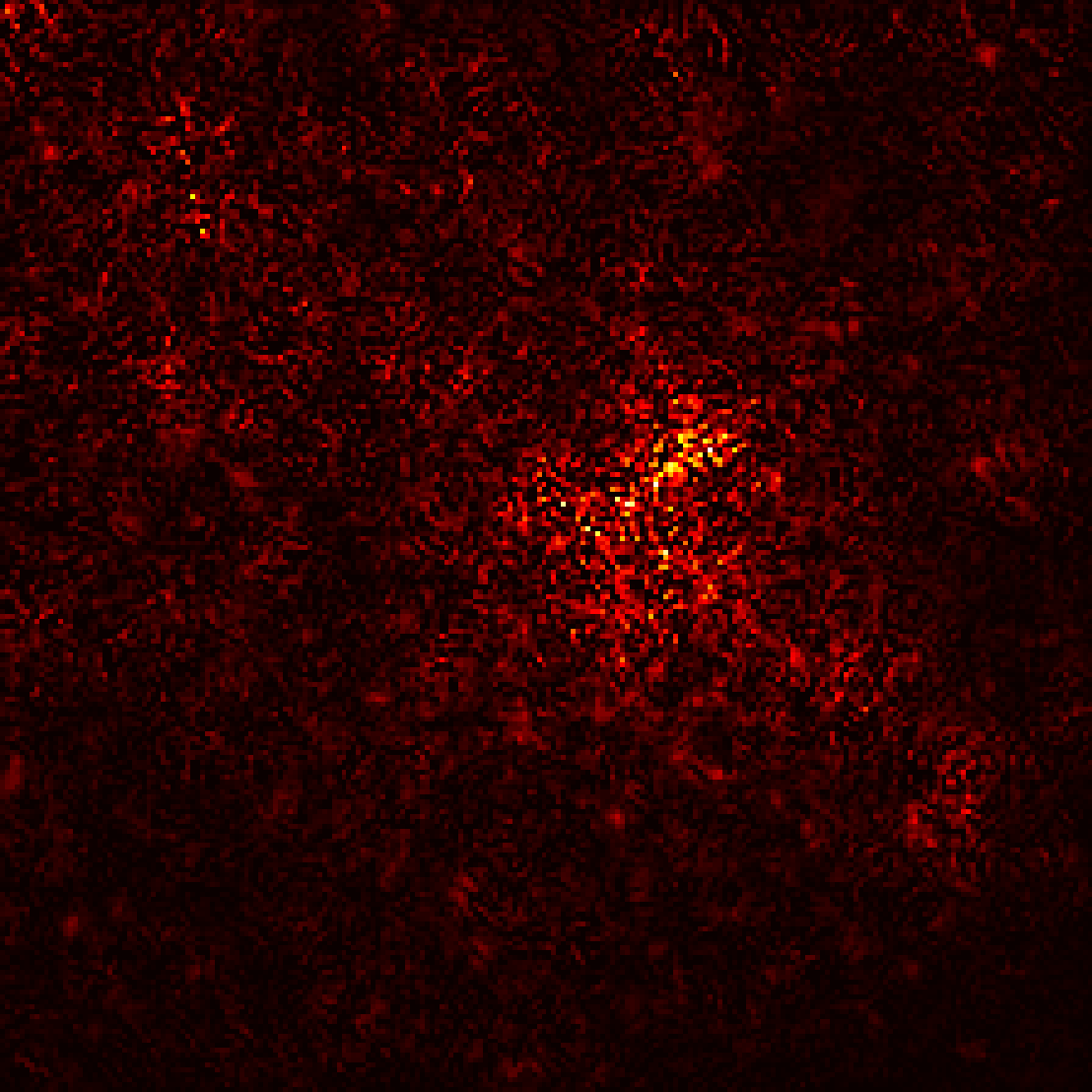} & 
  \includegraphics[scale=\scale]{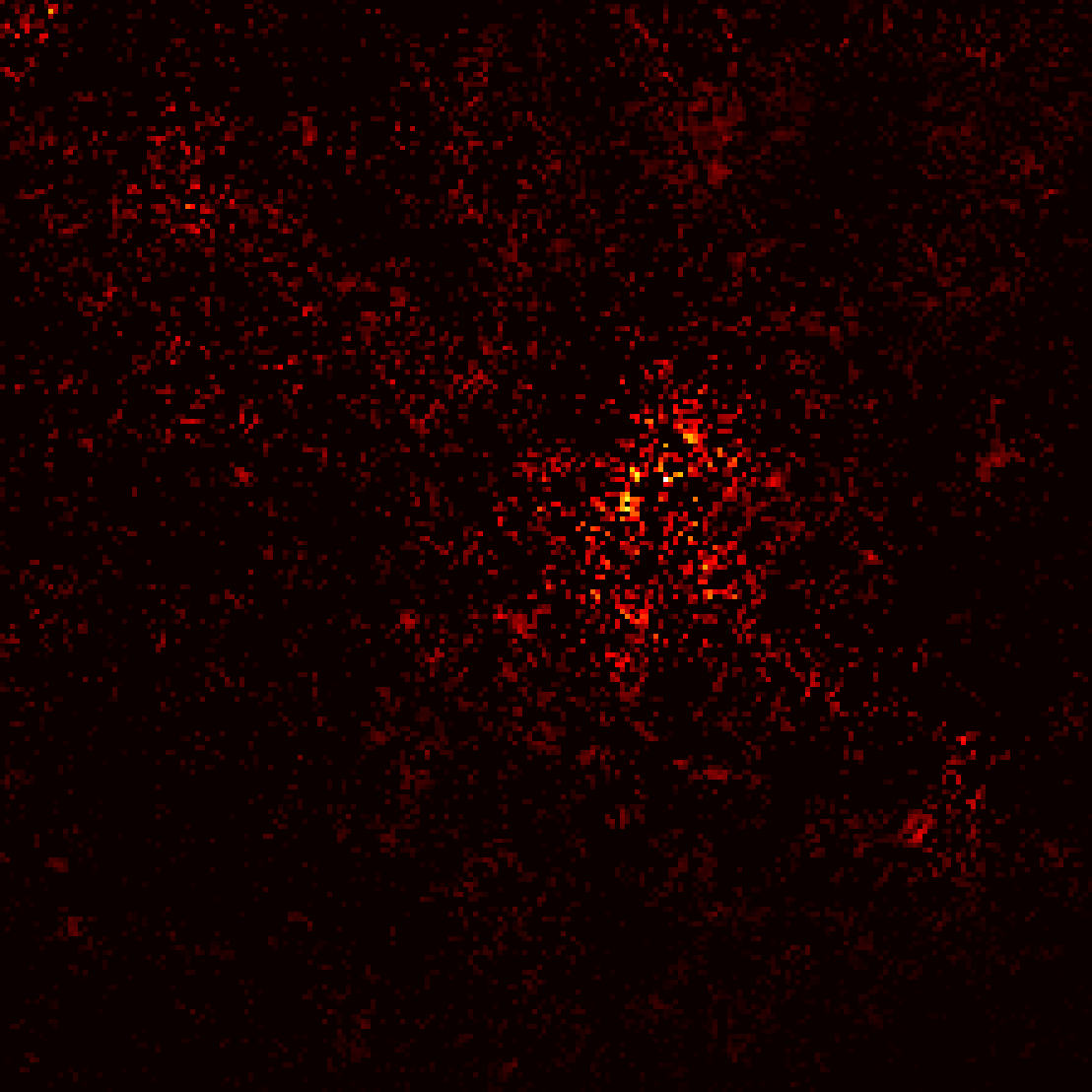} & 
  \includegraphics[scale=\scale]{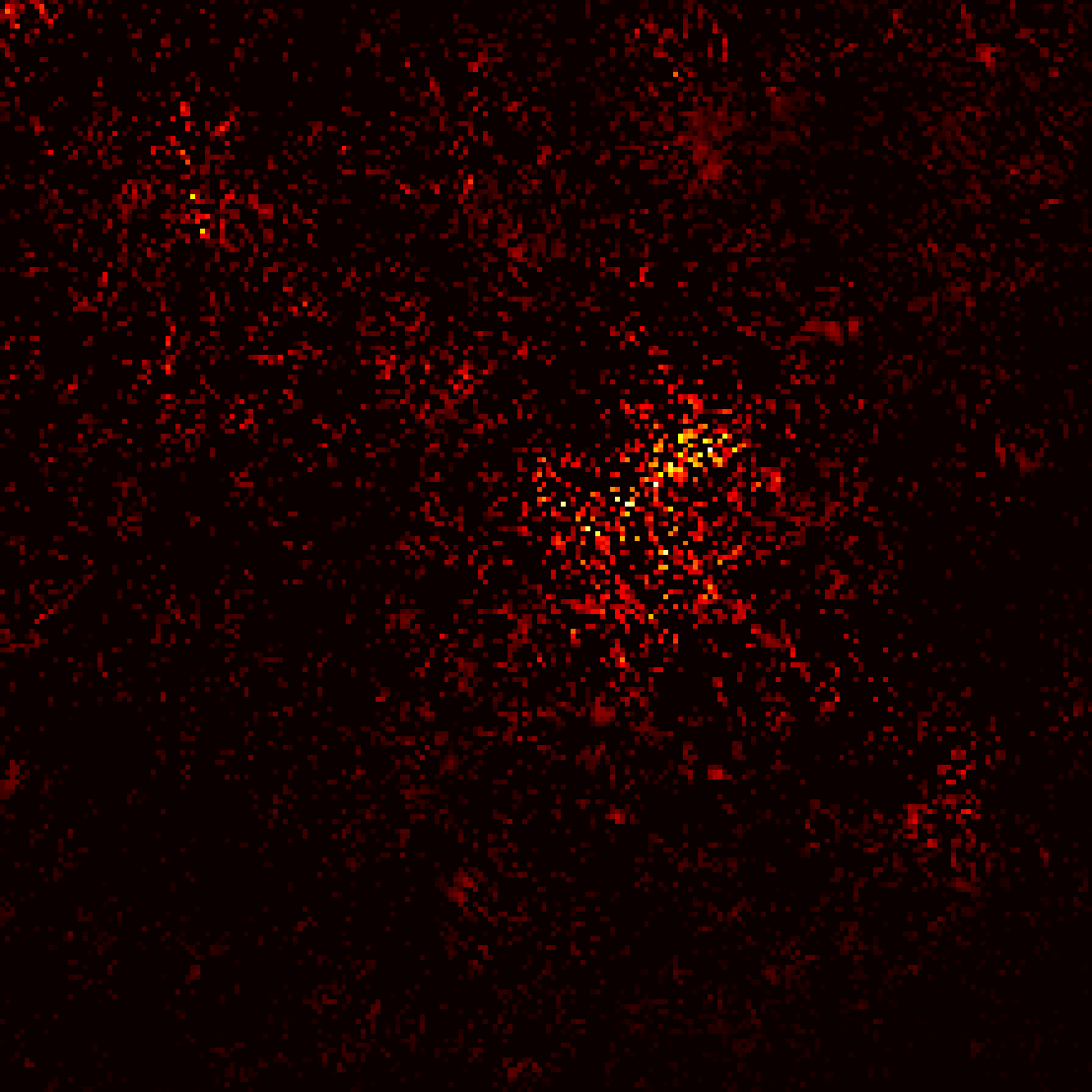} \\
  
  \includegraphics[scale=\scale]{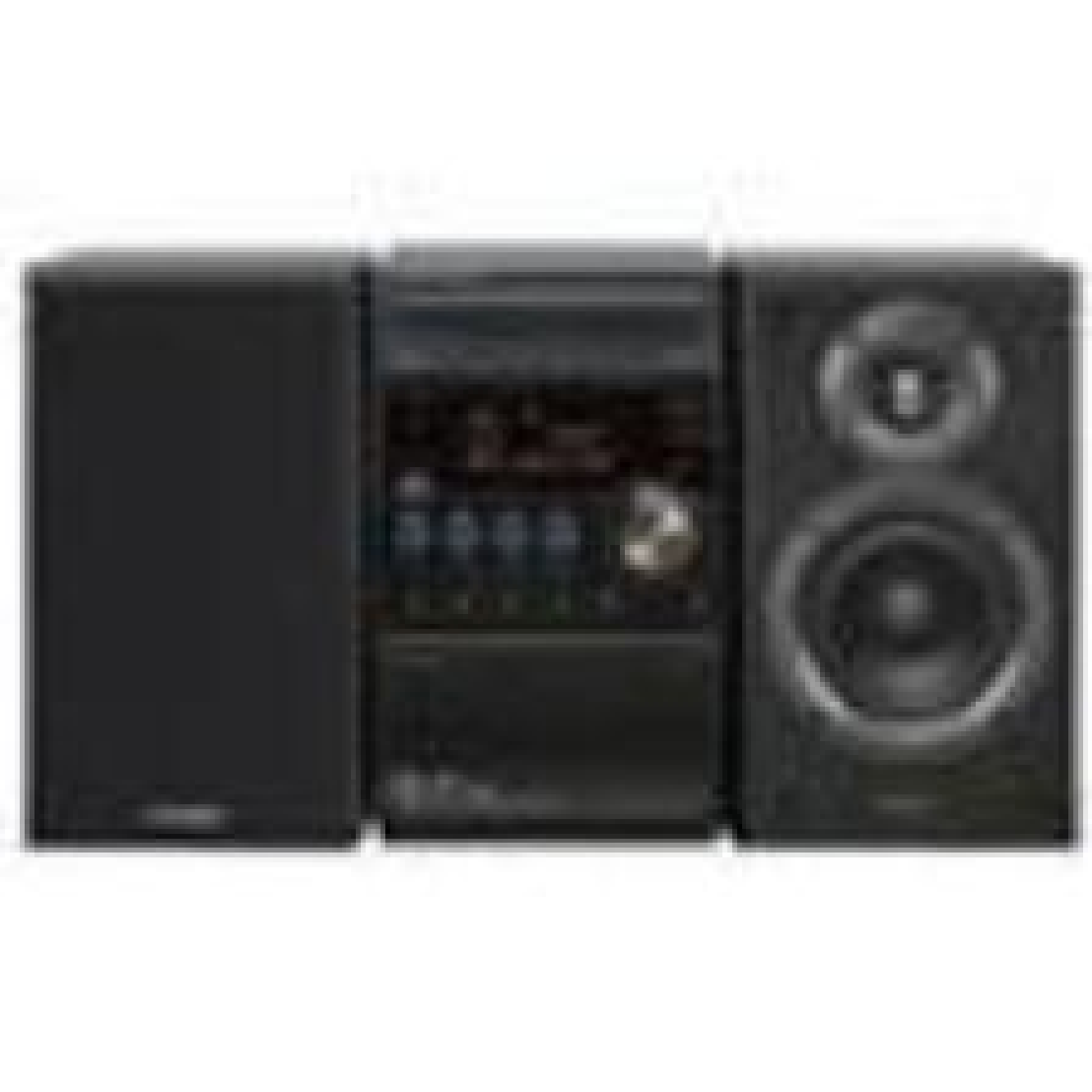} &
  \includegraphics[scale=\scale]{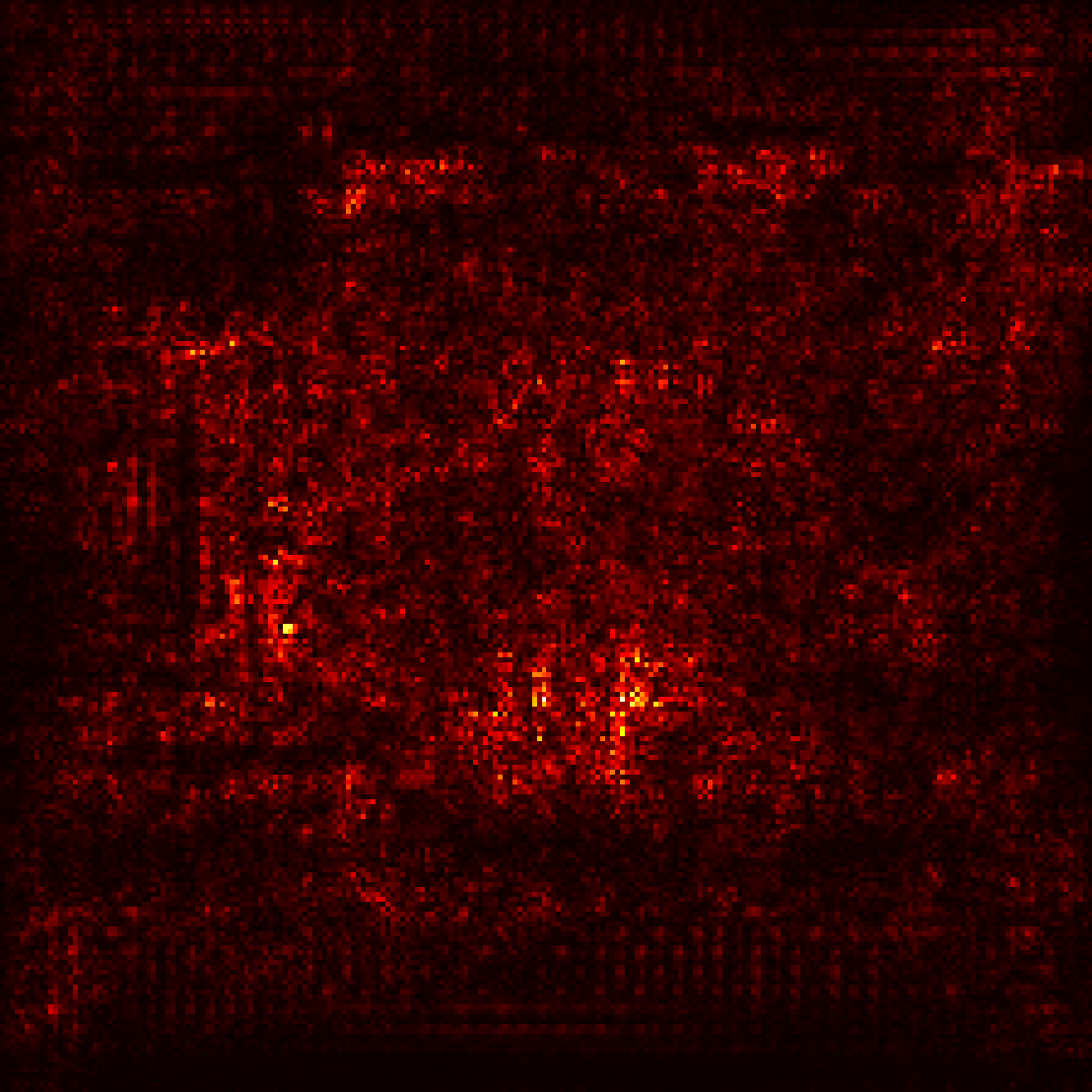} & 
  \includegraphics[scale=\scale]{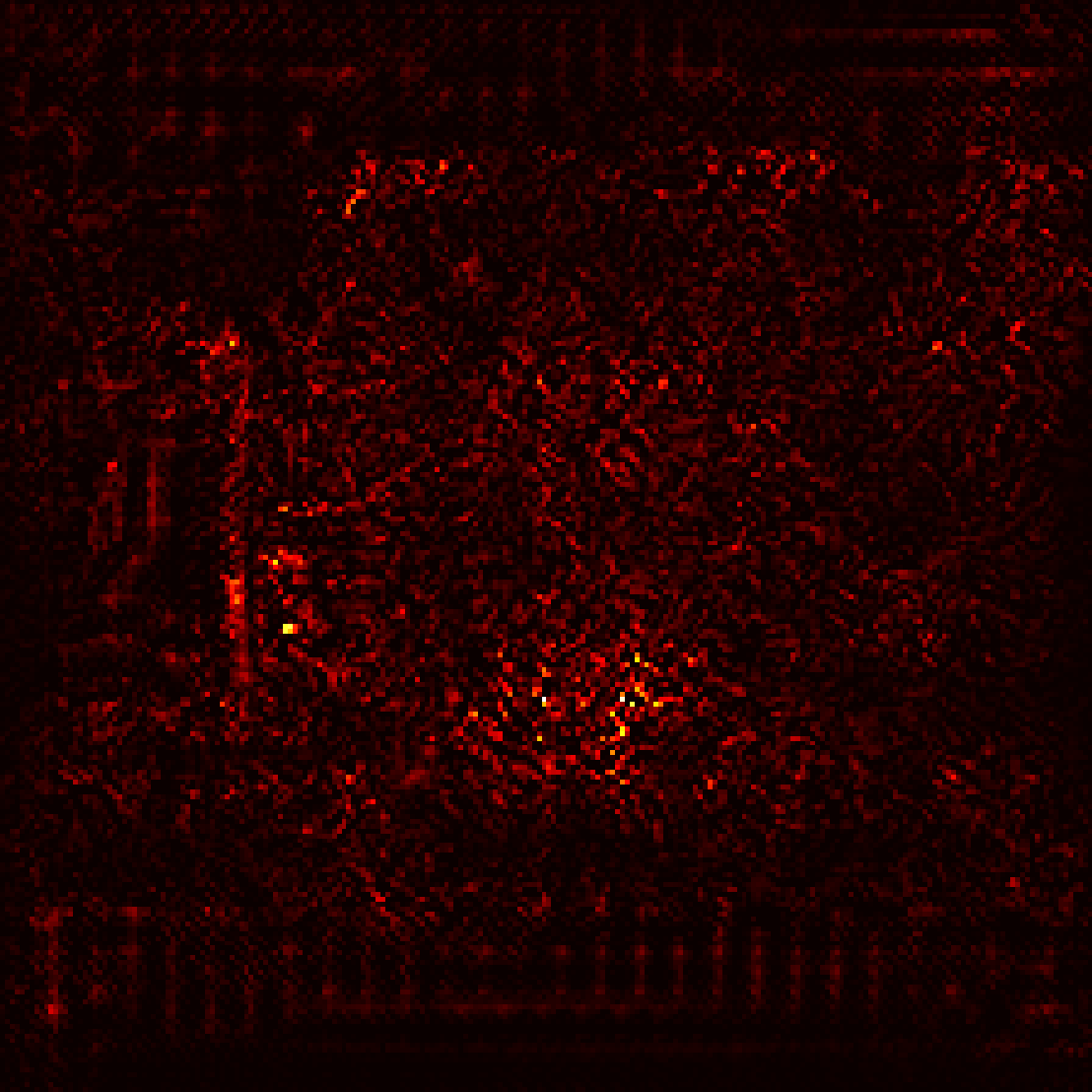} & 
  \includegraphics[scale=\scale]{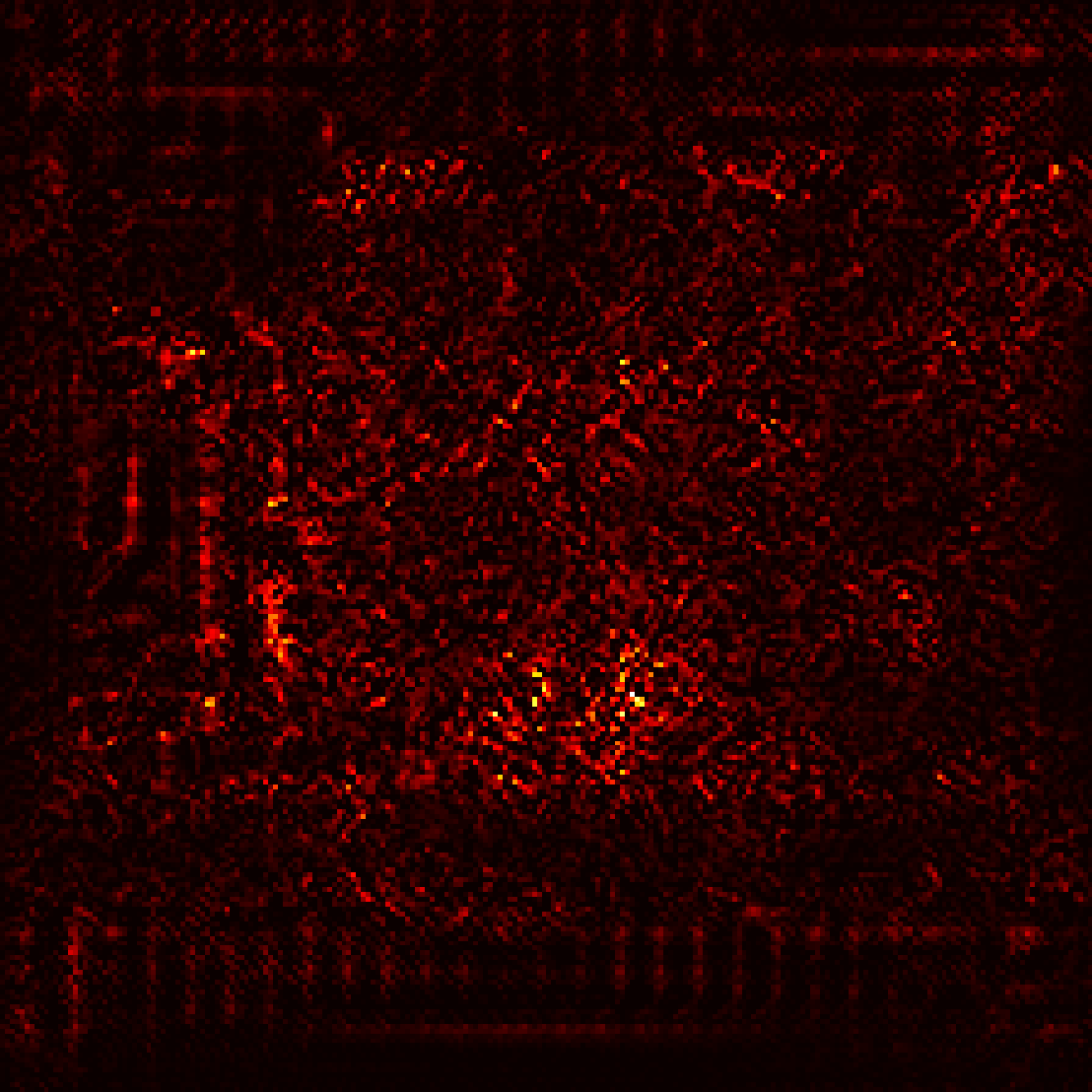} & 
  \includegraphics[scale=\scale]{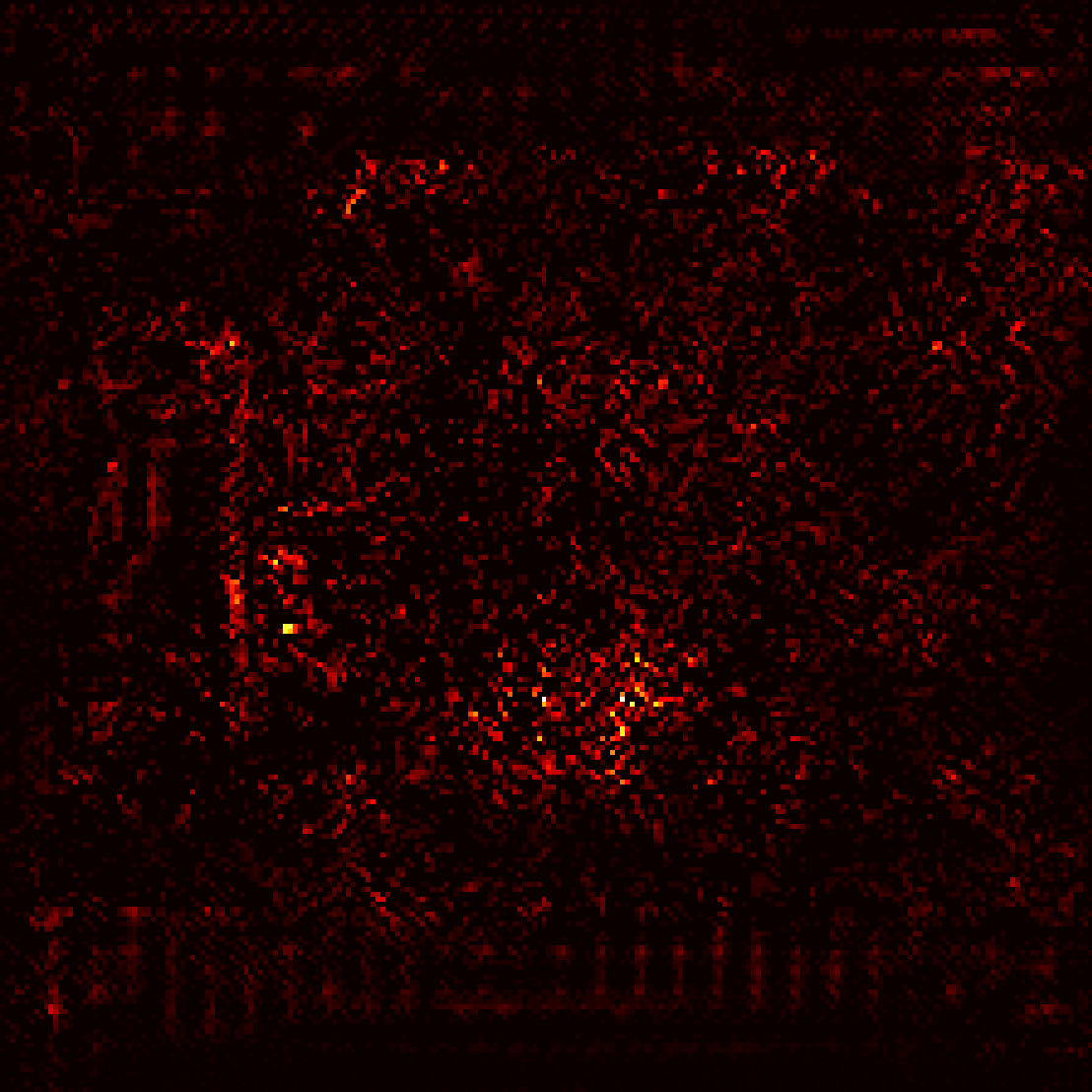} & 
  \includegraphics[scale=\scale]{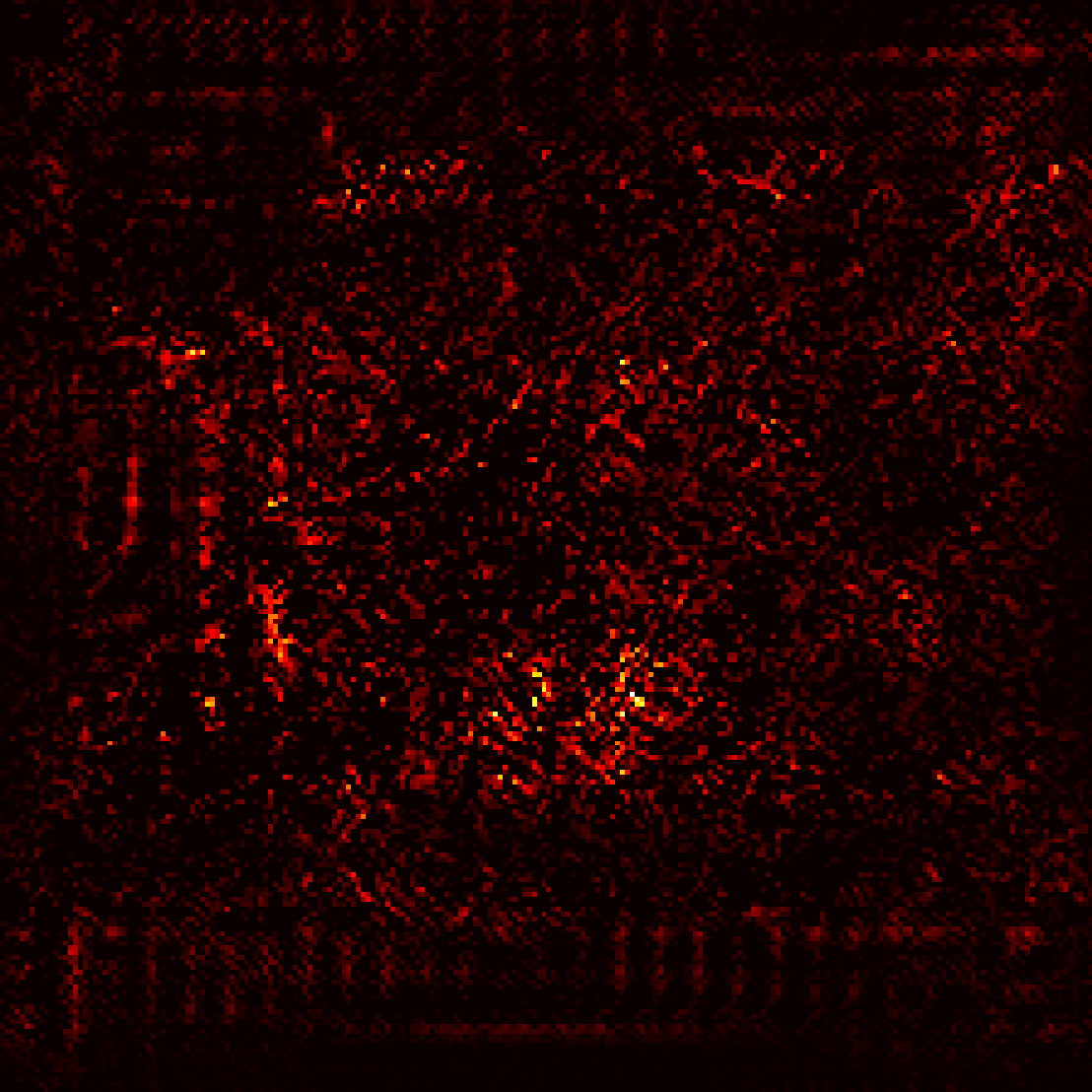} \\
  \end{tabular}
  \caption{Comparison saliency maps ResNet-18 Imagenette.}
  \label{fig: comparison saliency maps resnet18 imagenette}
\end{figure}

Following the same approach found in the literature of saliency maps \cite{Simonyan2014, Springenberg2015, Smilkov2017, Sundarajan2017}, first a visual inspection is carried out to compare the proposed visualizations with the standard saliency map. It is common practice to show only a few examples of the visualizations to perform the comparisons. However, none of the aforementioned articles explain how these examples are selected. Therefore, it could be the case that a particular visualization is better for a certain class or example. To prevent this problem, in this paper a correctly classified example has been randomly selected from the test set. To enhance readability, only two examples are shown in this section (Figures~\ref{fig: comparison saliency maps cnn cifar10} and \ref{fig: comparison saliency maps resnet18 imagenette}). Refer to \ref{sec:signed saliency map examples} to check the rest of the images.

It is noticeable that the proposed techniques reduce the noise and produce sharper visualization than the original saliency map. The shape of the objects that the neural network is classifying is more defined and exhibits a higher level of detail. Notwithstanding, even though this evaluation is typical in the literature, it does not prove them better. It could be the case that a noisier visualization is more faithful to what the neural network has learned to focus on. This is why a quantitative evaluation is required.

\subsection{Quantitative evaluation}
There have been some efforts in the literature to formulate metrics that measure the effectiveness of local interpretability techniques. While Ancona~et~al.~\cite{Ancona2018} develop on the desirable properties of interpretability methods,~\cite{Petsiuk2018} and~\cite{hookerBenchmarkInterpretabilityMethods2019} actually propose a metric to compare techniques. Specifically,~\cite{Petsiuk2018} suggests using a metric called deletion that removes pixels in descending order of importance ---according to the technique under evaluation--- and recomputes the probability of the correct output for each fraction of deleted pixels. Deleted pixels are either replaced with a constant value (e.g., black or gray) or random noise. Hooker~et~al.~\cite{hookerBenchmarkInterpretabilityMethods2019} claim that it is necessary to retrain the model after deleting pixels to maintain the same distribution in the training and the test sets. However, retraining affects the network's weights and the metric no longer provides a good estimate of how the original model behaves if some pixels are occluded.

\begin{figure}[p]
    \centering
    \subfloat[CIFAR-10 CNN]{\includegraphics[width=0.475\textwidth]{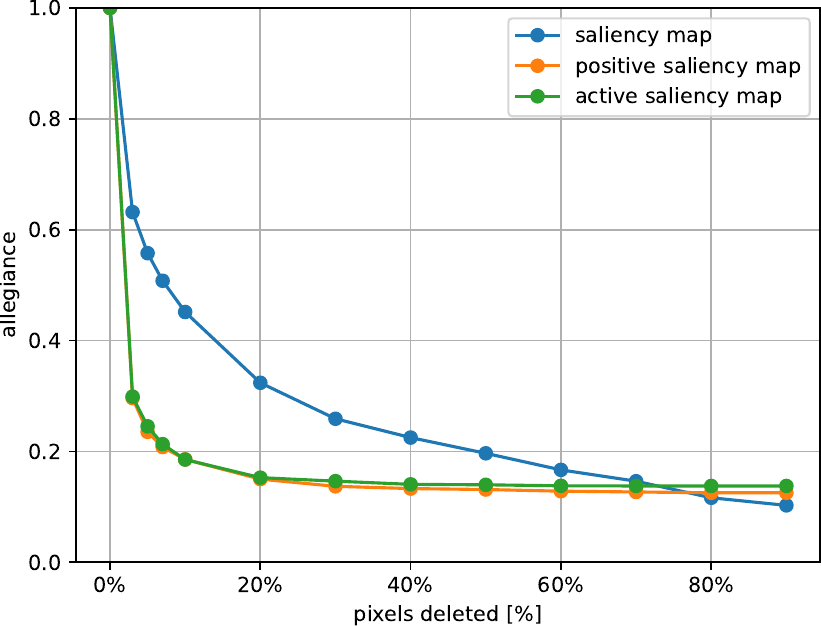}} \hfill
    \subfloat[CIFAR-10 ResNet-18]{\includegraphics[width=0.475\textwidth]{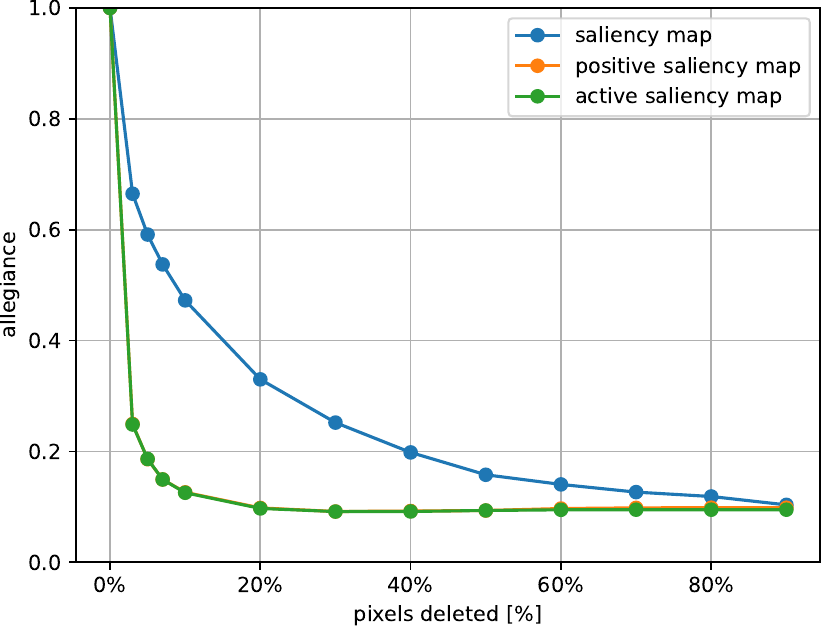}} \\
    \subfloat[Imagenette CNN]{\includegraphics[width=0.475\textwidth]{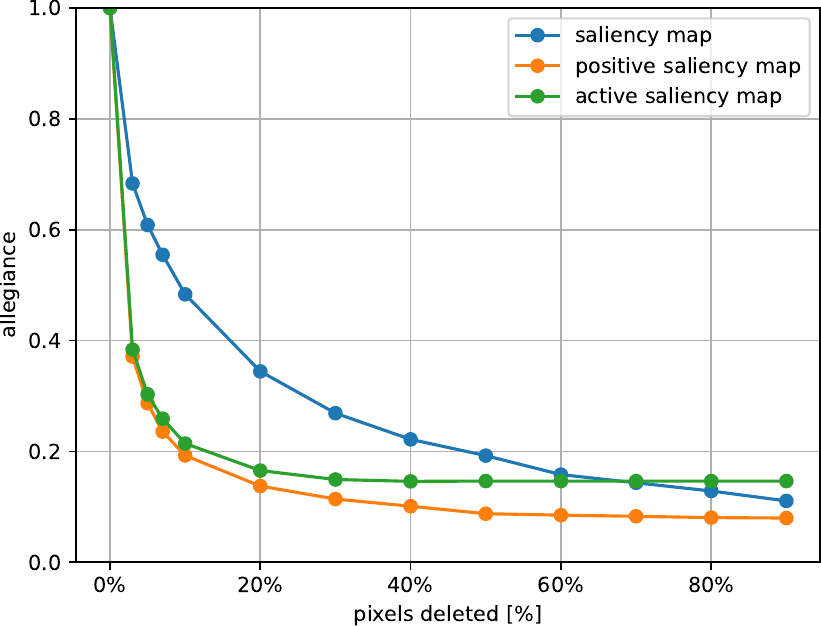}} \hfill
    \subfloat[Imagenette ResNet-18]{\includegraphics[width=0.475\textwidth]{visualizations/graphs/cifar10/resnet18_False/val/0.pdf}} \\
    \subfloat[Imagenette ResNet-18 pre-trained]{\includegraphics[width=0.475\textwidth]{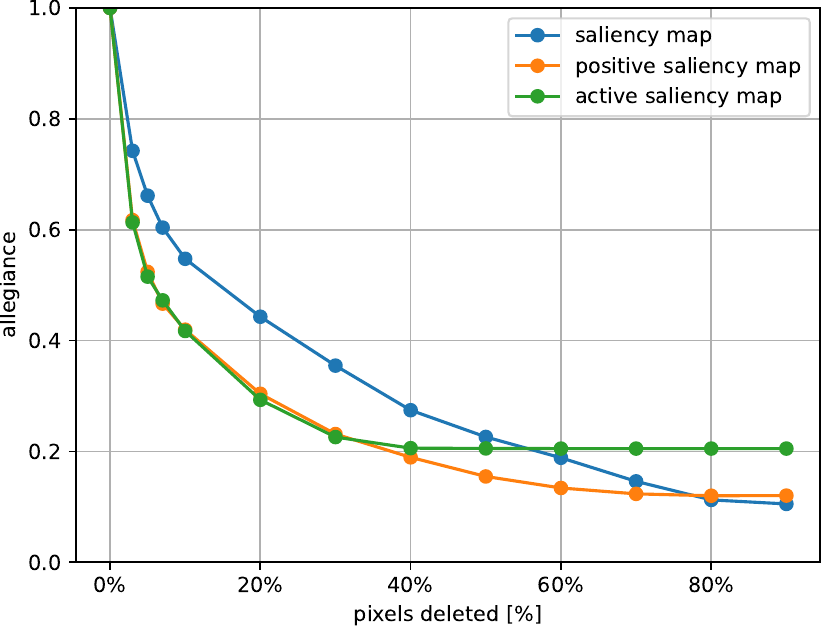}} \hfill
    \subfloat[Imagenette ConvNeXt pre-trained]{\includegraphics[width=0.475\textwidth]{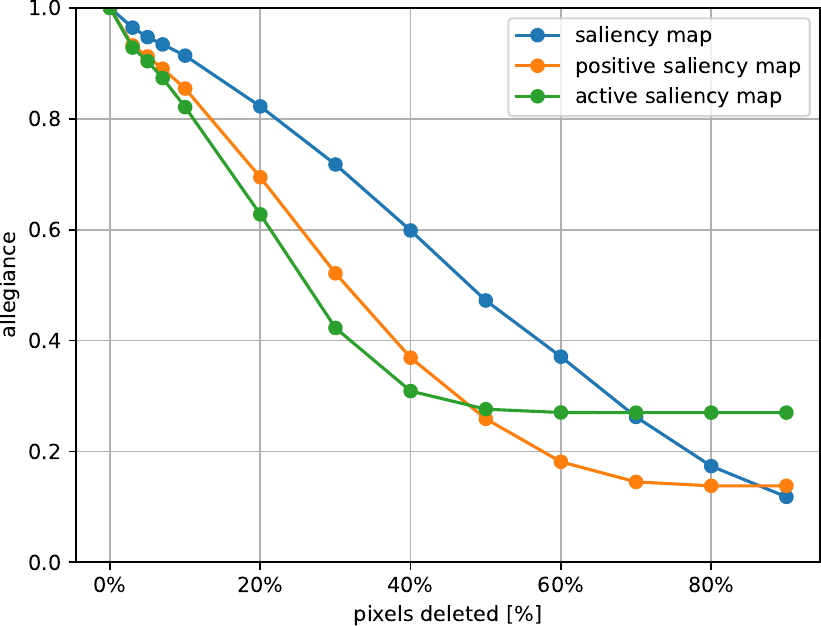}} \\
    \caption{Black-Deletion Benchmark.}
    \label{fig: black-deletion benchmark}
\end{figure}

\begin{figure}[p]
    \centering
    \subfloat[CIFAR-10 CNN]{\includegraphics[width=0.475\textwidth]{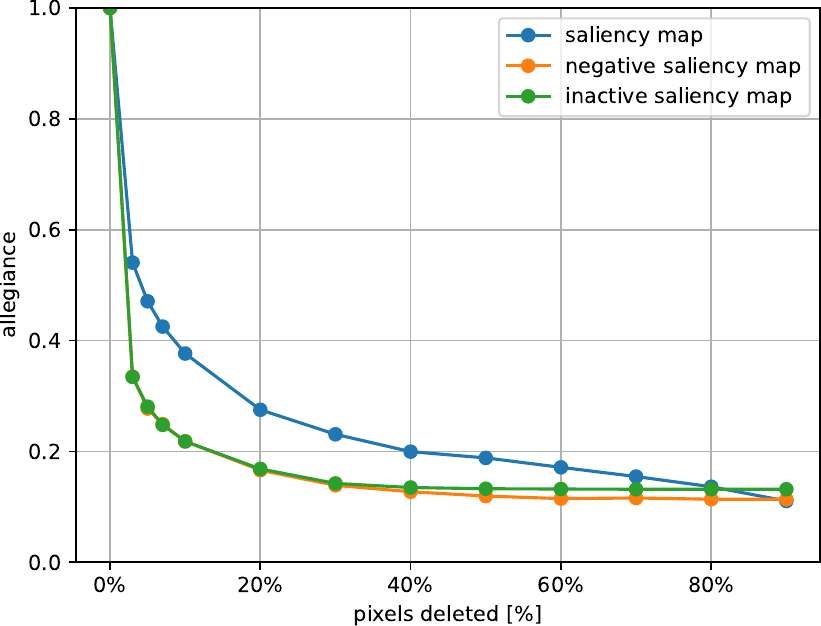}} \hfill
    \subfloat[CIFAR-10 ResNet-18]{\includegraphics[width=0.475\textwidth]{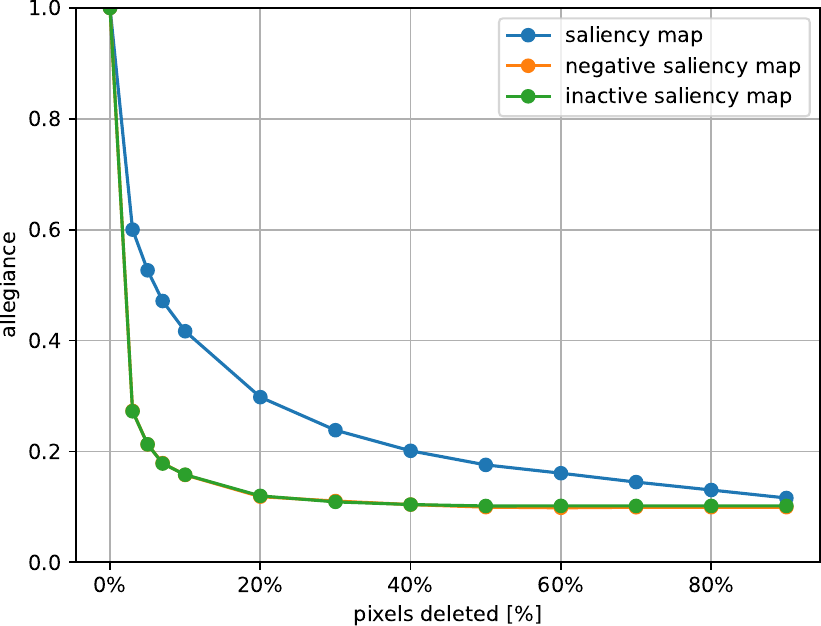}} \\
    \subfloat[Imagenette CNN]{\includegraphics[width=0.475\textwidth]{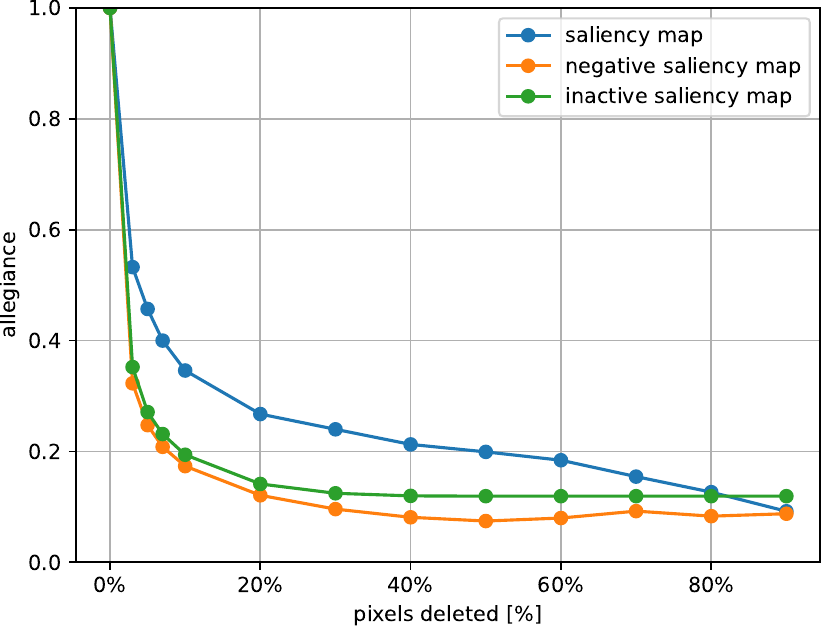}} \hfill
    \subfloat[Imagenette ResNet-18]{\includegraphics[width=0.475\textwidth]{visualizations/graphs/cifar10/resnet18_False/val/1.pdf}} \\
    \subfloat[Imagenette ResNet-18 pre-trained]{\includegraphics[width=0.475\textwidth]{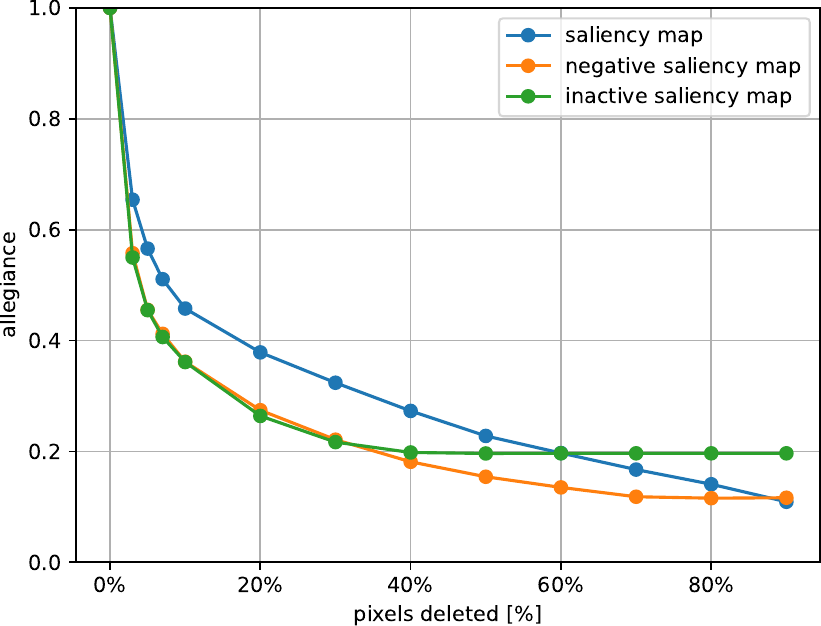}} \hfill
    \subfloat[Imagenette ConvNeXt pre-trained]{\includegraphics[width=0.475\textwidth]{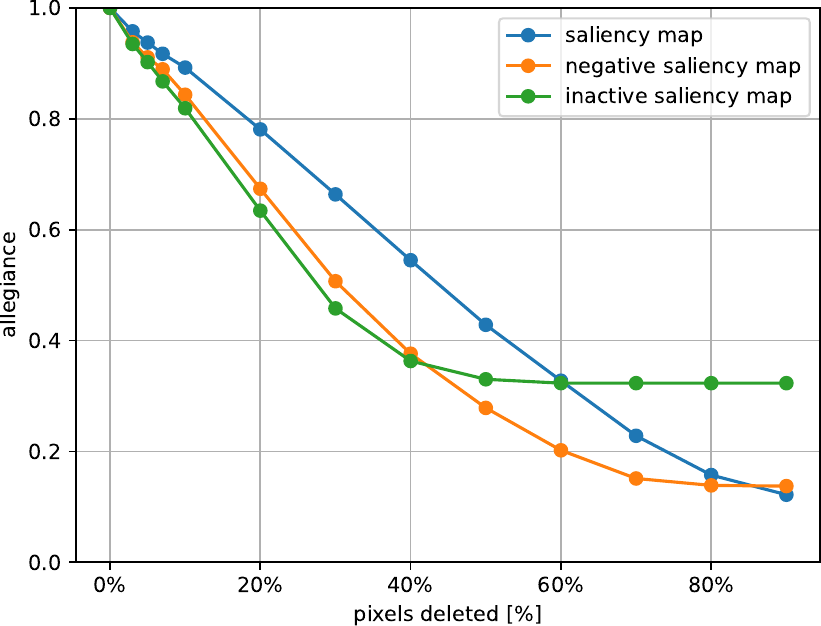}} \\
    \caption{White-Deletion Benchmark.}
    \label{fig: white-deletion benchmark}
\end{figure}

The main drawback of the deletion metric is that the value of a pixel cannot be actually deleted. No matter what value pixels are replaced with, they will still affect the internal computations of the network. The replacement value introduces unknown biases unless pixels are separated in two different sets: the ones that should be brighter to improve the classification score of the original predictions (i.e., those identified by positive or active saliency maps) and the ones that should be darker (i.e., pixels in negative or inactive saliency maps). Thanks to this distinction, the meaning of replacing a pixel with white (white-deletion) or black (black-deletion) becomes instantly clear. For positive or active pixels, using white would tend to improve the classification score of the predicted class, whereas zeroing them out should severely harm the original classification. The opposite is expected to happen with negative and inactive pixels.

Both black- (\autoref{fig: black-deletion benchmark}) and white-deletion (\autoref{fig: white-deletion benchmark}) measure the change in the predicted classes with respect to the original classification, which we have decided to coin \emph{allegiance}. Using the test set ---the results for the training set can be found in \ref{sec:black-deletion and white-deletion for training set}---, pixels are removed in descending order of importance in blocks of~10\% as suggested in~\cite{hookerBenchmarkInterpretabilityMethods2019}, except for the initial interval, in which we deem it necessary to study the response in more detail. The behavior observed in the graphs corresponds to what we expected. The decrease in allegiance for black-deletion is greater for active and positive saliency maps than for the standard implementation. Likewise, the decrease in allegiance for white-deletion is greater for inactive and negative saliency maps. Apparently, this confirms the hypothesis that the pixels identified are more important to the network when the sign of the gradients is taken into account.

\begin{table}
  \centering
  \small
  \caption{AUC for black deletions in saliency maps.}
  \label{tab:auc for black deletions in saliency maps}
  \begin{tabular*}{\textwidth}{L{2cm} @{\extracolsep{\fill}} *{8}{c}}
    \toprule
    {} & \multicolumn{2}{c}{CIFAR-10} & \multicolumn{4}{c}{Imagenette} \\
    \cline{2-3} \cline{4-7}
    {} & CNN & ResNet-18 & CNN & ResNet-18 & ResNet-18 pre-trained & ConvNeXt\\  
    \midrule
    Original & 0.23 & 0.23 & 0.24 & 0.26 & 0.28 & 0.49 \\
    Positive & 0.14 & 0.11 & 0.12 & 0.14 & 0.21 & 0.37 \\
    Active   & 0.15 & 0.11 & 0.16 & 0.14 & 0.24 & 0.39 \\ 
    \bottomrule
  \end{tabular*}
\end{table}

\begin{table}
  \centering
  \small
  \caption{AUC for white deletions in saliency maps.}
  \label{tab:auc for white deletions in saliency maps}
  \begin{tabular*}{\textwidth}{L{2cm} @{\extracolsep{\fill}} *{8}{c}}
    \toprule
    {} & \multicolumn{2}{c}{CIFAR-10} & \multicolumn{4}{c}{Imagenette} \\
    \cline{2-3} \cline{4-7}
    {} & CNN & ResNet-18 & CNN & ResNet-18 & ResNet-18 pre-trained & ConvNeXt\\  
    \midrule
    Original & 0.21 & 0.22 & 0.21 & 0.23 & 0.26 & 0.46 \\
    Negative & 0.14 & 0.12 & 0.11 & 0.18 & 0.20 & 0.37 \\ 
    Inactive & 0.15 & 0.12 & 0.14 & 0.16 & 0.23 & 0.42 \\
    \bottomrule
  \end{tabular*}
\end{table}

It is important to note that that for active and inactive saliency maps the allegiance stops decreasing after around 50\% of the pixels have been deleted. The reason is that many pixels have a value of zero because their derivative with respect to the original predicted class is not the largest (for the active saliency map) or the smallest (for the inactive). Hence, after all the non-zero pixels from the active and inactive saliency maps have been deleted there is nothing else to remove. The same happens for the positive and negative saliency maps at approximately 80\%.

To provide concrete numbers, the area under the curve is shown in \autoref{tab:auc for black deletions in saliency maps} for black-deletion and in \autoref{tab:auc for white deletions in saliency maps} for white-deletion. The results support the hypothesis that the proposed saliency maps better identify those pixels that, when made brighter or darker as appropriate, increase the confidence of the originally predicted class. Interestingly, although the improvement over the standard saliency map is clear, it is surprising how positive and negative saliency maps sometimes work better than active and inactive. It could still be due to the use of the extremes (i.e., either black or white) as replacement values, instead of slightly darker or brighter variants of the original pixel colors. Nevertheless, the results are still more interpretable than those provided by other metrics proposed in the literature because the effect of the alteration on the image is now known.

\section{Conclusion and future work}
\label{sec:conclusions and future research}
There is more information hidden in the gradients of a saliency map than is usually exploited, both in the sign of the individual pixels and in the gradients with respect to the incorrect classes. Separating pixels according to these dimensions could pave the way to improving the quality of the insights extracted, not only from saliency maps but also from other local interpretability techniques based on gradients.

Furthermore, instead of arbitrarily choosing black to occlude pixels as it is typically done, the proposed approach allows to better understand the effect of replacing pixels with black or white, which can positively or negatively contribute to the classification score depending on the gradient sign. Analyzing the faithfulness of the different variations of the saliency map from this point of view is left as future work.

\bibliographystyle{elsarticle-num} 
{\small\bibliography{references}}

\appendix

\section{Signed saliency map examples}
\label{sec:signed saliency map examples}

This section shows a comparison of the different saliency maps for an example of each class from each the dataset.

\begin{figure}[H]
  \centering
  \footnotesize
  \newcommand{\scale}{0.20}
  \setlength{\tabcolsep}{2pt}
  \begin{tabular}{cccccc}
  Image & Original & Positive & Negative & Active & Inactive \\
  
  \includegraphics[scale=\scale]{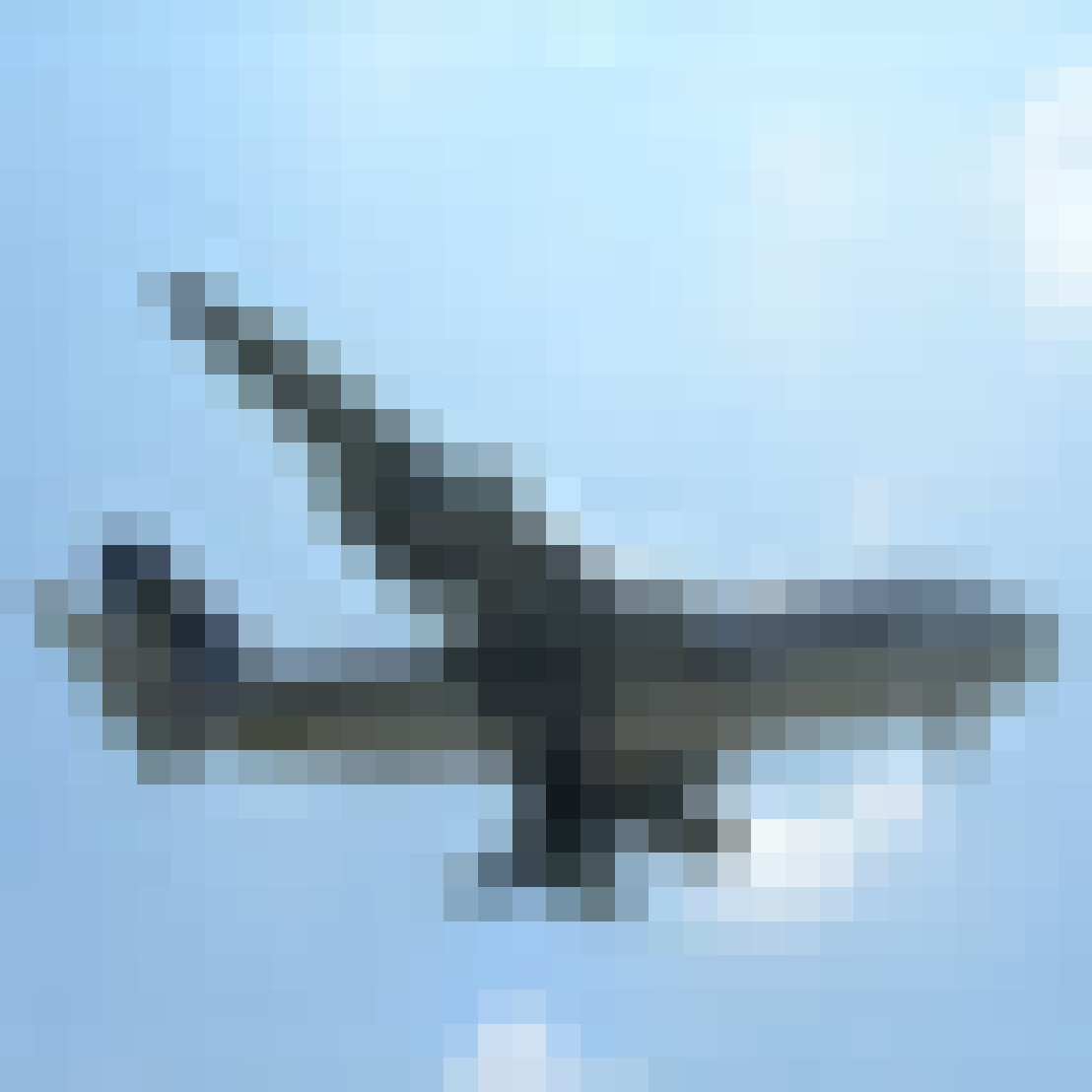} &
  \includegraphics[scale=\scale]{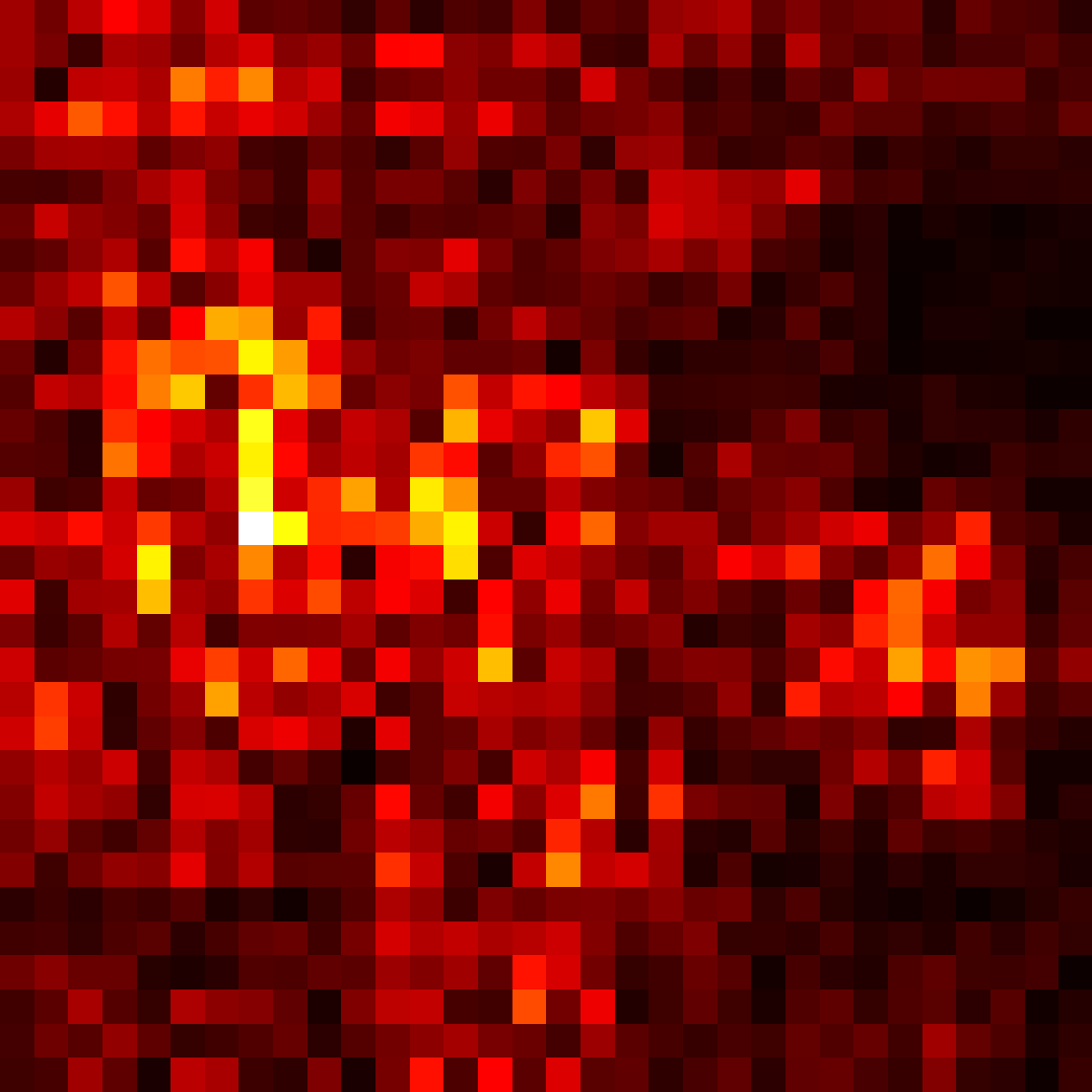} & 
  \includegraphics[scale=\scale]{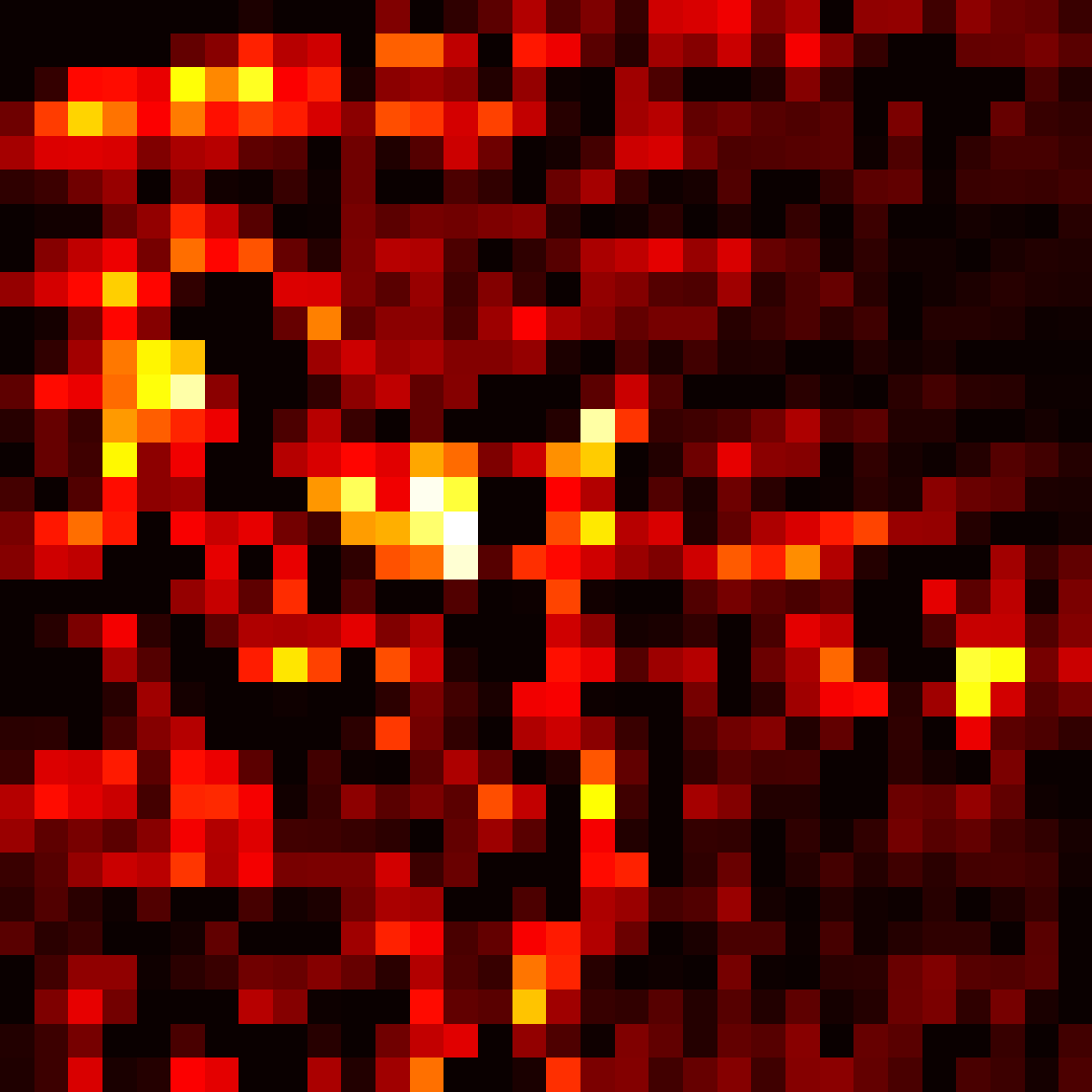} & 
  \includegraphics[scale=\scale]{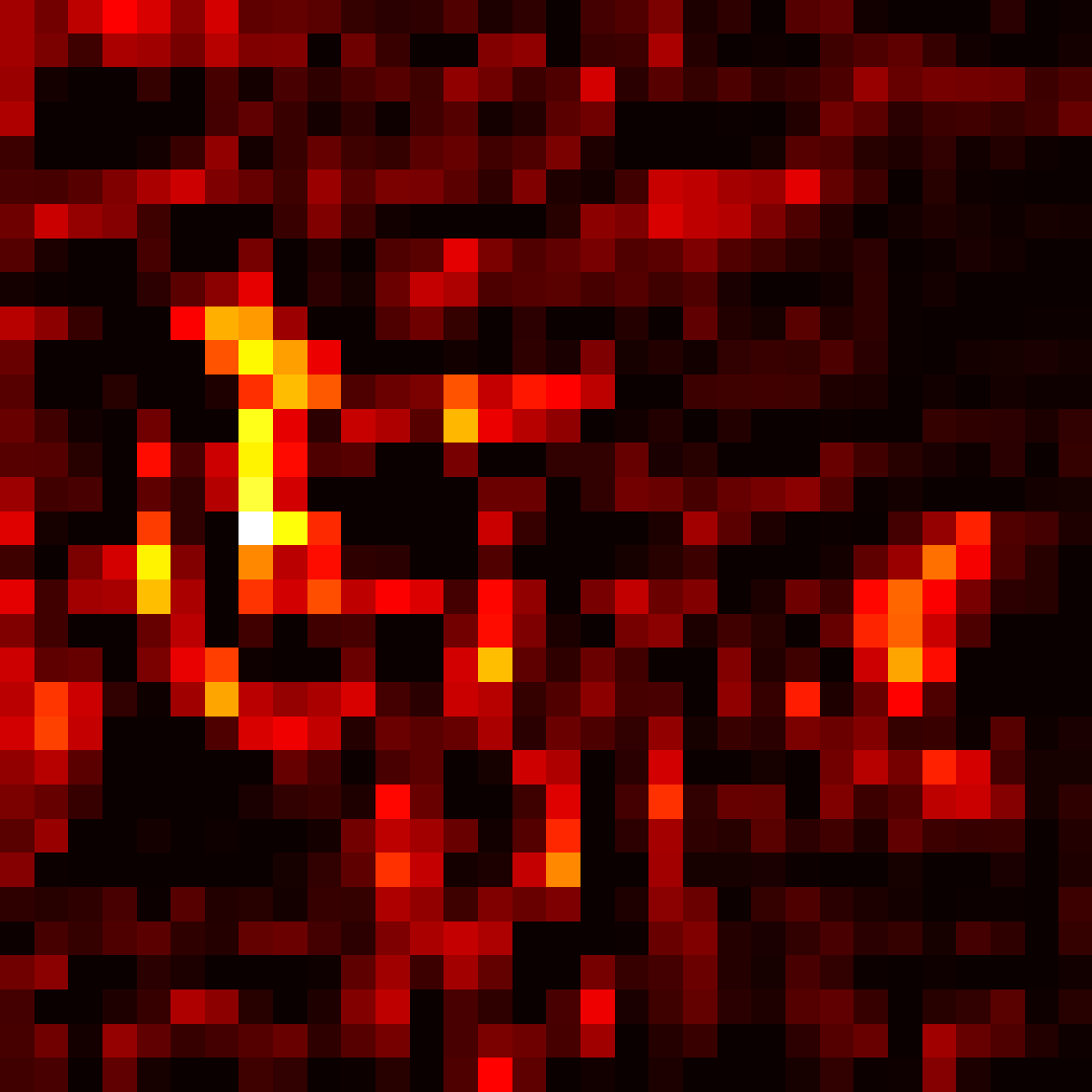} & 
  \includegraphics[scale=\scale]{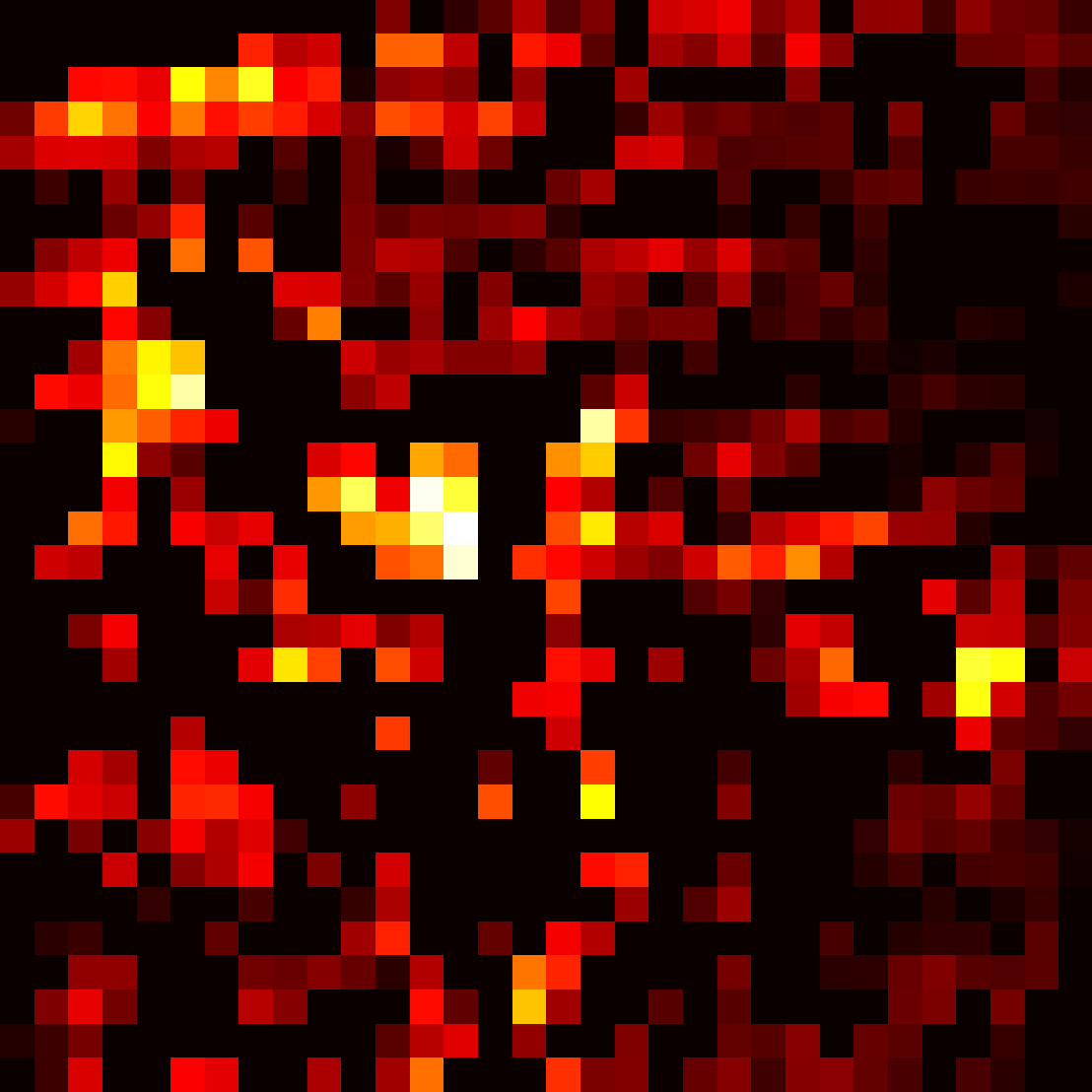} & 
  \includegraphics[scale=\scale]{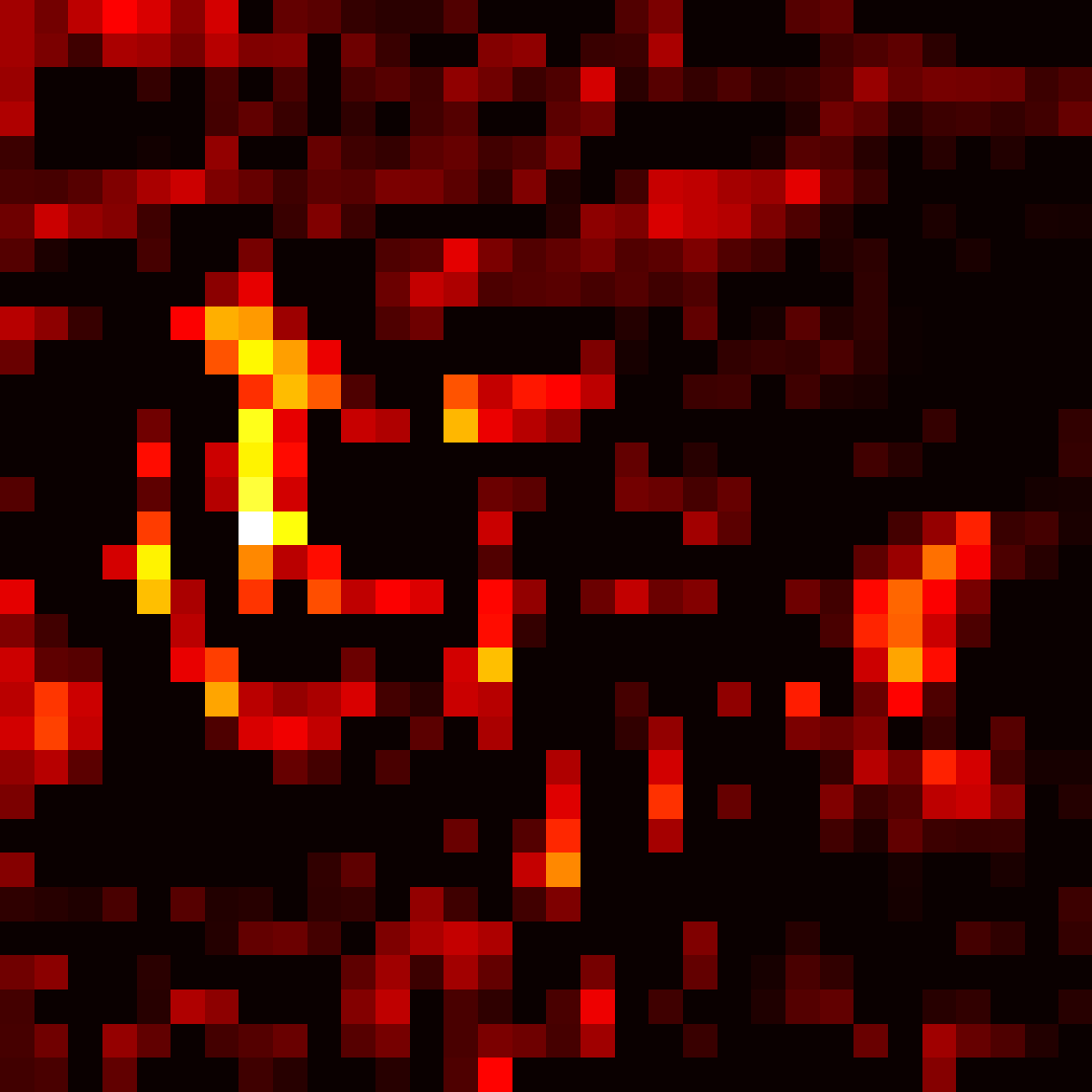} \\
  
  \includegraphics[scale=\scale]{visualizations/images/cifar10/examples/1.png} &
  \includegraphics[scale=\scale]{visualizations/images/cifar10/saliency_map/1.png} & 
  \includegraphics[scale=\scale]{visualizations/images/cifar10/positive_saliency_map/1.png} & 
  \includegraphics[scale=\scale]{visualizations/images/cifar10/negative_saliency_map/1.png} & 
  \includegraphics[scale=\scale]{visualizations/images/cifar10/active_saliency_map/1.png} & 
  \includegraphics[scale=\scale]{visualizations/images/cifar10/inactive_saliency_map/1.png} \\
  
  \includegraphics[scale=\scale]{visualizations/images/cifar10/examples/2.png} &
  \includegraphics[scale=\scale]{visualizations/images/cifar10/saliency_map/2.png} & 
  \includegraphics[scale=\scale]{visualizations/images/cifar10/positive_saliency_map/2.png} & 
  \includegraphics[scale=\scale]{visualizations/images/cifar10/negative_saliency_map/2.png} & 
  \includegraphics[scale=\scale]{visualizations/images/cifar10/active_saliency_map/2.png} & 
  \includegraphics[scale=\scale]{visualizations/images/cifar10/inactive_saliency_map/2.png} \\
  
  \includegraphics[scale=\scale]{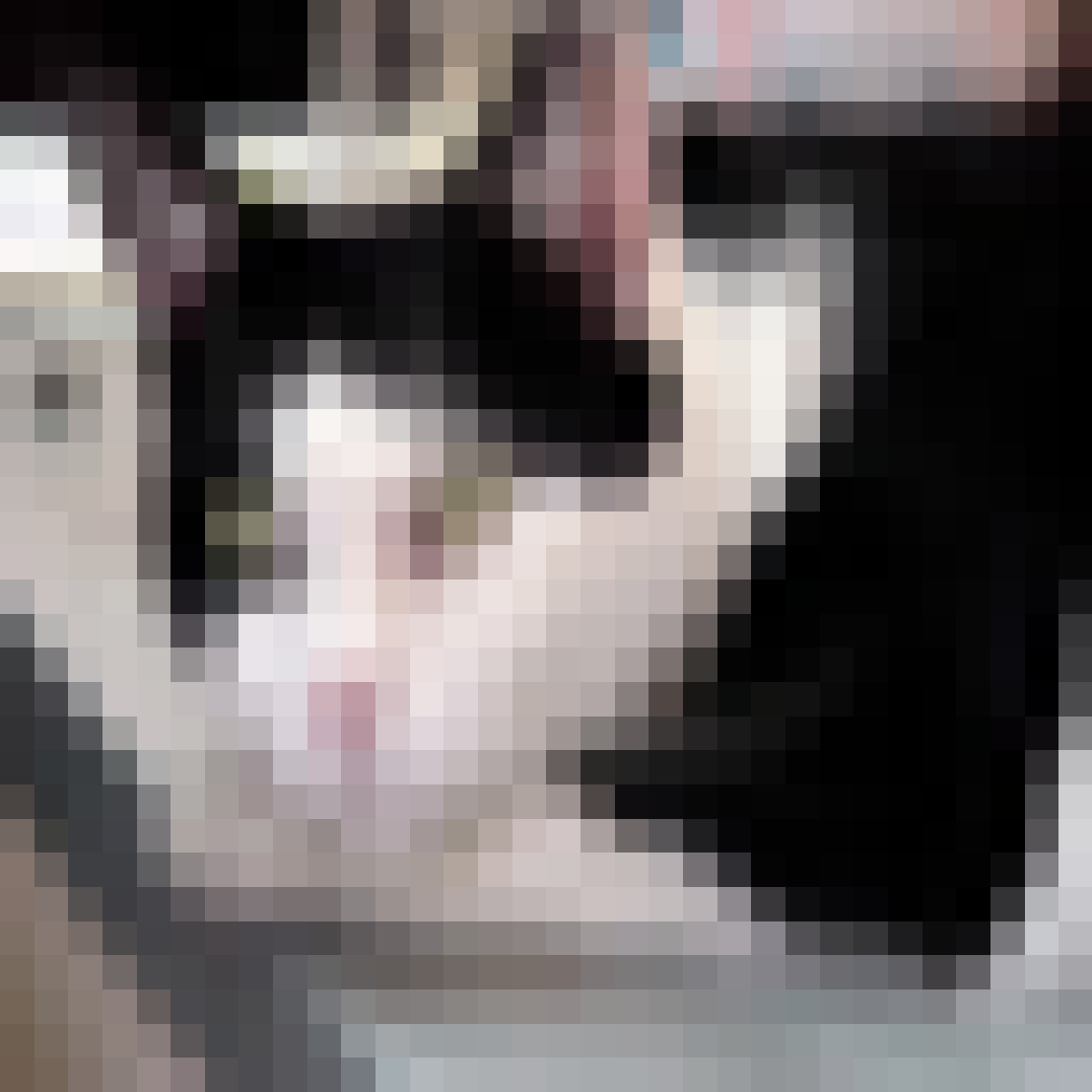} &
  \includegraphics[scale=\scale]{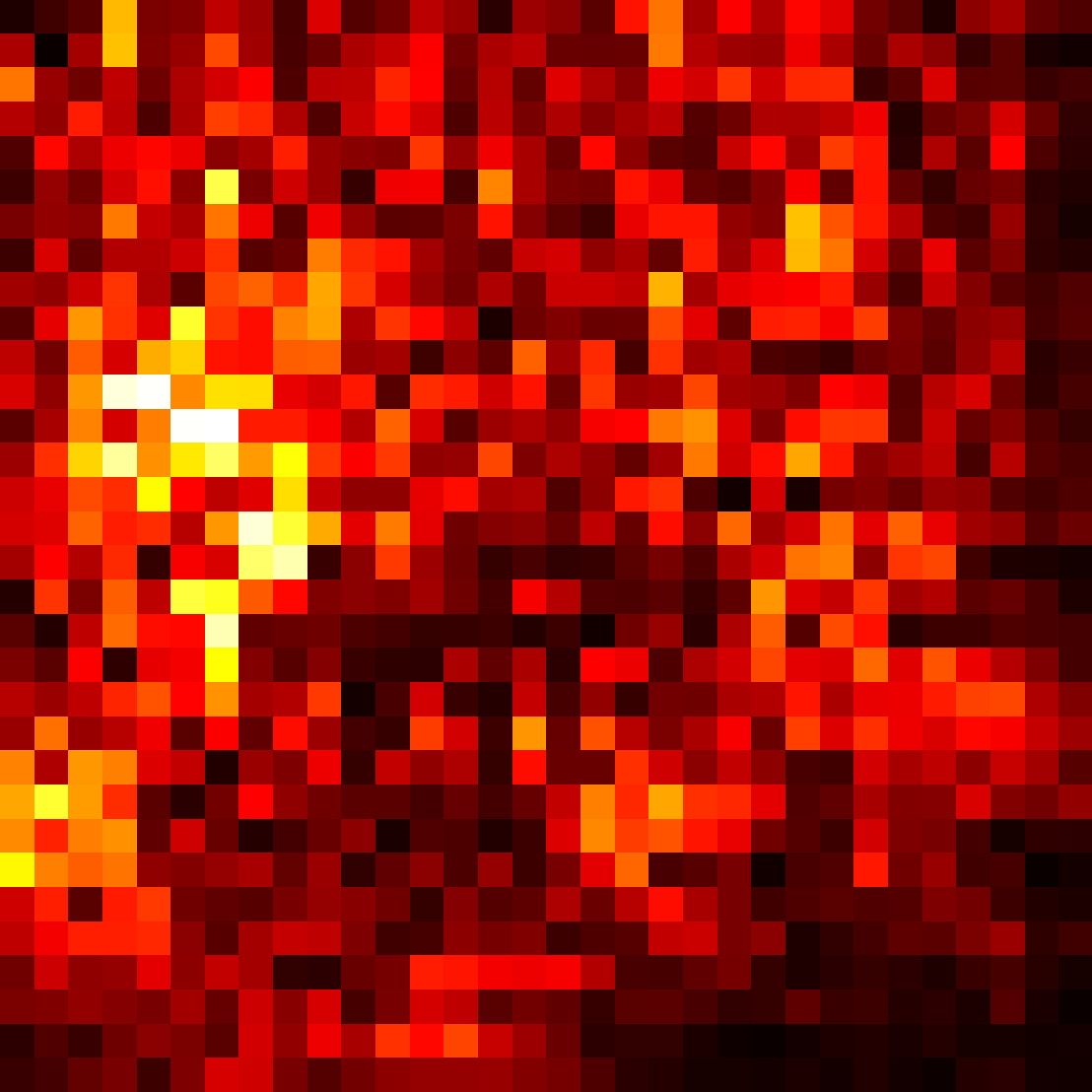} & 
  \includegraphics[scale=\scale]{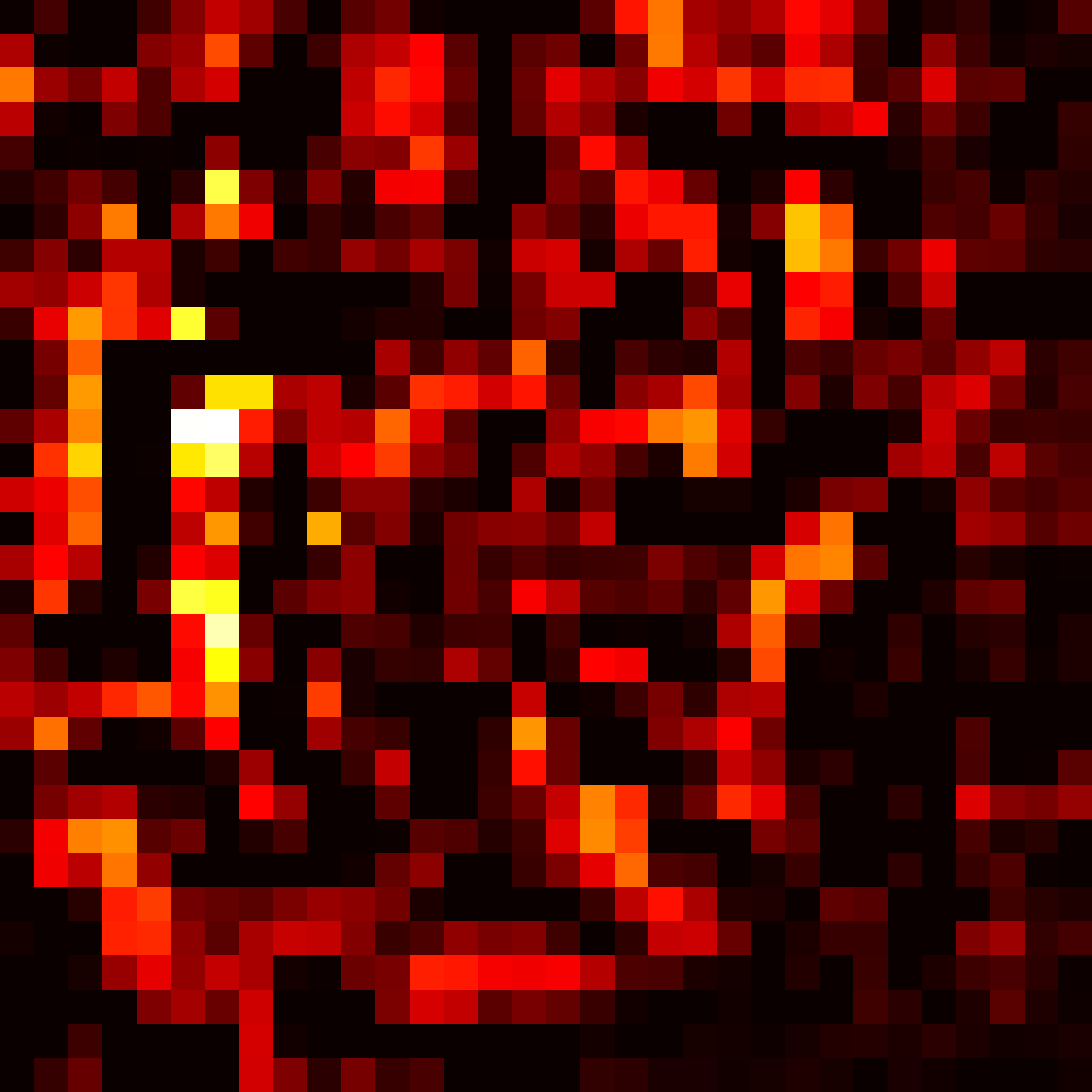} & 
  \includegraphics[scale=\scale]{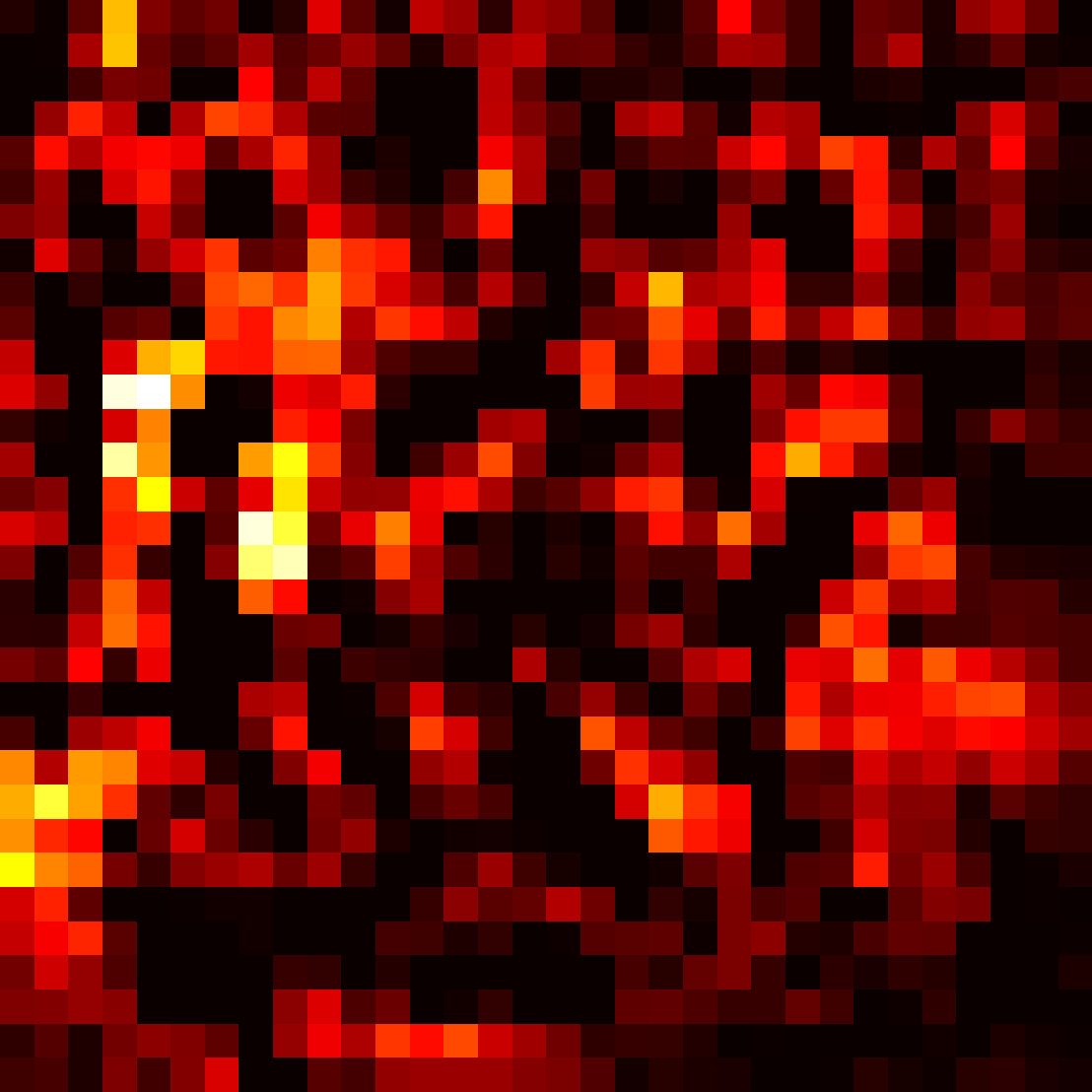} & 
  \includegraphics[scale=\scale]{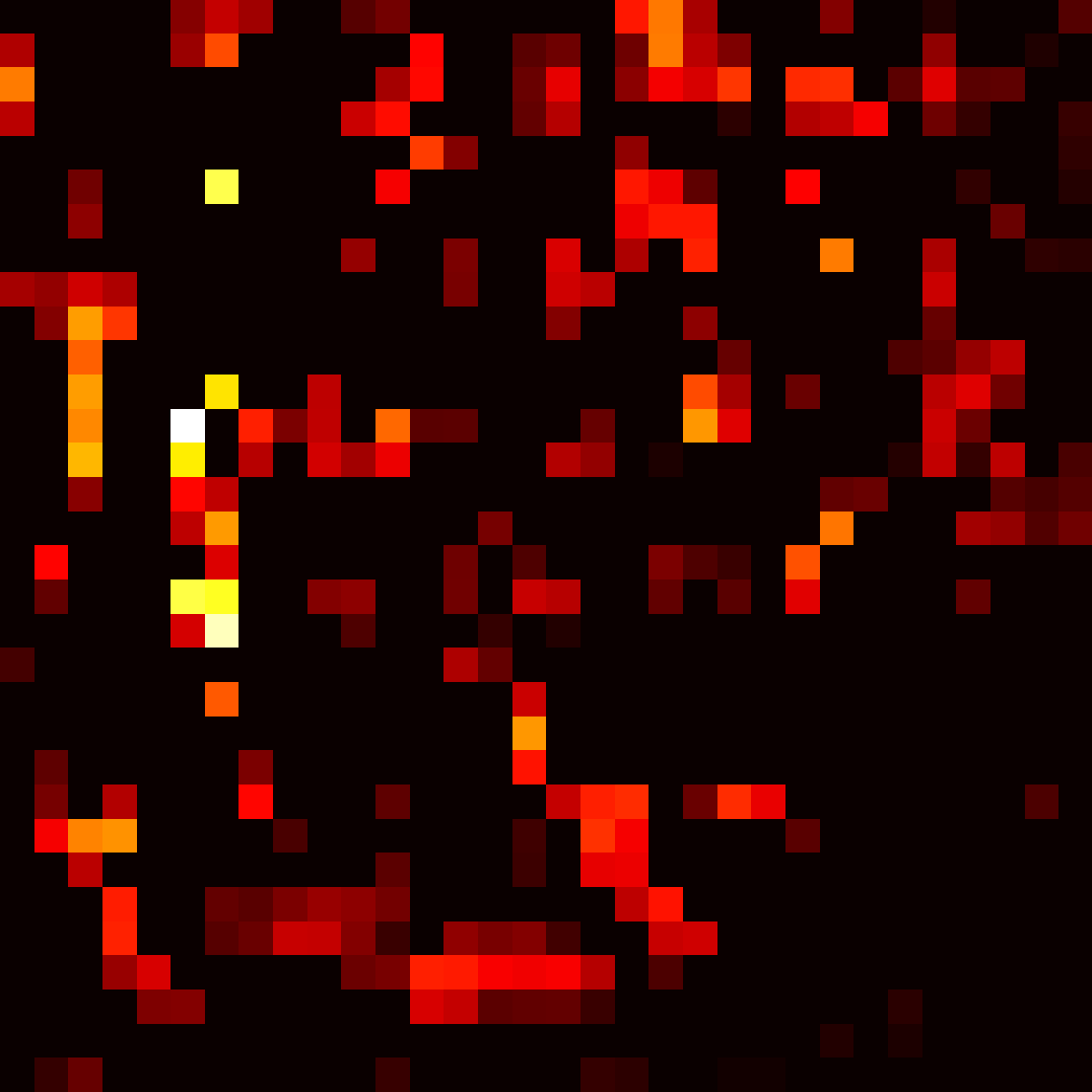} & 
  \includegraphics[scale=\scale]{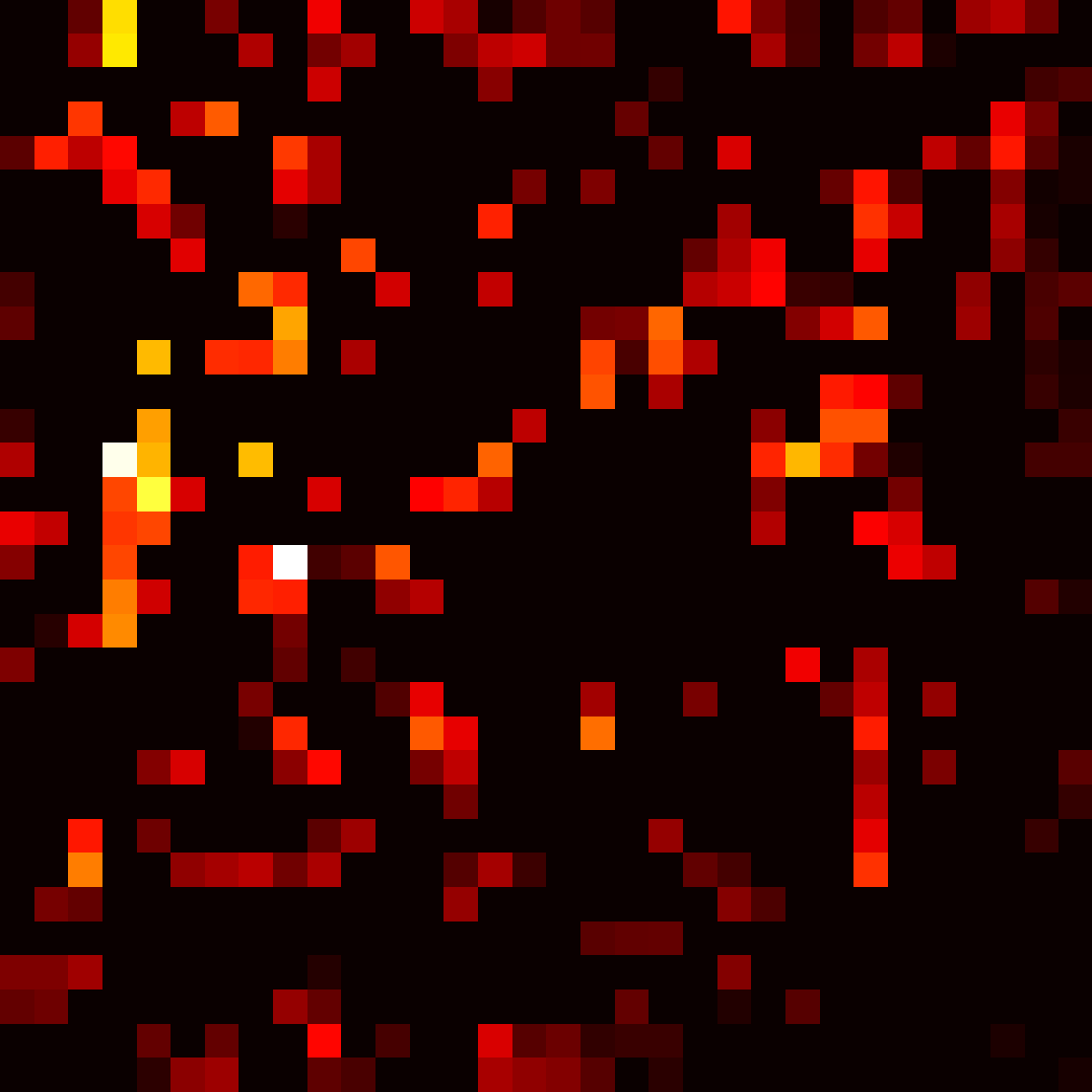} \\
  
  \includegraphics[scale=\scale]{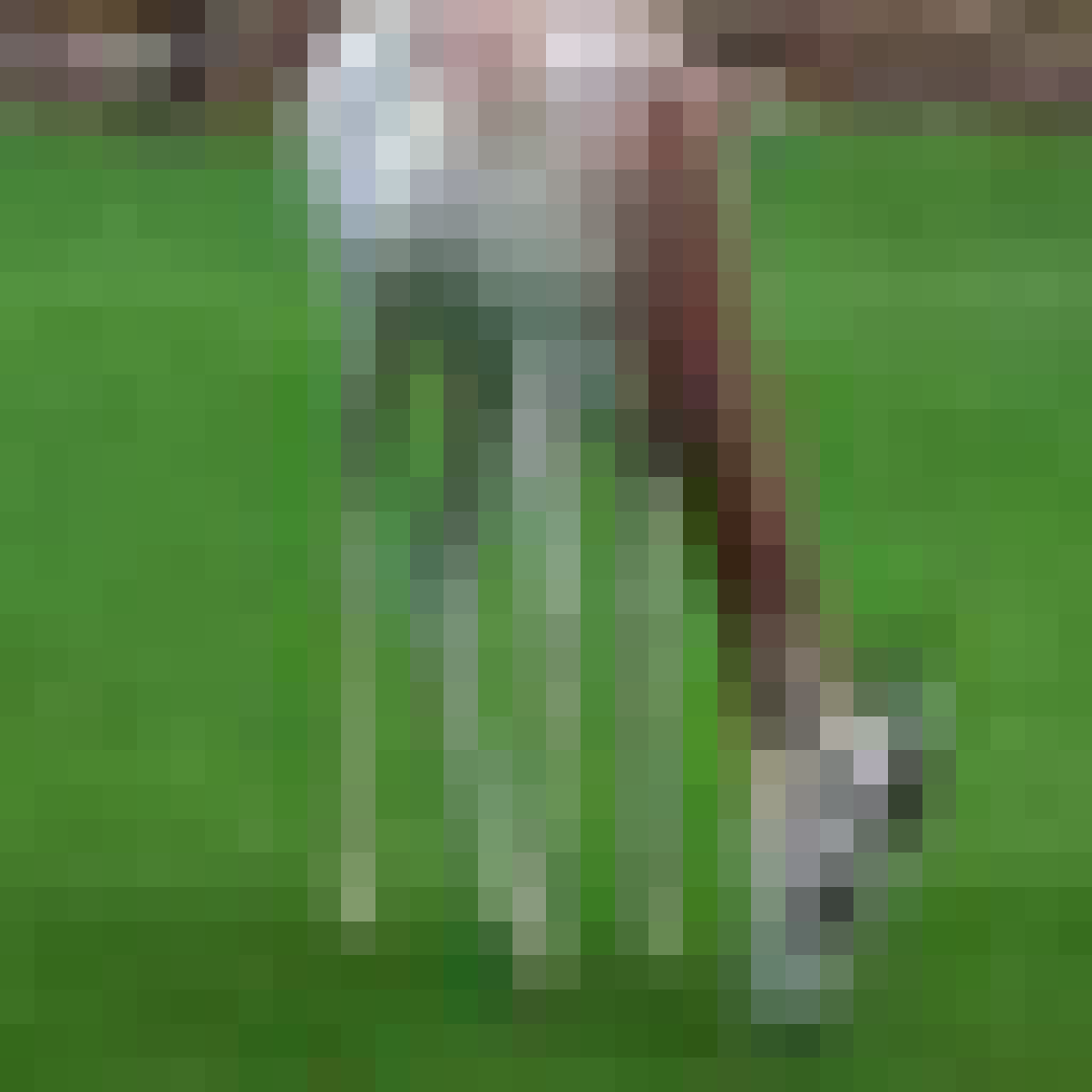} &
  \includegraphics[scale=\scale]{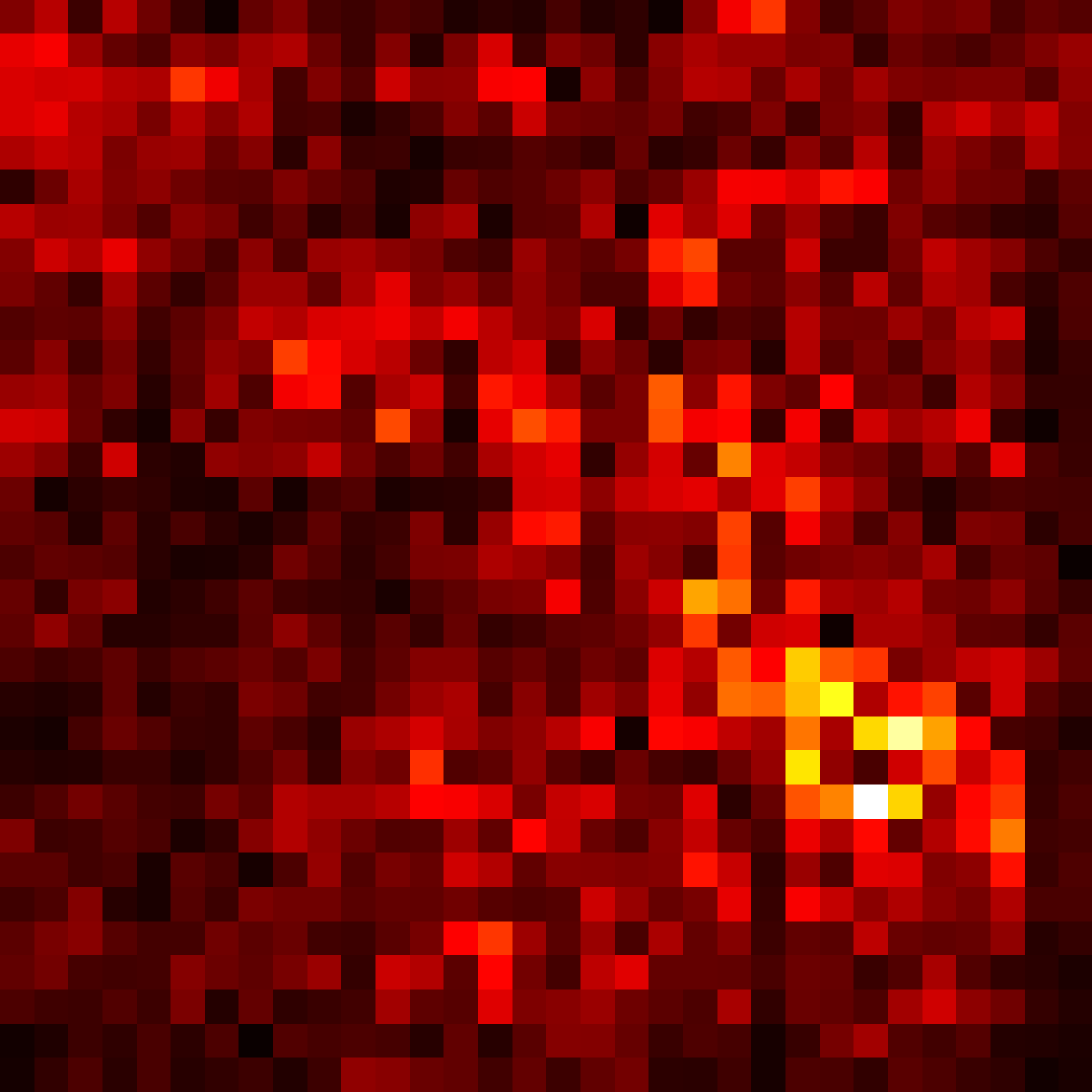} & 
  \includegraphics[scale=\scale]{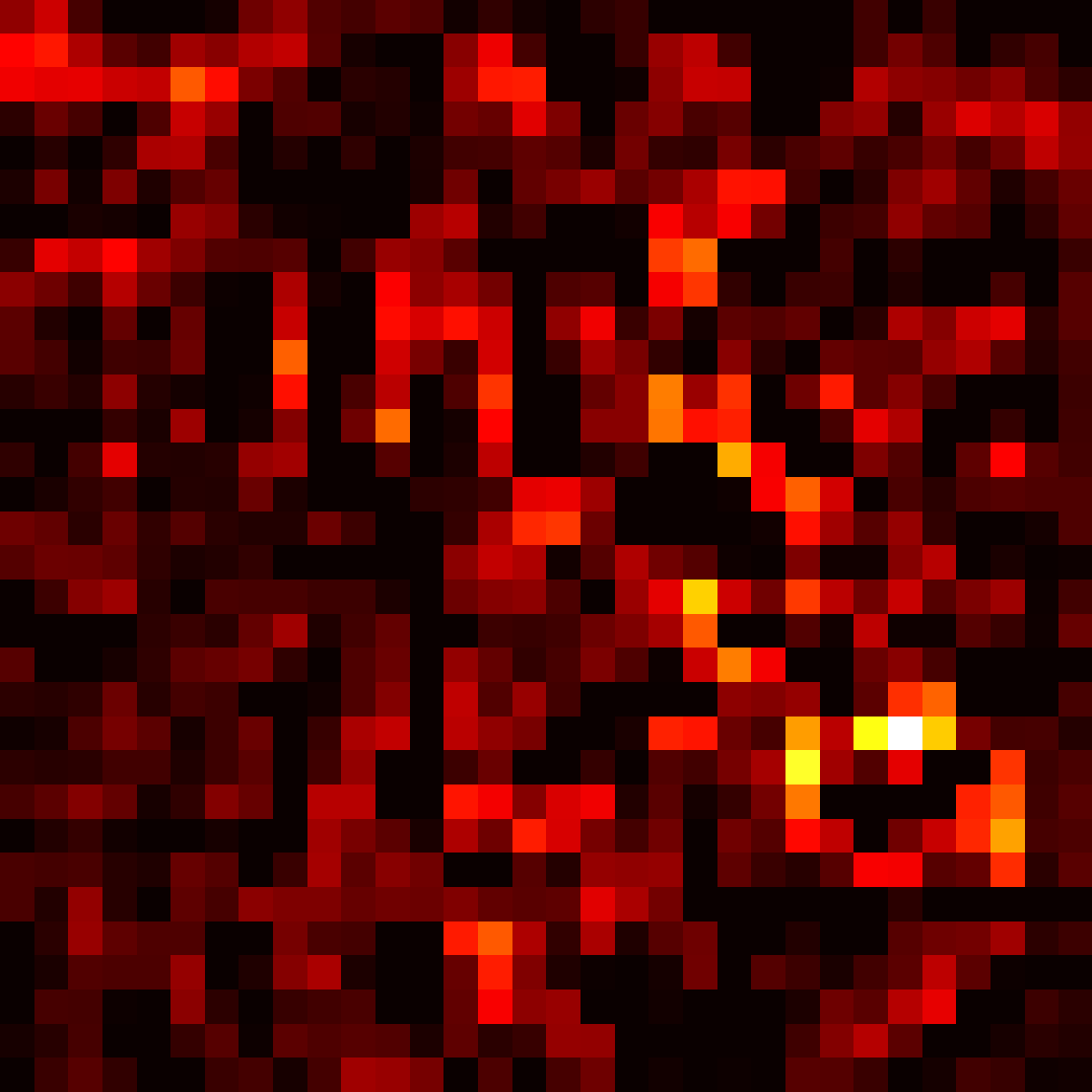} & 
  \includegraphics[scale=\scale]{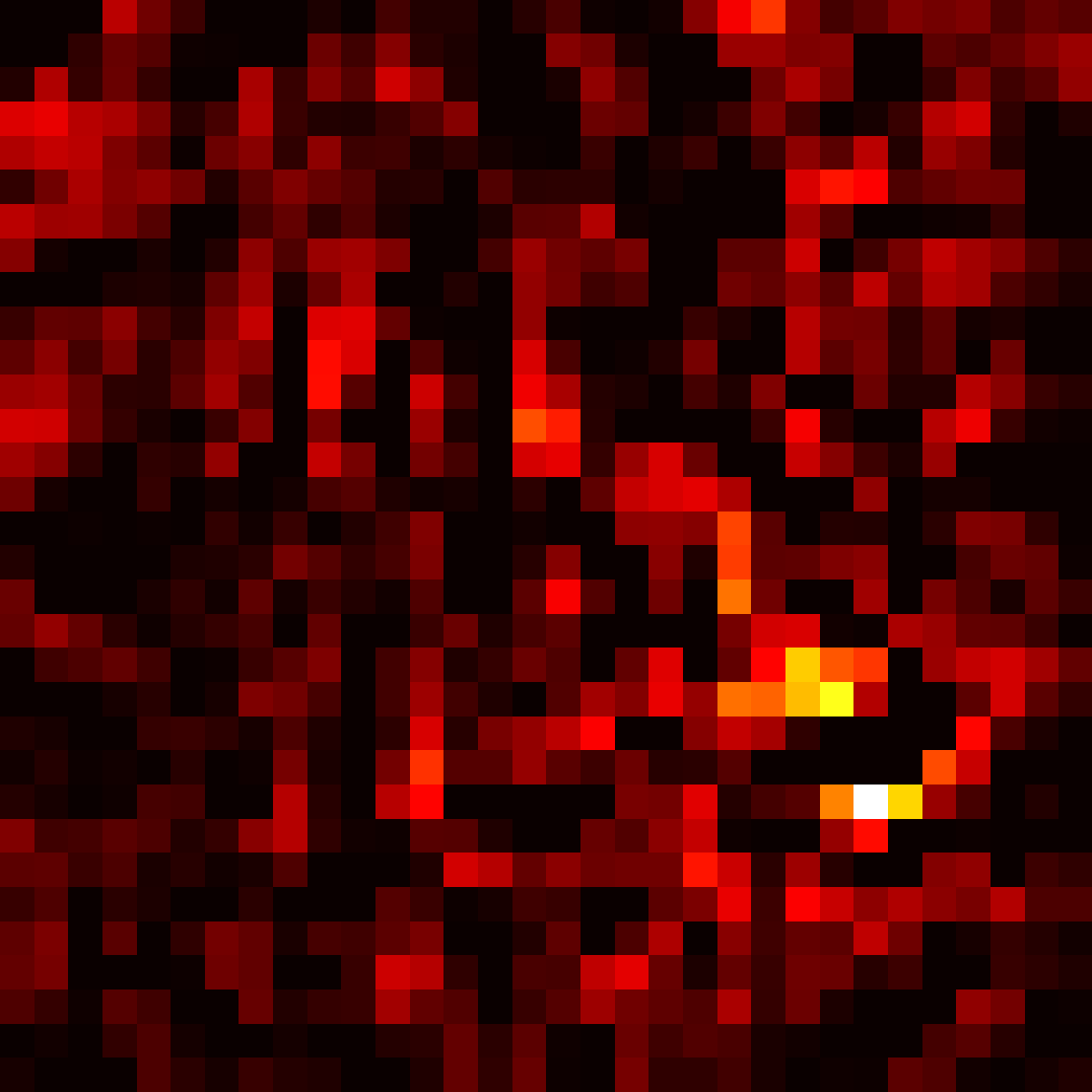} & 
  \includegraphics[scale=\scale]{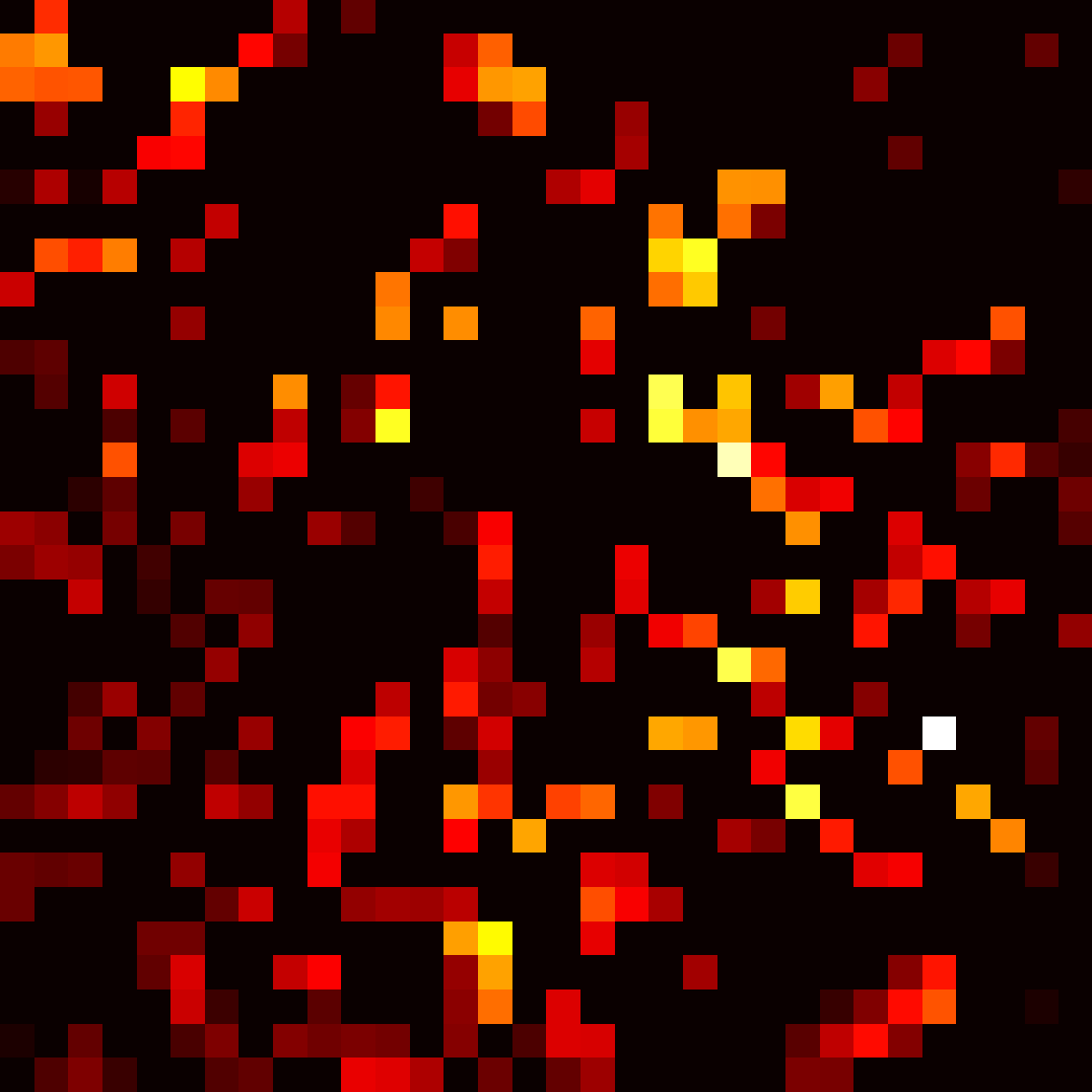} & 
  \includegraphics[scale=\scale]{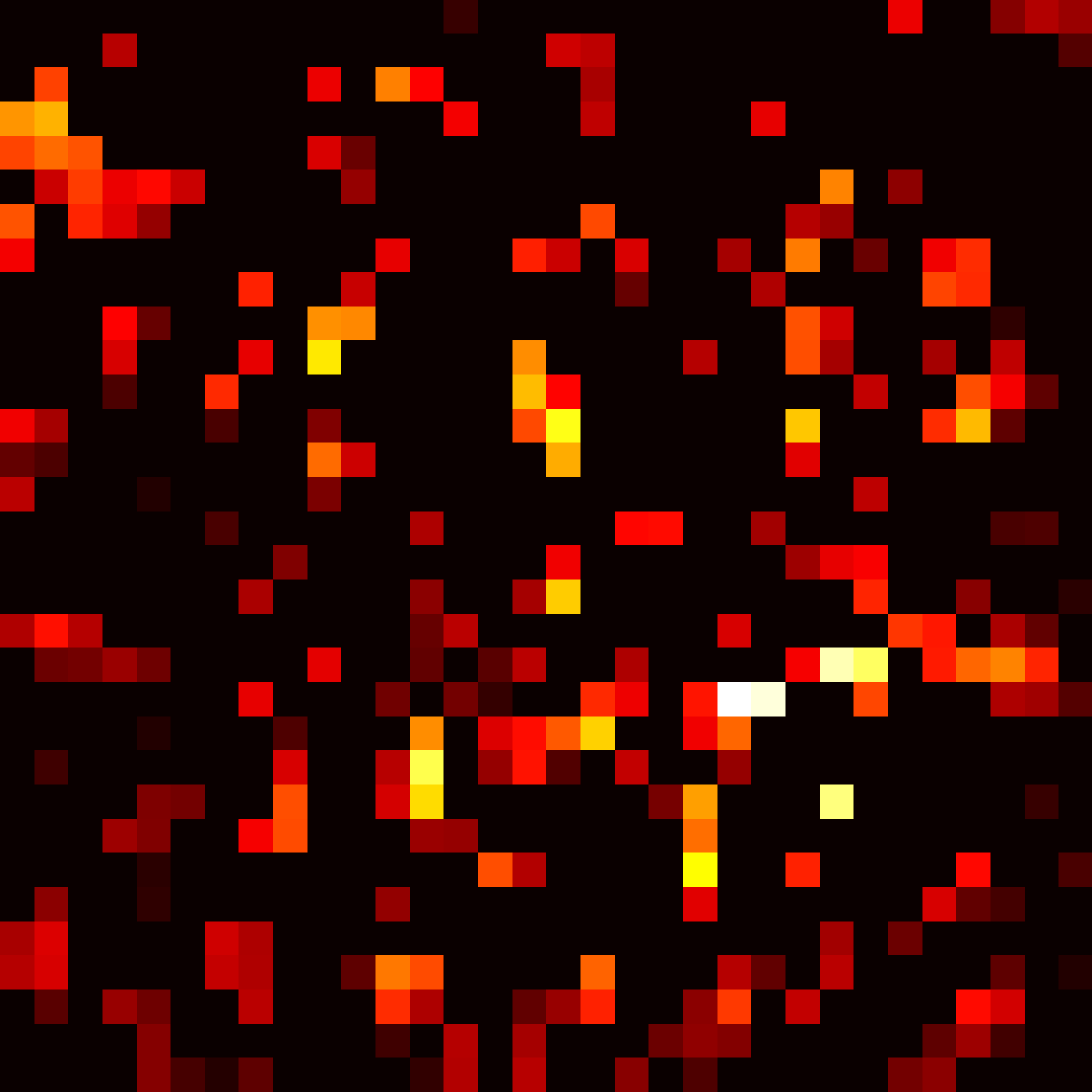} \\
  
  \includegraphics[scale=\scale]{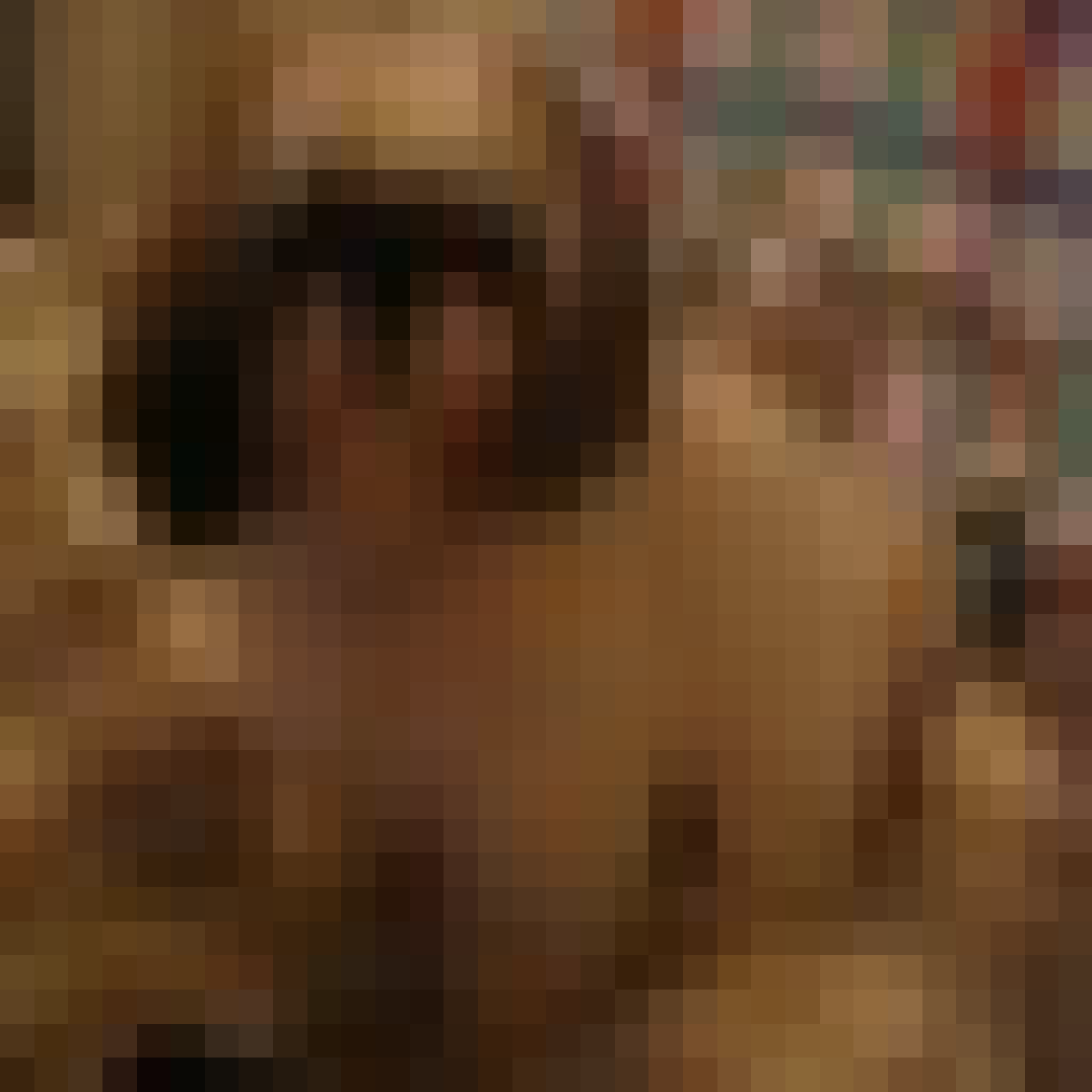} &
  \includegraphics[scale=\scale]{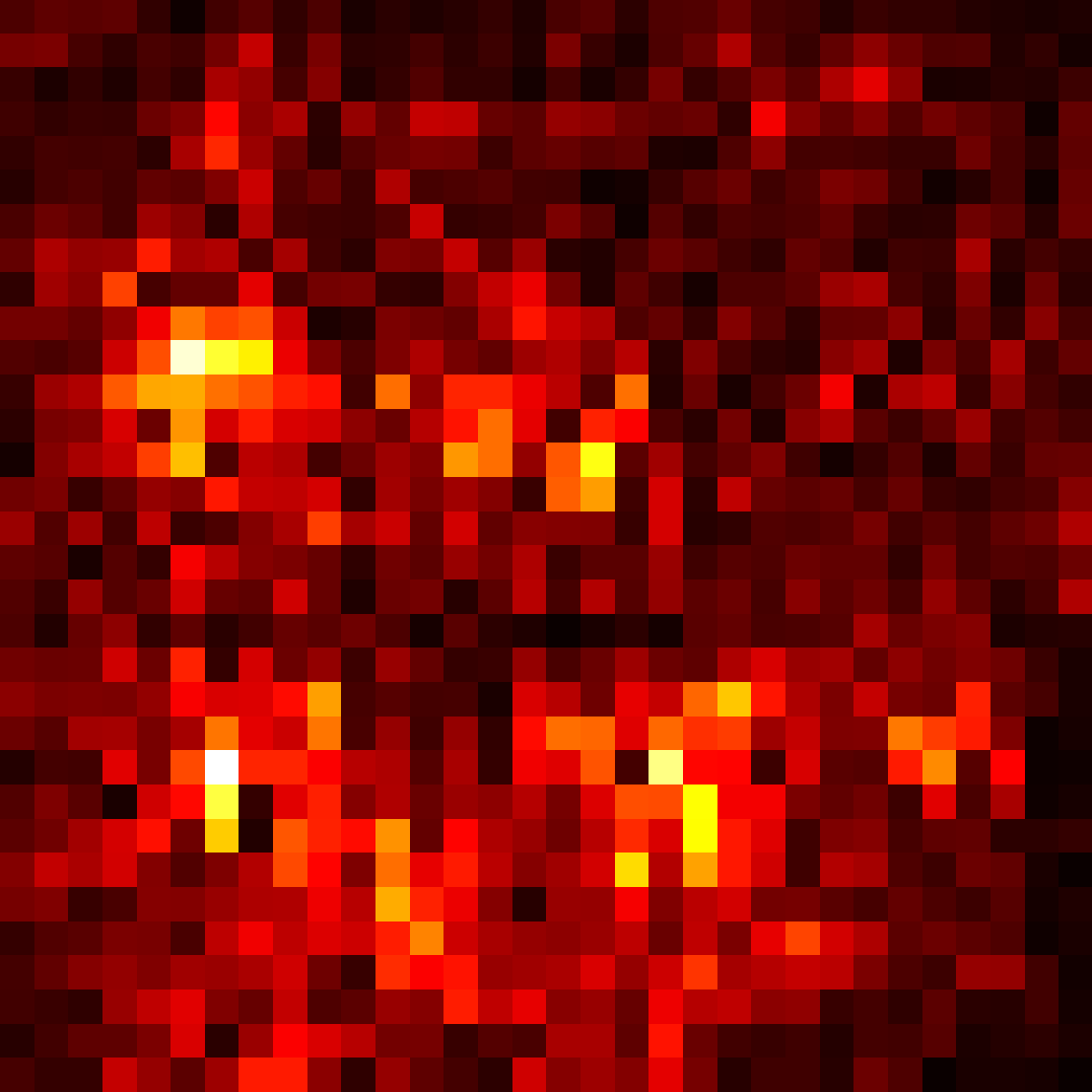} & 
  \includegraphics[scale=\scale]{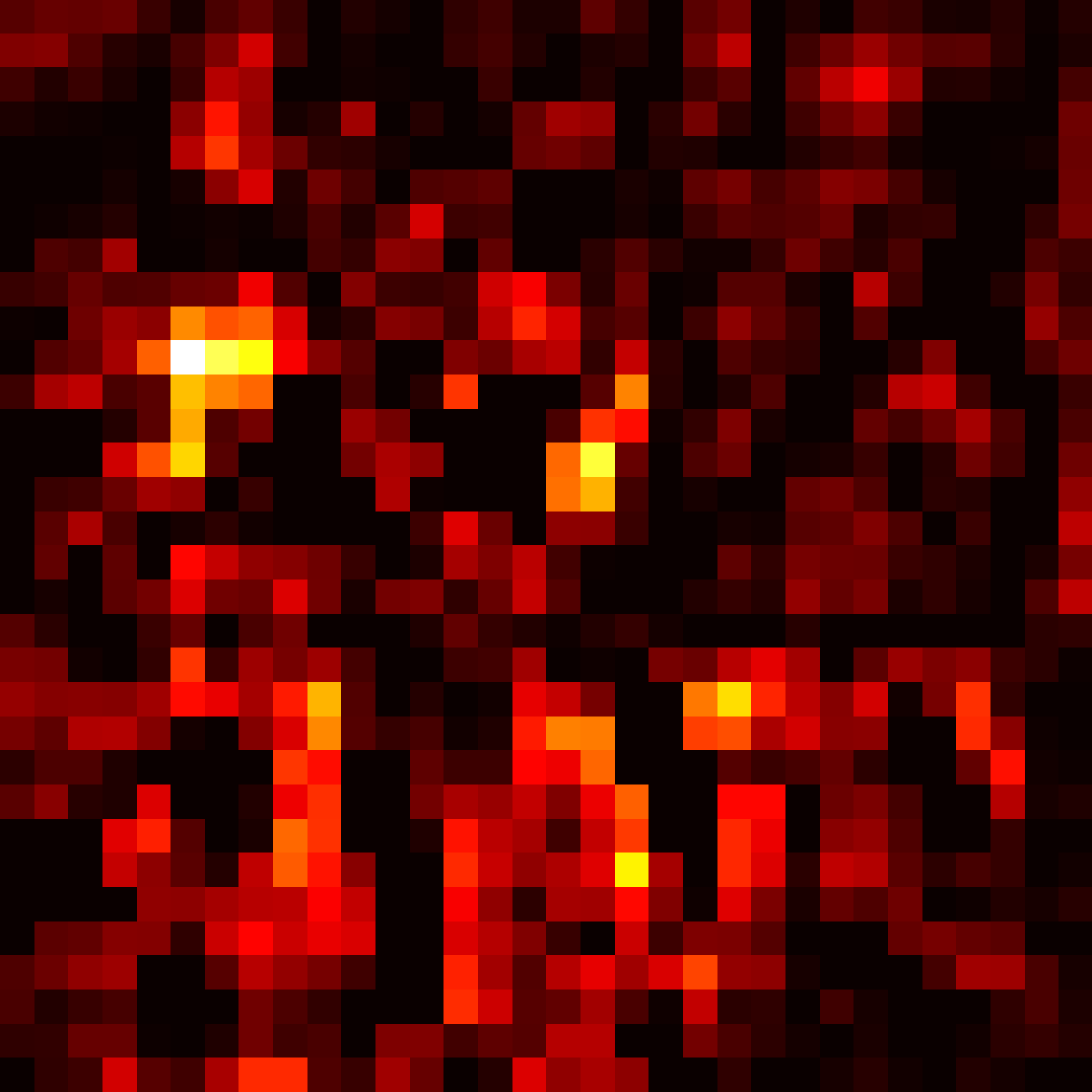} & 
  \includegraphics[scale=\scale]{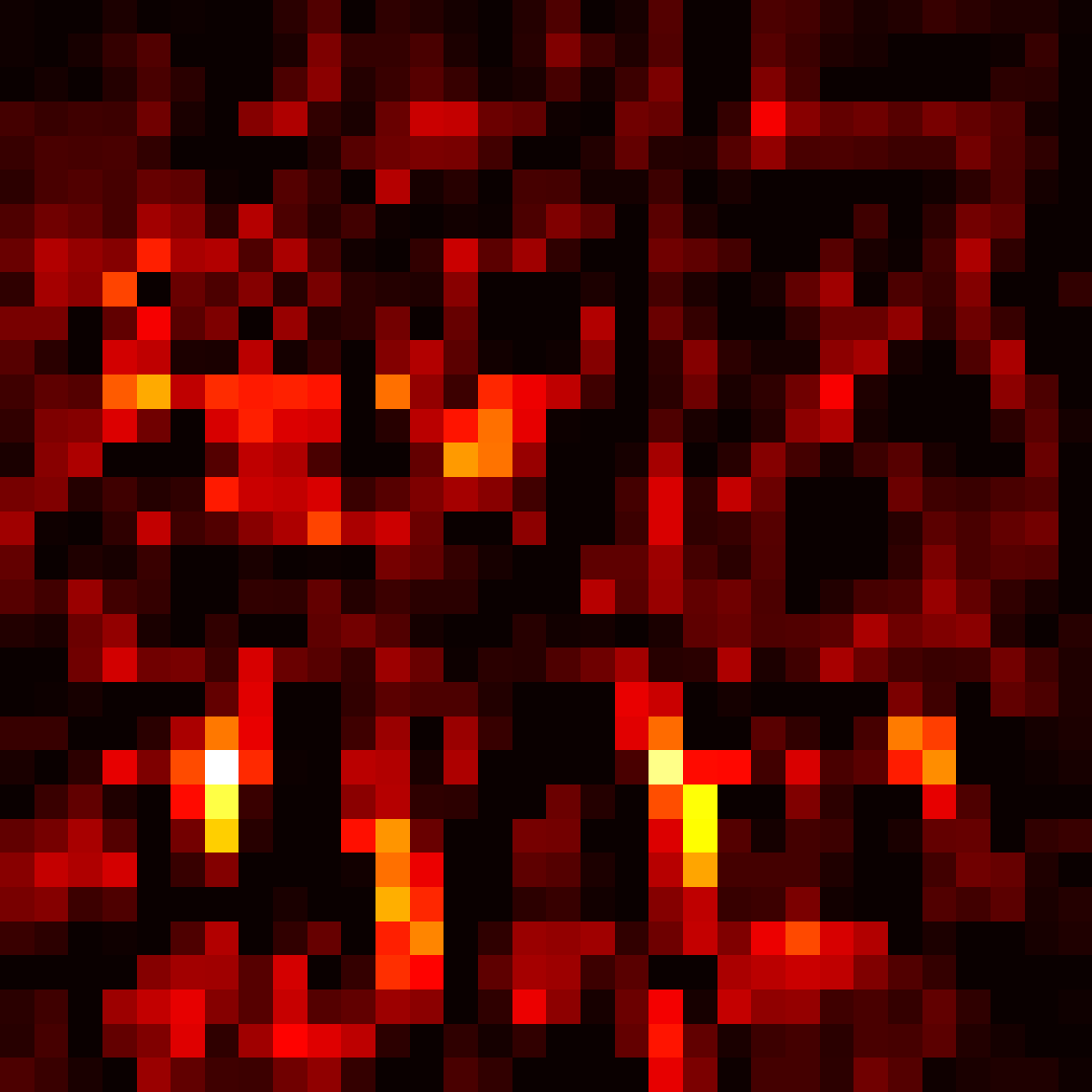} & 
  \includegraphics[scale=\scale]{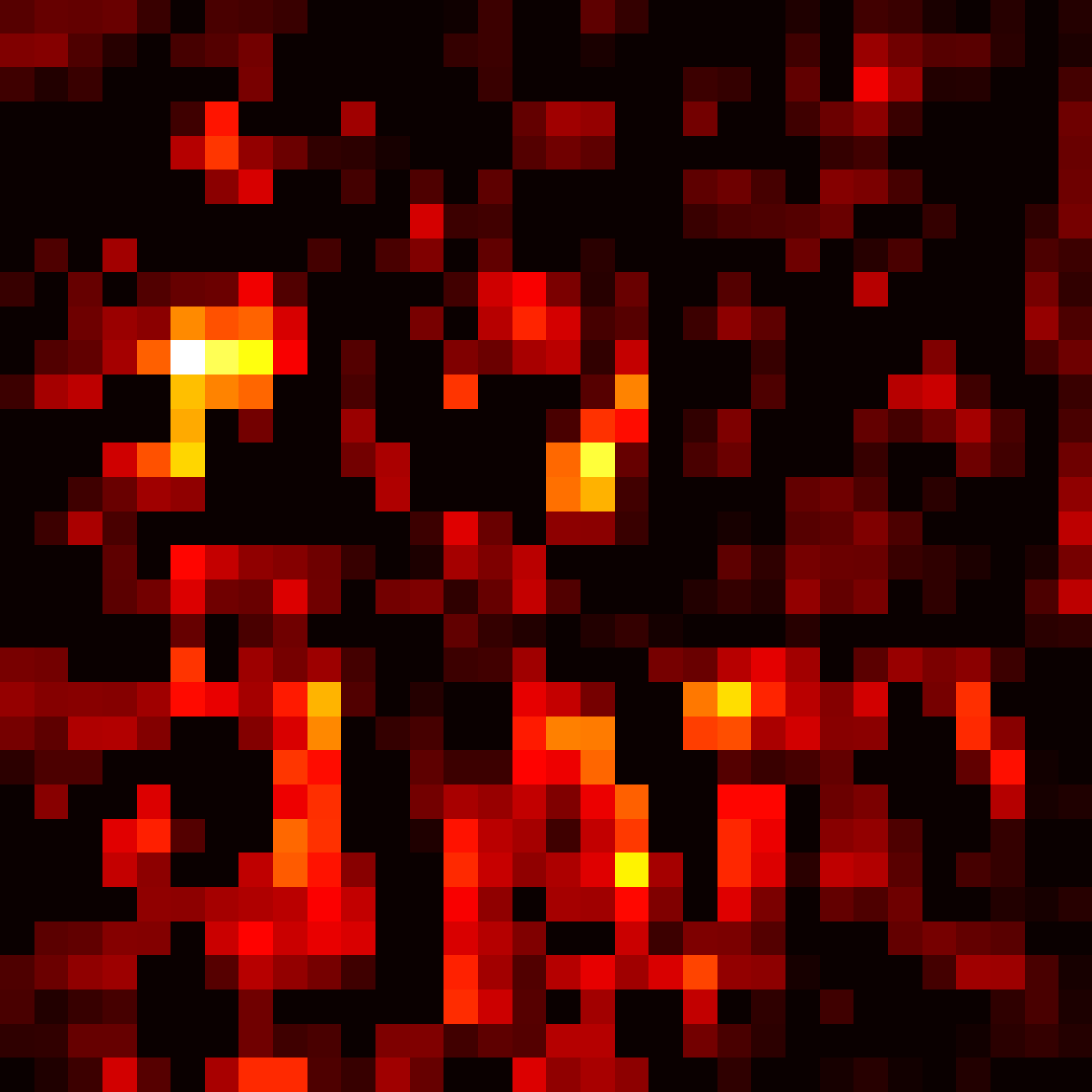} & 
  \includegraphics[scale=\scale]{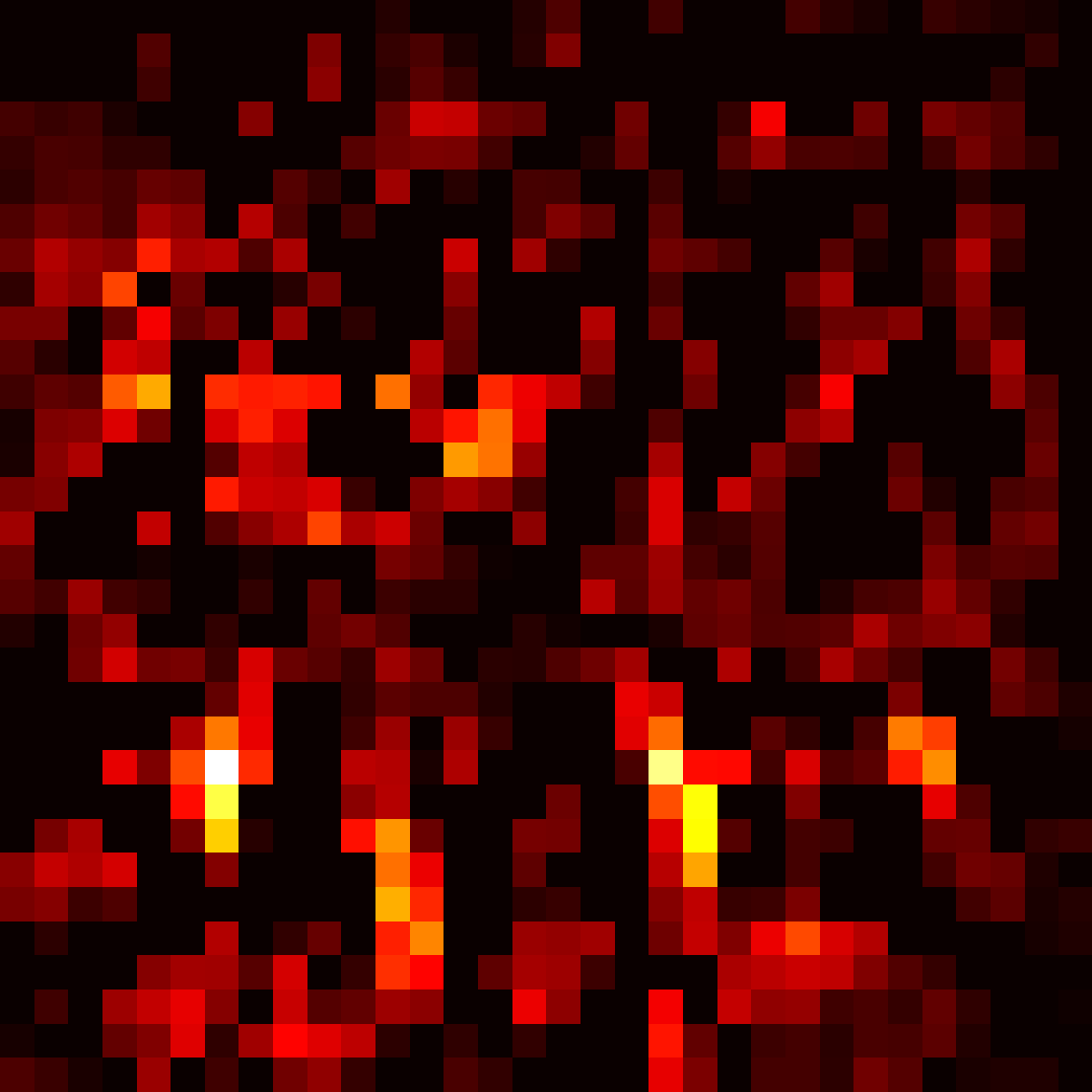} \\
  
  \includegraphics[scale=\scale]{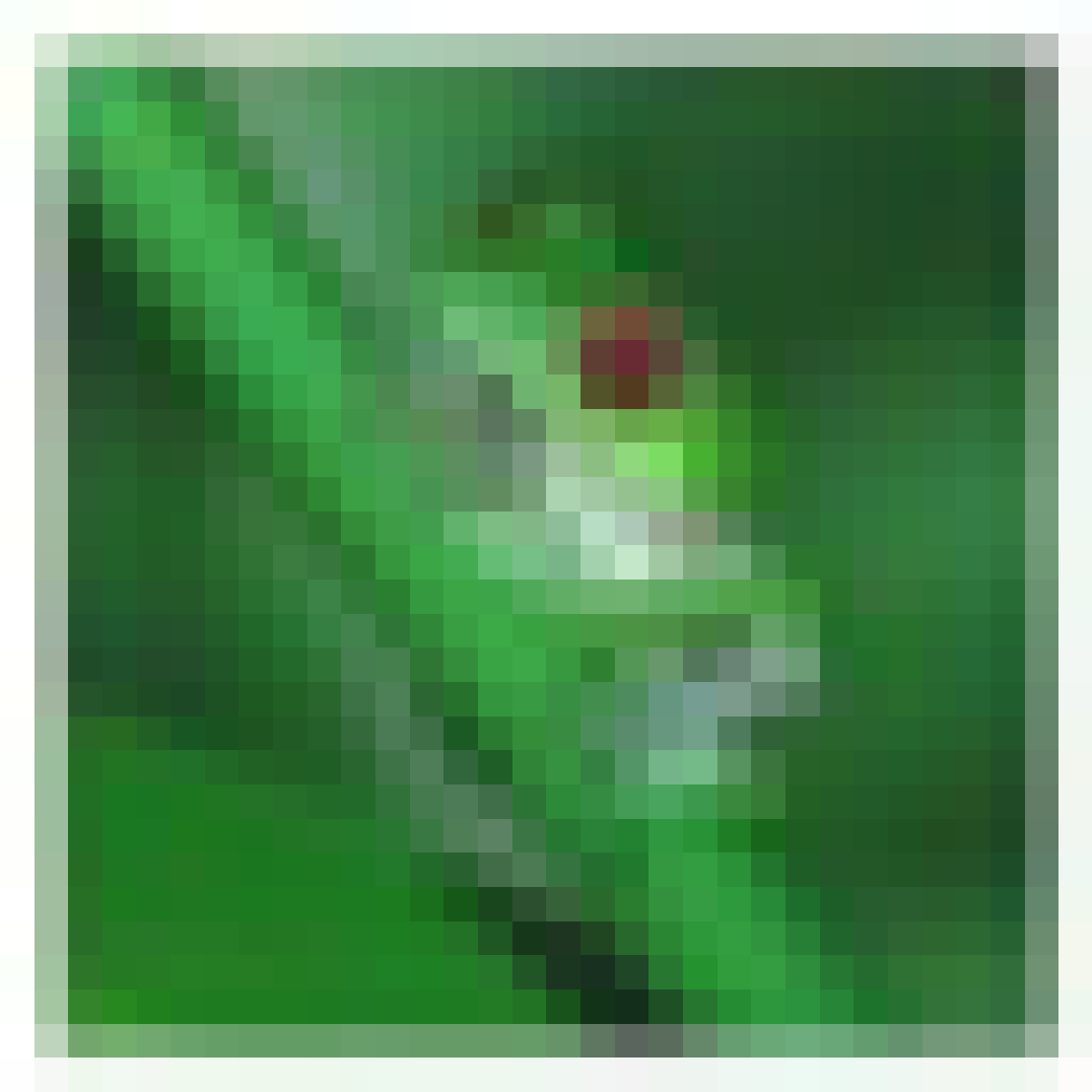} &
  \includegraphics[scale=\scale]{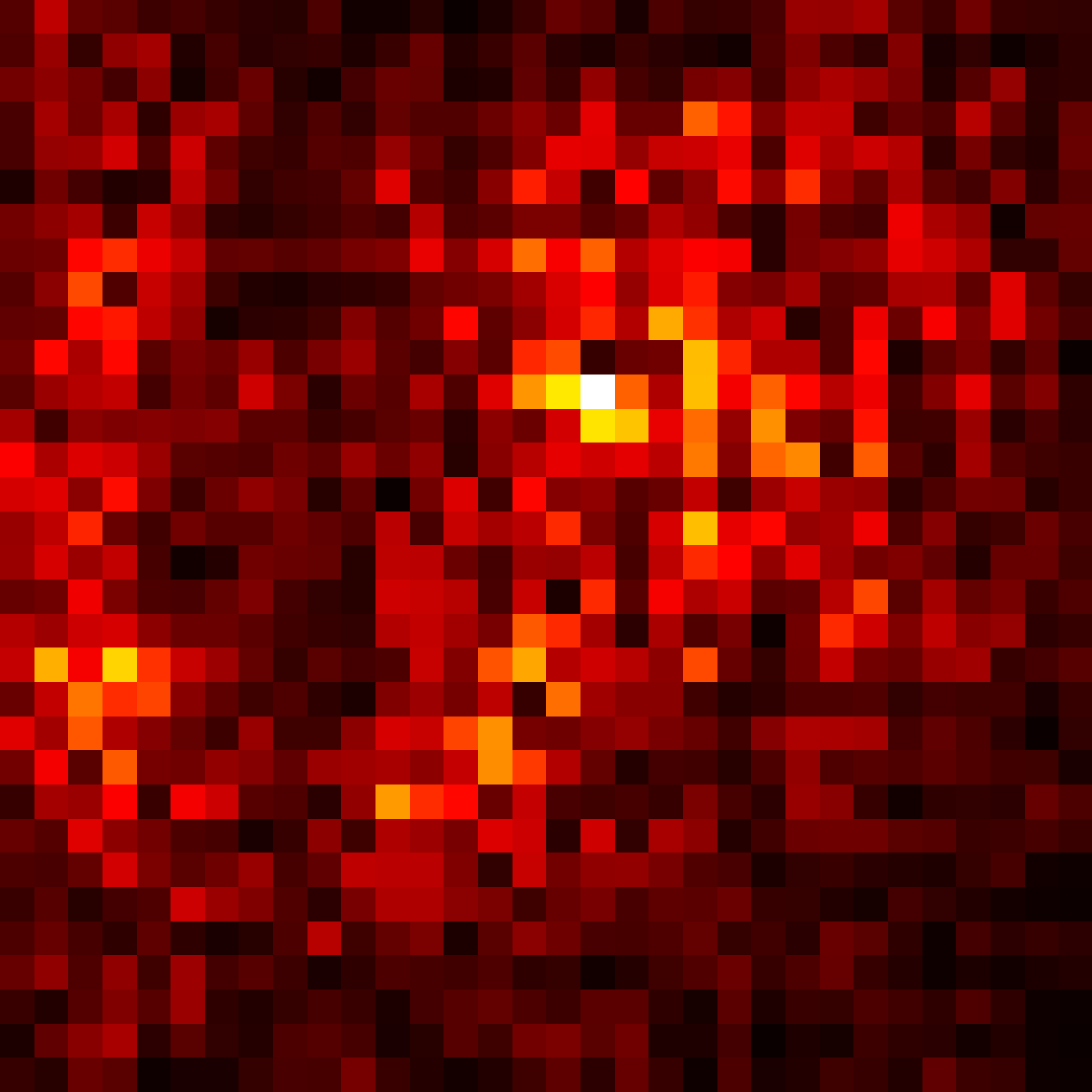} & 
  \includegraphics[scale=\scale]{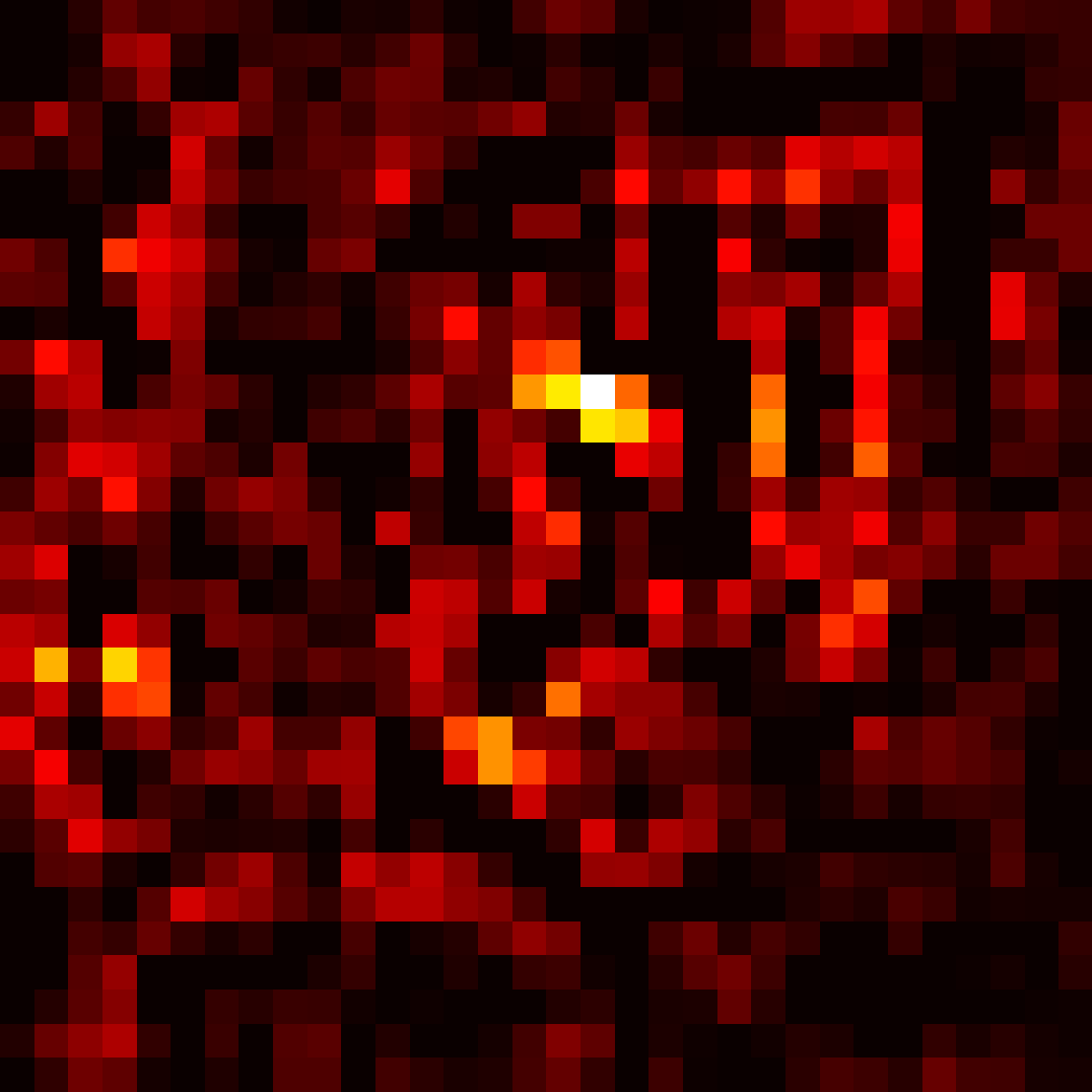} & 
  \includegraphics[scale=\scale]{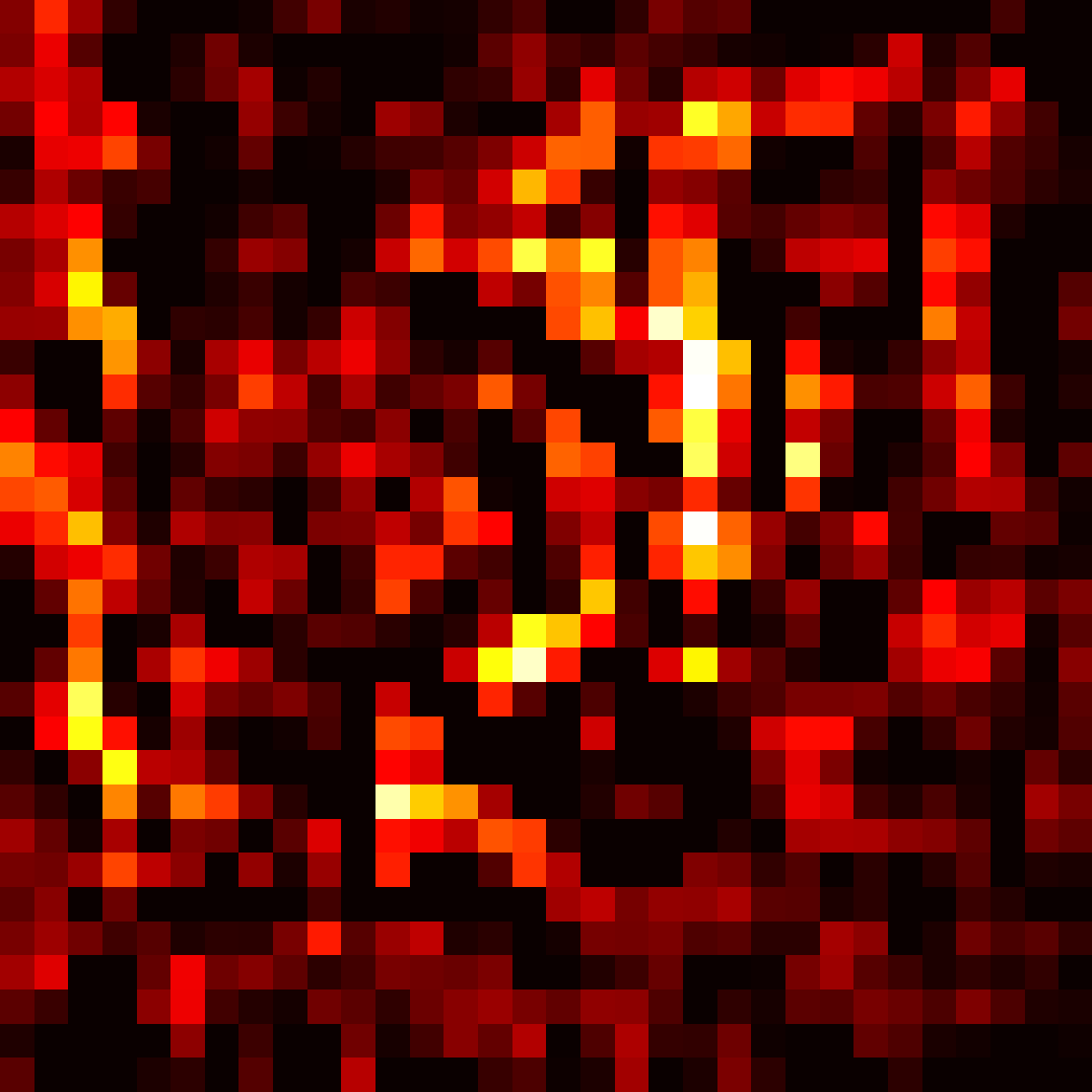} & 
  \includegraphics[scale=\scale]{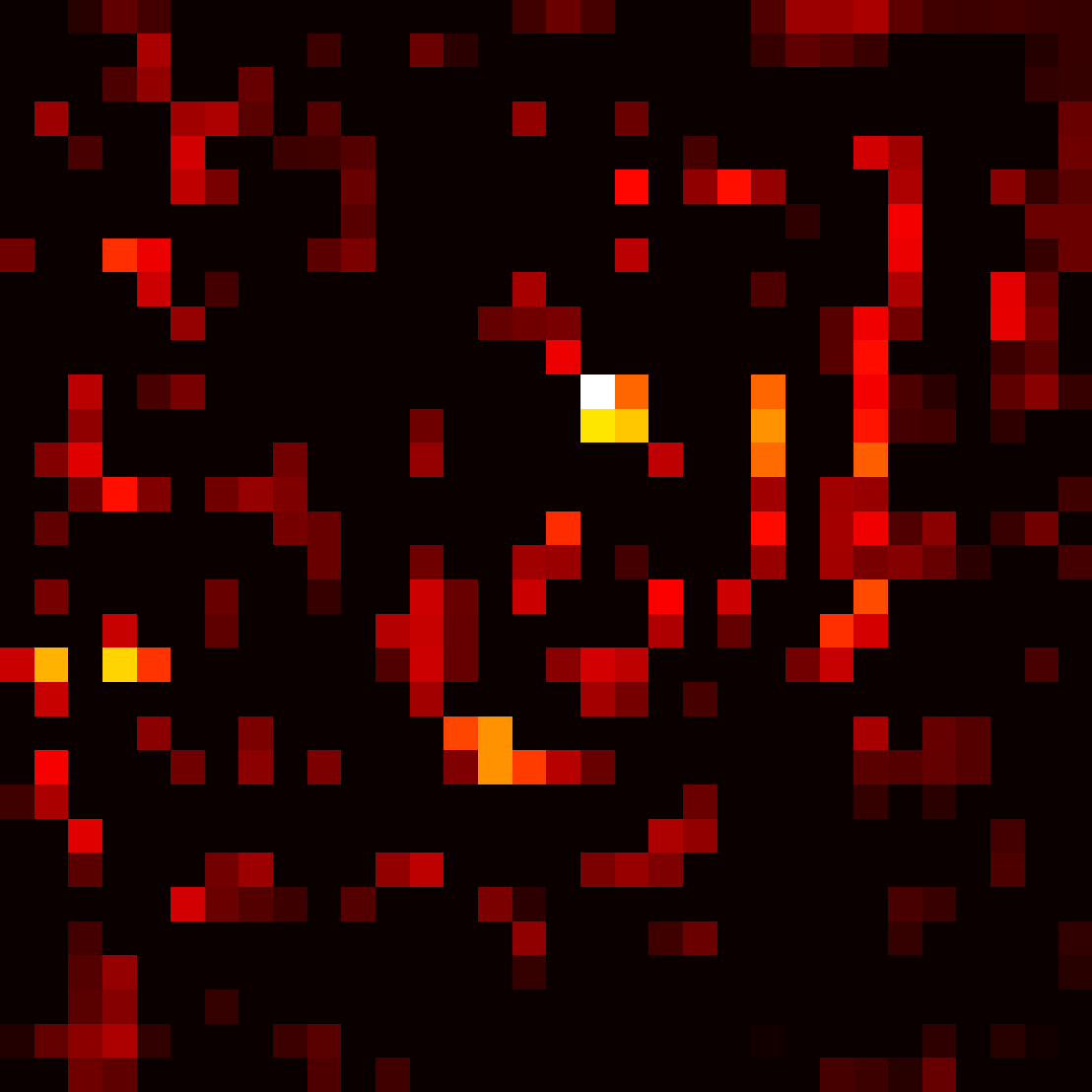} & 
  \includegraphics[scale=\scale]{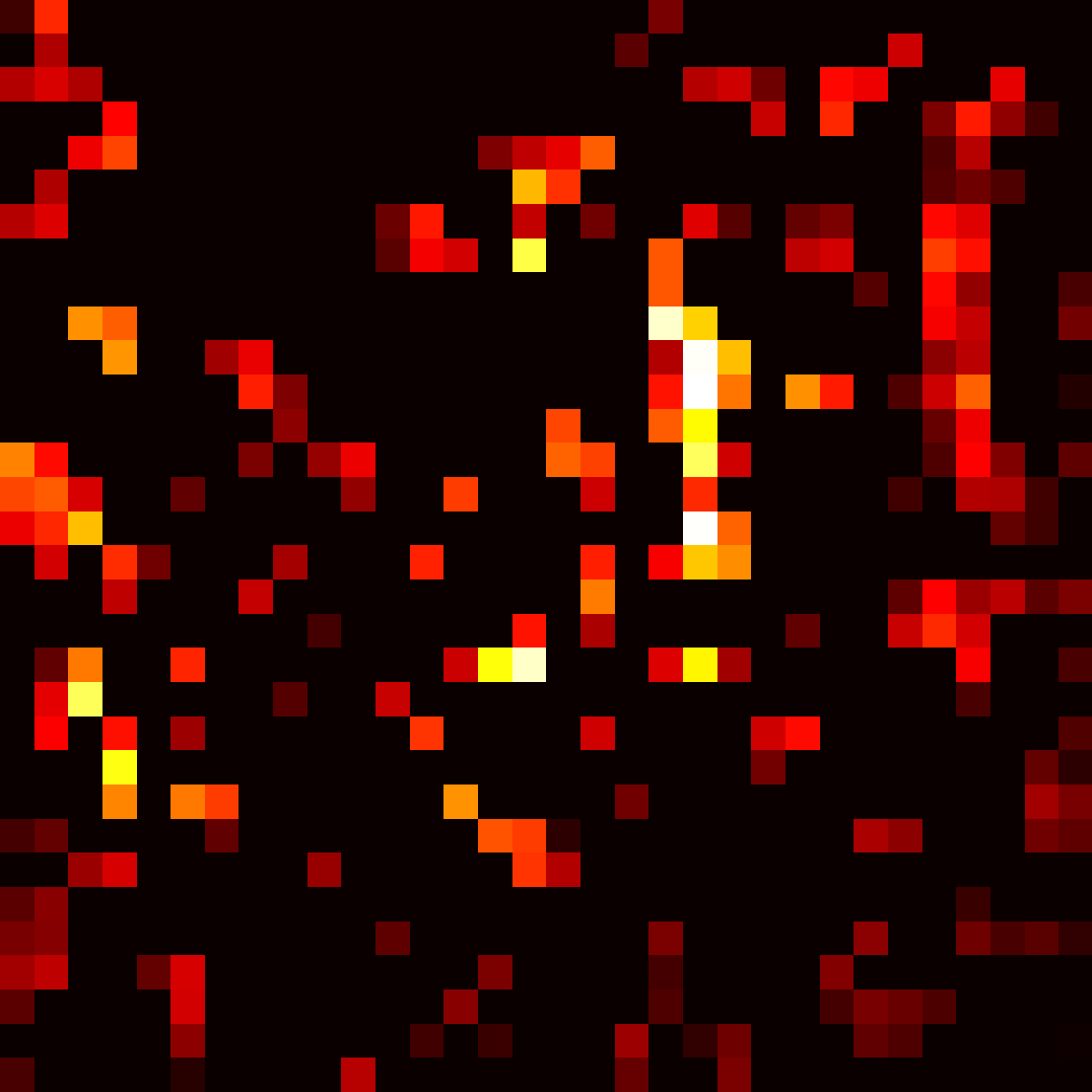} \\
  
  \includegraphics[scale=\scale]{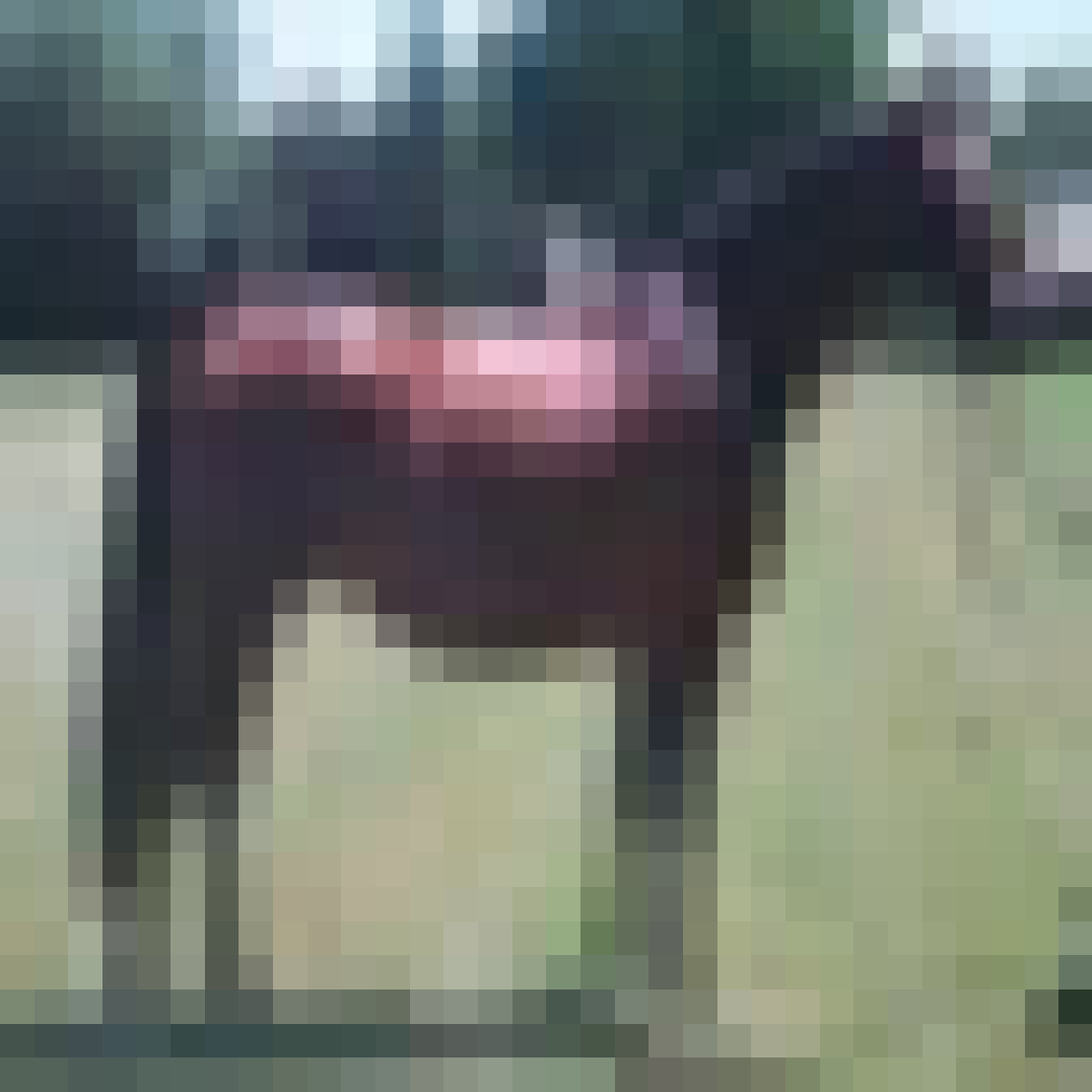} &
  \includegraphics[scale=\scale]{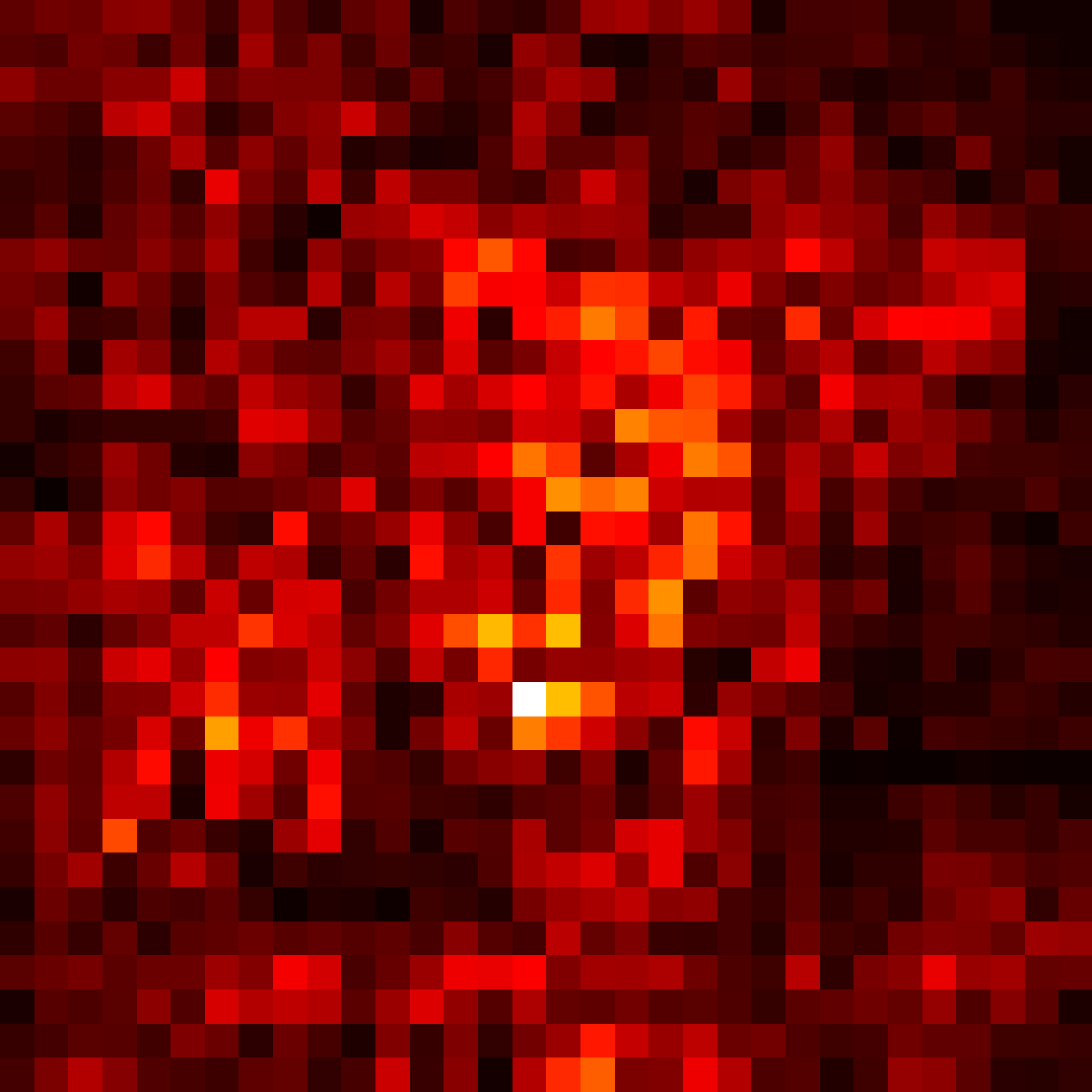} & 
  \includegraphics[scale=\scale]{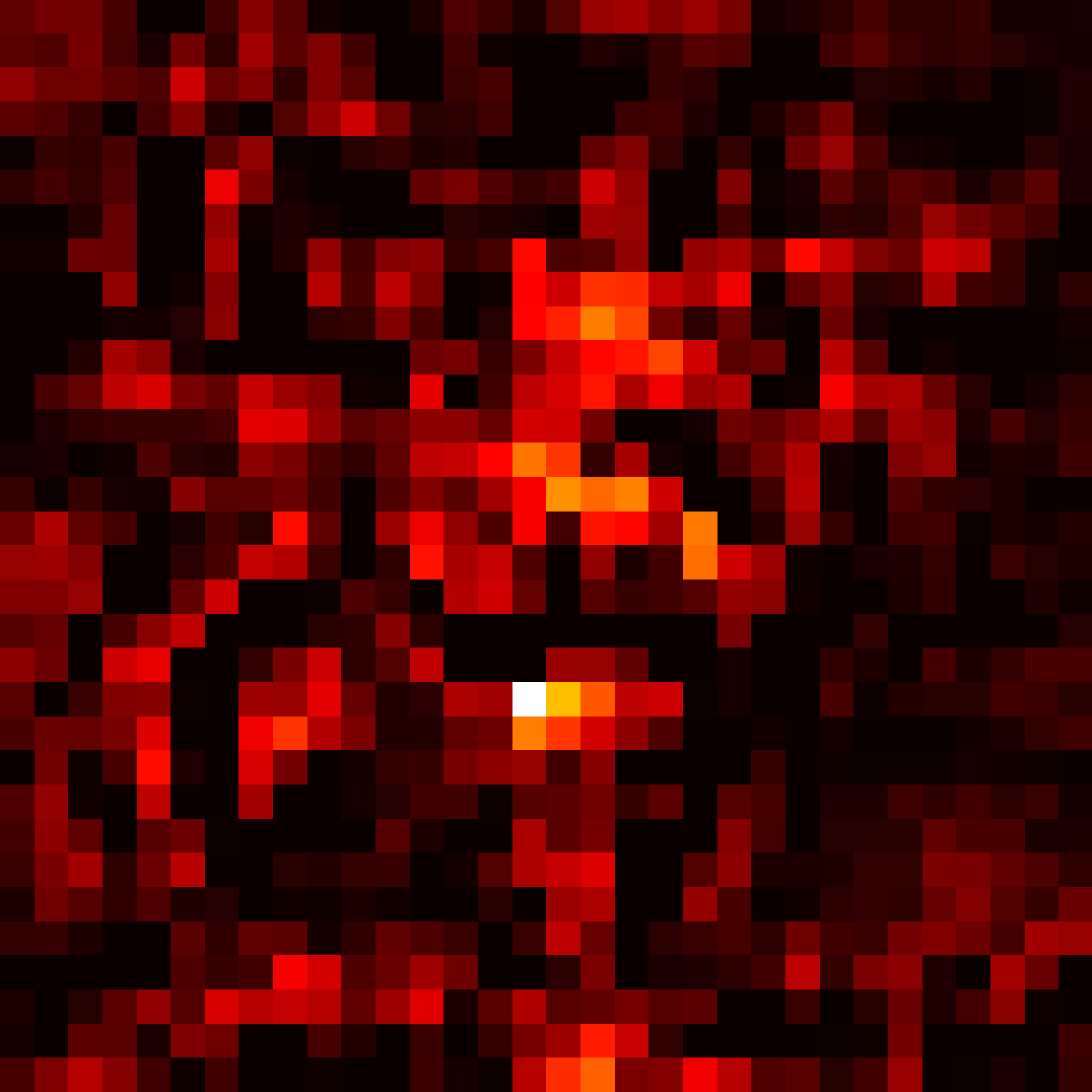} & 
  \includegraphics[scale=\scale]{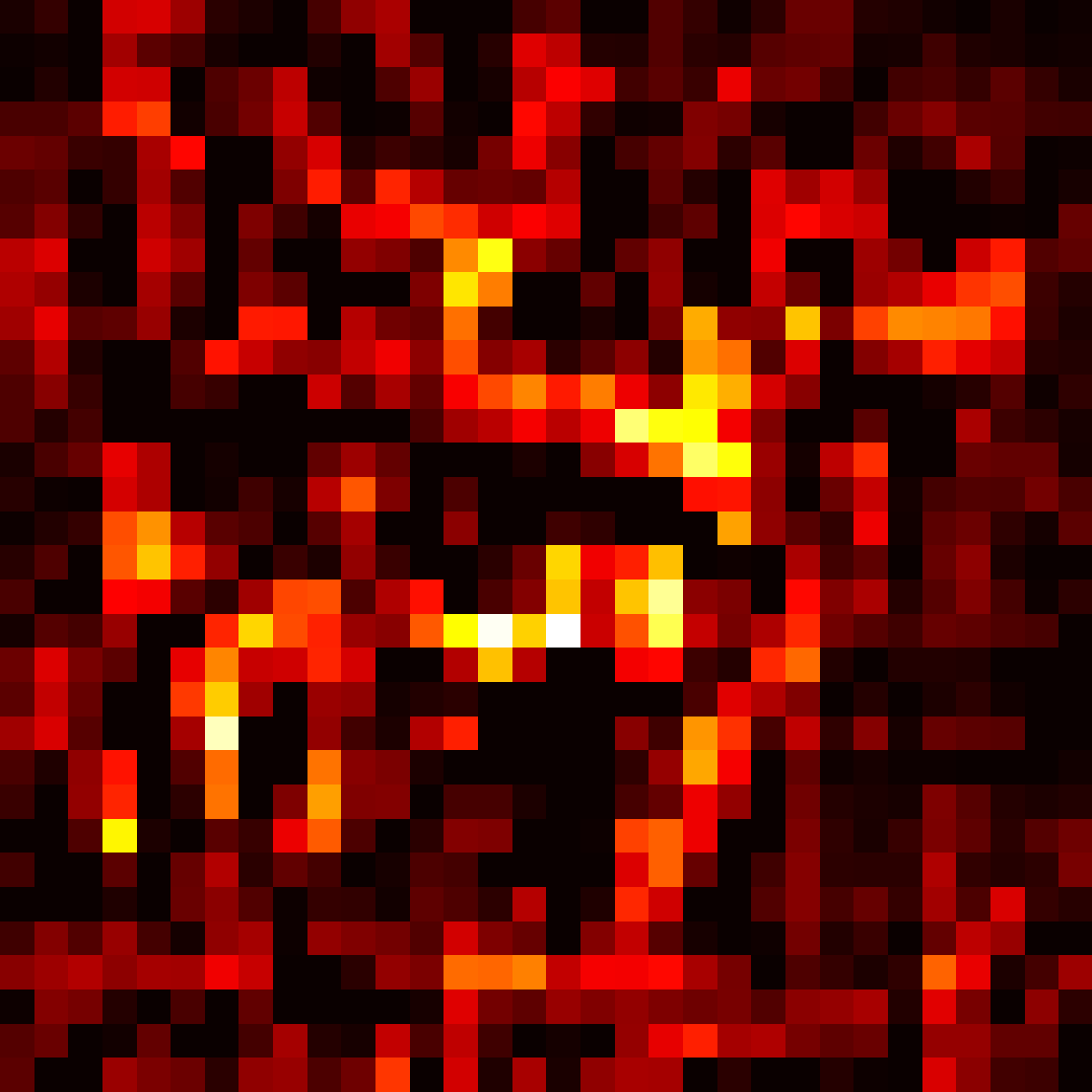} & 
  \includegraphics[scale=\scale]{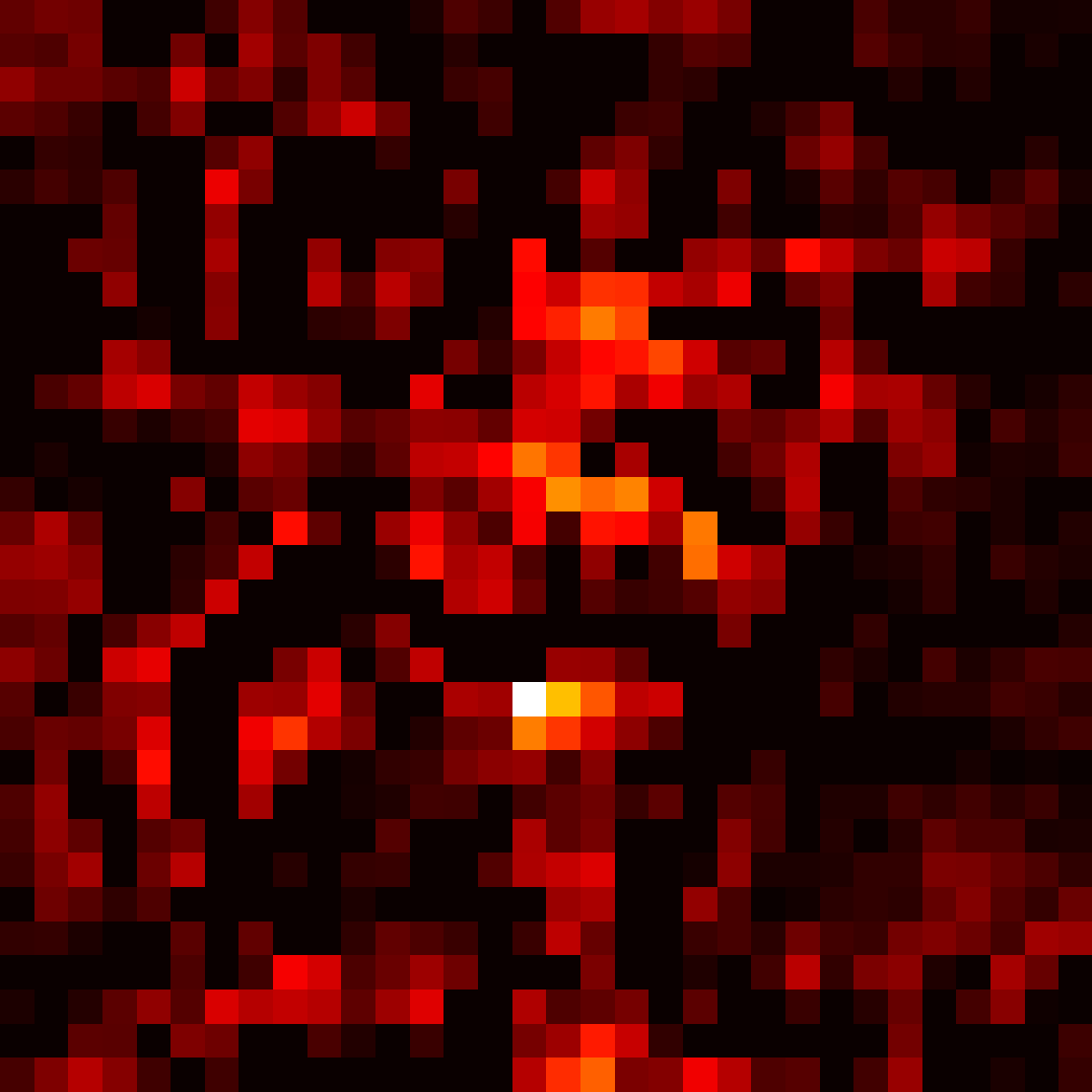} & 
  \includegraphics[scale=\scale]{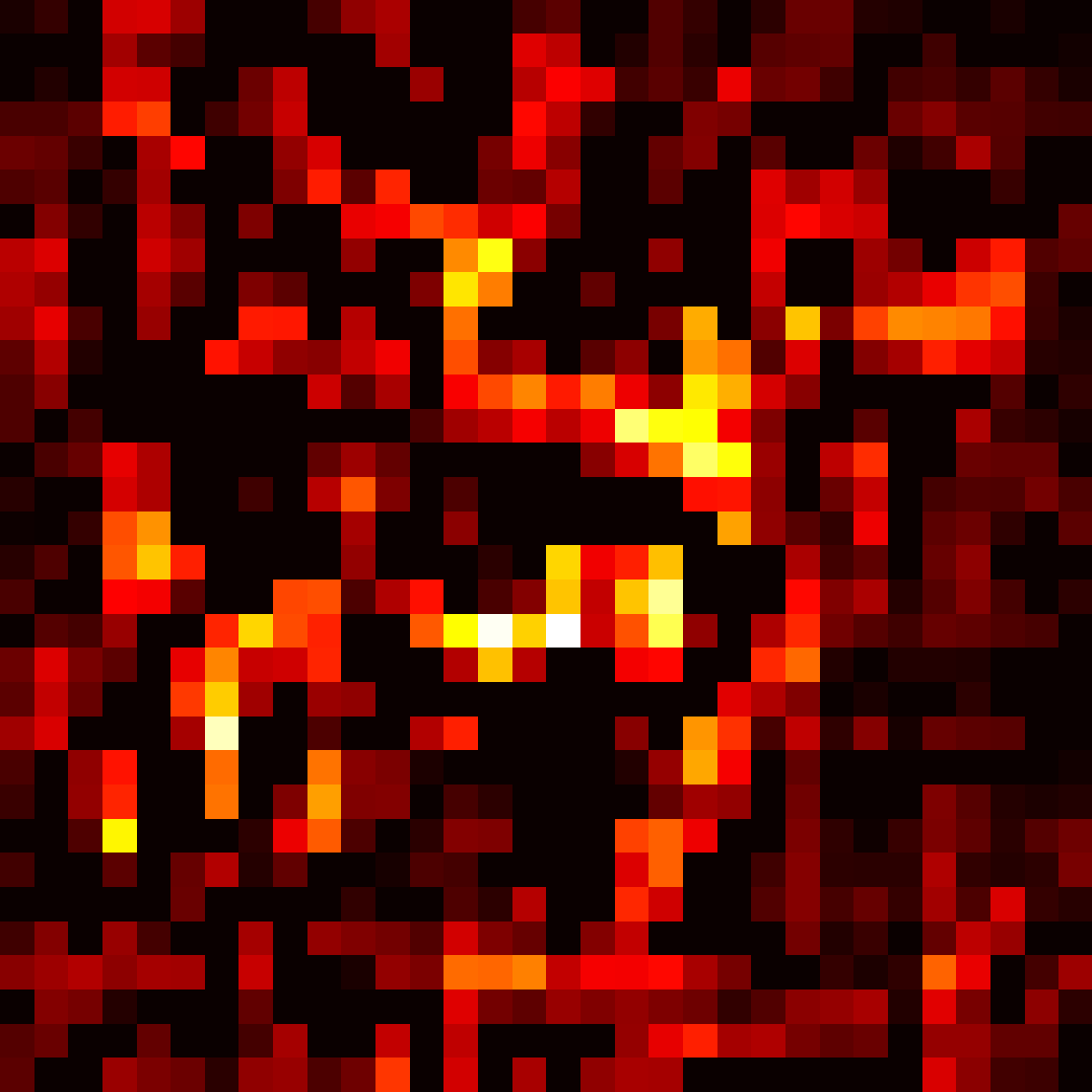} \\
  
  \includegraphics[scale=\scale]{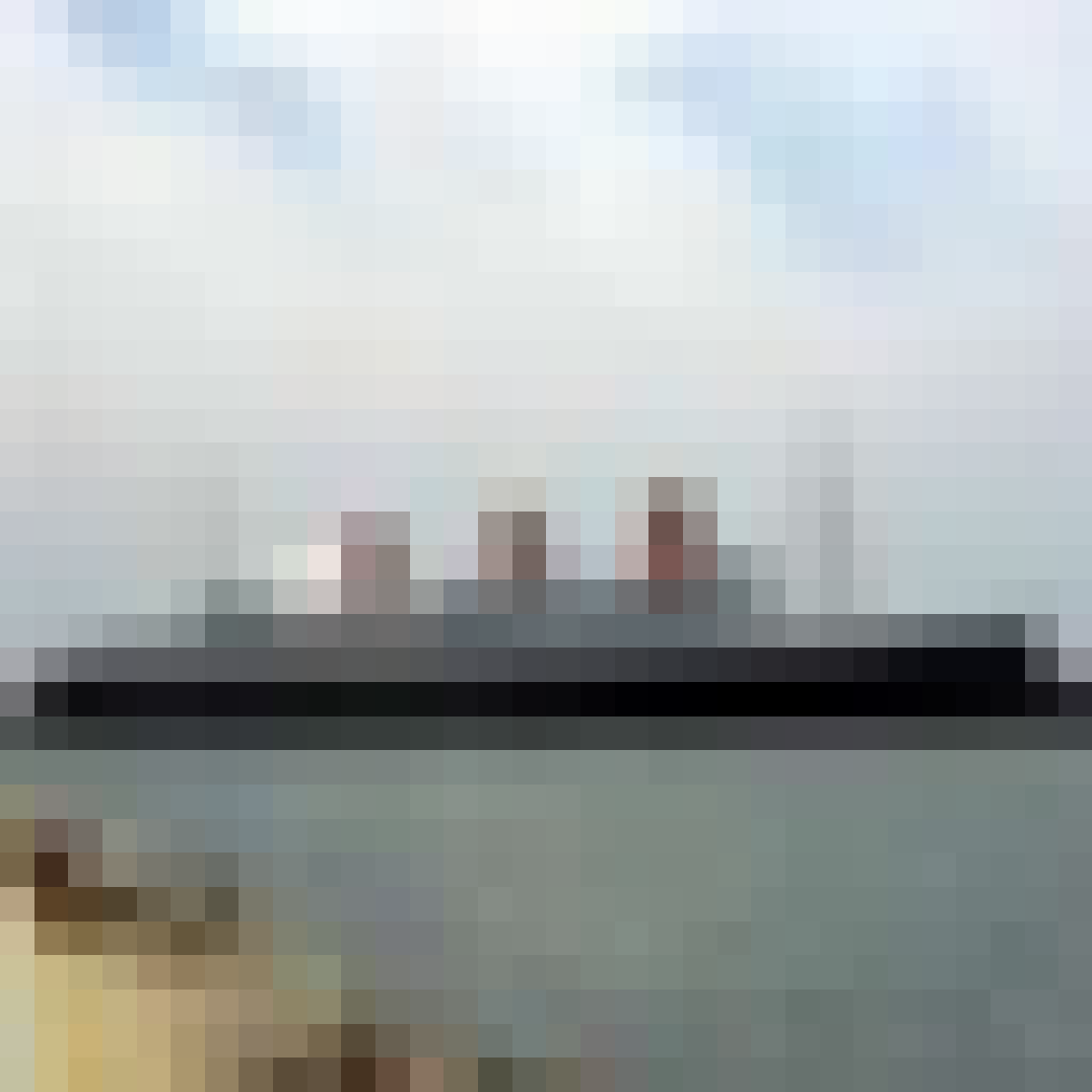} &
  \includegraphics[scale=\scale]{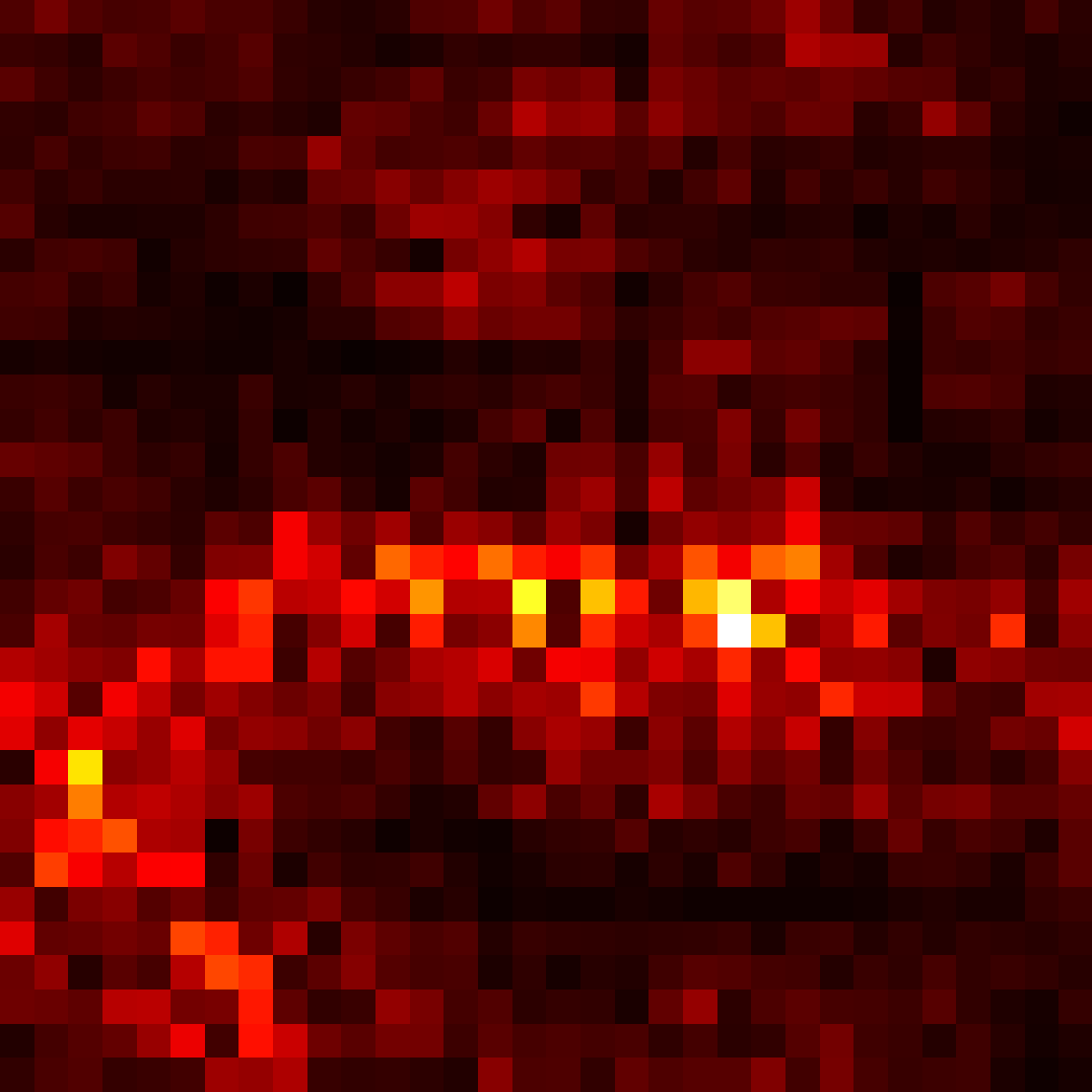} & 
  \includegraphics[scale=\scale]{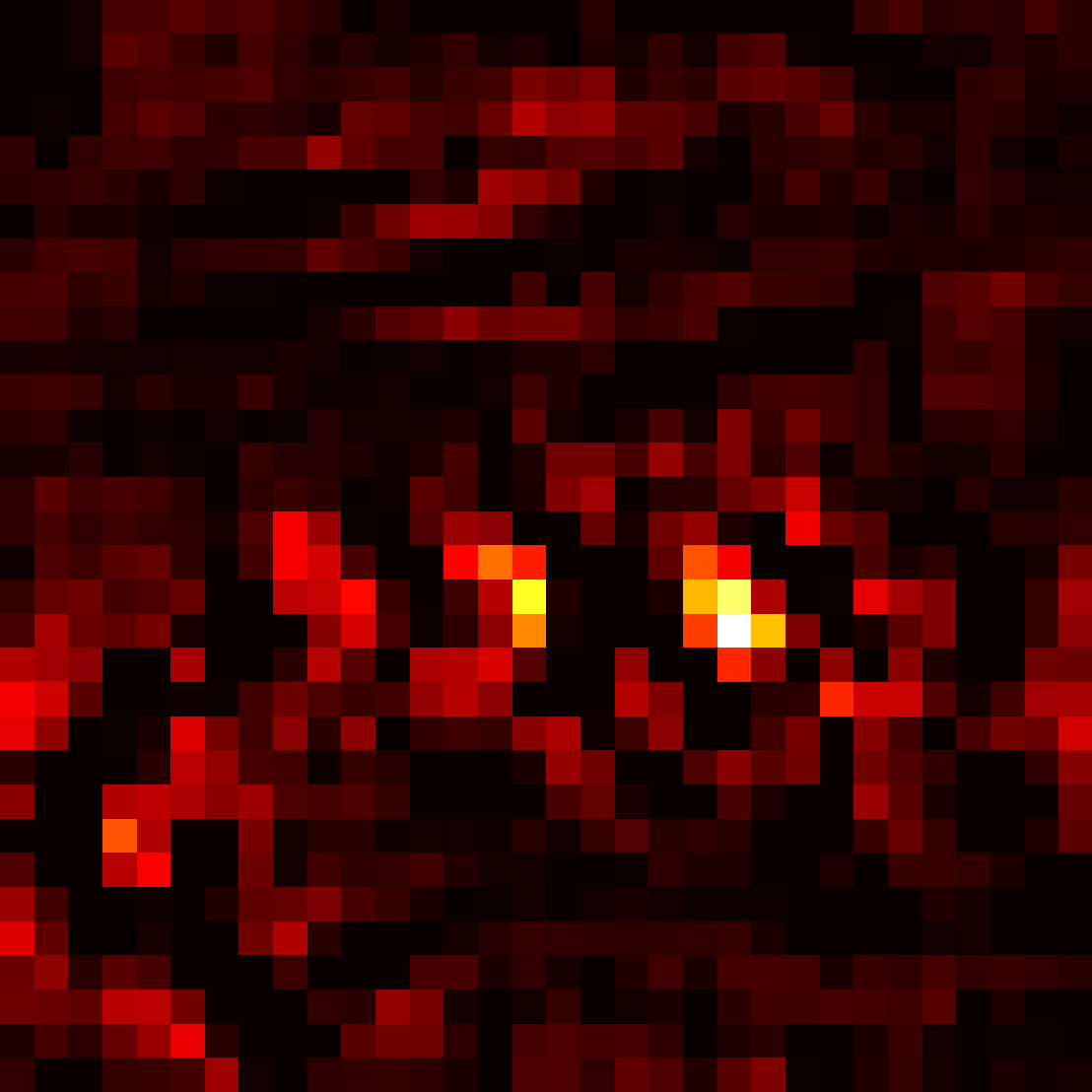} & 
  \includegraphics[scale=\scale]{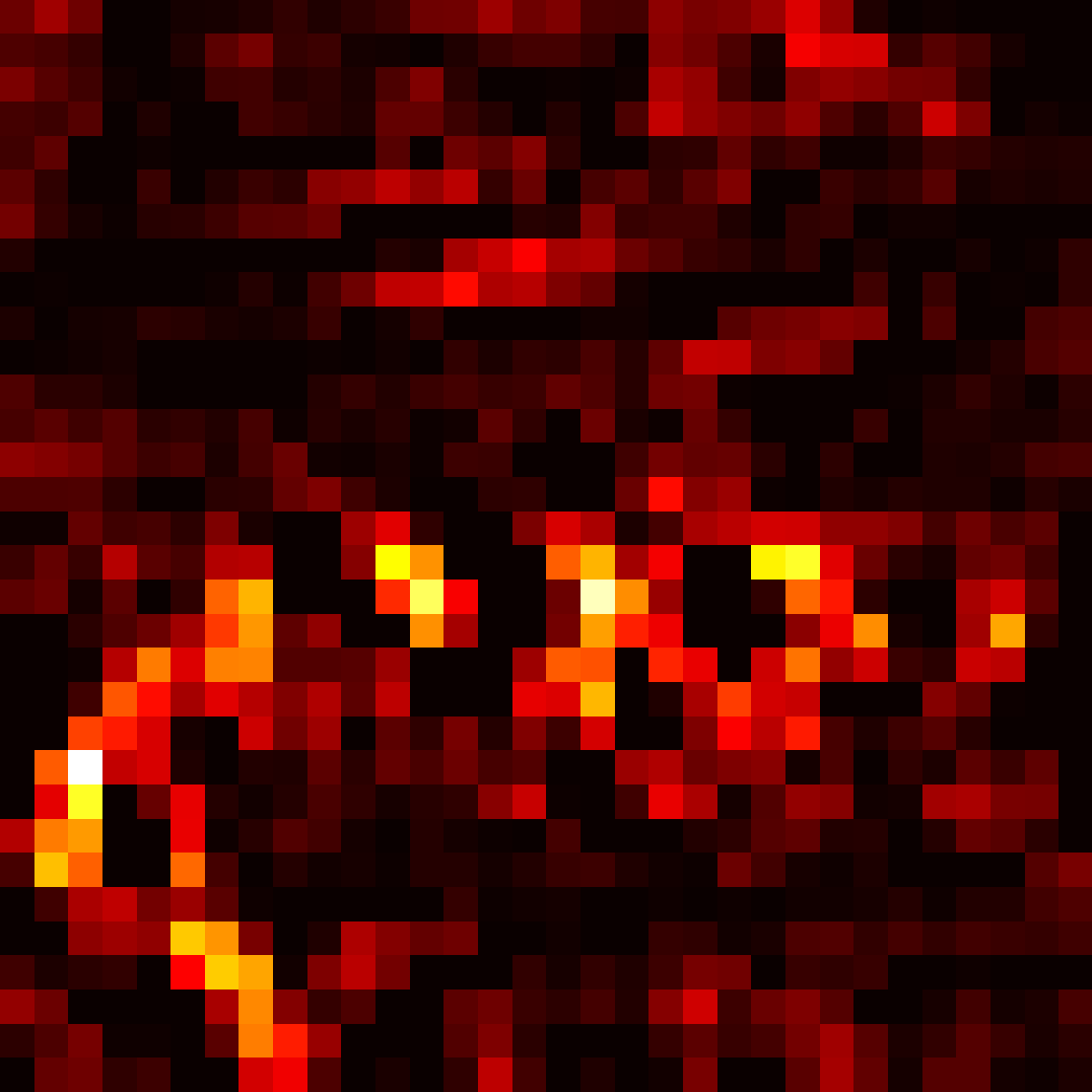} & 
  \includegraphics[scale=\scale]{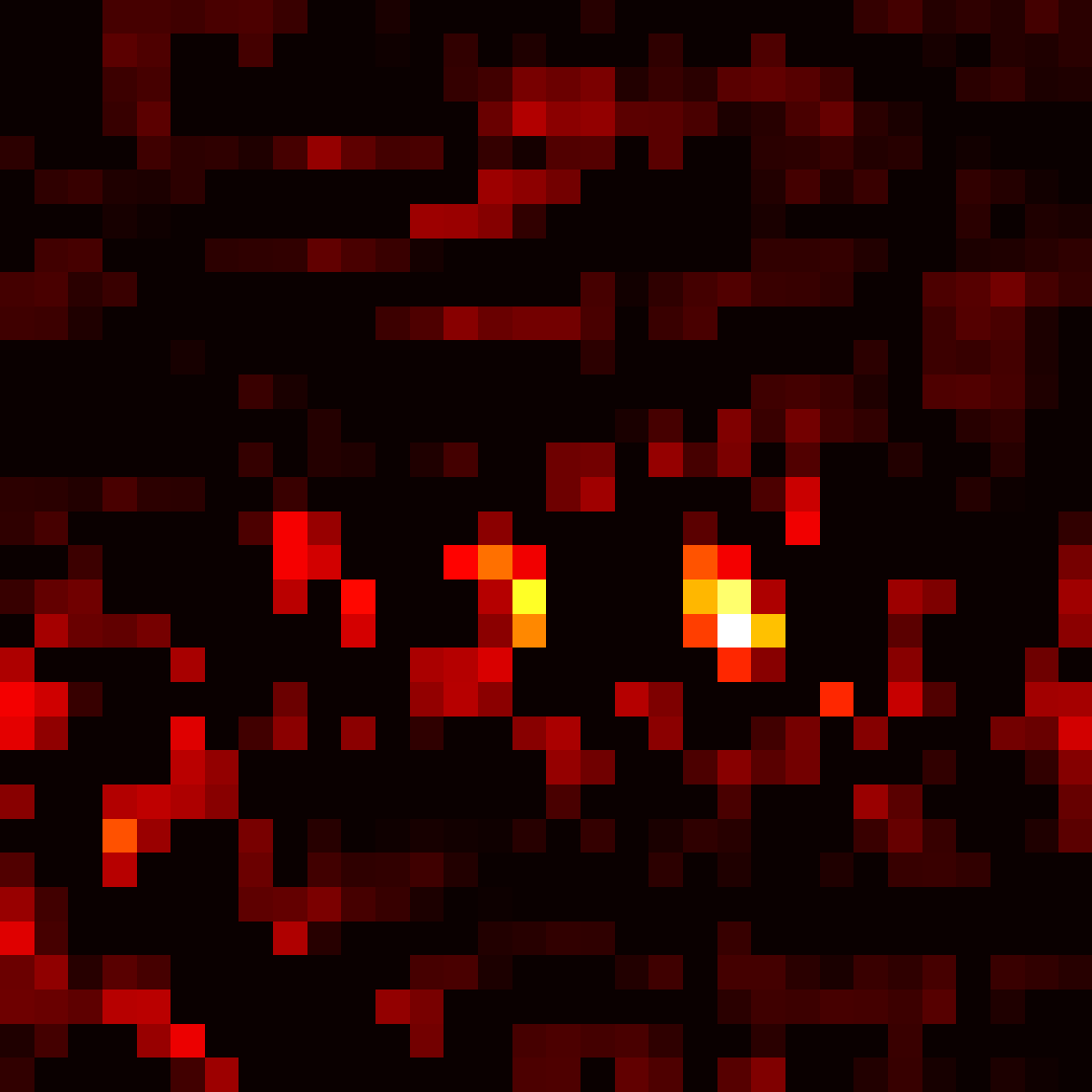} & 
  \includegraphics[scale=\scale]{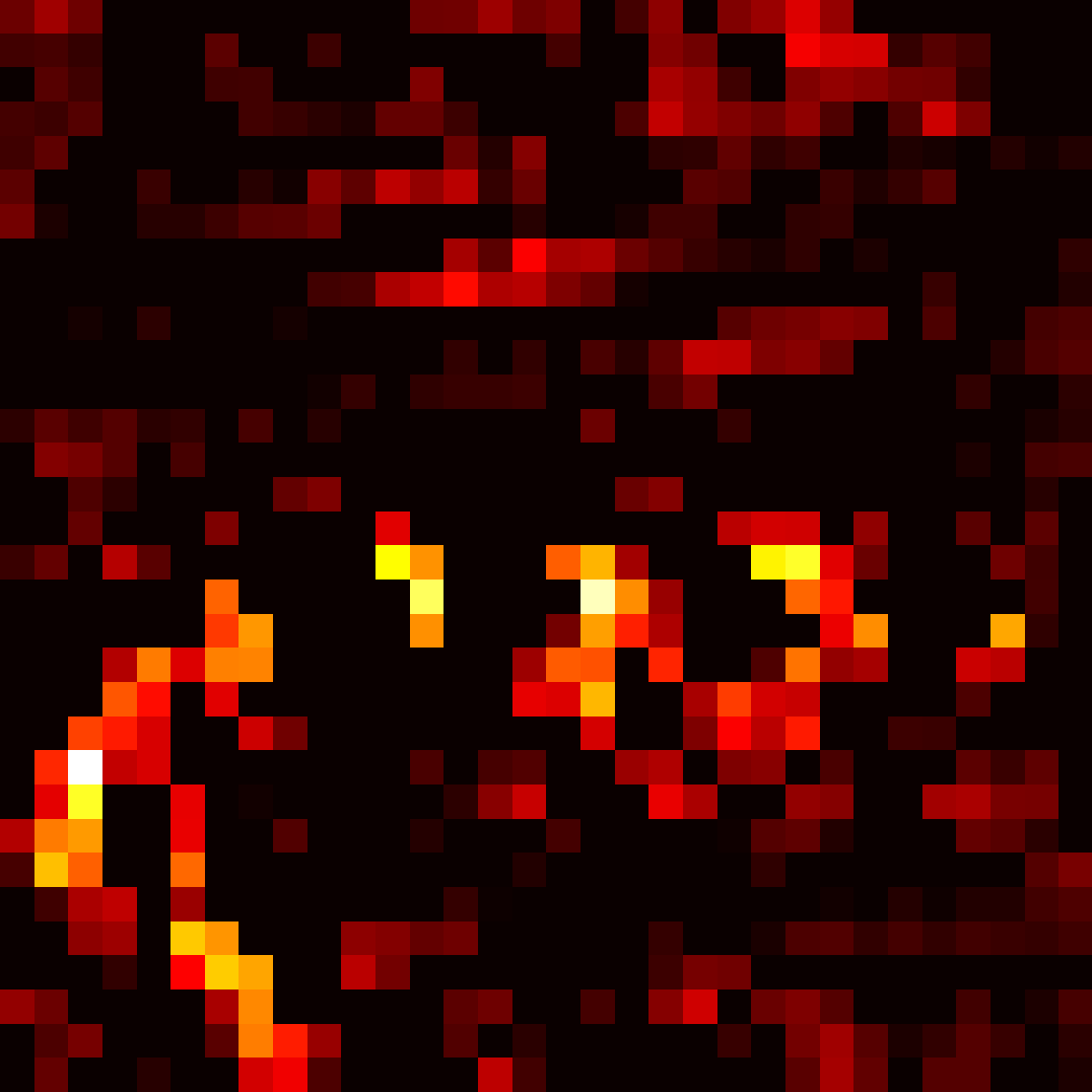} \\
  
  \includegraphics[scale=\scale]{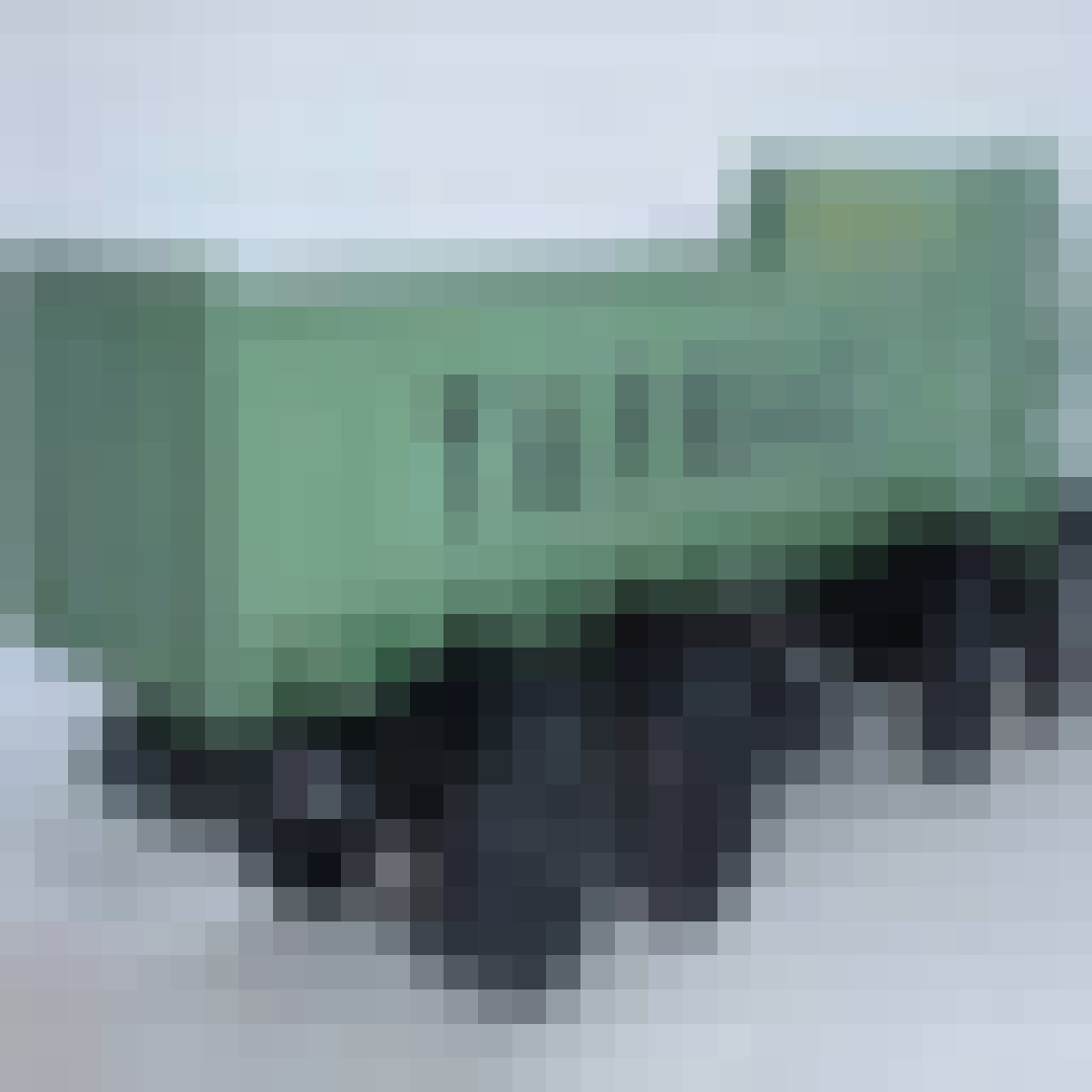} &
  \includegraphics[scale=\scale]{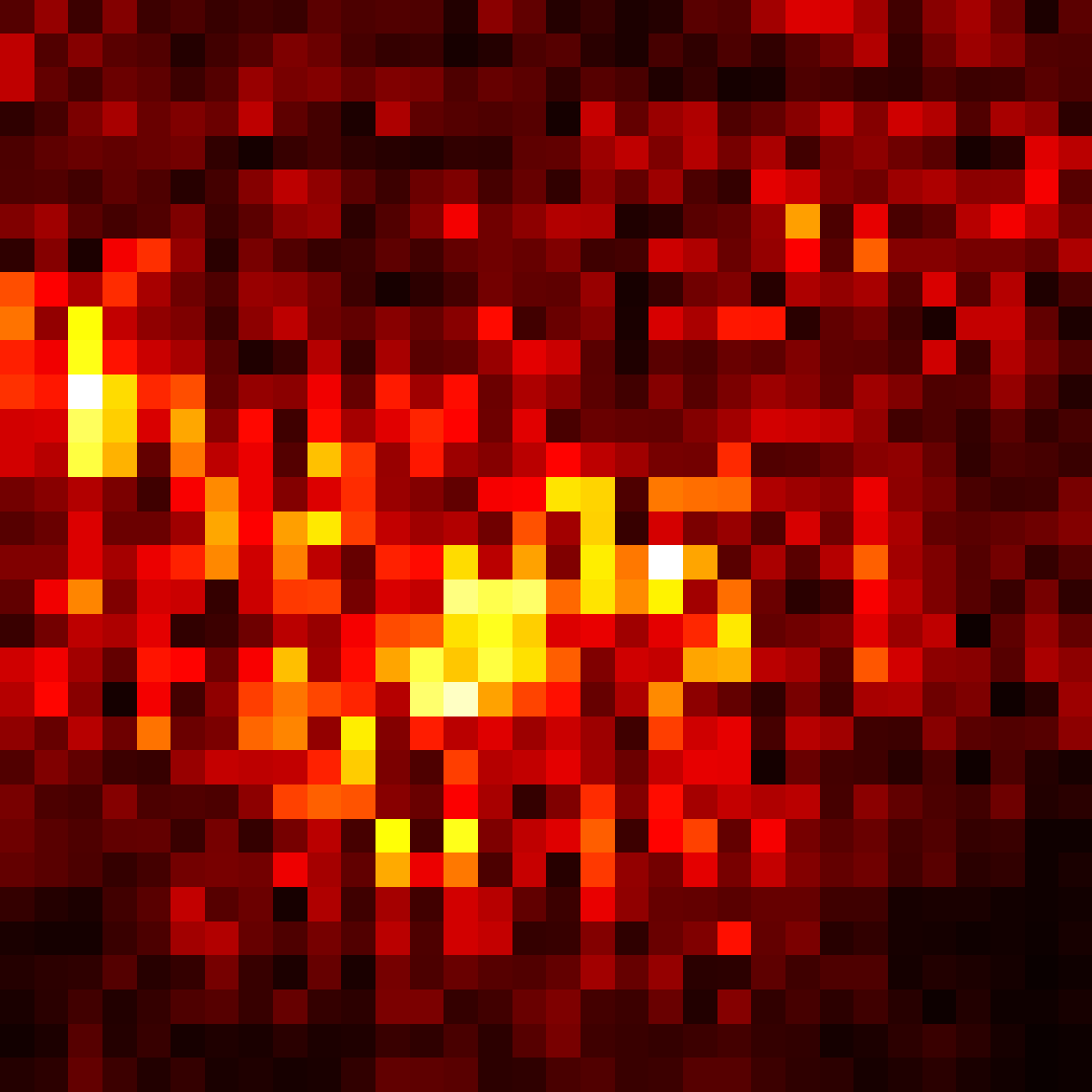} & 
  \includegraphics[scale=\scale]{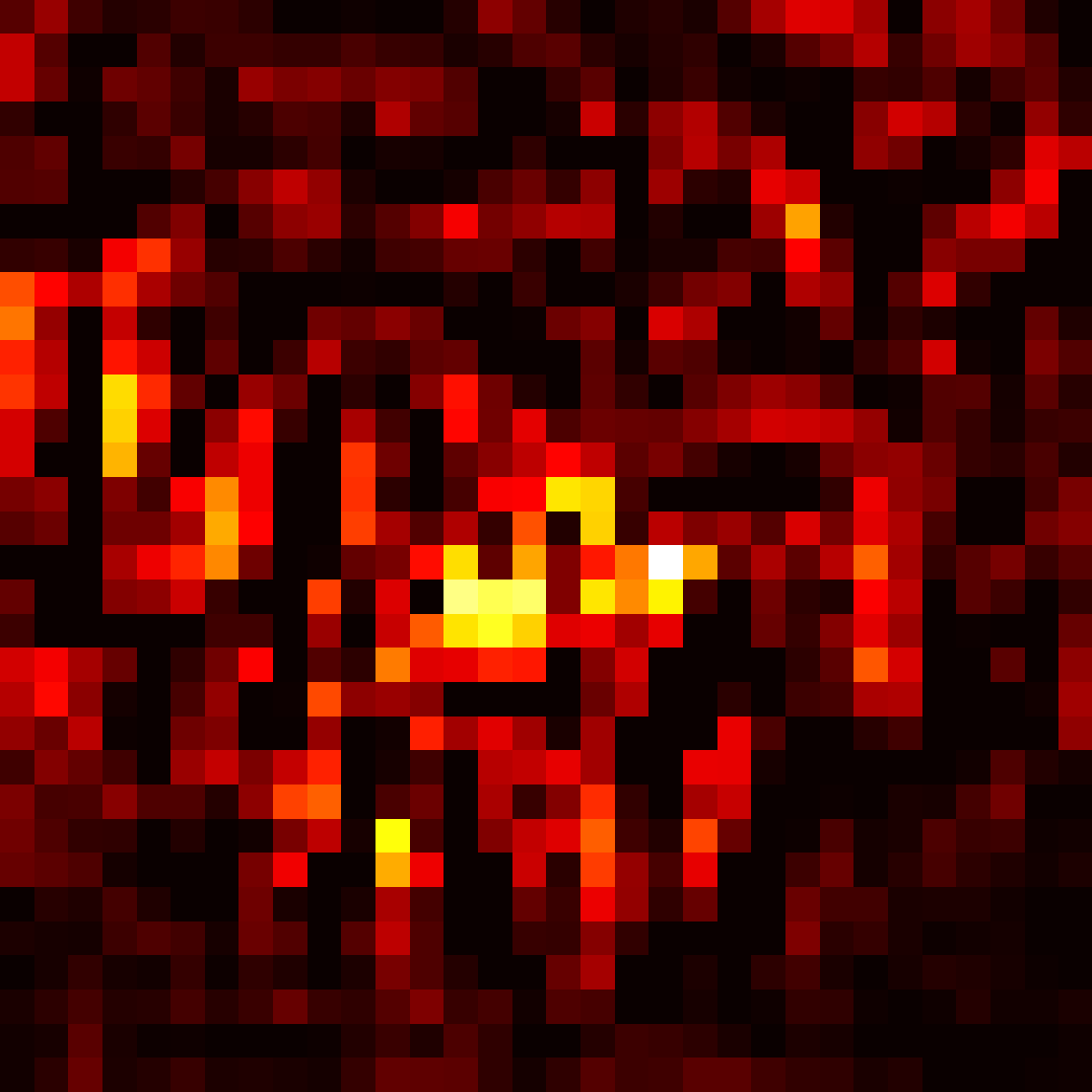} & 
  \includegraphics[scale=\scale]{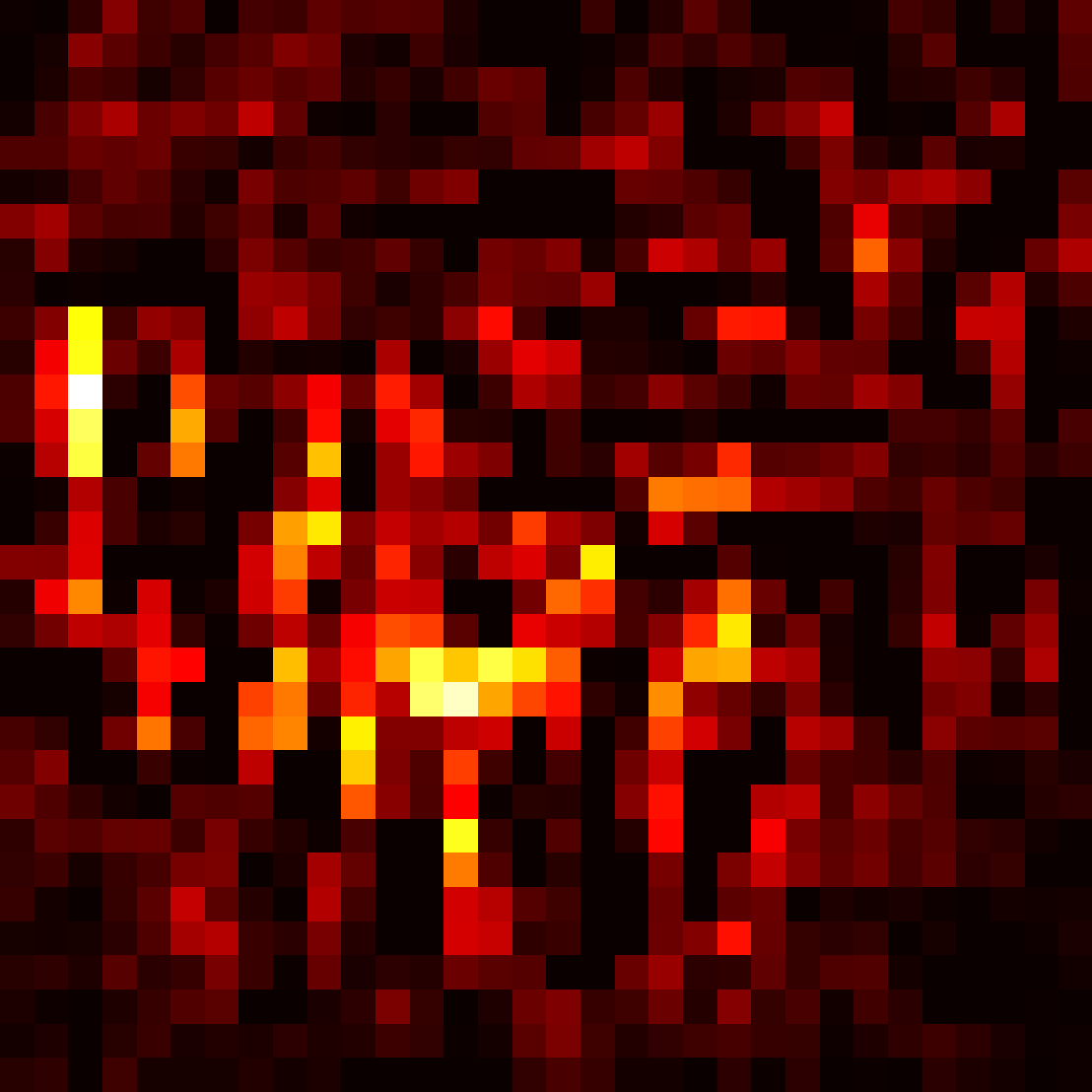} & 
  \includegraphics[scale=\scale]{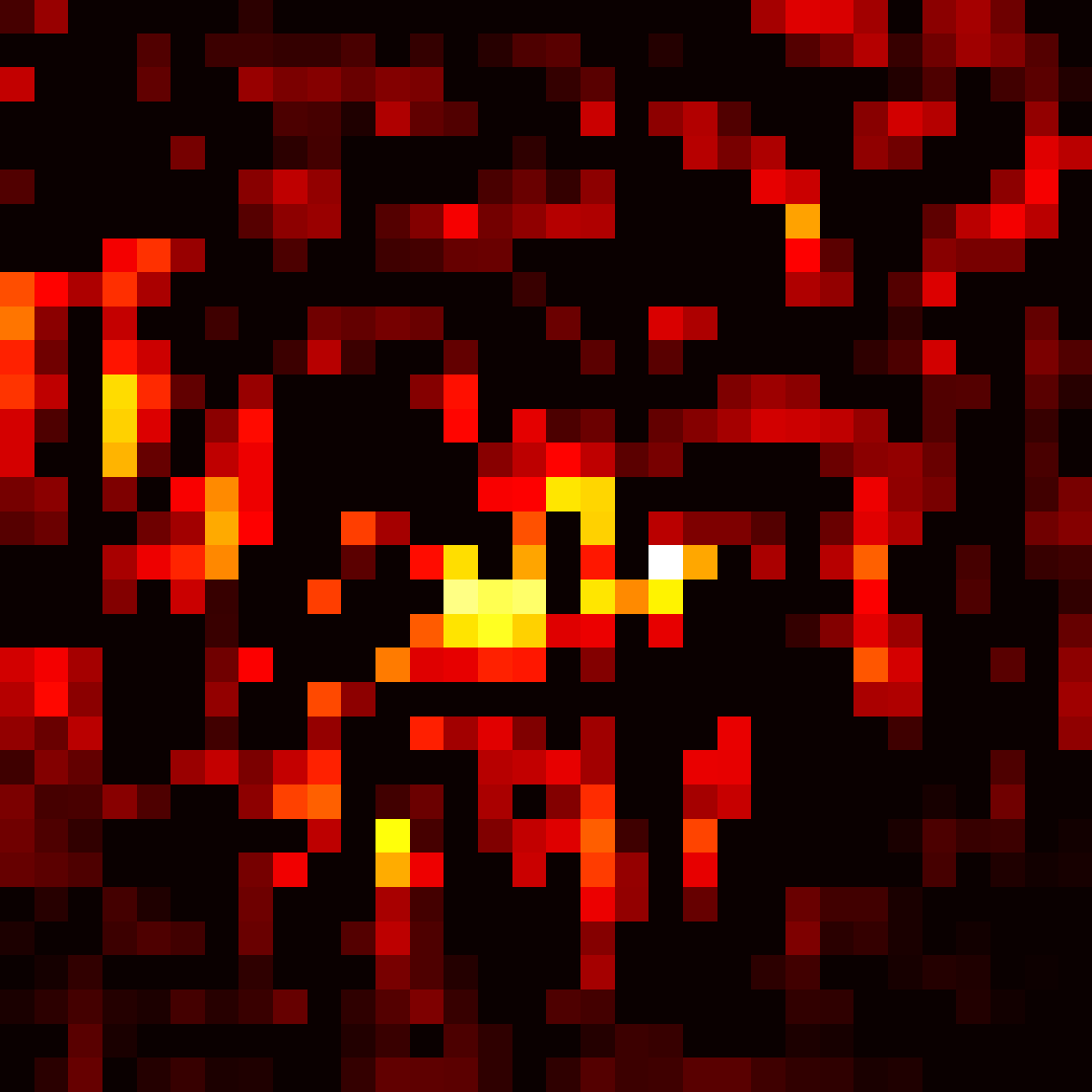} & 
  \includegraphics[scale=\scale]{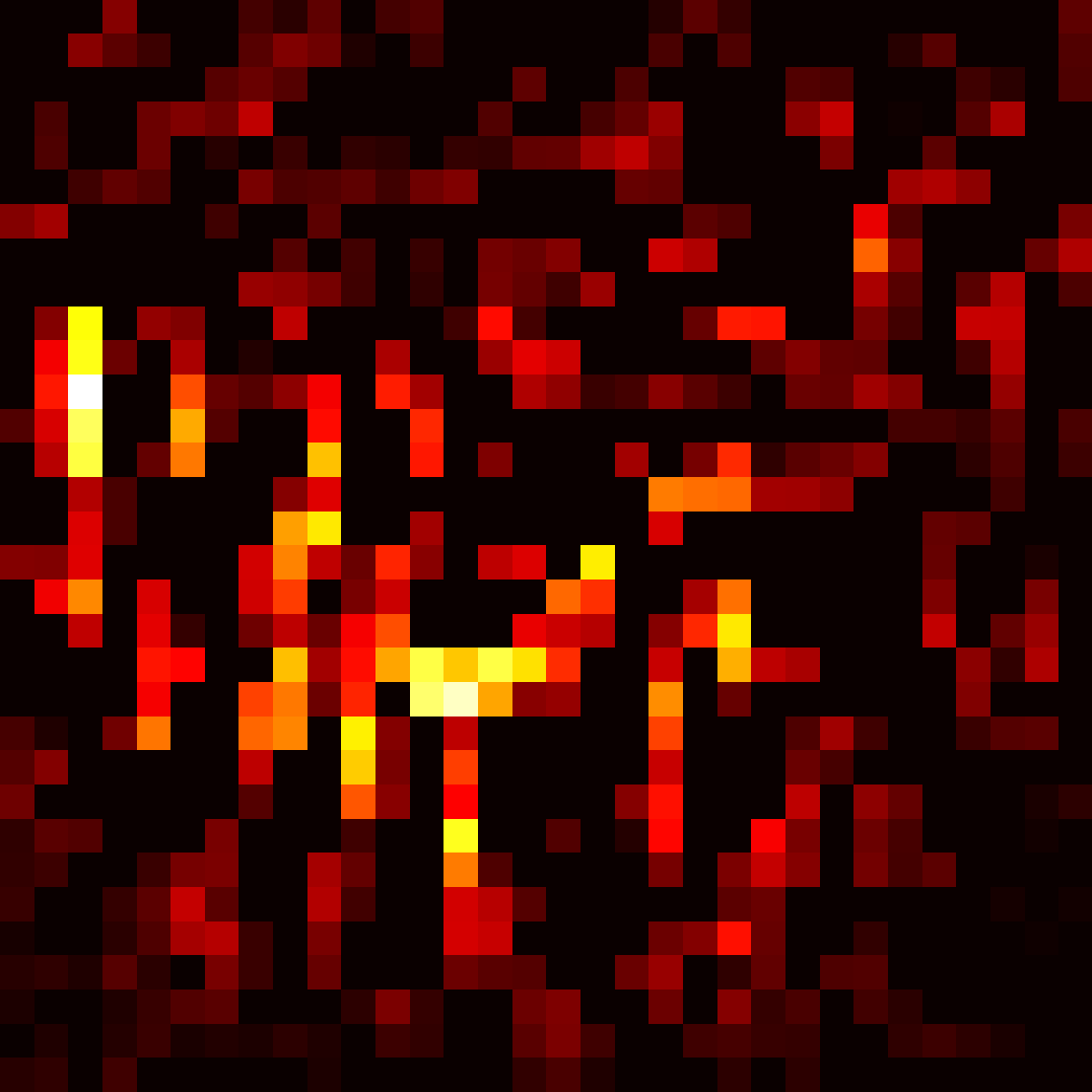} \\
  \end{tabular}
  \caption{Comparison saliency maps CNN CIFAR-10.}
\end{figure}

\begin{figure}[H]
  \centering
  \footnotesize
  \newcommand{\scale}{0.20}
  \setlength{\tabcolsep}{2pt}
  \begin{tabular}{cccccc}
  Image & Original & Positive & Negative & Active & Inactive \\
  
  \includegraphics[scale=\scale]{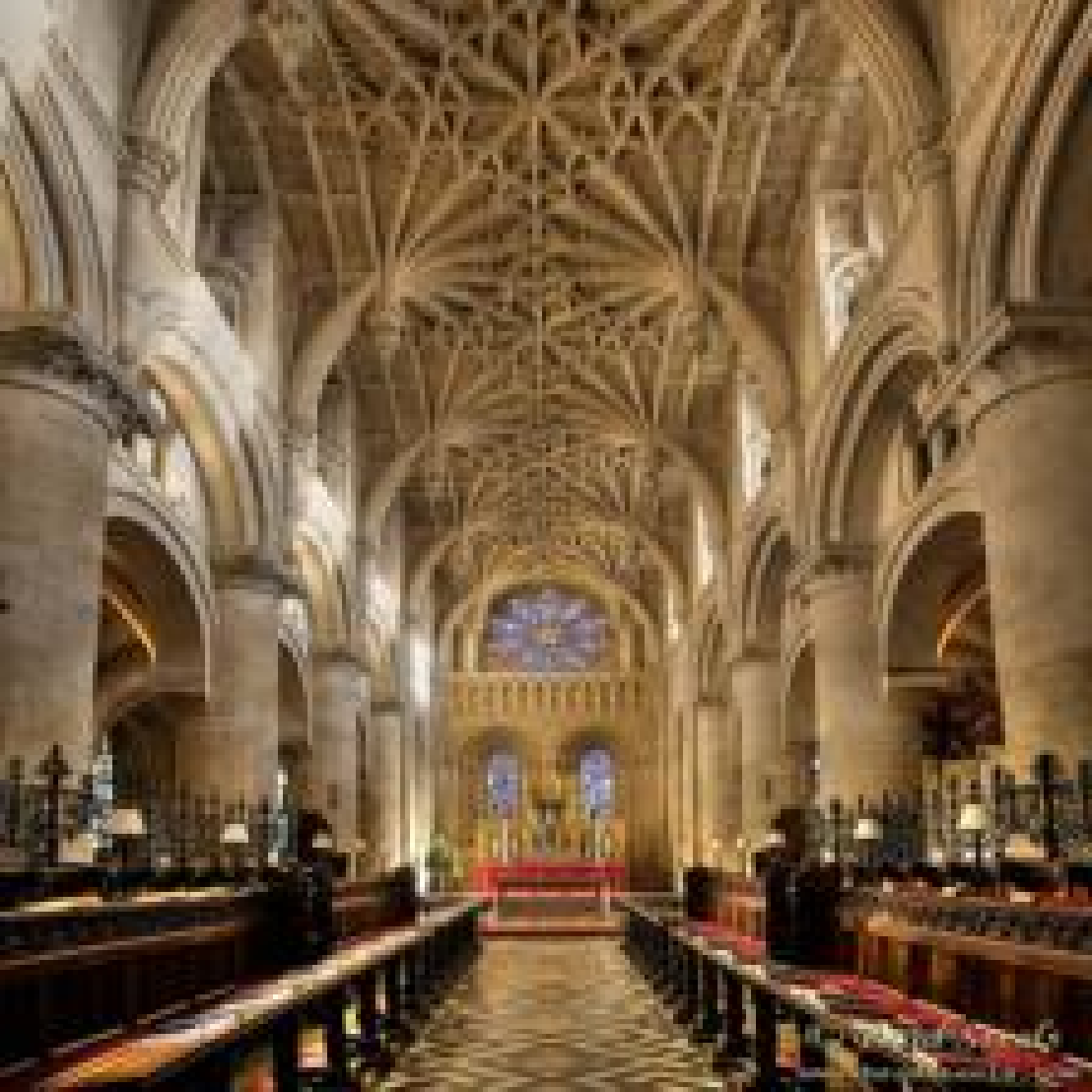} &
  \includegraphics[scale=\scale]{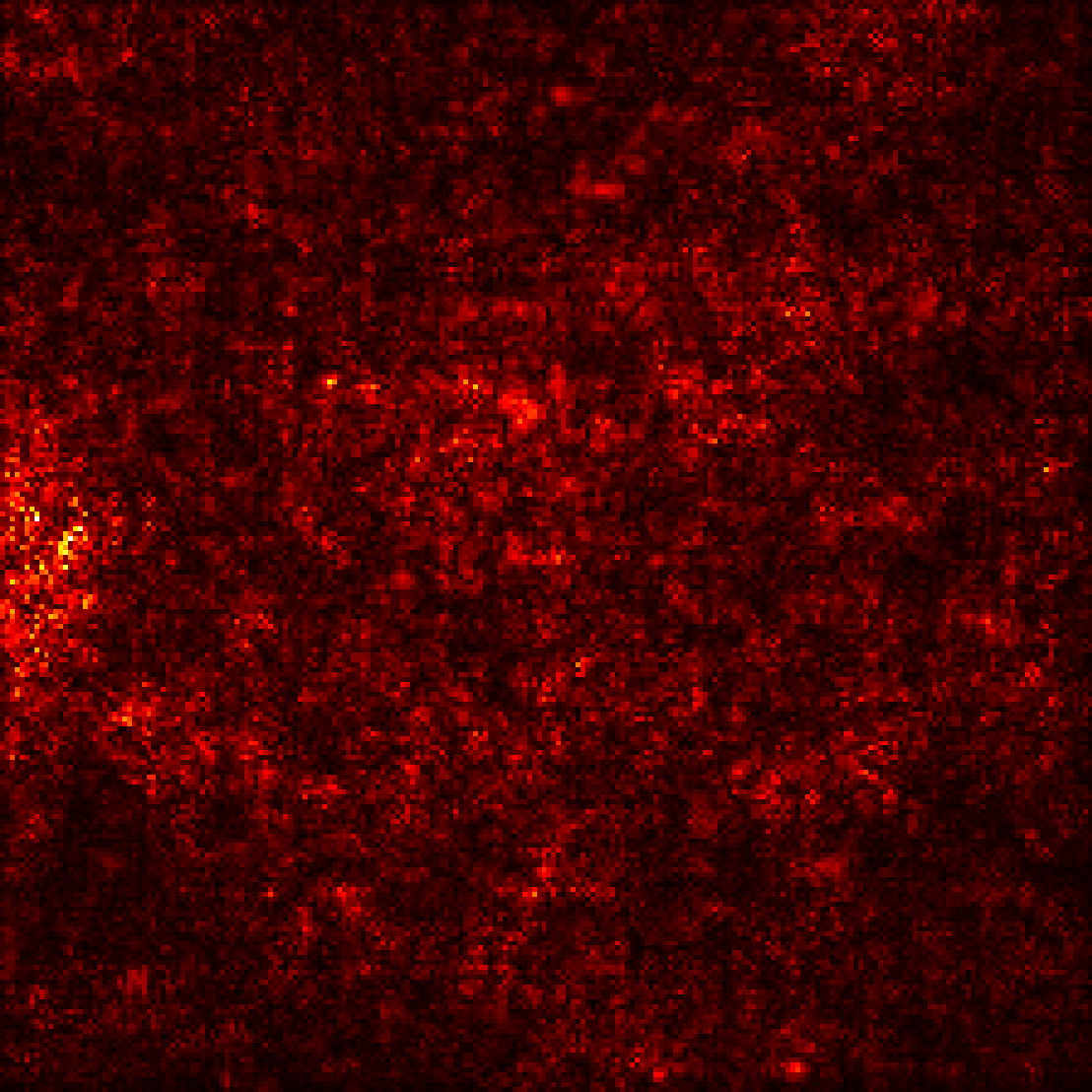} & 
  \includegraphics[scale=\scale]{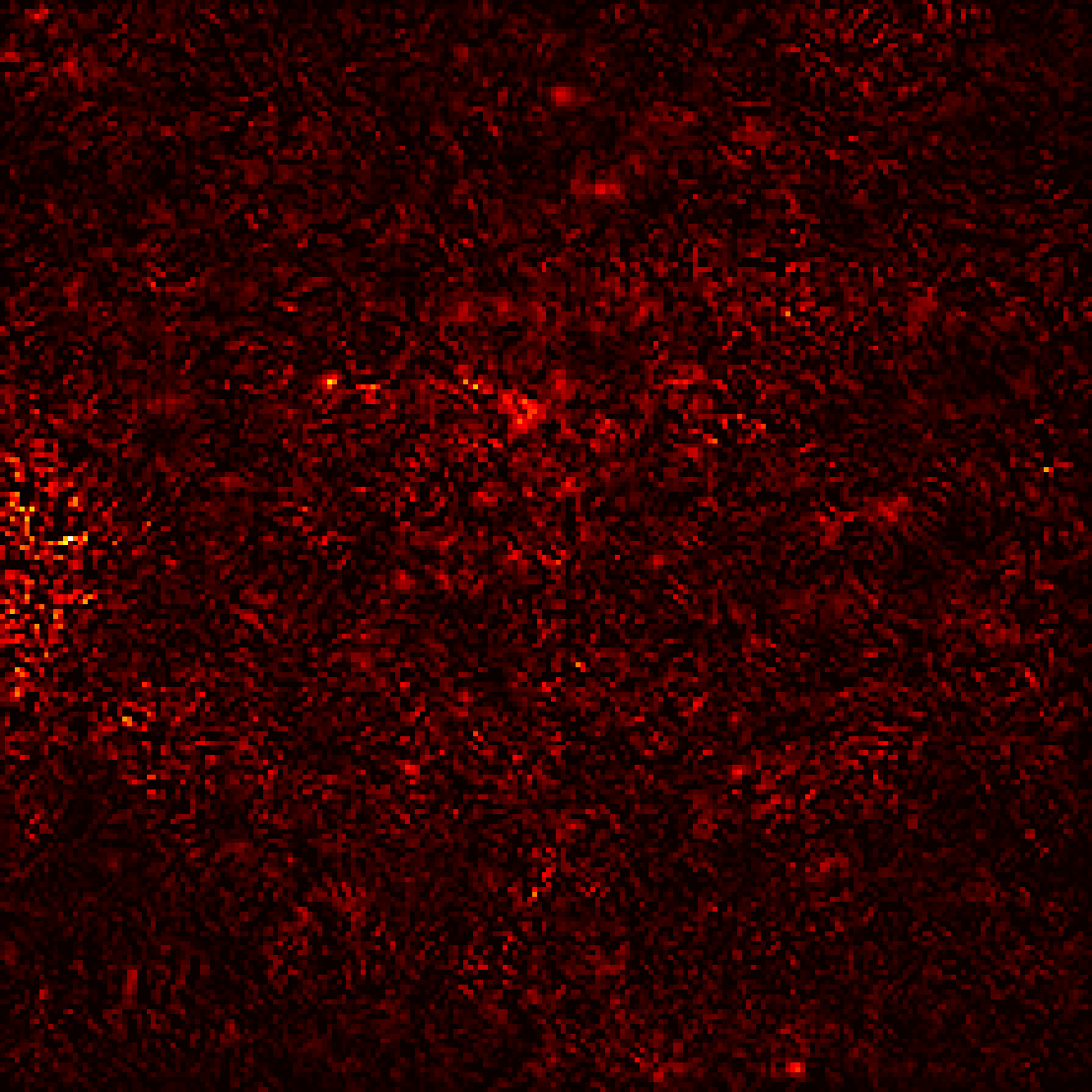} & 
  \includegraphics[scale=\scale]{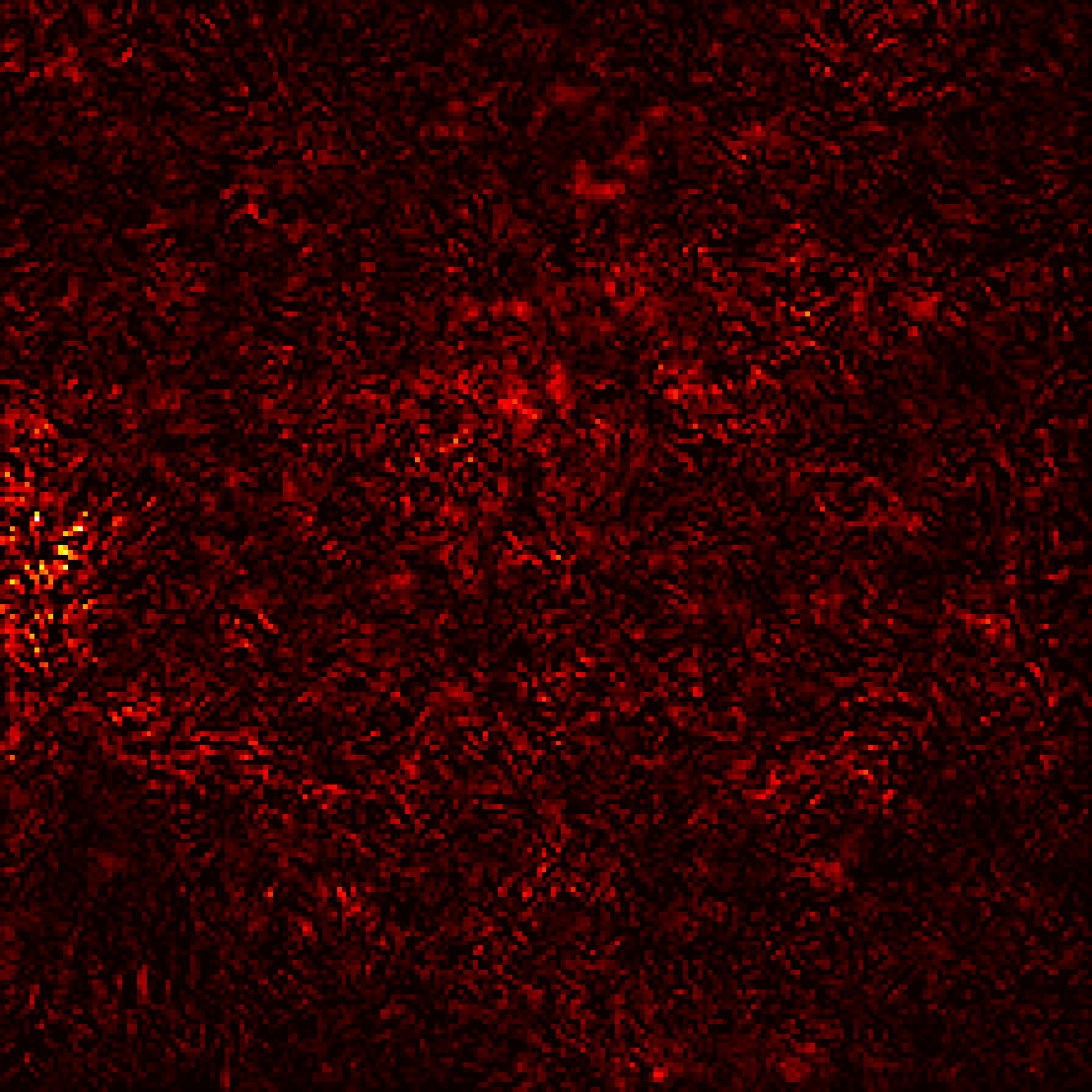} & 
  \includegraphics[scale=\scale]{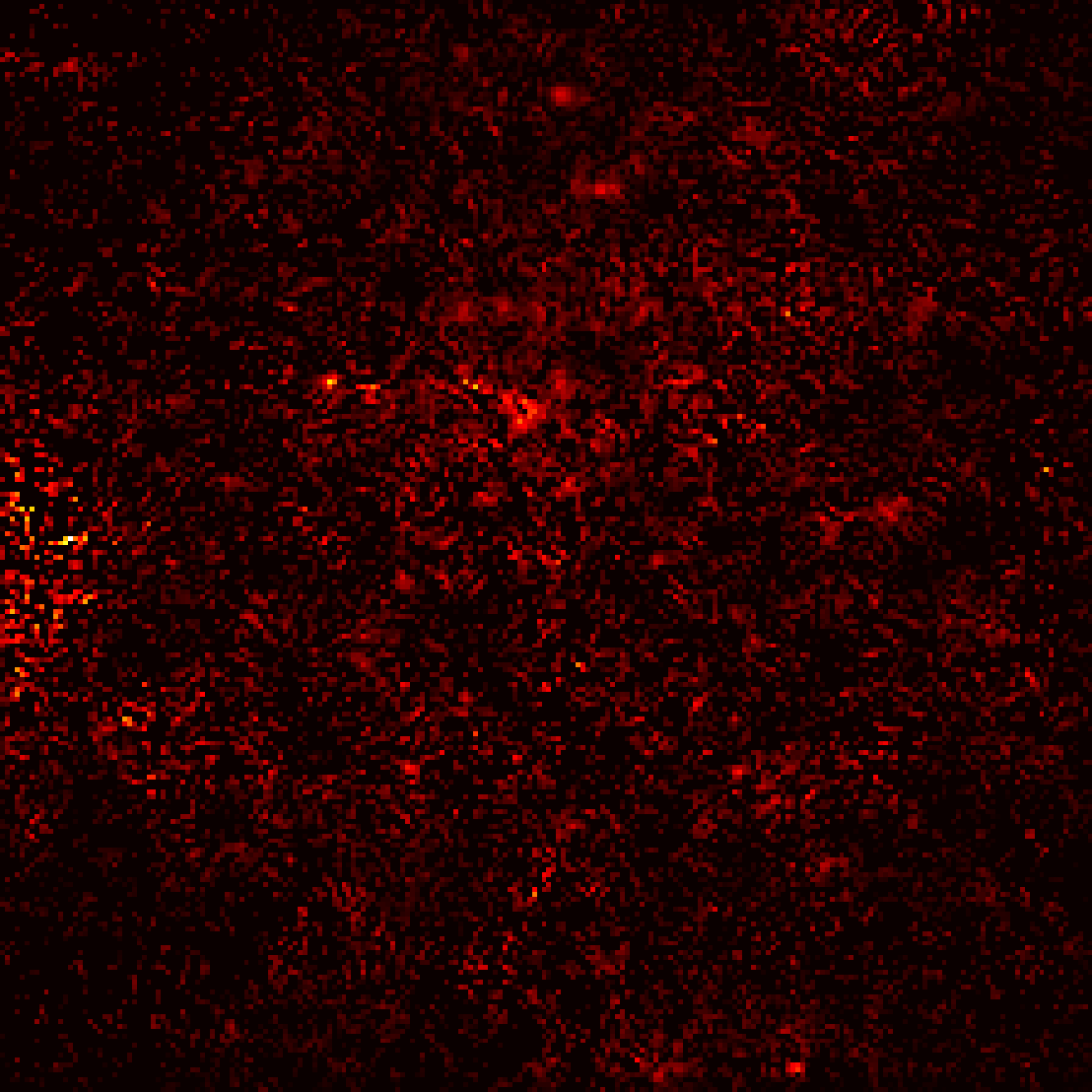} & 
  \includegraphics[scale=\scale]{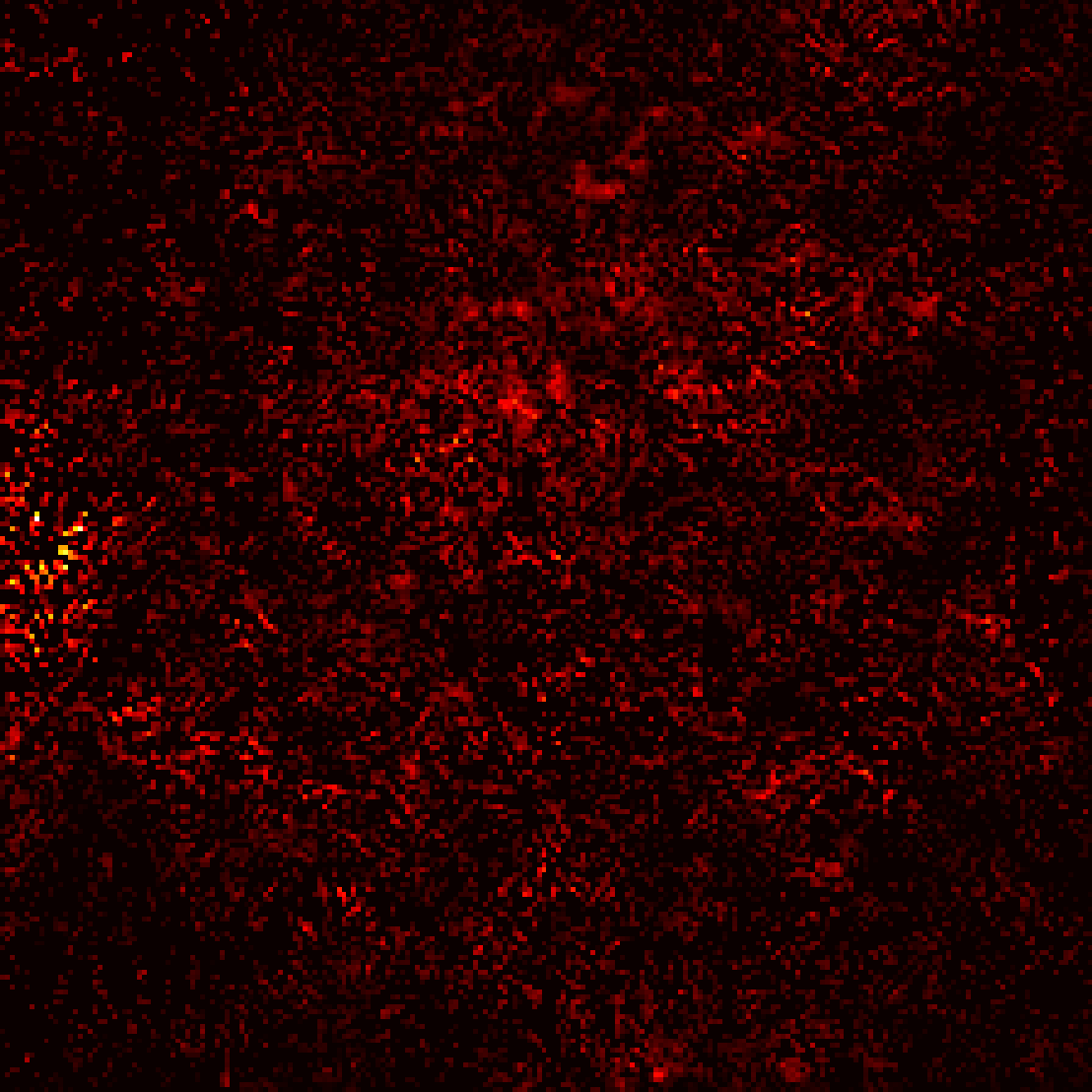} \\
  
  \includegraphics[scale=\scale]{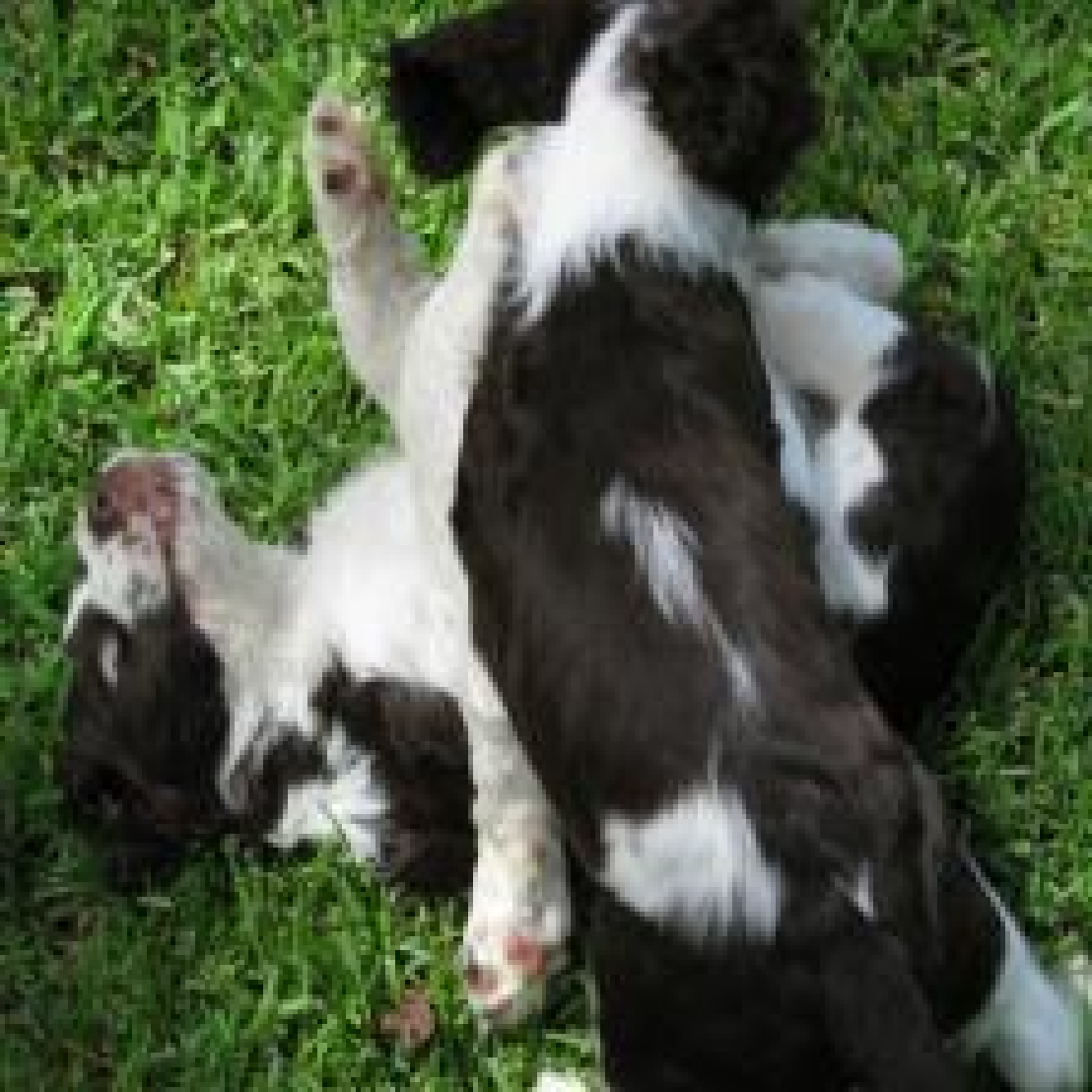} &
  \includegraphics[scale=\scale]{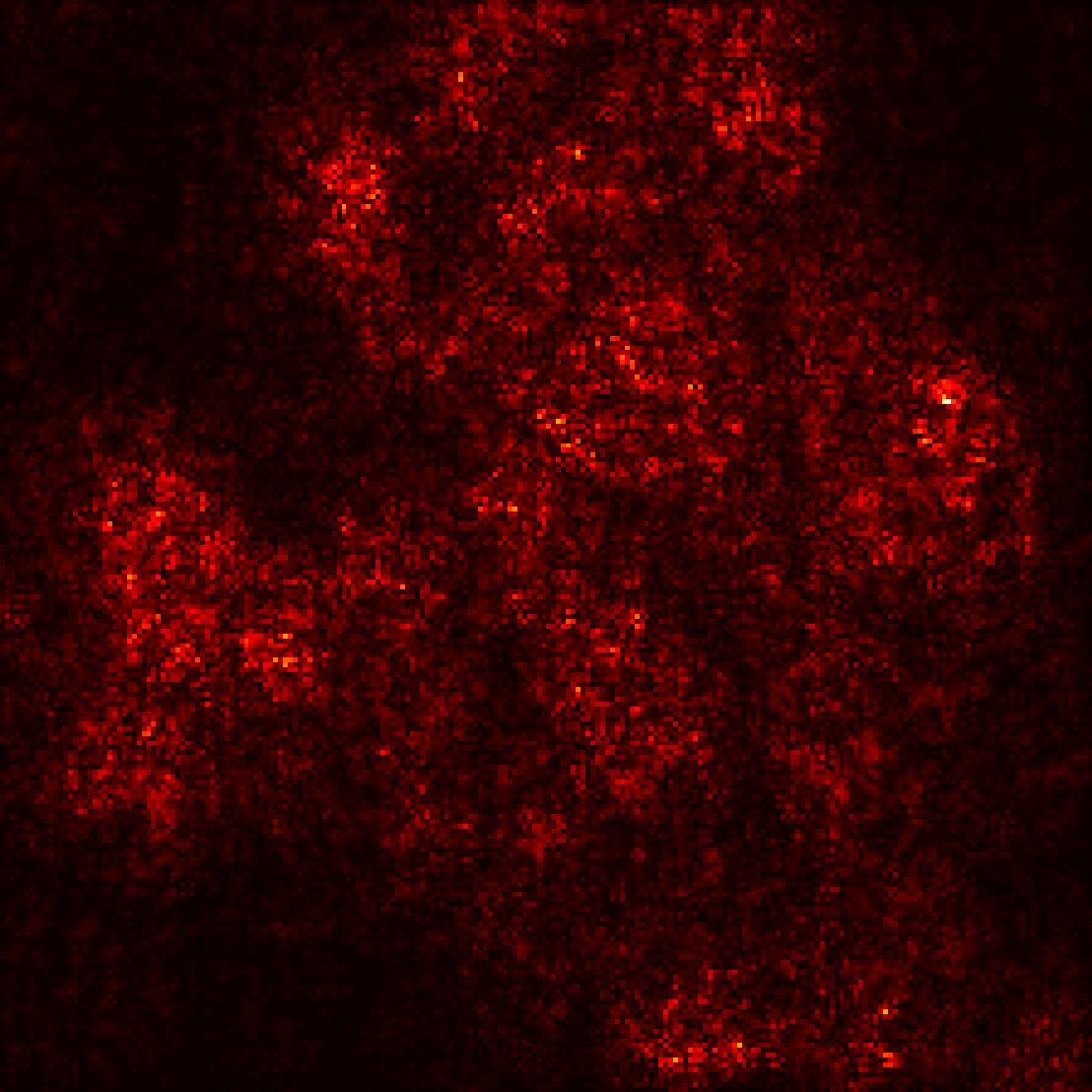} & 
  \includegraphics[scale=\scale]{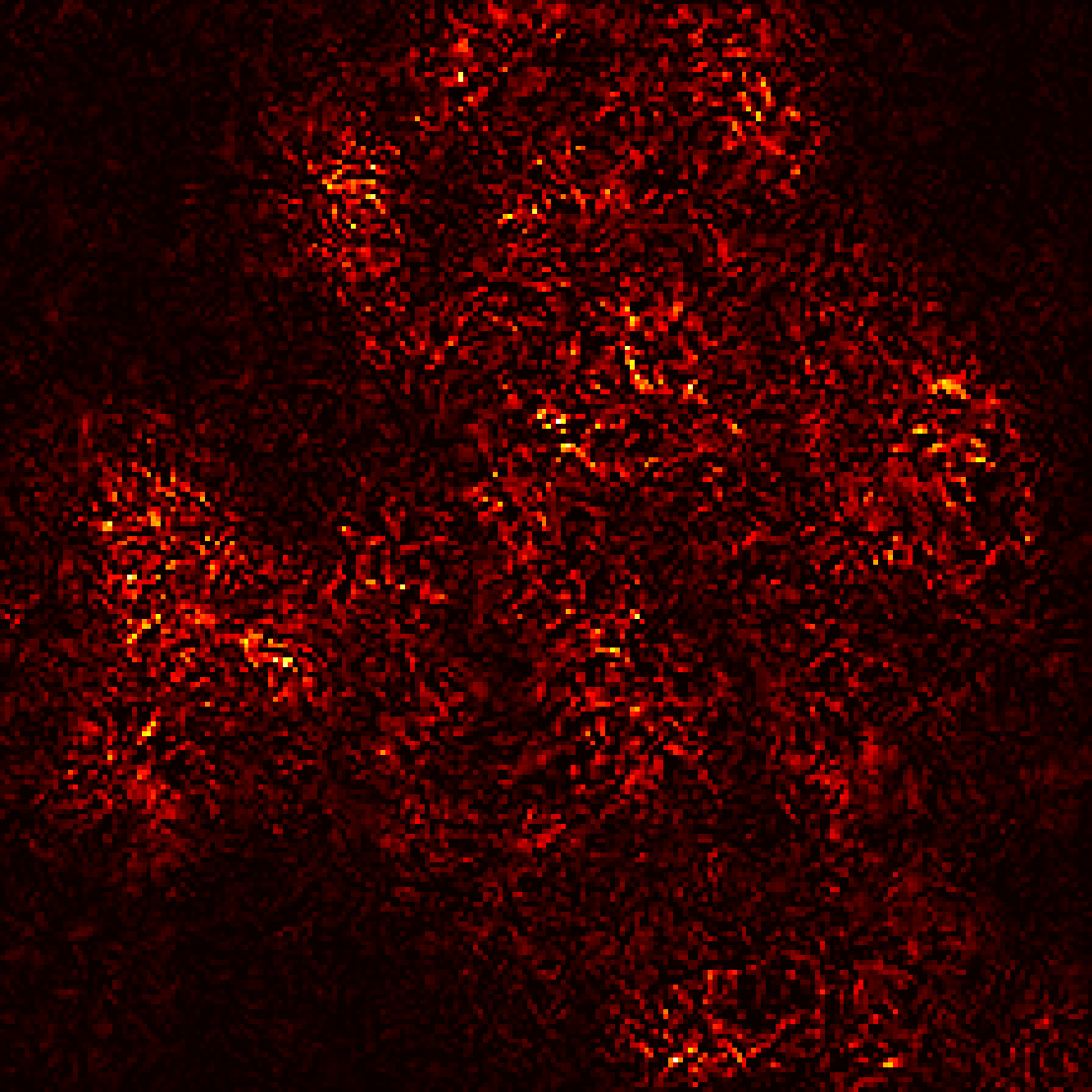} & 
  \includegraphics[scale=\scale]{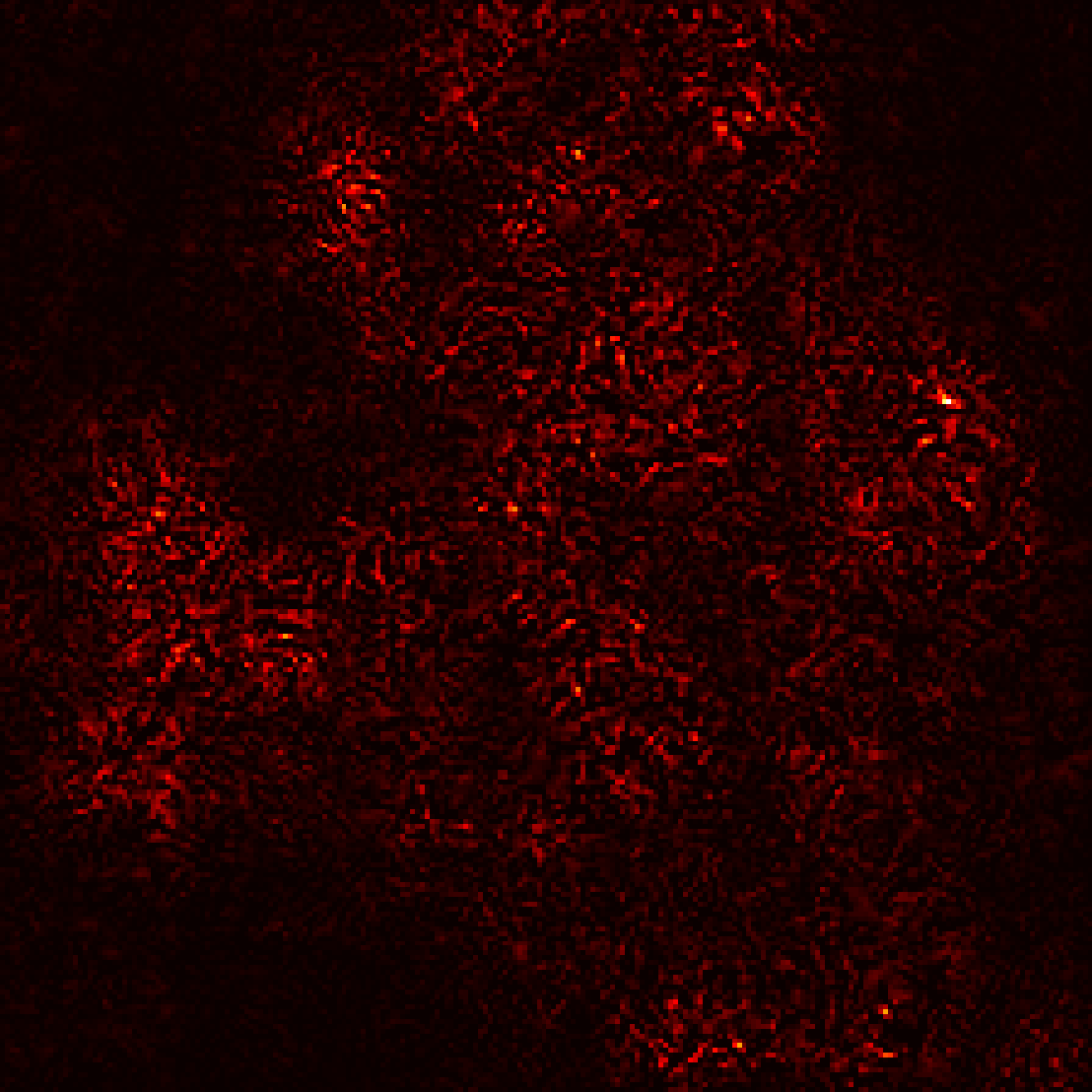} & 
  \includegraphics[scale=\scale]{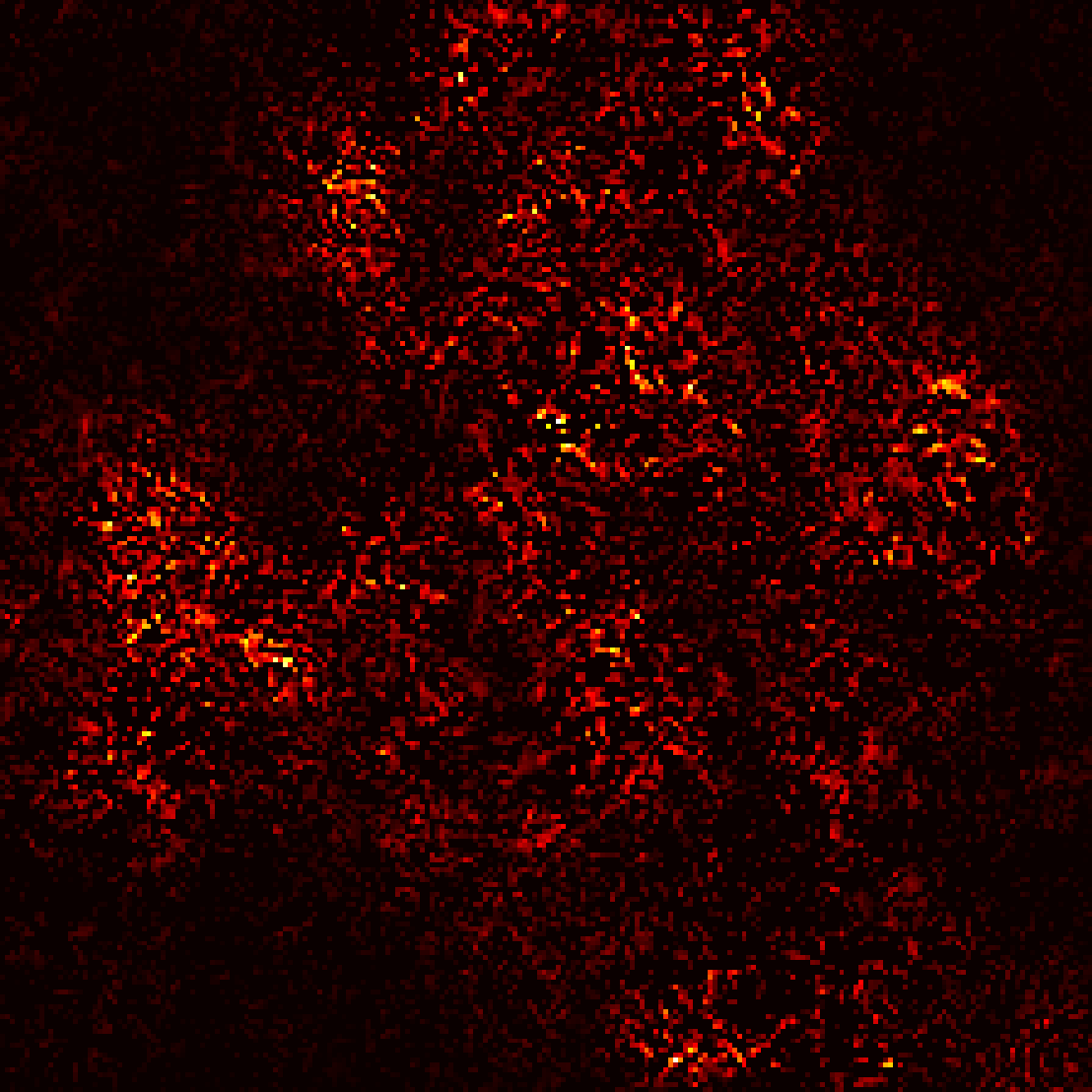} & 
  \includegraphics[scale=\scale]{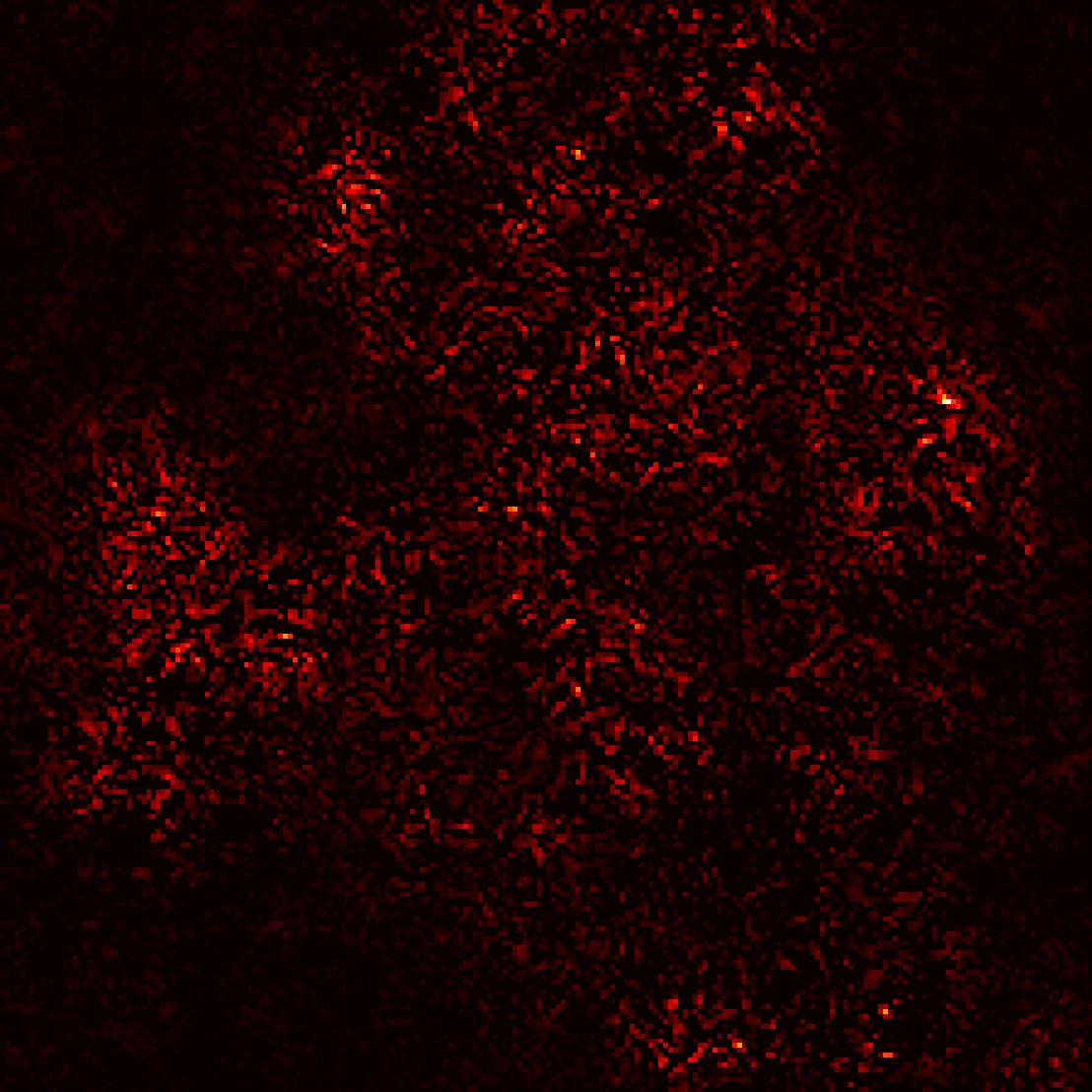} \\
  
  \includegraphics[scale=\scale]{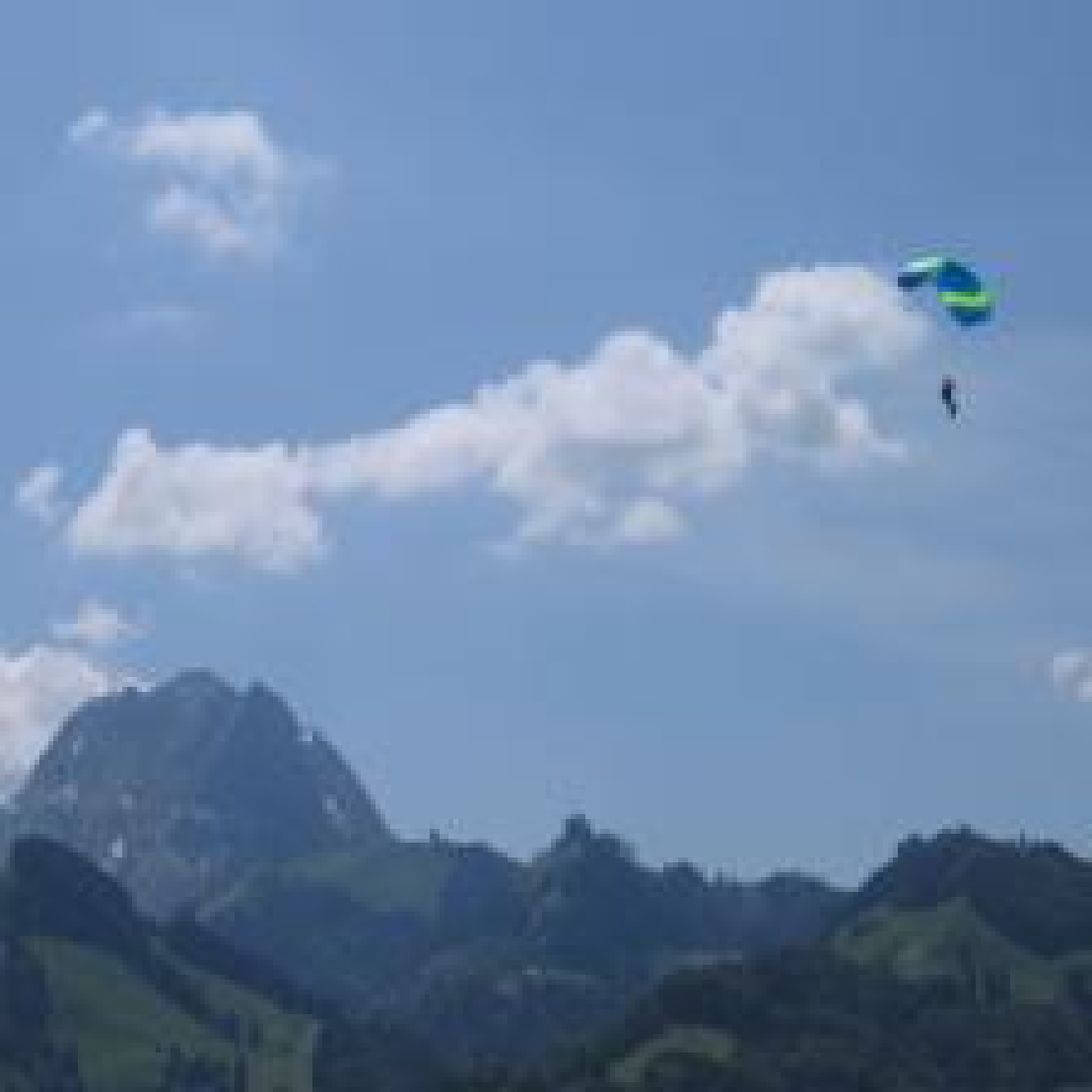} &
  \includegraphics[scale=\scale]{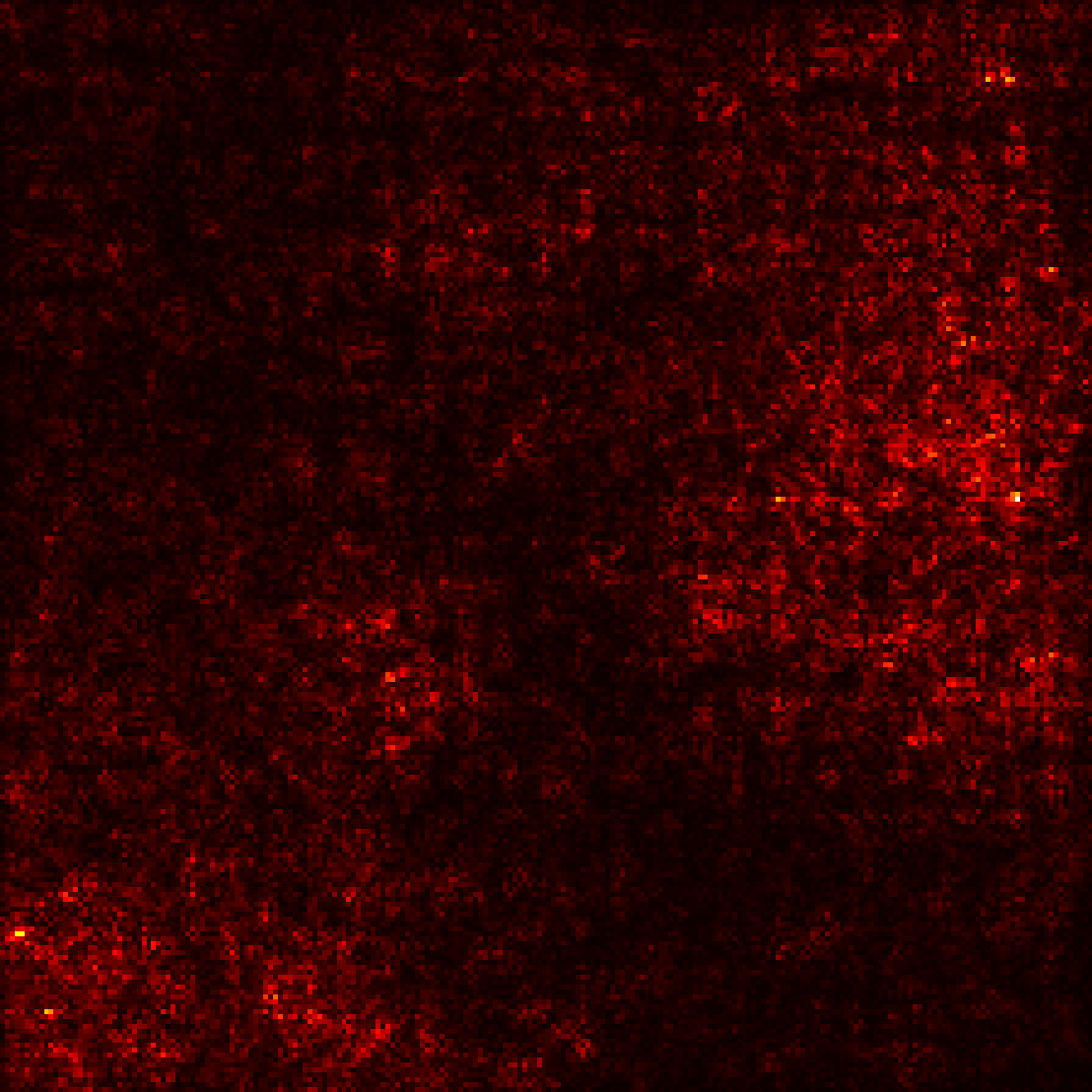} & 
  \includegraphics[scale=\scale]{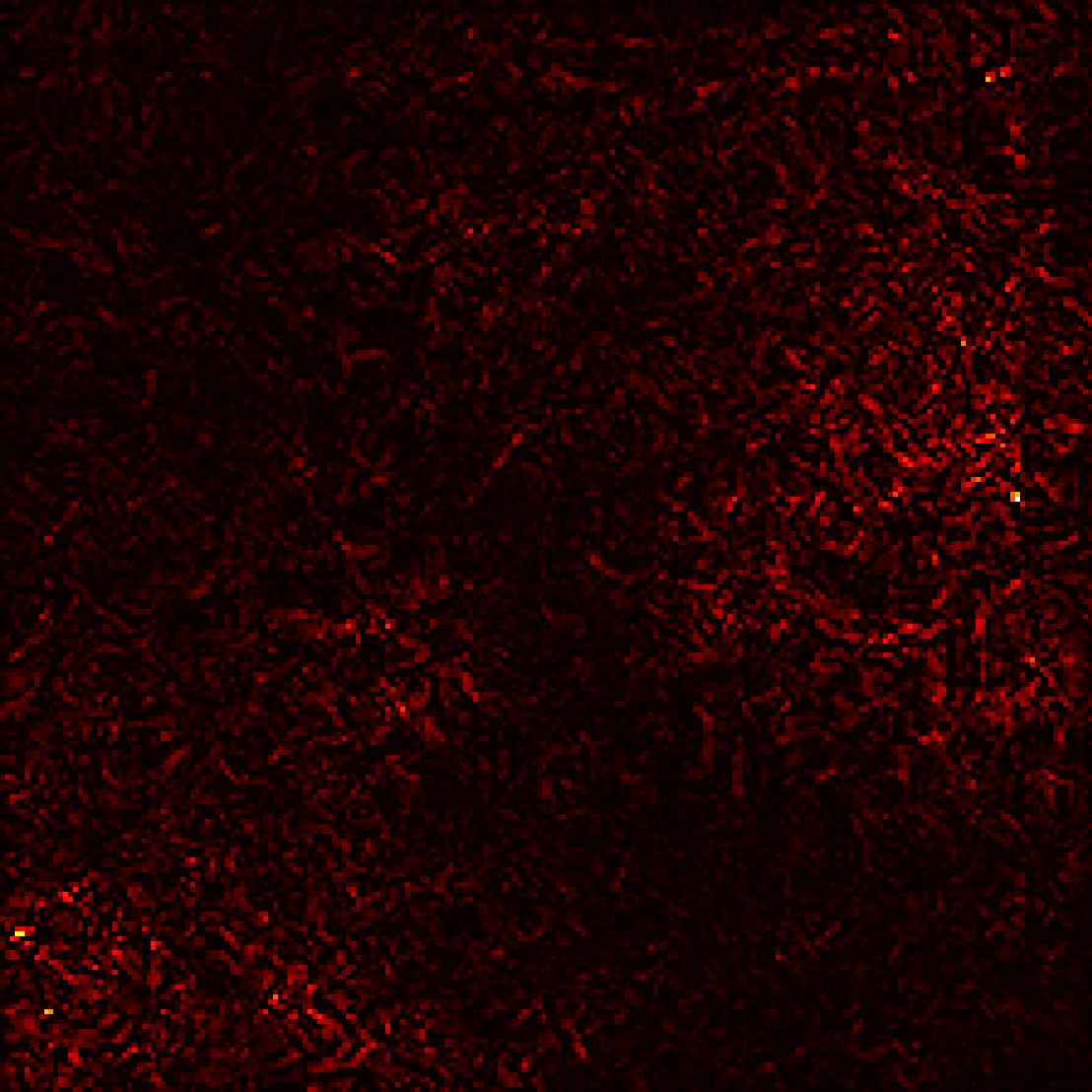} & 
  \includegraphics[scale=\scale]{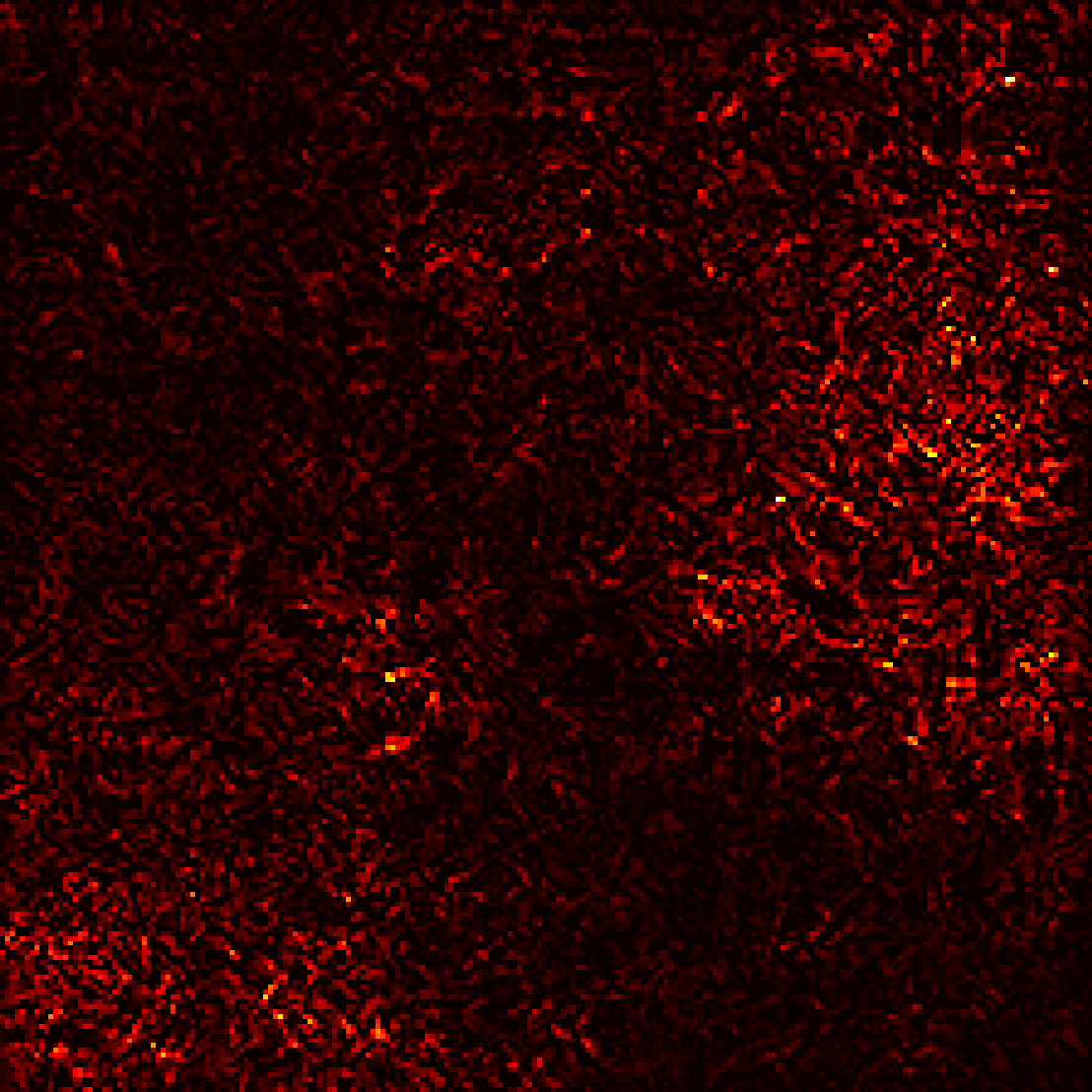} & 
  \includegraphics[scale=\scale]{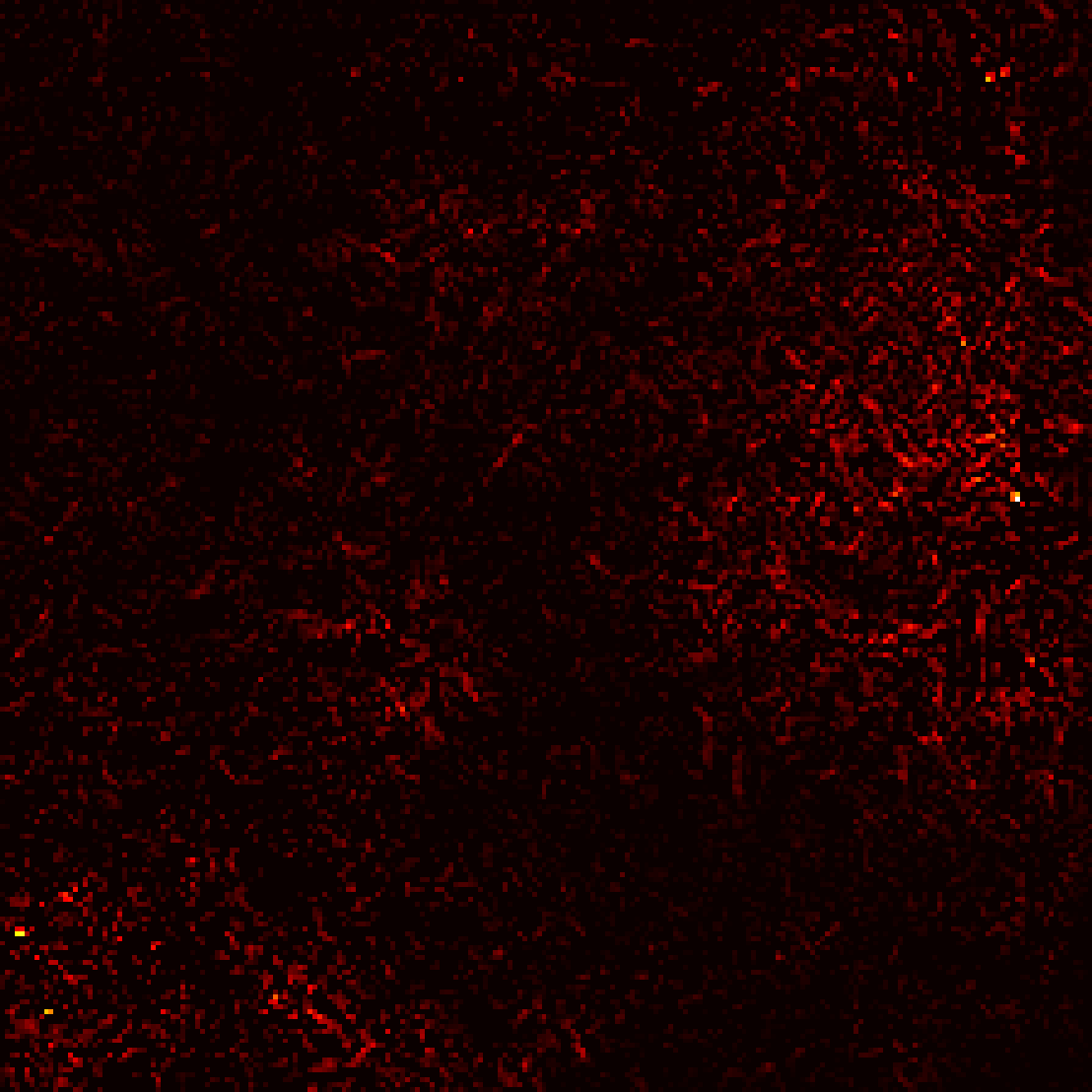} & 
  \includegraphics[scale=\scale]{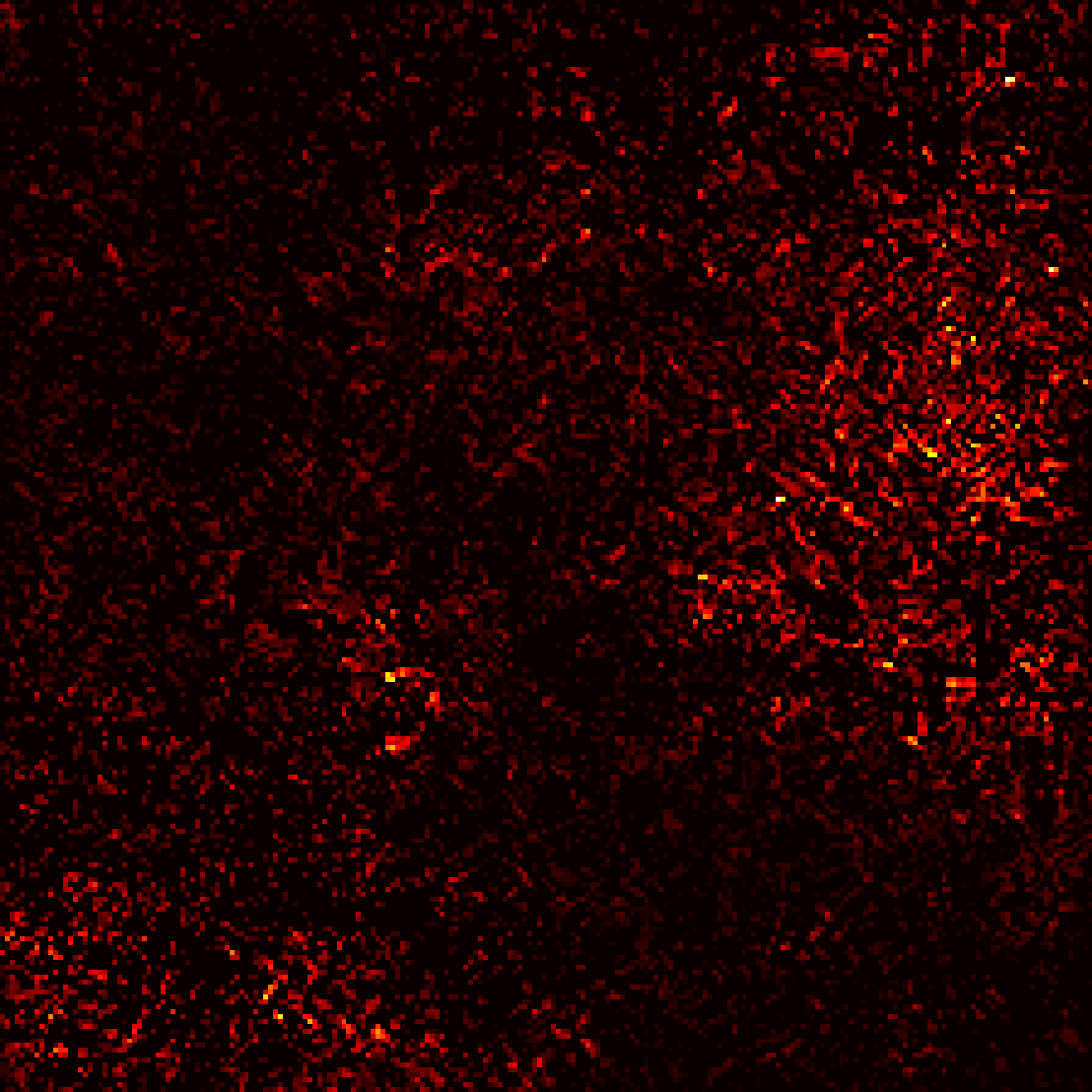} \\
  
  \includegraphics[scale=\scale]{visualizations/images/imagenette/examples/3.png} &
  \includegraphics[scale=\scale]{visualizations/images/imagenette/saliency_map/3.png} & 
  \includegraphics[scale=\scale]{visualizations/images/imagenette/positive_saliency_map/3.png} & 
  \includegraphics[scale=\scale]{visualizations/images/imagenette/negative_saliency_map/3.png} & 
  \includegraphics[scale=\scale]{visualizations/images/imagenette/active_saliency_map/3.png} & 
  \includegraphics[scale=\scale]{visualizations/images/imagenette/inactive_saliency_map/3.png} \\
  
  \includegraphics[scale=\scale]{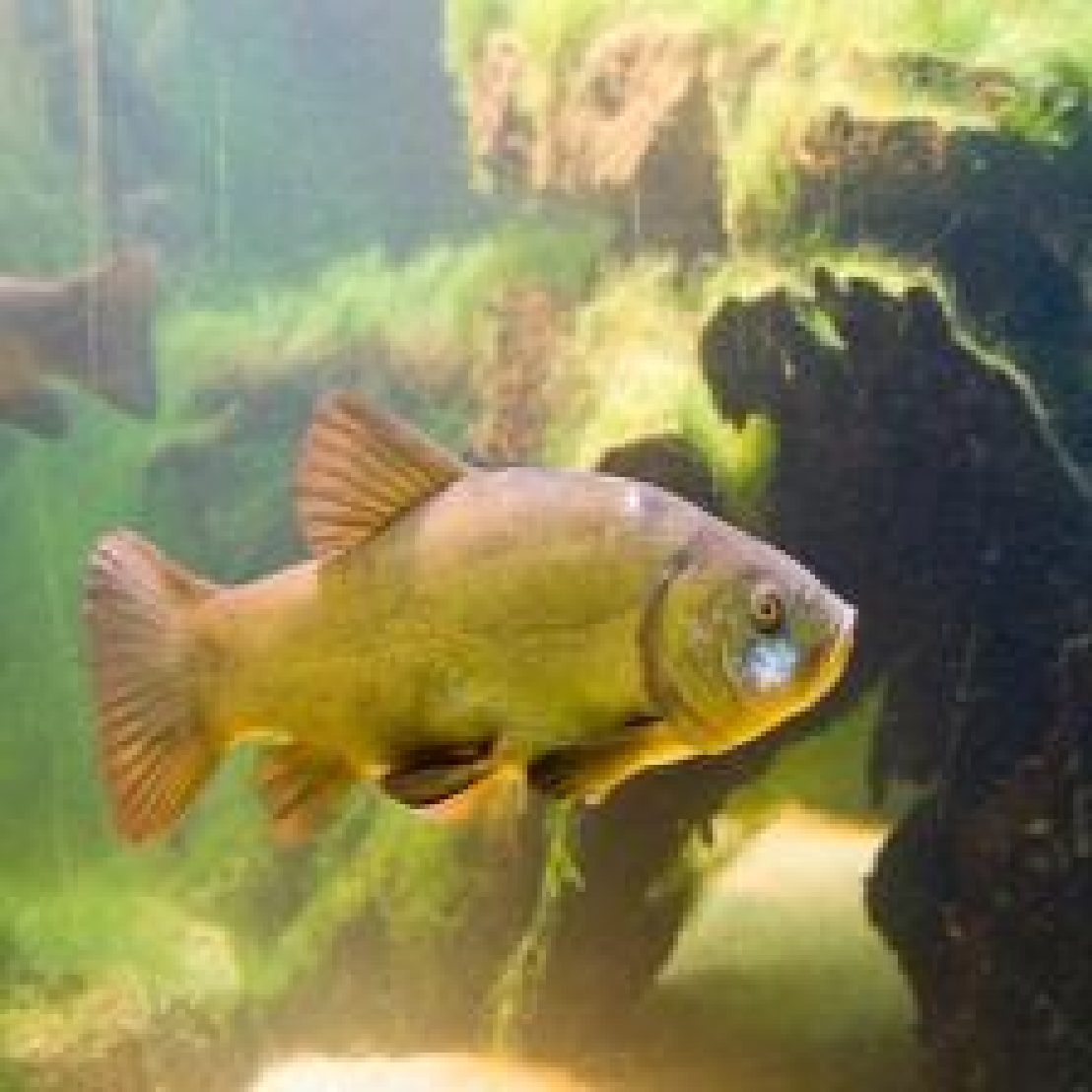} &
  \includegraphics[scale=\scale]{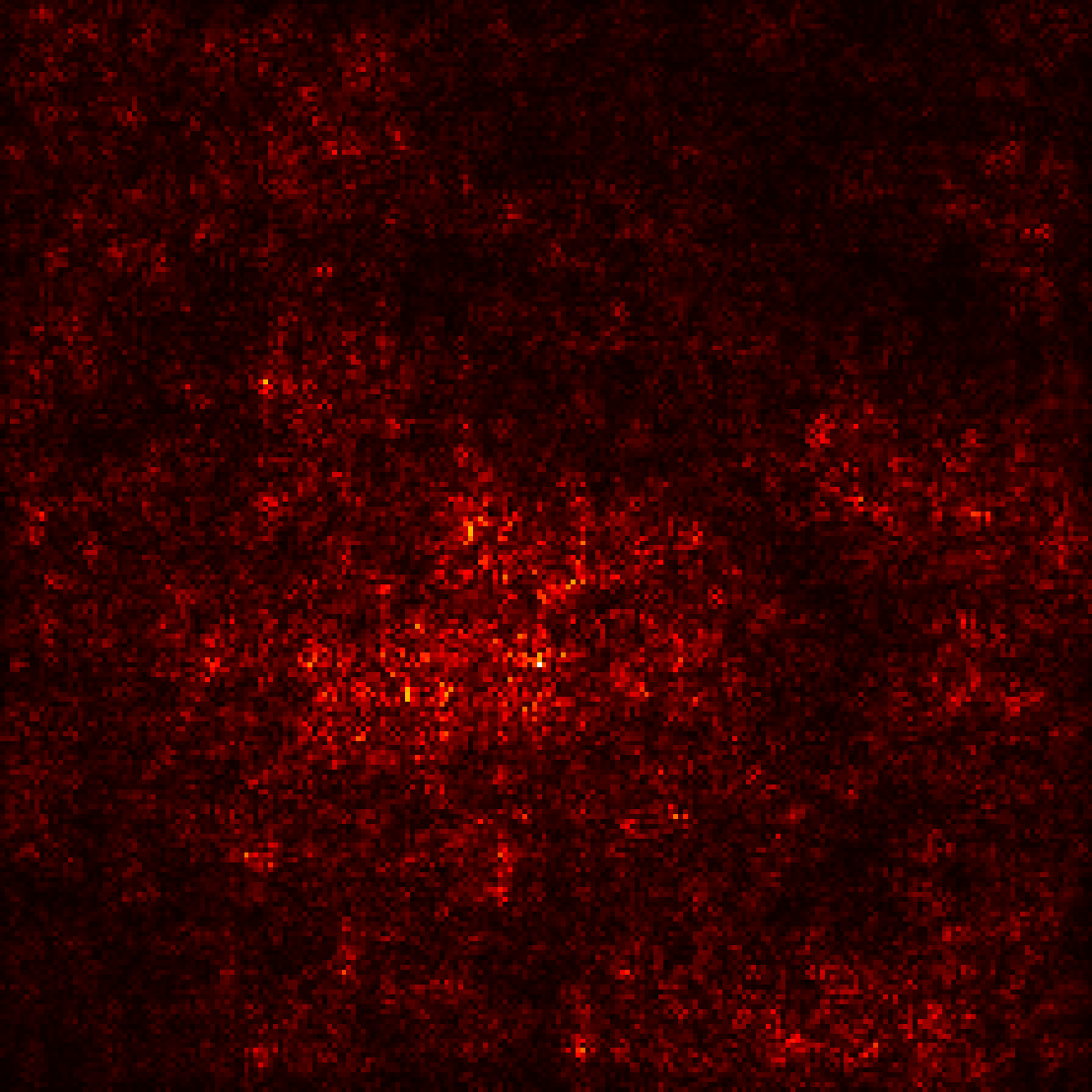} & 
  \includegraphics[scale=\scale]{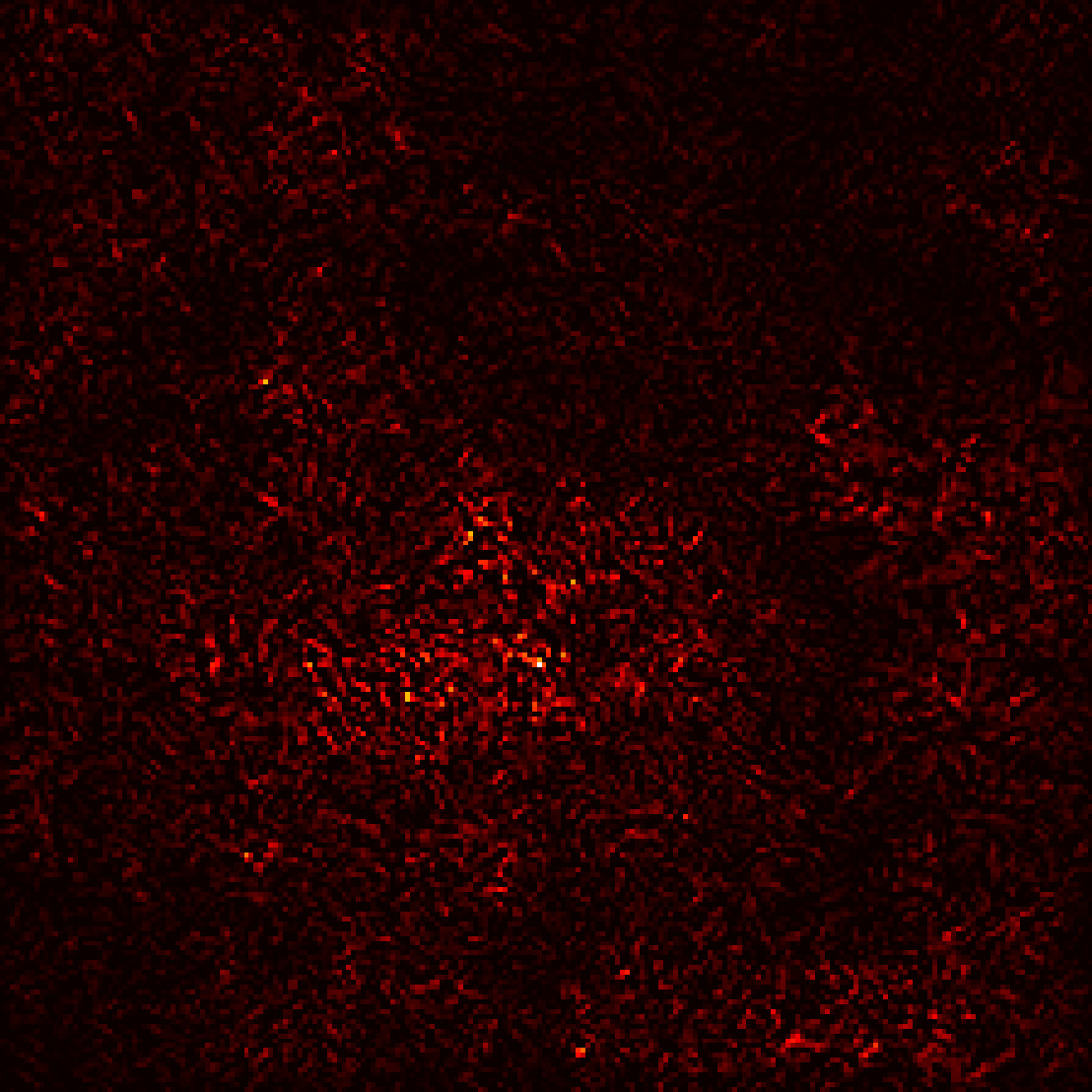} & 
  \includegraphics[scale=\scale]{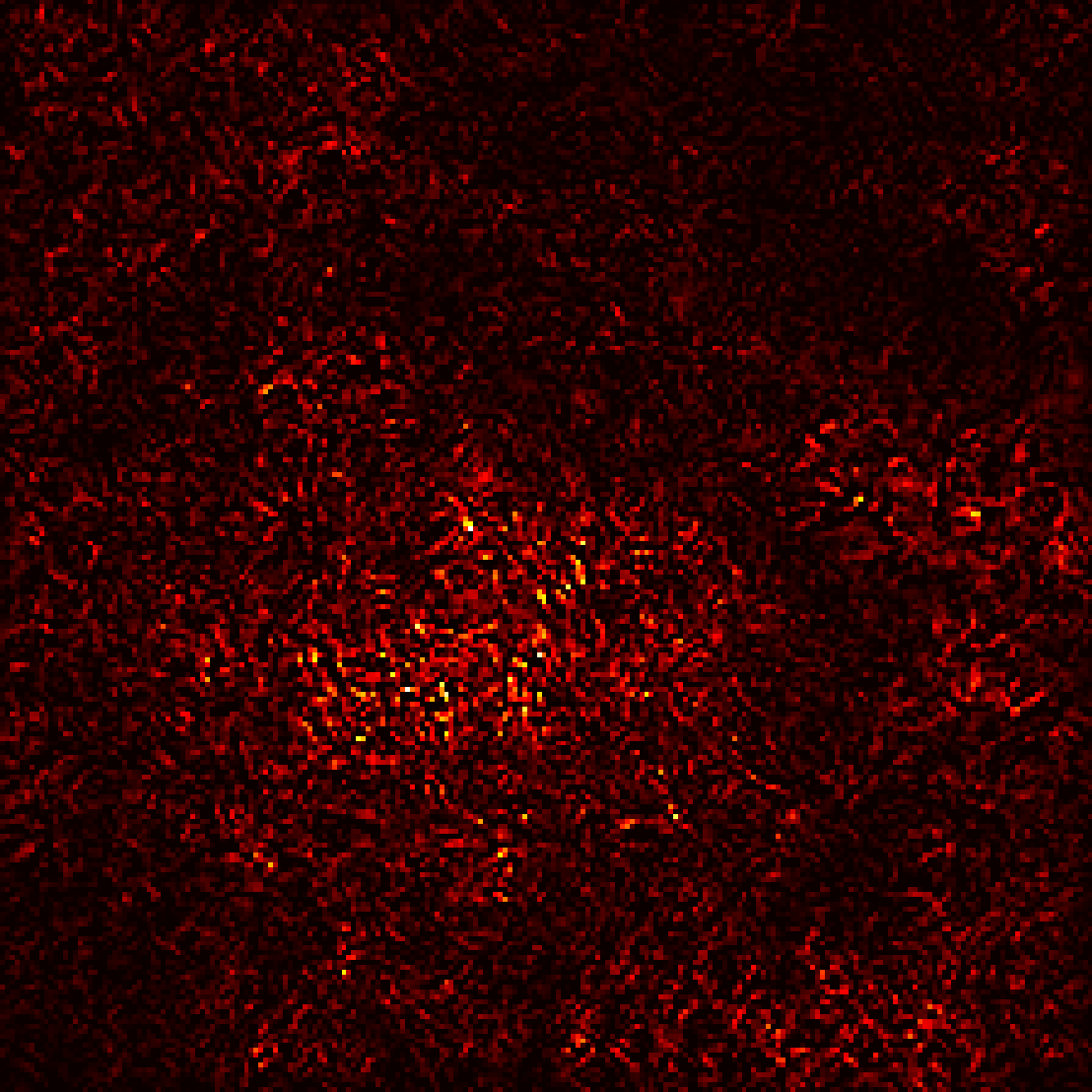} & 
  \includegraphics[scale=\scale]{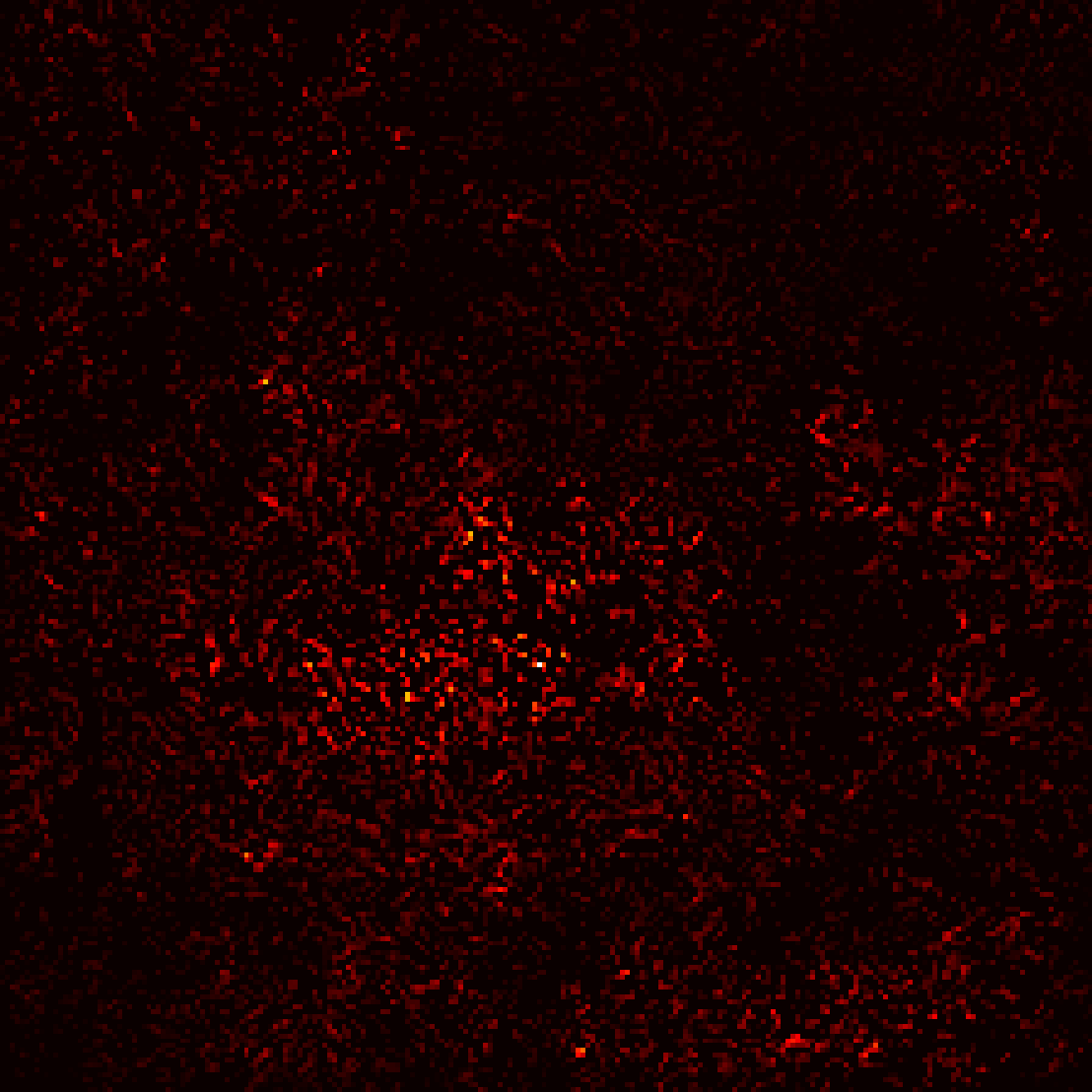} & 
  \includegraphics[scale=\scale]{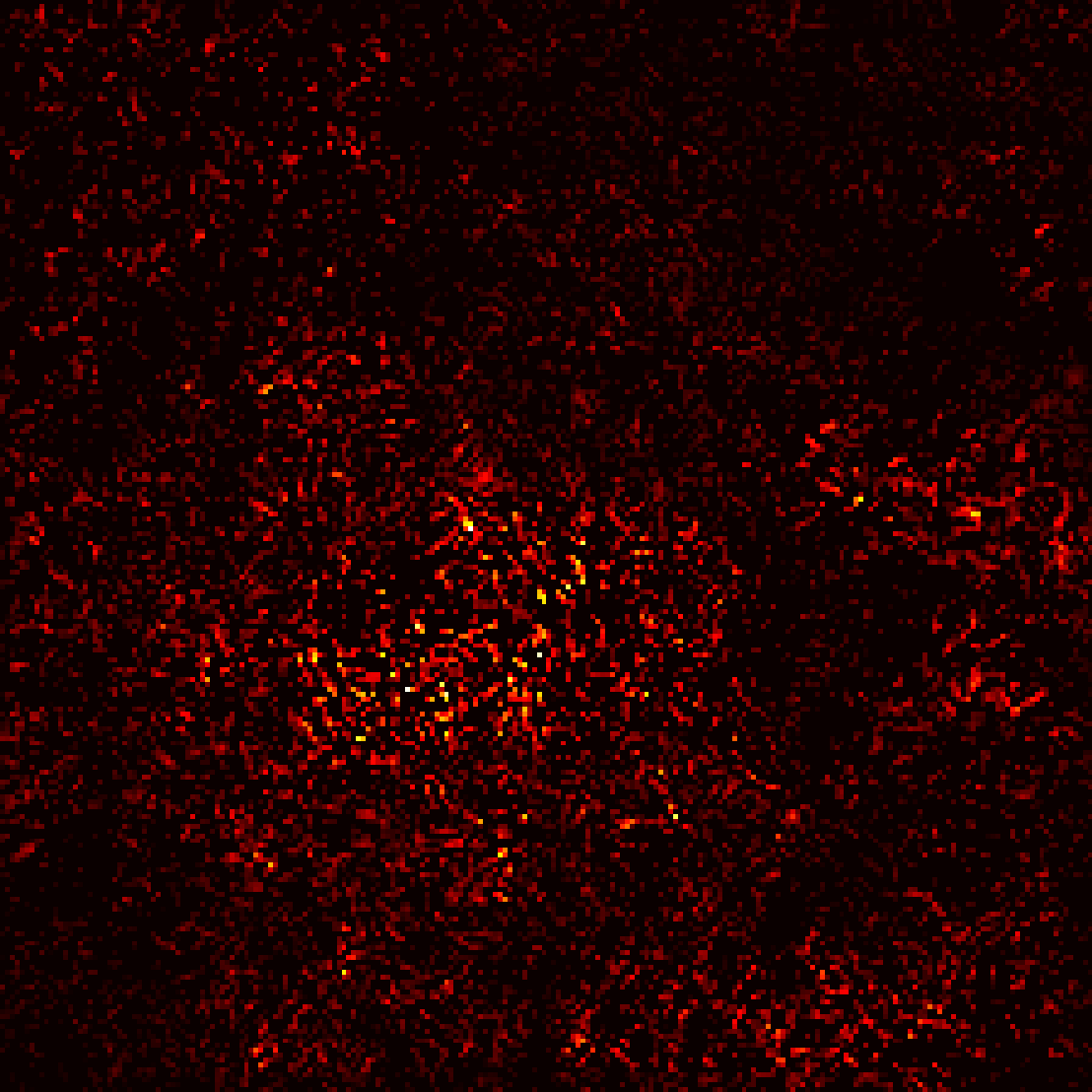} \\
  
  \includegraphics[scale=\scale]{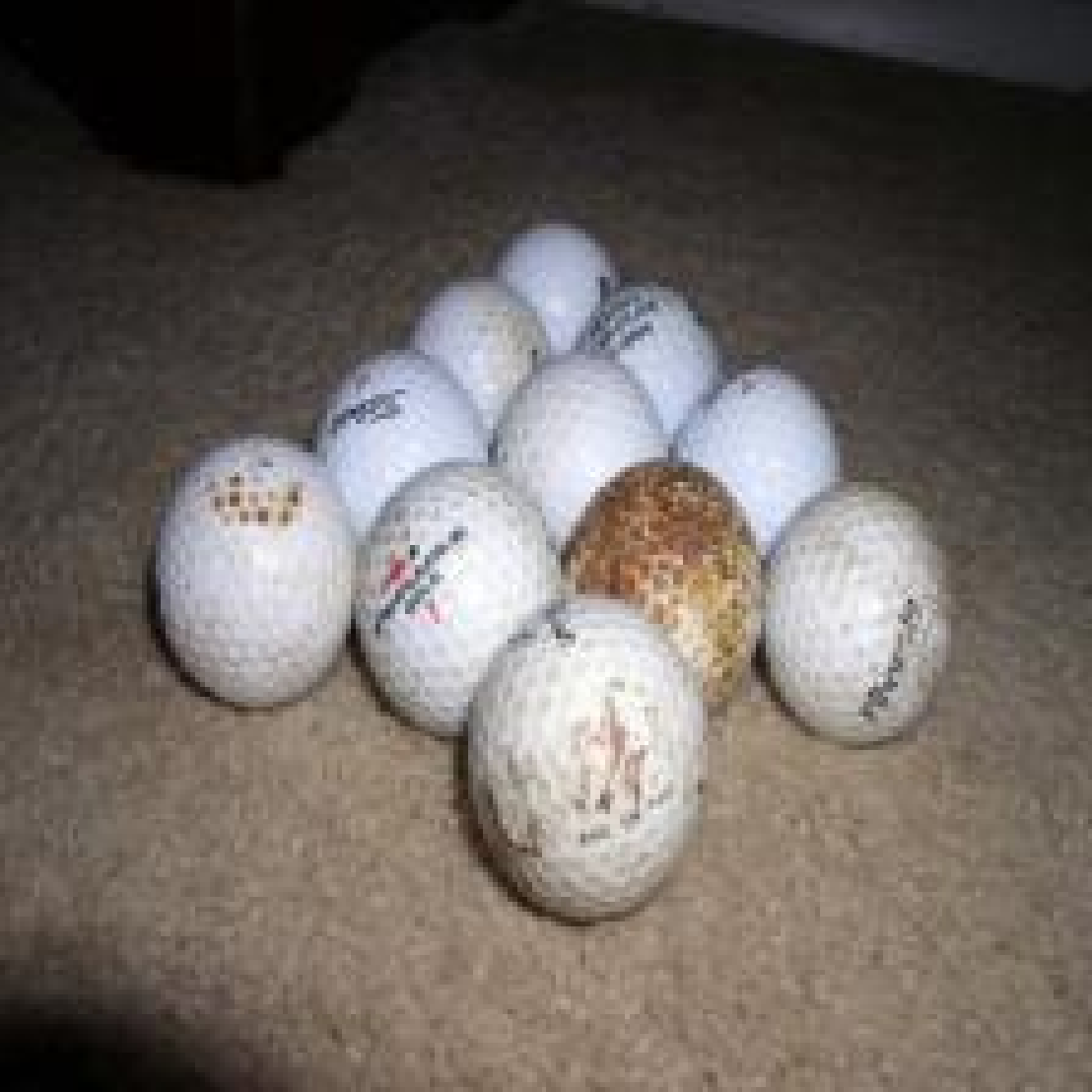} &
  \includegraphics[scale=\scale]{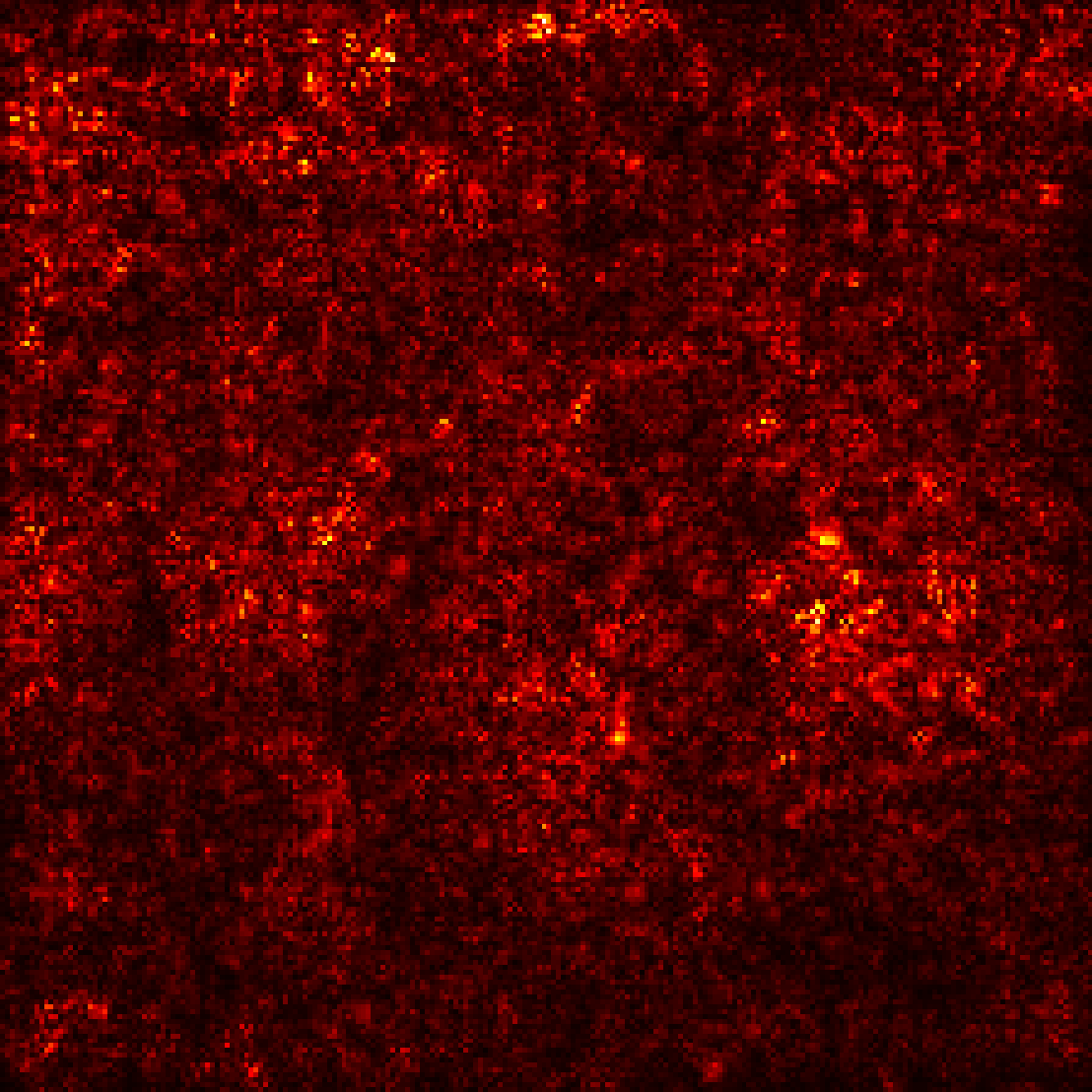} & 
  \includegraphics[scale=\scale]{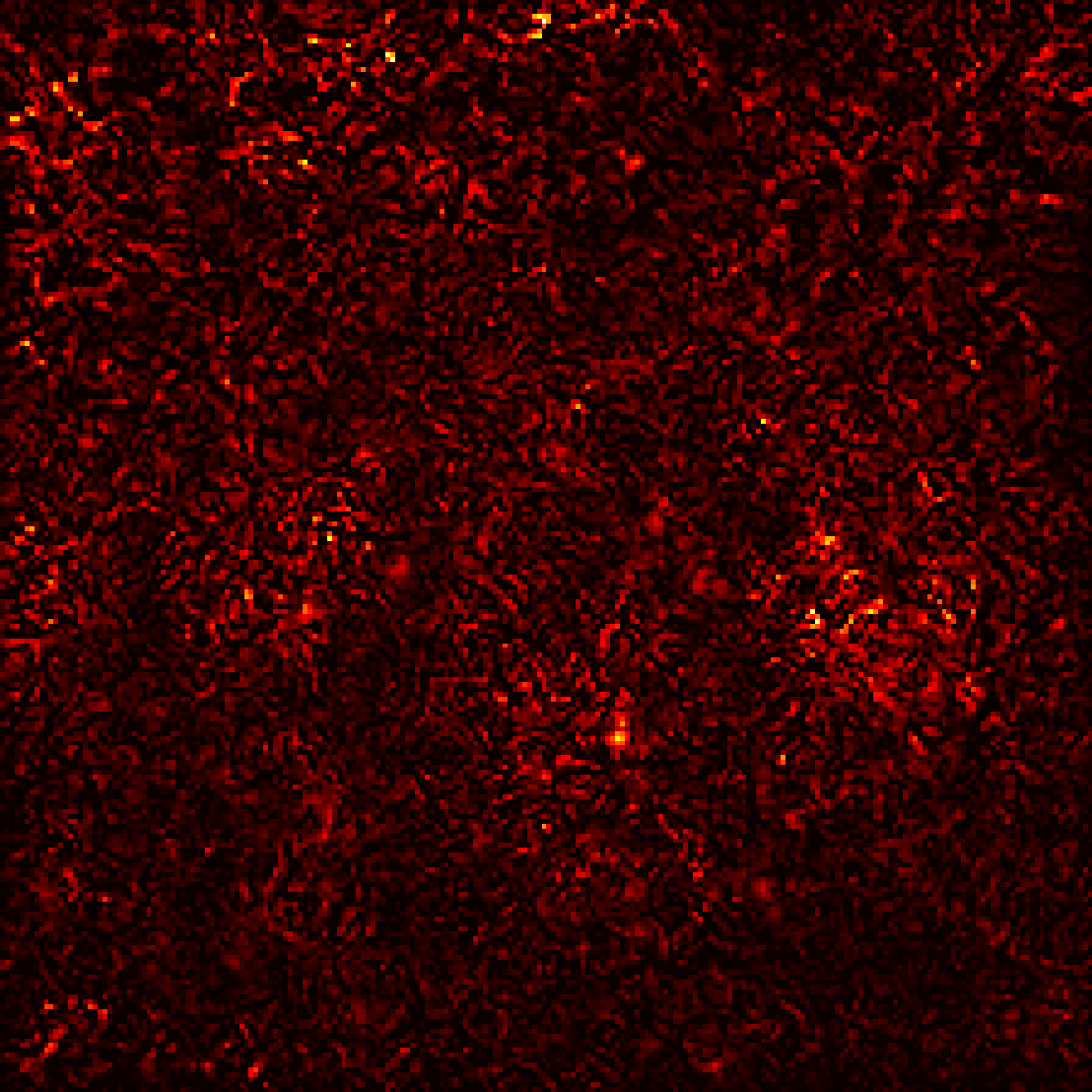} & 
  \includegraphics[scale=\scale]{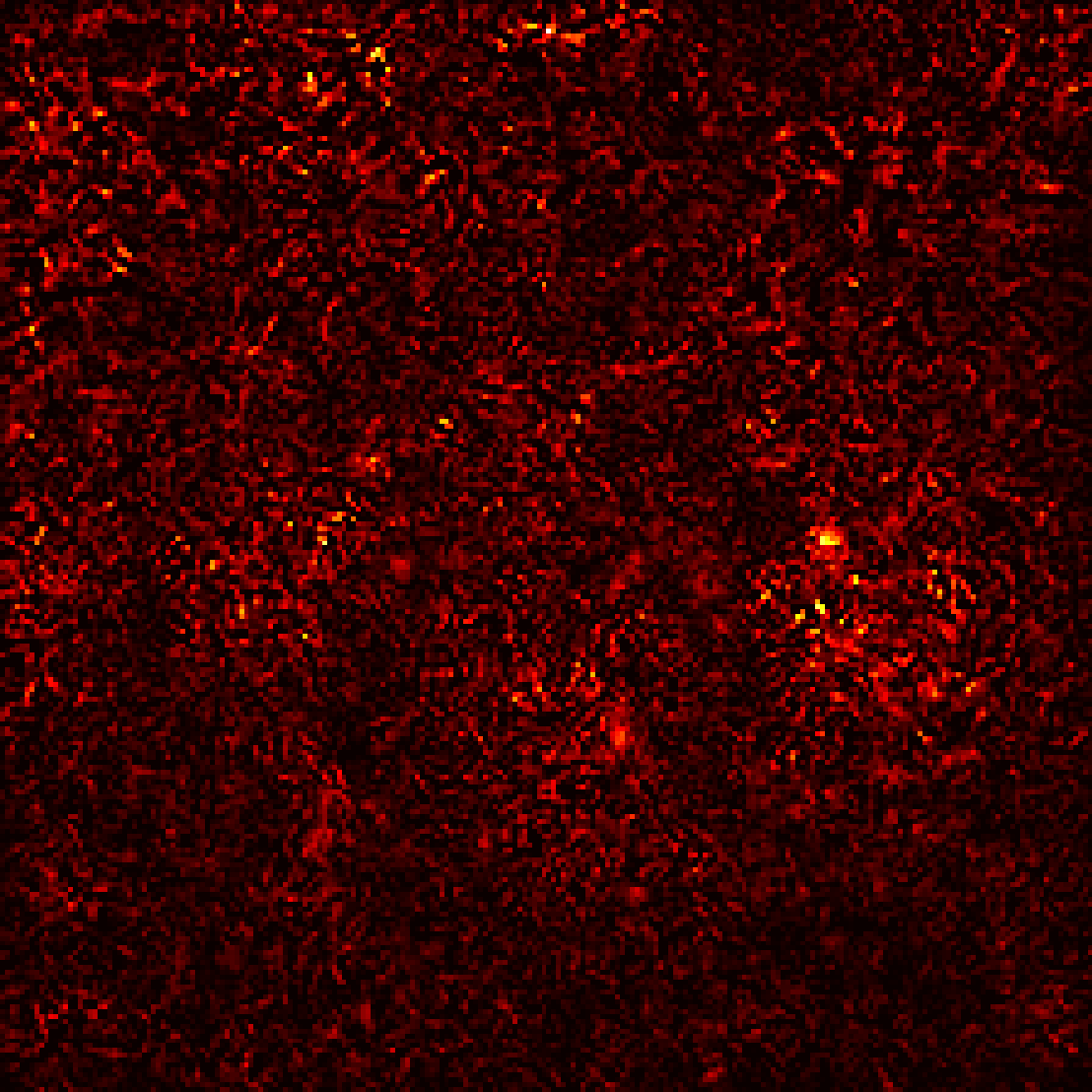} & 
  \includegraphics[scale=\scale]{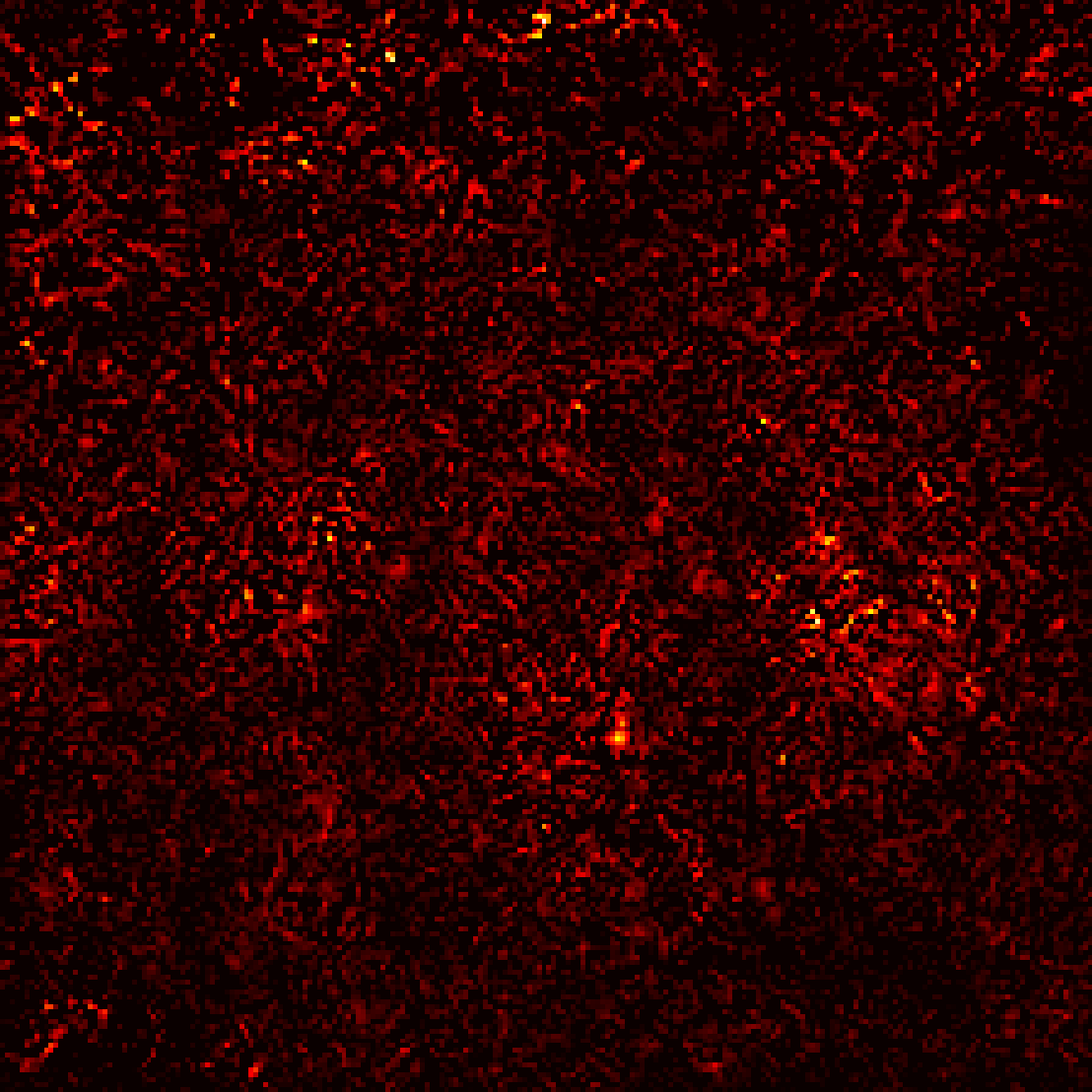} & 
  \includegraphics[scale=\scale]{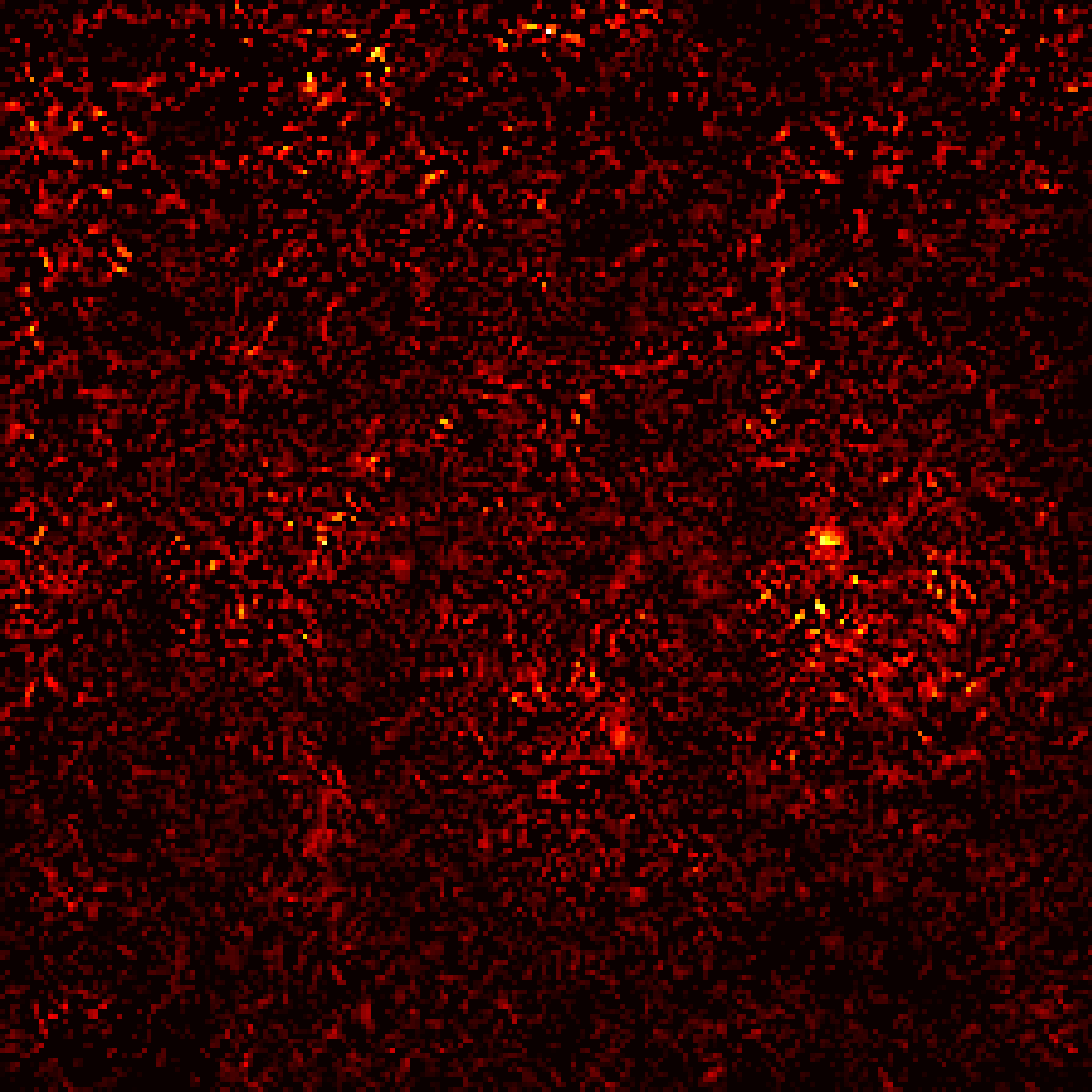} \\
  
  \includegraphics[scale=\scale]{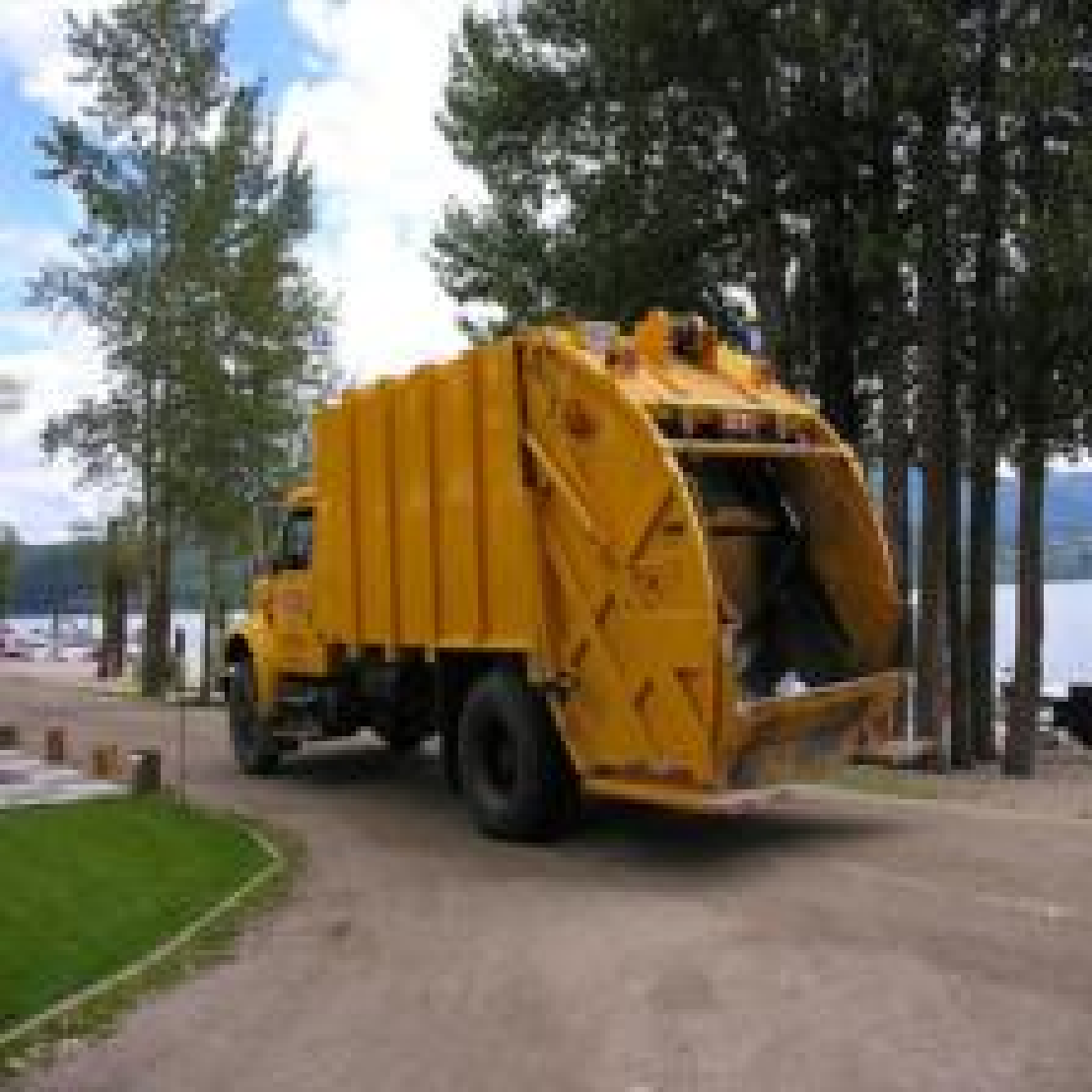} &
  \includegraphics[scale=\scale]{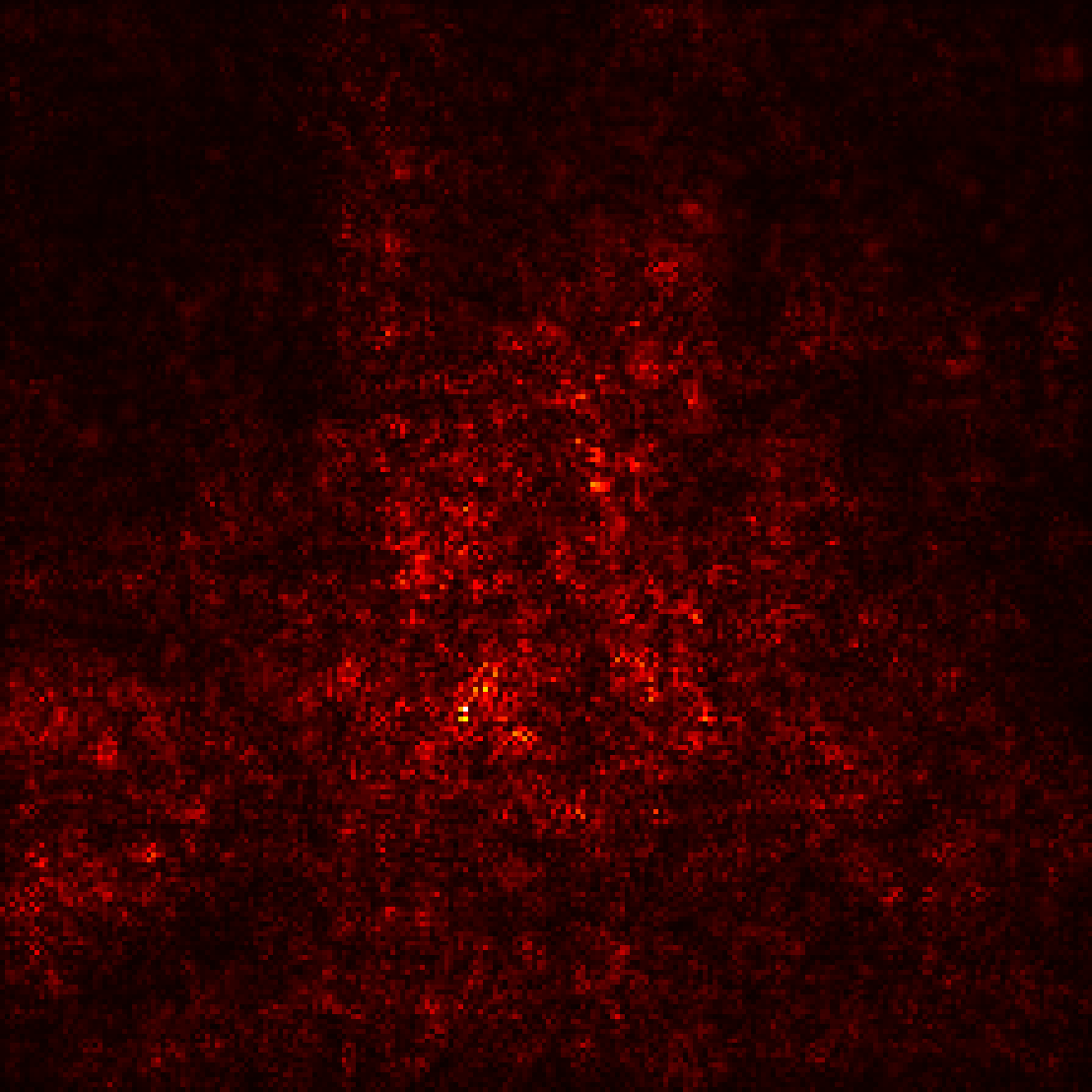} & 
  \includegraphics[scale=\scale]{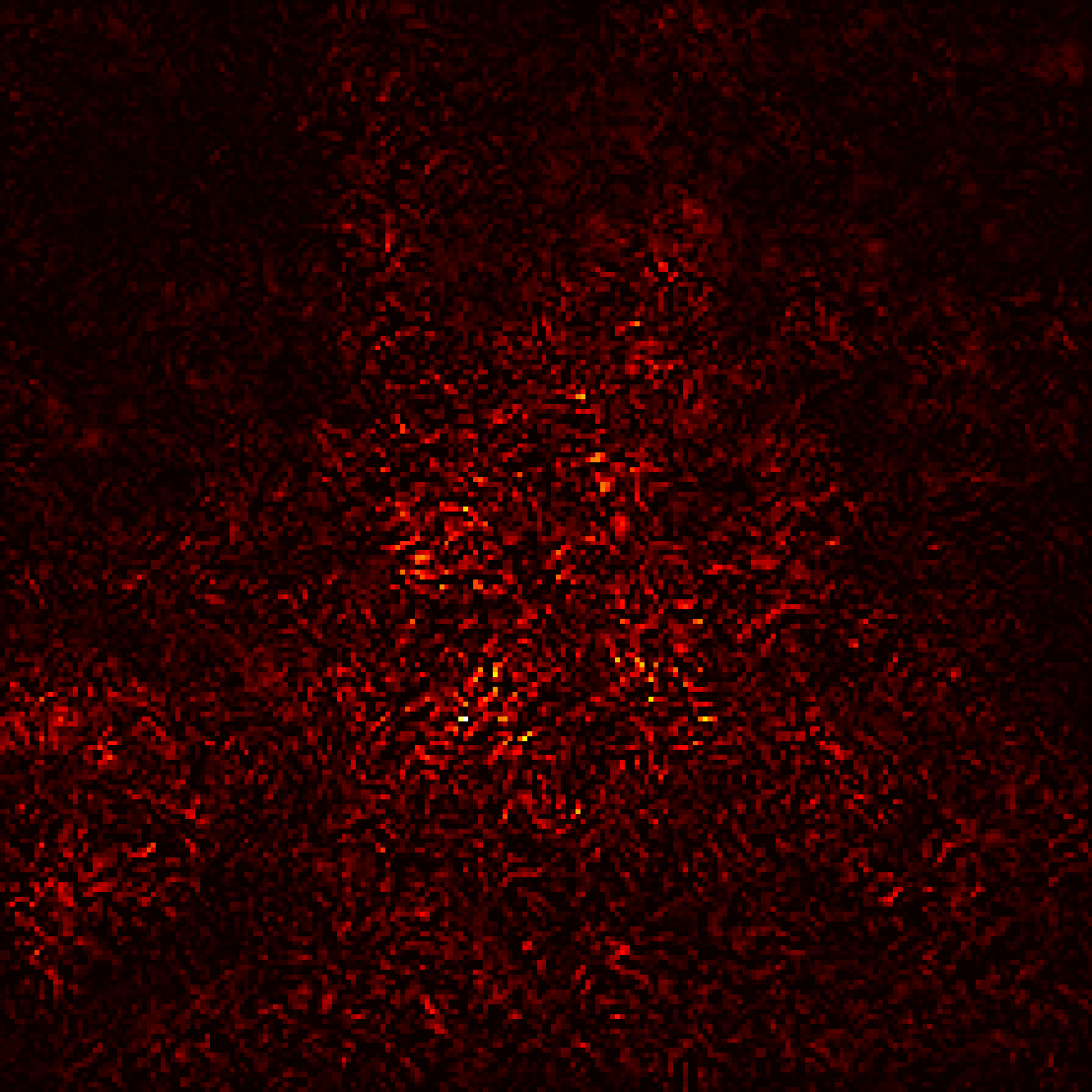} & 
  \includegraphics[scale=\scale]{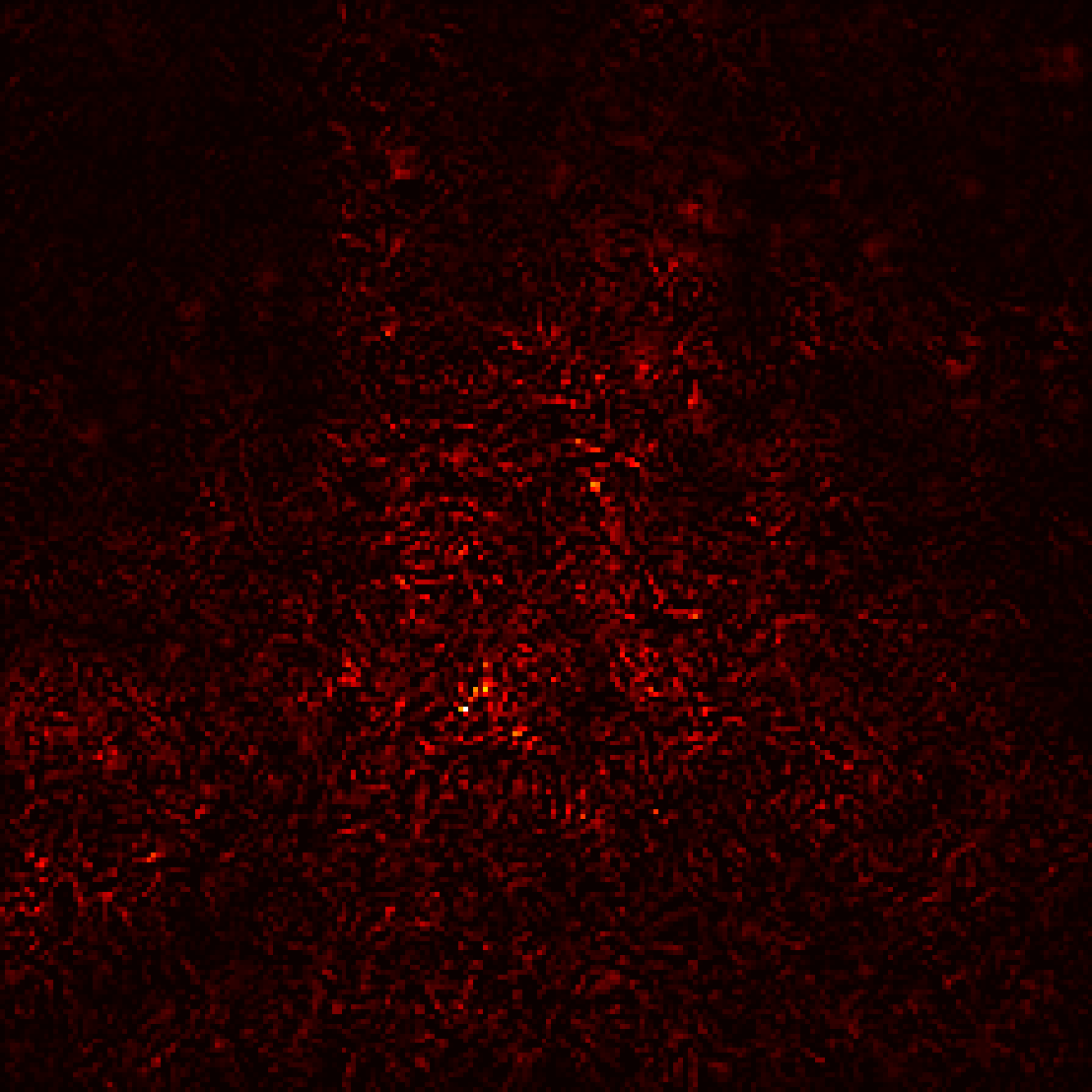} & 
  \includegraphics[scale=\scale]{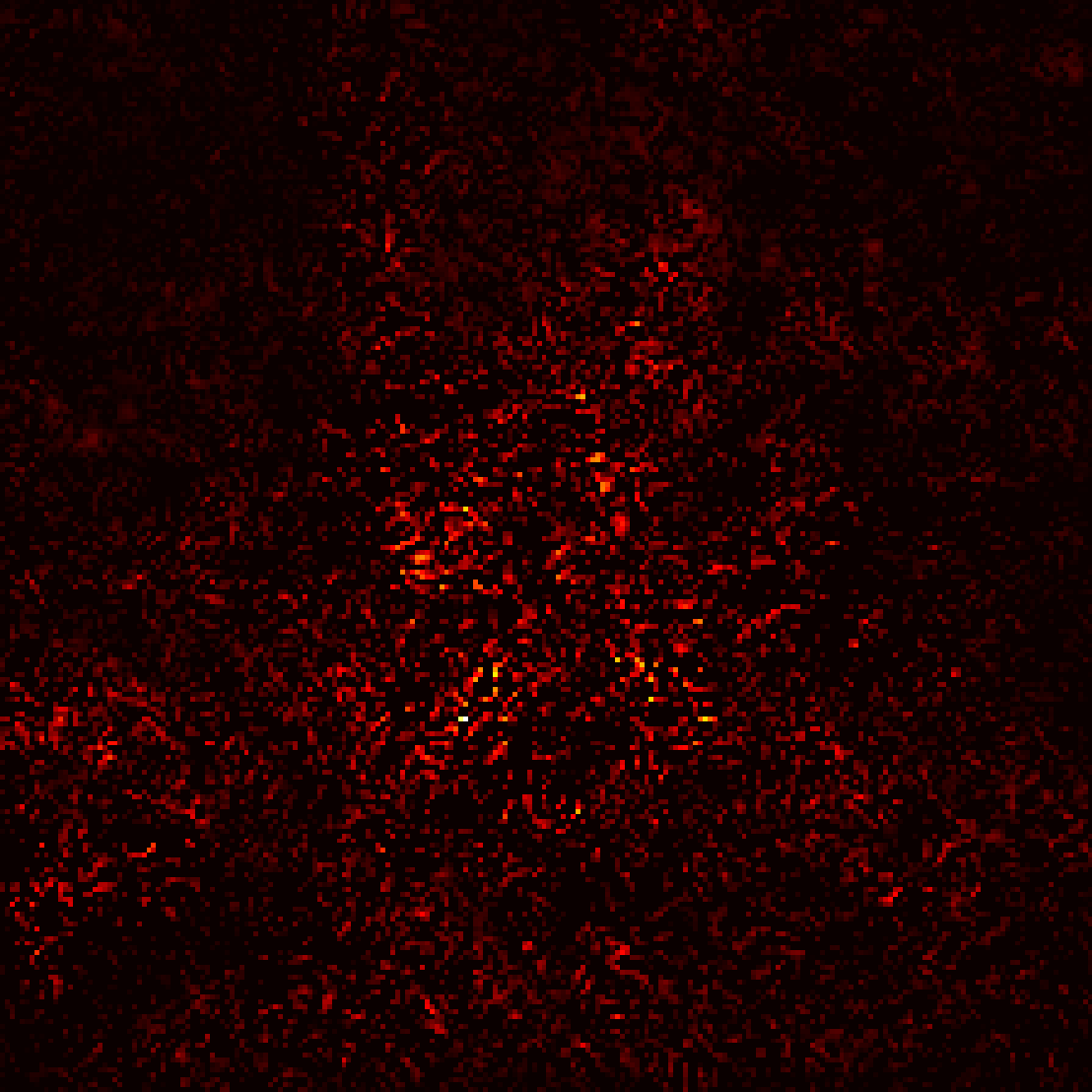} & 
  \includegraphics[scale=\scale]{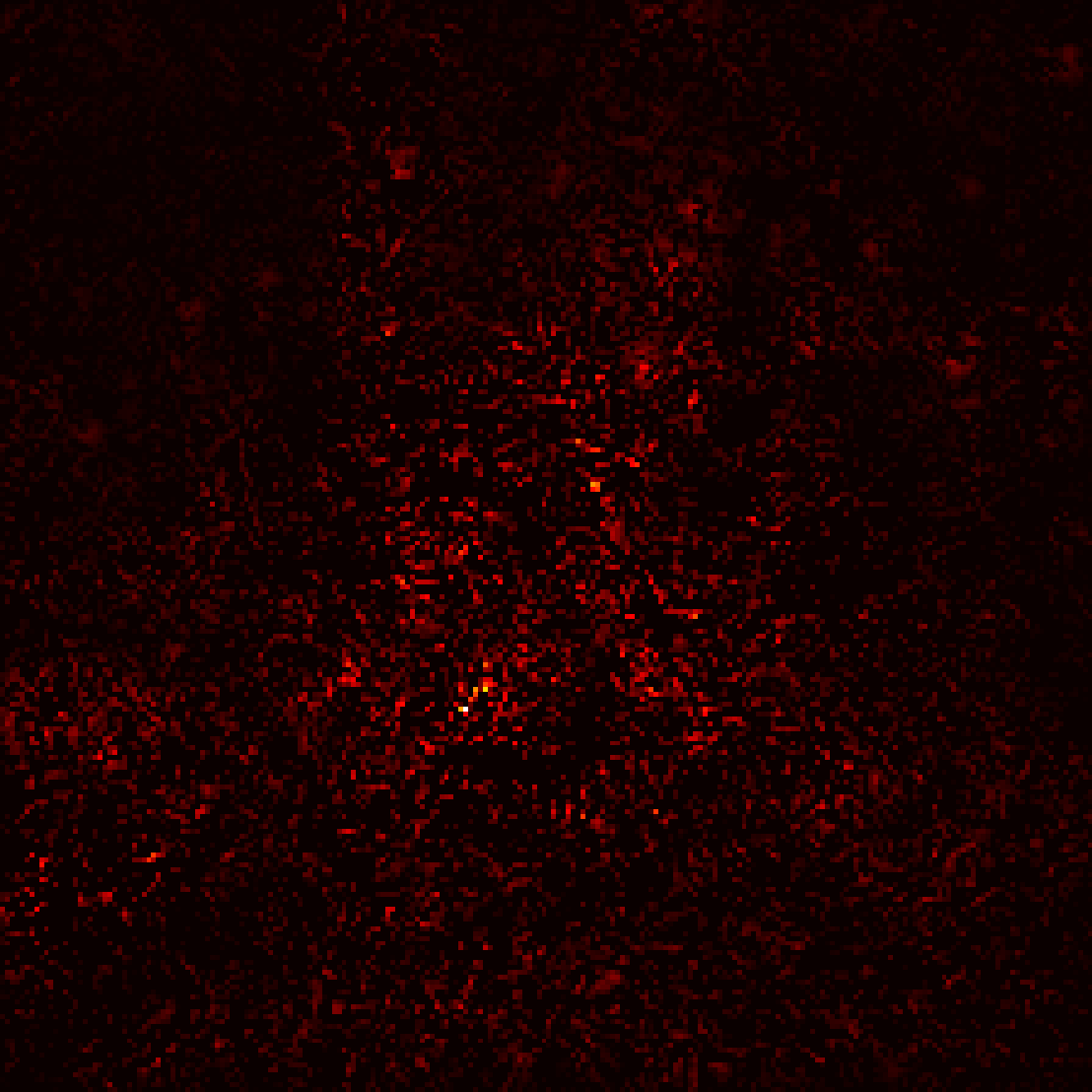} \\
  
  \includegraphics[scale=\scale]{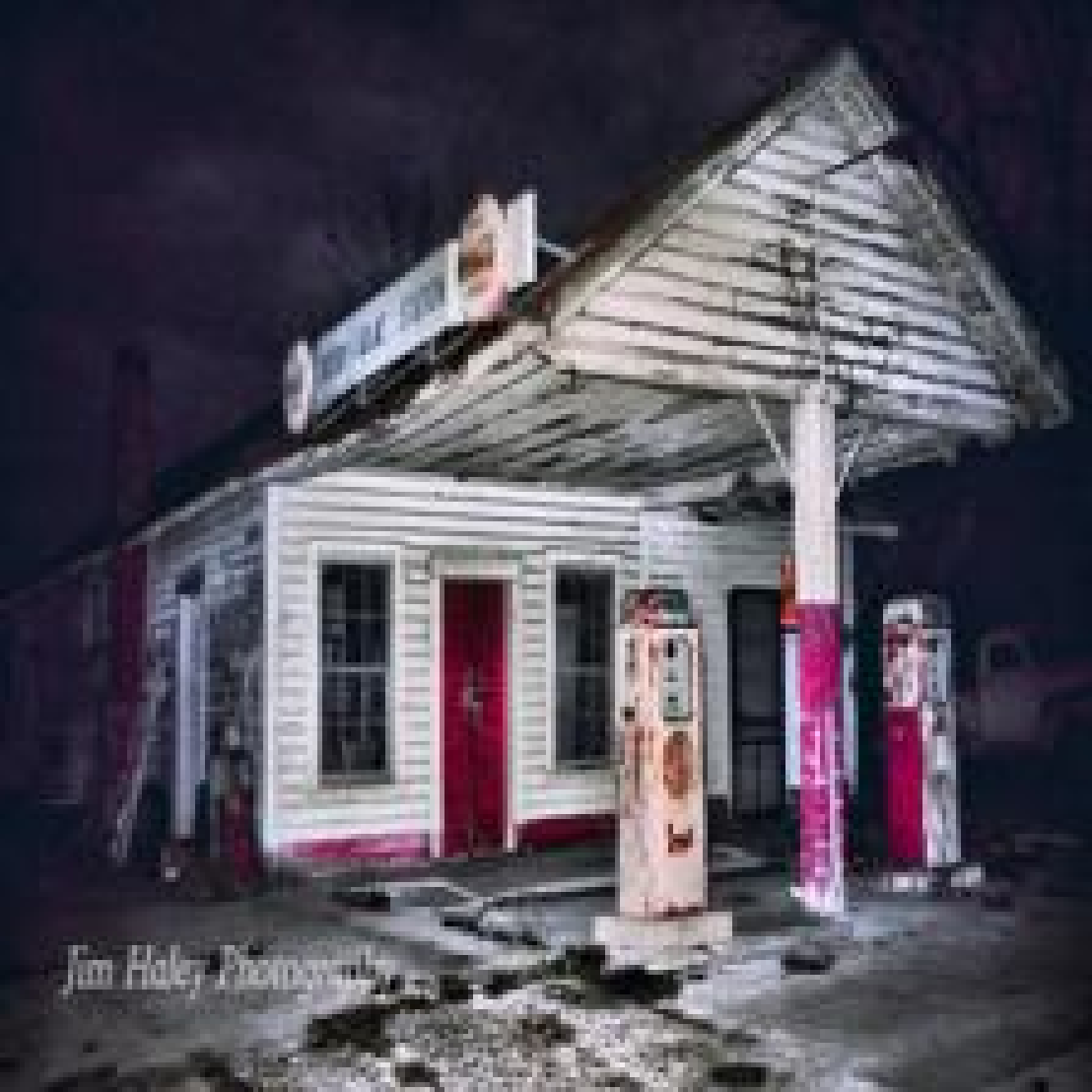} &
  \includegraphics[scale=\scale]{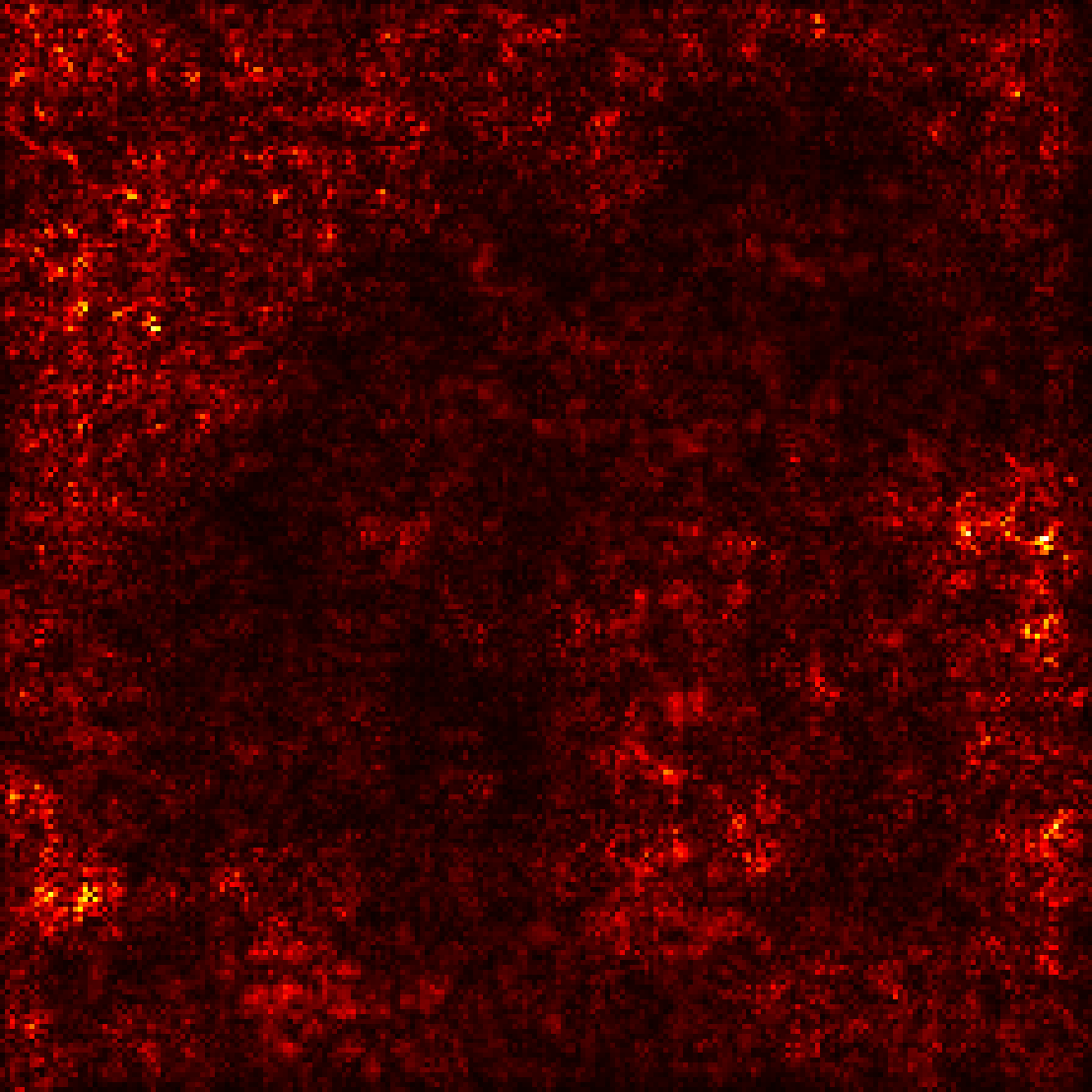} & 
  \includegraphics[scale=\scale]{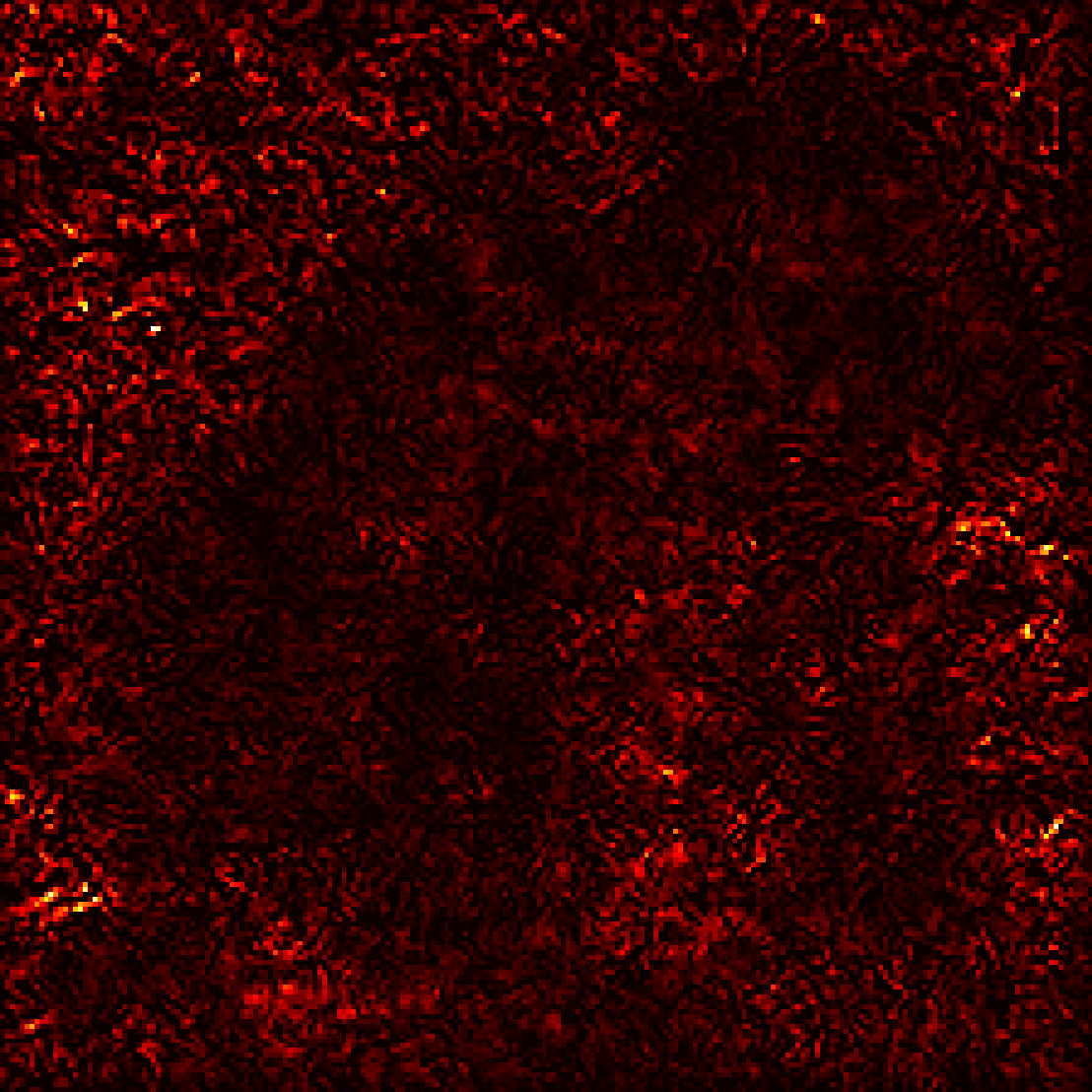} & 
  \includegraphics[scale=\scale]{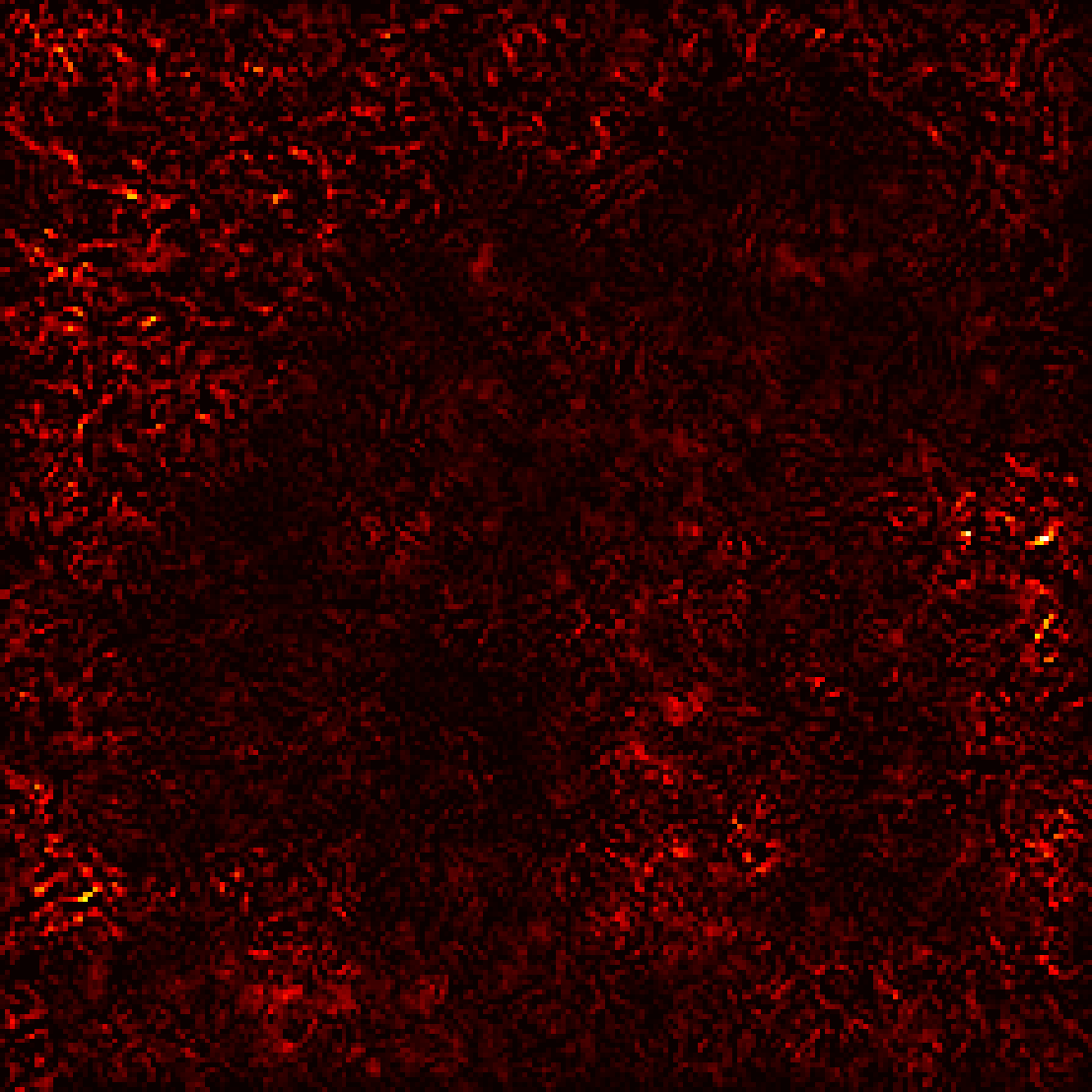} & 
  \includegraphics[scale=\scale]{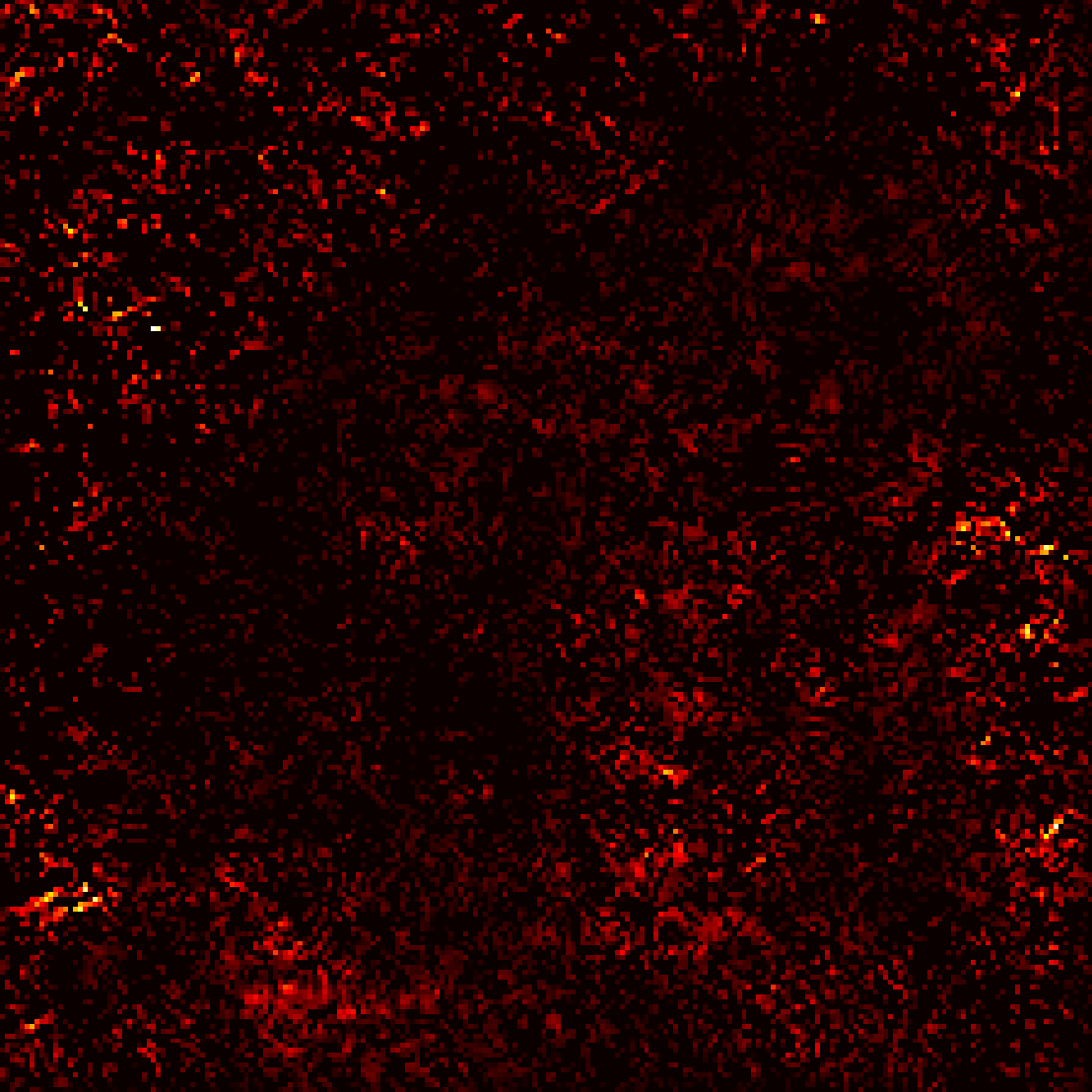} & 
  \includegraphics[scale=\scale]{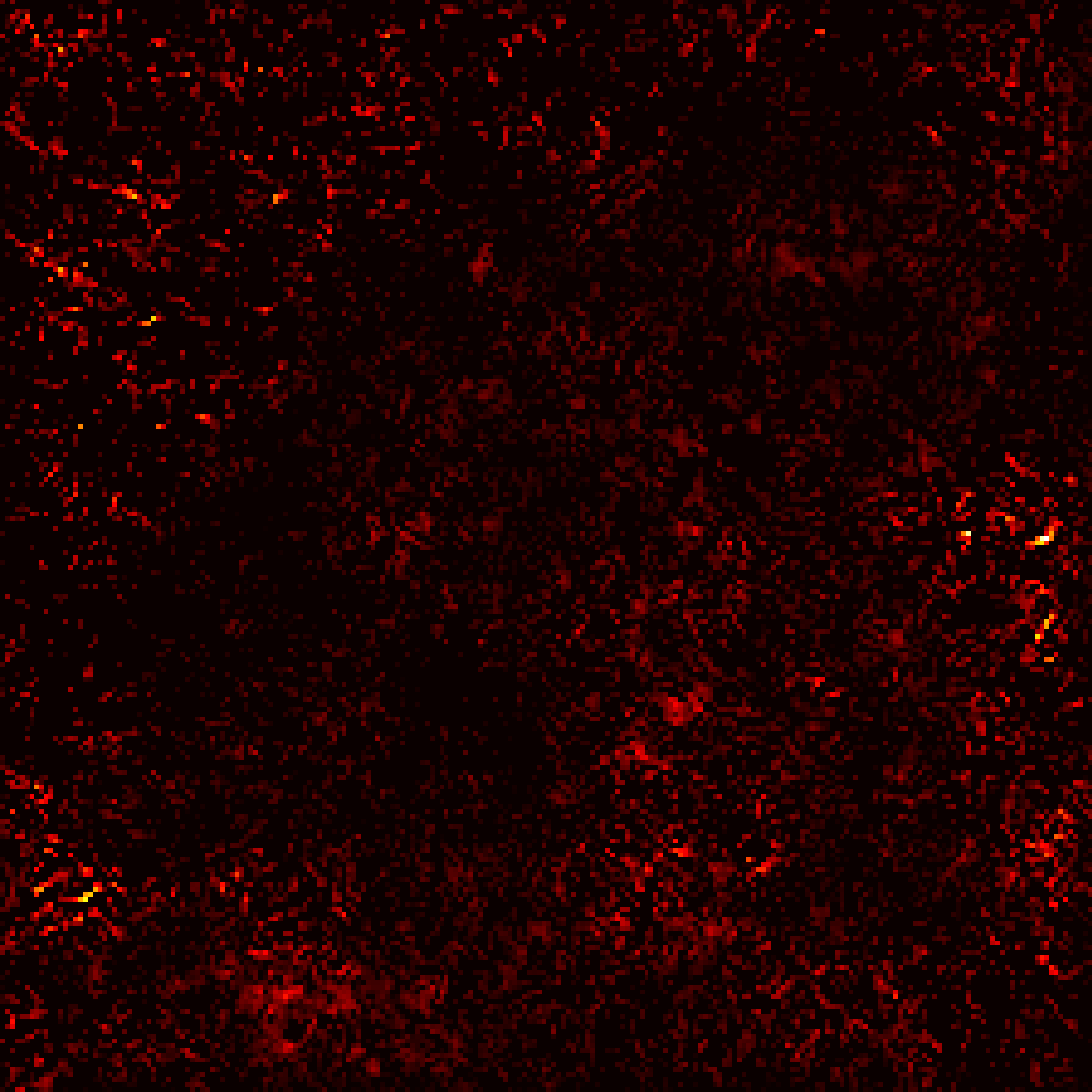} \\
  
  \includegraphics[scale=\scale]{visualizations/images/imagenette/examples/8.png} &
  \includegraphics[scale=\scale]{visualizations/images/imagenette/saliency_map/8.png} & 
  \includegraphics[scale=\scale]{visualizations/images/imagenette/positive_saliency_map/8.png} & 
  \includegraphics[scale=\scale]{visualizations/images/imagenette/negative_saliency_map/8.png} & 
  \includegraphics[scale=\scale]{visualizations/images/imagenette/active_saliency_map/8.png} & 
  \includegraphics[scale=\scale]{visualizations/images/imagenette/inactive_saliency_map/8.png} \\
  
  \includegraphics[scale=\scale]{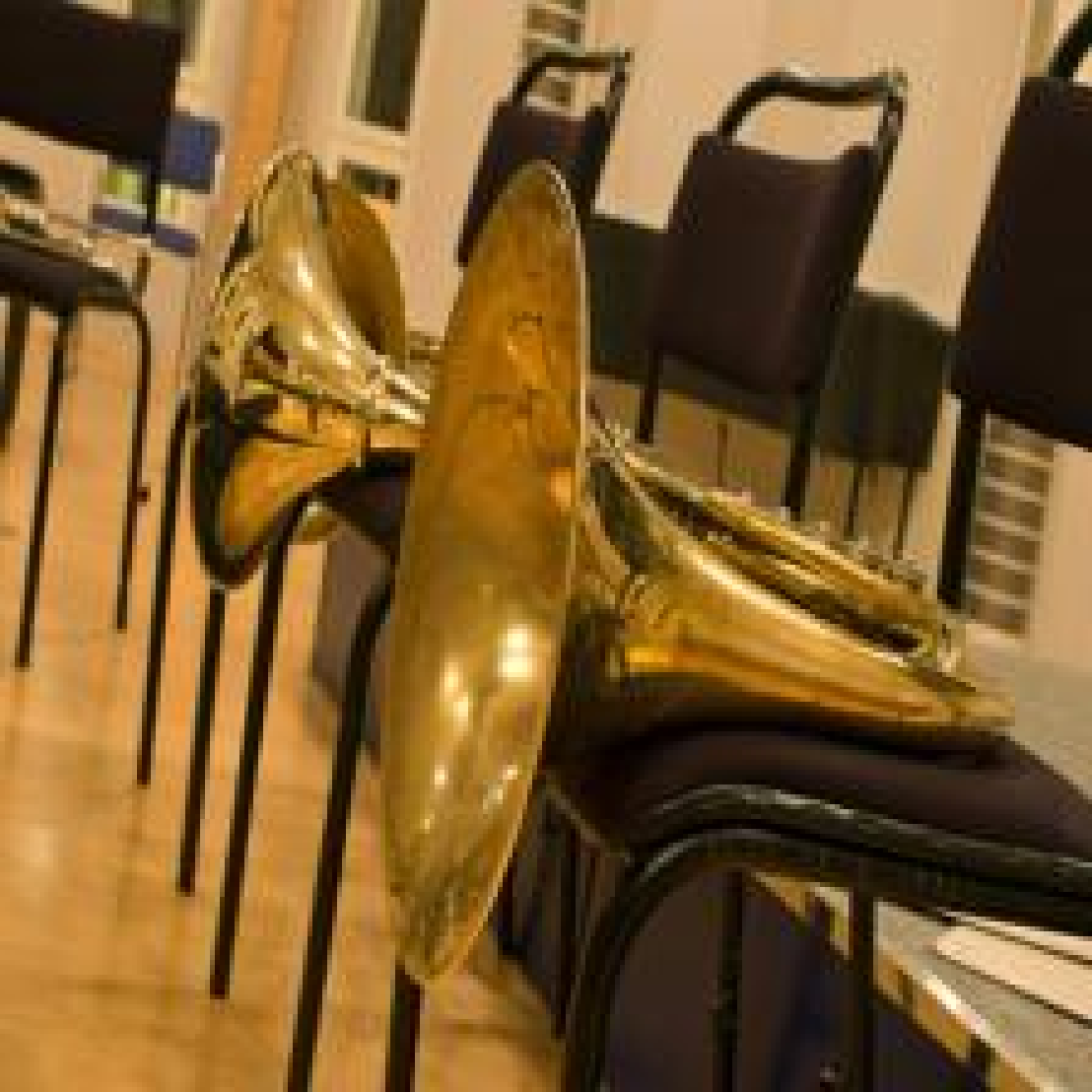} &
  \includegraphics[scale=\scale]{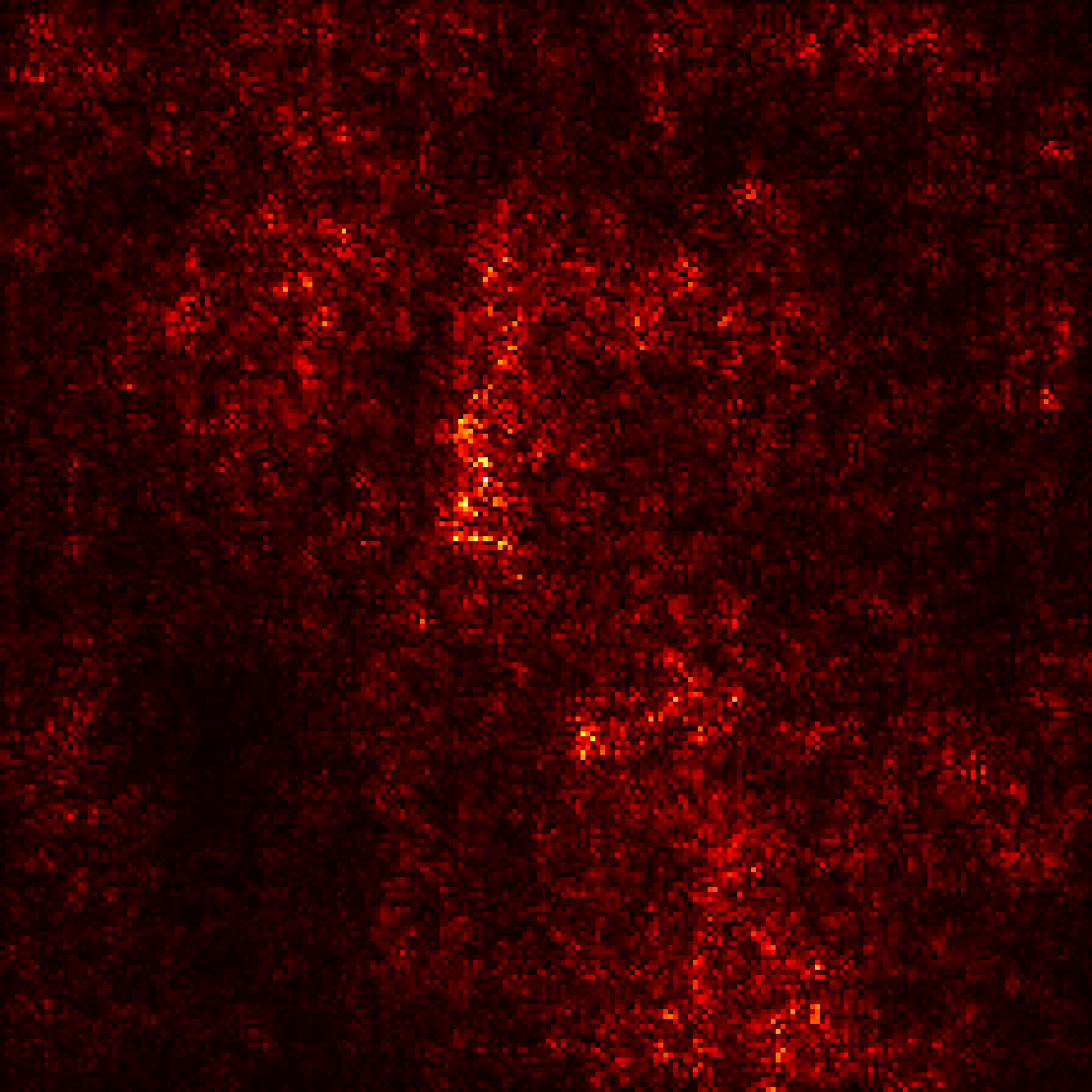} & 
  \includegraphics[scale=\scale]{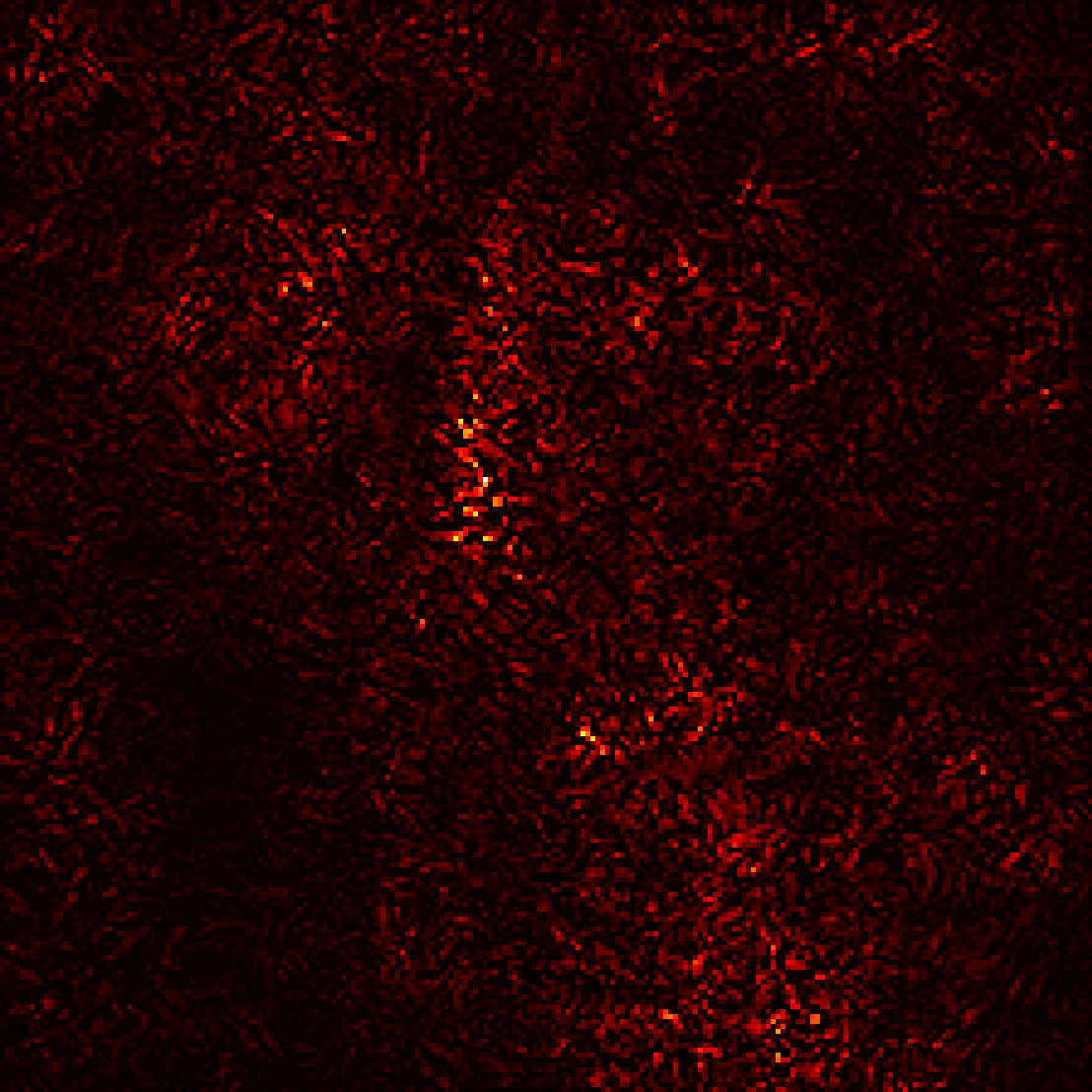} & 
  \includegraphics[scale=\scale]{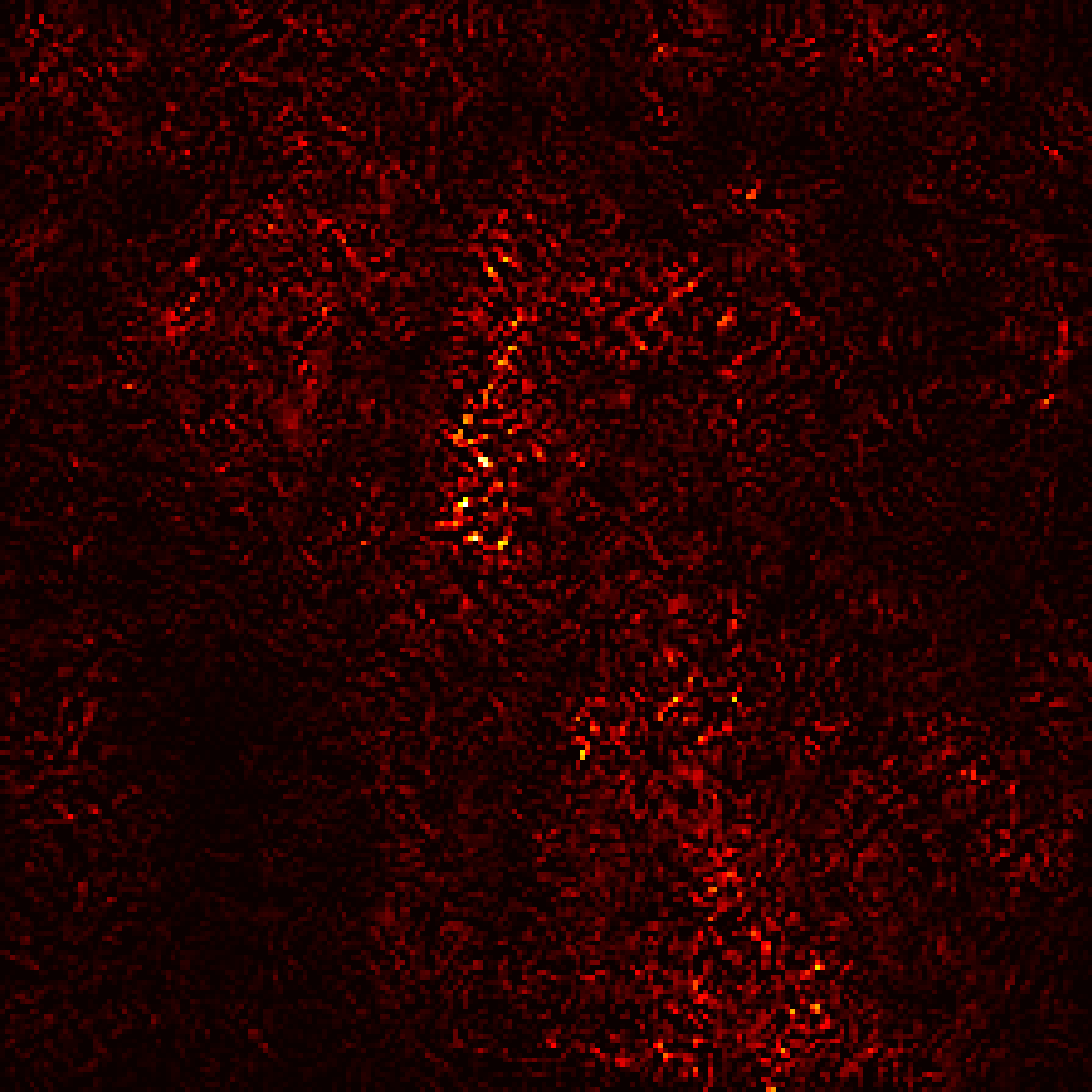} & 
  \includegraphics[scale=\scale]{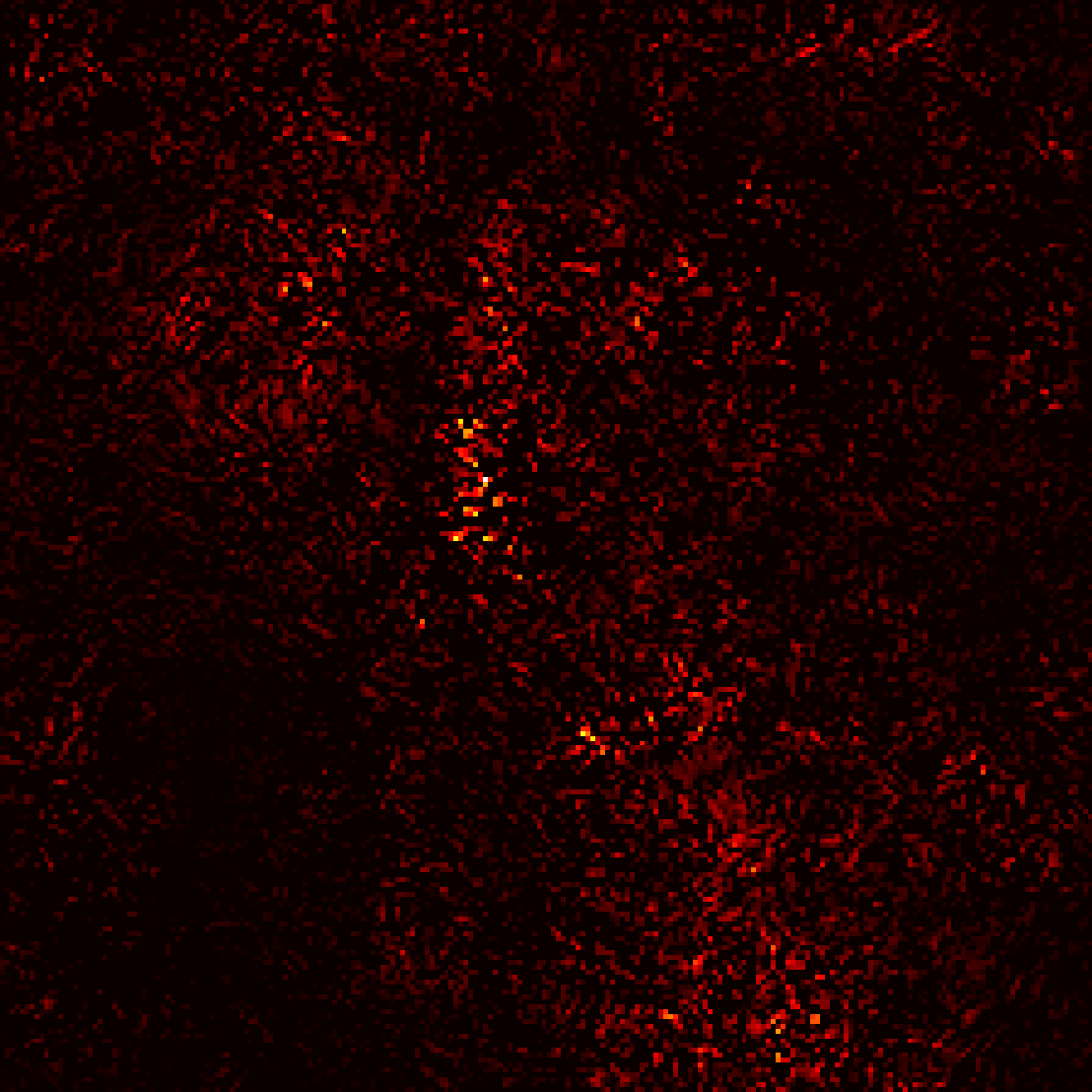} & 
  \includegraphics[scale=\scale]{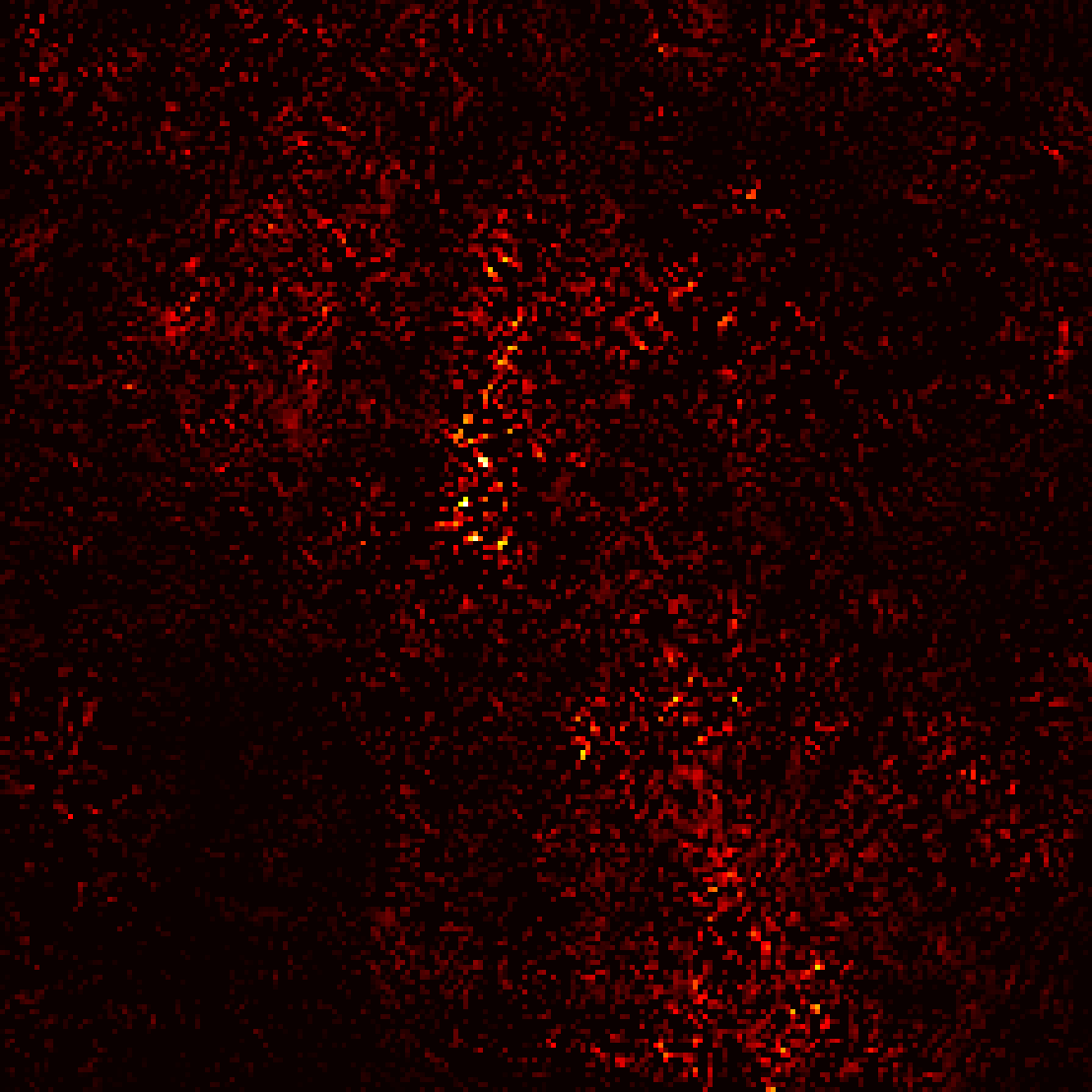} \\
  \end{tabular}
  \caption{Comparison saliency maps ResNet18 Imagenette.}
\end{figure}

\section{Black-deletion and white-deletion on the training set}
\label{sec:black-deletion and white-deletion for training set}

This section shows the black- and white-deletion on the training set for each dataset. It can be appreciated that the results are similar to the ones showed in Section~\ref{sec:experiments}.

\begin{figure}[p]
    \centering
    \subfloat[CIFAR-10 CNN]{\includegraphics[width=0.475\textwidth]{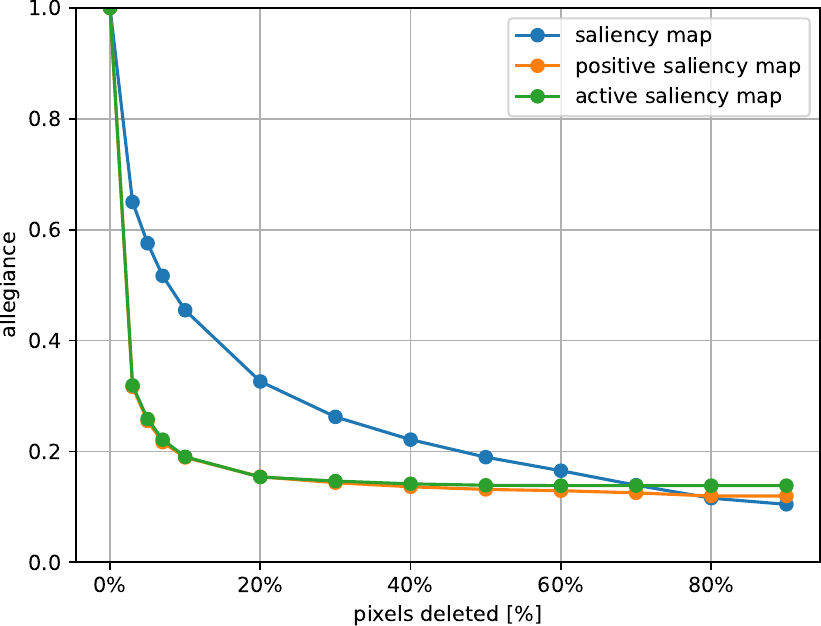}} \hfill
    \subfloat[CIFAR-10 ResNet-18]{\includegraphics[width=0.475\textwidth]{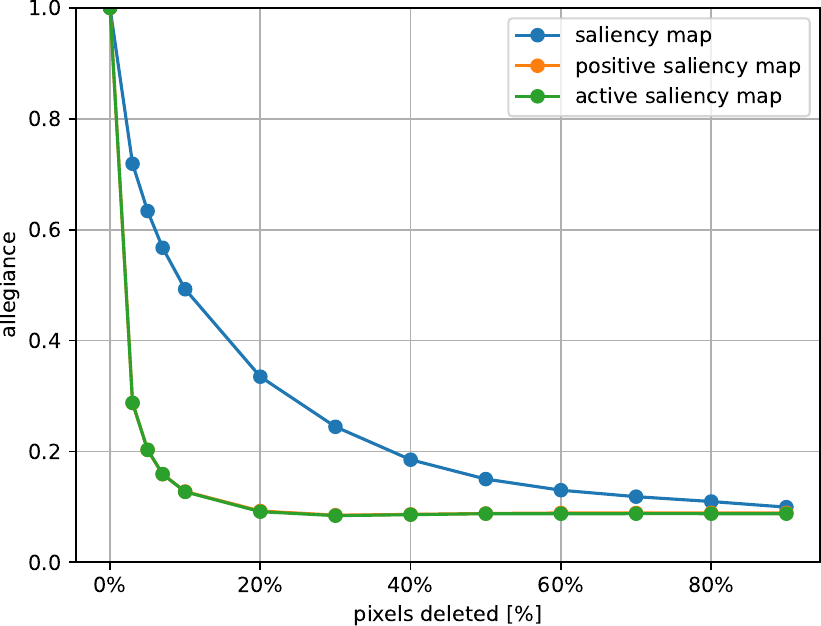}} \\
    \subfloat[Imagenette CNN]{\includegraphics[width=0.475\textwidth]{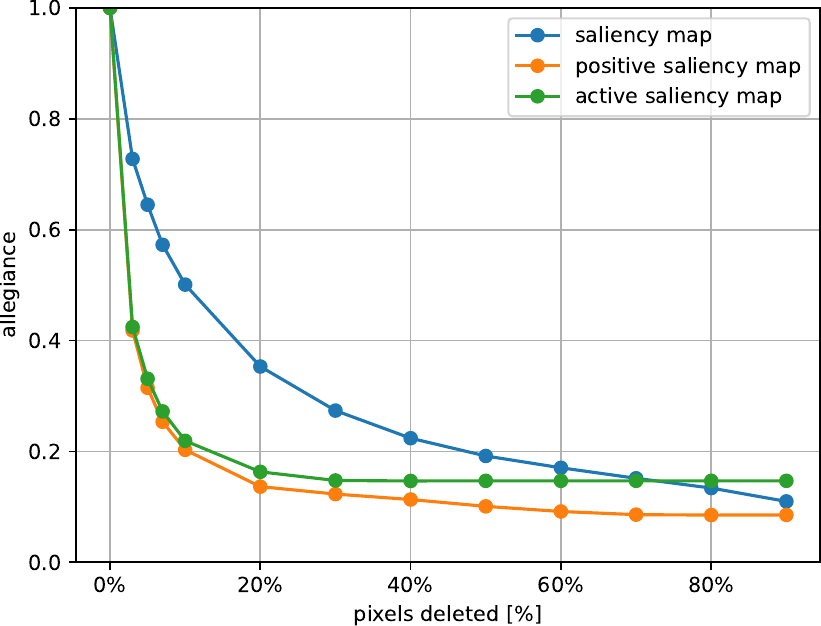}} \hfill
    \subfloat[Imagenette ResNet-18]{\includegraphics[width=0.475\textwidth]{visualizations/graphs/cifar10/resnet18_False/train/0.pdf}} \\
    \subfloat[Imagenette ResNet-18 pre-trained]{\includegraphics[width=0.475\textwidth]{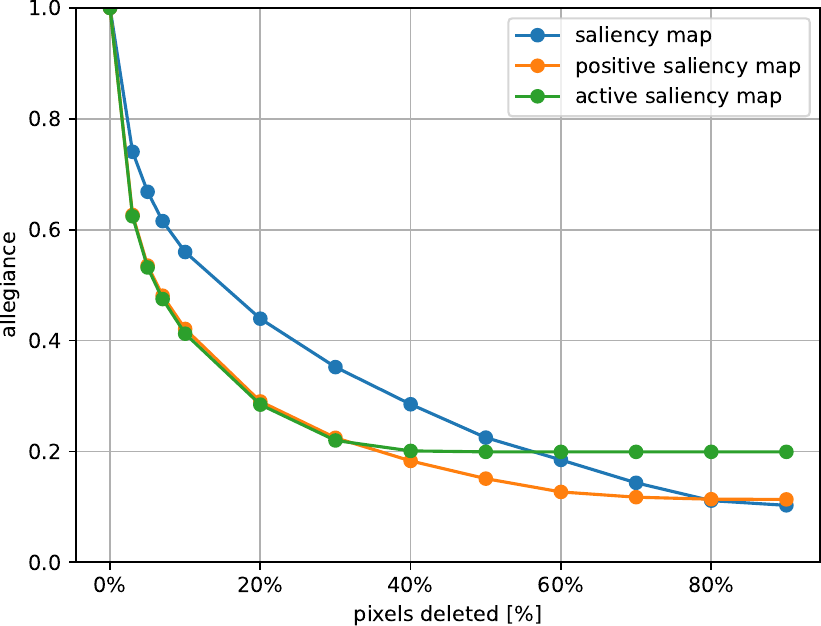}} \hfill
    \subfloat[Imagenette ConvNeXt pre-trained]{\includegraphics[width=0.475\textwidth]{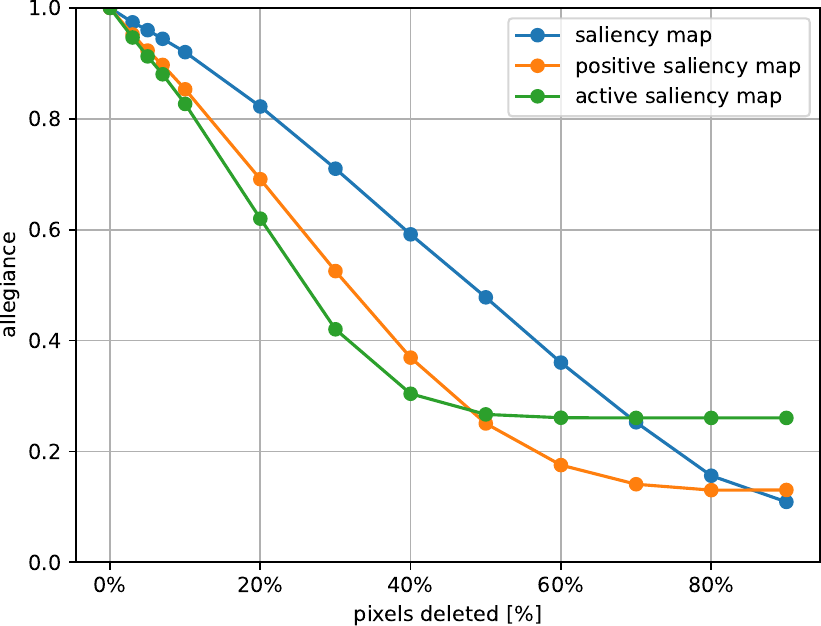}} \\
    \caption{Black-Deletion Benchmark on the training set.}
    \label{fig: black-deletion benchmark on the training set}
\end{figure}

\begin{figure}[p]
    \centering
    \subfloat[CIFAR-10 CNN]{\includegraphics[width=0.475\textwidth]{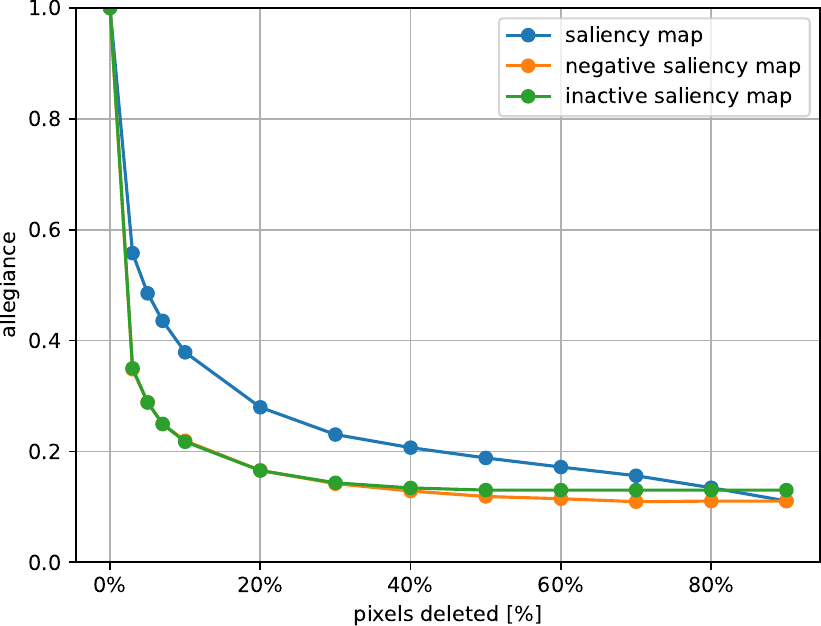}} \hfill
    \subfloat[CIFAR-10 ResNet-18]{\includegraphics[width=0.475\textwidth]{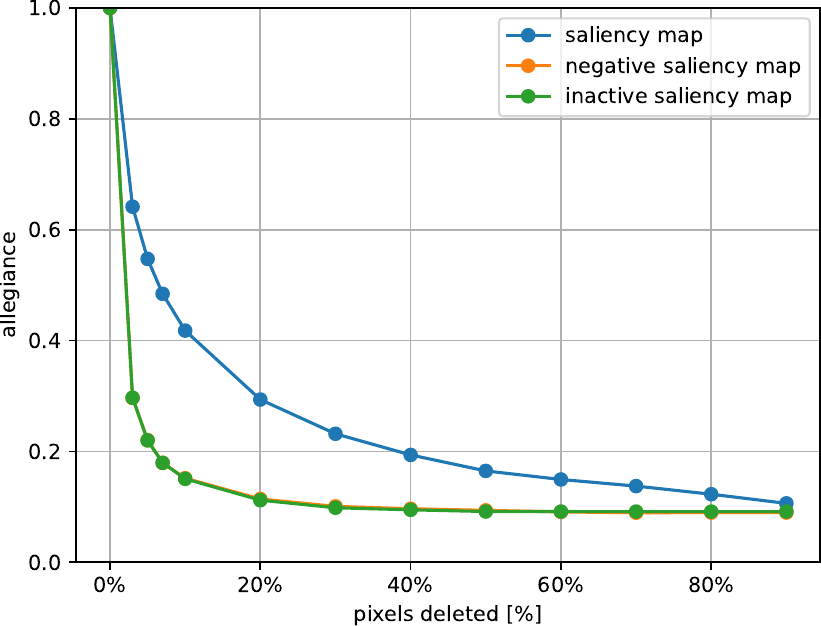}} \\
    \subfloat[Imagenette CNN]{\includegraphics[width=0.475\textwidth]{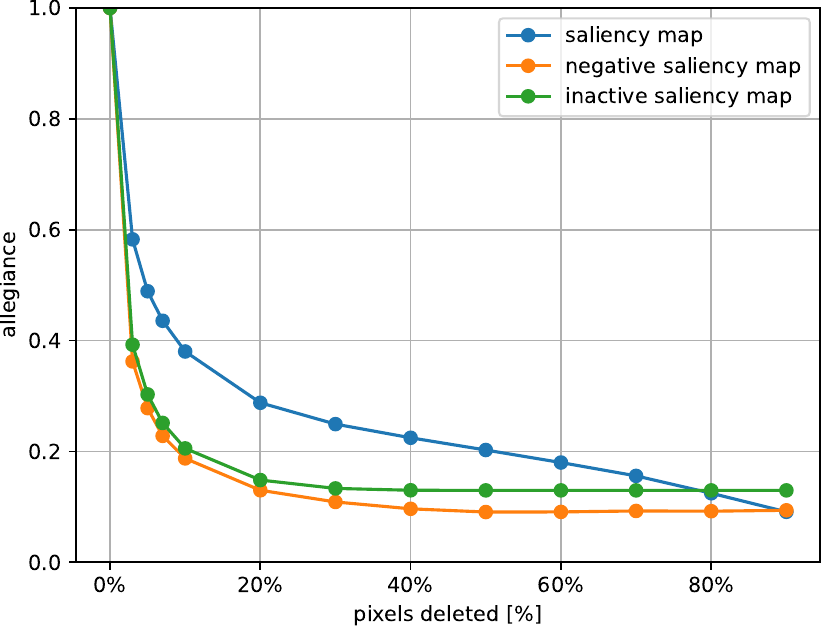}} \hfill
    \subfloat[Imagenette ResNet-18]{\includegraphics[width=0.475\textwidth]{visualizations/graphs/cifar10/resnet18_False/train/1.pdf}} \\
    \subfloat[Imagenette ResNet-18 pre-trained]{\includegraphics[width=0.475\textwidth]{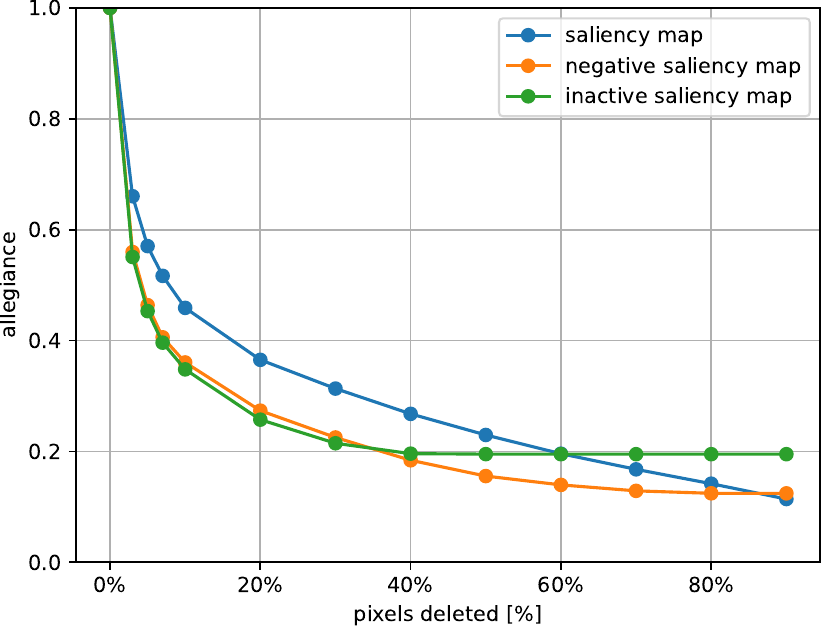}} \hfill
    \subfloat[Imagenette ConvNeXt pre-trained]{\includegraphics[width=0.475\textwidth]{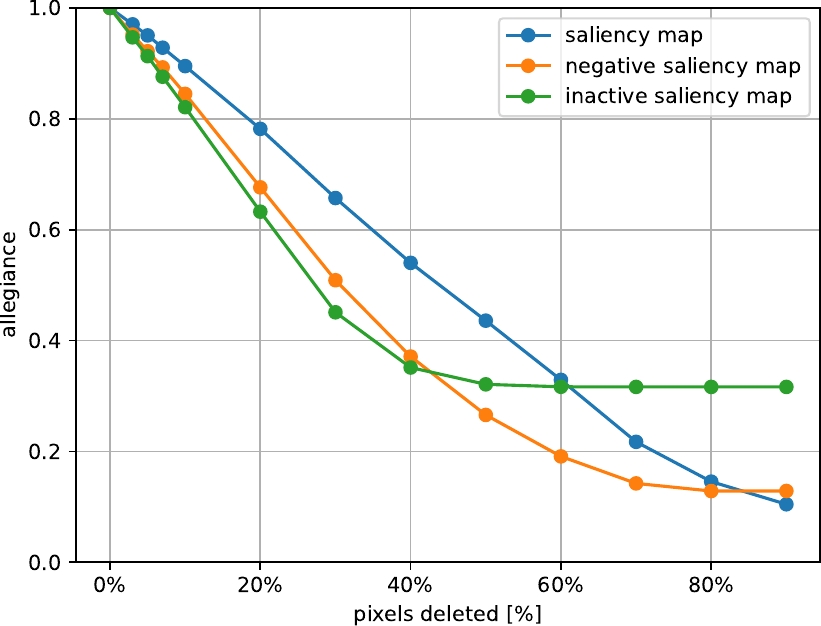}} \\
    \caption{White-Deletion Benchmark on the training set.}
    \label{fig: white-deletion benchmark on the training set}
\end{figure}





\end{document}